\documentclass[11pt, oneside]{Thesis} 

\graphicspath{{Pictures/}} 

\usepackage[square, numbers, comma, sort&compress]{natbib} 
\usepackage{subcaption}
\usepackage{enumitem}
\usepackage{multirow}
\usepackage{threeparttable}
\usepackage{hyperref}
\usepackage[colorlinks]{hyperref}
\hypersetup{urlcolor=black, colorlinks=false} 
\title{\ttitle} 

\begin{document}

\frontmatter 

\setstretch{1.5} 

\fancyhead{} 
\rhead{\thepage} 
\lhead{} 

\pagestyle{fancy} 

\newcommand{\HRule}{\rule{\linewidth}{0.5mm}} 

\hypersetup{pdftitle={Alzheimer's Disease Diagnosis Based on Cognitive Methods in Virtual Environments and Emotions Analysis}}
\hypersetup{pdfsubject=\subjectname}
\hypersetup{pdfauthor=Juan Manuel Fernandez Montenegro}
\hypersetup{pdfkeywords=Computer Vision; Machine Learning; Virtual Reality; Alzheimer's Disease}


\begin{titlepage}
\begin{center}

\textsc{\LARGE Kingston University}\\[1.5cm] 
\textsc{\Large Doctoral Thesis}\\[0.5cm] 

\HRule \\[0.4cm] 
{\huge \bfseries Alzheimer's Disease Diagnosis Based on Cognitive Methods in Virtual Environments and Emotions Analysis}\\[0.4cm] 
\HRule \\[1.5cm] 

\begin{minipage}{0.5\textwidth}
\begin{flushleft}
\emph{Author:}\\
{Juan Manuel Fernández Montenegro} 
\end{flushleft}
\end{minipage}
\begin{minipage}{0.4\textwidth}
\begin{flushright}
\emph{Supervisor:} \\
{Dr. Vasileios Argyriou\\Prof. Mary Chambers} 
\end{flushright}
\end{minipage}\\[3cm]

\textit{in the}\\[0.4cm]
\href{http://sec.kingston.ac.uk/research/research-centres/digital-information-research-centre/} {Digital Information Research Centre} \\
\href{http://sec.kingston.ac.uk/}{Faculty of Science, Engineering and Computing}\\[1.5cm] 

{\today}\\[2.5cm] 

\small This Thesis is being submitted in partial fulfilment of the requirements of Kingston\\
 University for the Degree of Doctor of Philosophy (Ph.D.)\\[1cm] 

\vfill
\end{center}

\end{titlepage}


\Copyright{

\addtocontents{toc}{\vspace{1em}} 

\begin{itemize}
\item[\tiny{$\blacksquare$}] The author of this thesis (including any appendices and/or schedules to this thesis) owns any copyright and rights in it (the "Copyright") and he has given to Kingston University certain rights to use such Copyright for any administrative, promotional, educational and/or teaching purposes.
\item[\tiny{$\blacksquare$}] Copies of this thesis, either in full or in extracts and whether in hard or electronic copy, may be made only in accordance with the Copyright, Designs and Patents Act 1988 (as amended) and regulations issued under it or, where appropriate, in accordance with licensing agreements which the University has from time to time. This page must form part of any such copies made.
\item[\tiny{$\blacksquare$}] The ownership of certain Copyright, patents, designs, trademarks and other intellectual property (the "Intellectual Property") and any reproductions of copyright works, for example graphs and tables ("Reproductions"), which may be described in this thesis, may not be owned by the author and may be owned by third parties. Such Intellectual Property and Reproductions cannot and must not be made available for use without the prior written permission of the owner(s) of the relevant Intellectual Property and/or Reproductions.
\item[\tiny{$\blacksquare$}] The report may be freely copied and distributed provided the source is explicitly acknowledged and copies are not made or distributed for profit or commercial advantage, and that copies bear this notice and the full citation on the first page. To copy otherwise, to republish, to post on servers or to redistribute to lists, requires prior specific permission.
\item[\tiny{$\blacksquare$}] Further information on the conditions under which disclosure, publication, exploitation and commercialisation of this thesis, the Copyright and any Intellectual Property and/or Reproductions described in it may take place is available in the University IP Policy, in any relevant Thesis restriction declarations deposited in the University Library, The University Library's regulations and in The University's policy on presentation of Theses.\\
\end{itemize}


}

\clearpage 


\Declaration{

\addtocontents{toc}{\vspace{1em}} 

\begin{itemize}
\item[\tiny{$\blacksquare$}] This report is submitted as requirement for a Ph.D. Degree in the School of Computing and Information Systems (Faculty of Science, Engineering and Computing) at Kingston University. It is substantially the result of my work except where explicitly indicated in the text.
\item[\tiny{$\blacksquare$}] No portion of the work referred to in this report has been submitted in support of an application for another degree or qualification on this or any other UK or foreign examination board, university or other institute of learning.
\item[\tiny{$\blacksquare$}] The thesis work was conducted from October 2014 to September 2017 under the supervision of Dr. Vasileios Argytiou and Prof. Mary Chambers in the Digital Information Research Centre (DIRC) of Kingston University, London.\\
\end{itemize}


}

\clearpage 









\addtotoc{Abstract} 
\setstretch{1.2}
\abstract{\addtocontents{toc}{\vspace{1em}} 
Dementia is a syndrome characterised by the decline of different cognitive abilities. Alzheimer's Disease (AD) is the most common dementia affecting cognitive domains such as memory and learning, perceptual-motion or executive function. High rate of deaths and high cost for detection, treatments and patient’s care count amongst its consequences. Early detection of AD is considered of high importance for improving the quality of life of patients and their families. The aim of this thesis is to introduce novel non-invasive early diagnosis methods in order to speed the diagnosis, reduce the associated costs and make them widely accessible.

Novel AD's screening tests based on virtual environments using new immersive technologies combined with advanced Human Computer Interaction (HCI) systems are introduced. Four tests demonstrate the wide range of screening mechanisms based on cognitive domain impairments that can be designed using virtual environments. The proposed tests are focused on the evaluation of memory loss related to common objects, recent conversations and events; the diagnosis of problems in expressing and understanding language; the ability to recognise abnormalities; and to differentiate between virtual worlds and reality, and humans and machines.

The use of emotion recognition to analyse AD symptoms has been also proposed. A novel multimodal dataset was specifically created to remark the autobiographical memory deficits of AD patients. Data from this dataset is used to introduce novel descriptors for Electroencephalogram (EEG) and facial images data. EEG features are based on quaternions in order to keep the correlation information between sensors, whereas, for facial expression recognition, a preprocessing method for motion magnification and descriptors based on origami crease pattern algorithm are proposed to enhance facial micro-expressions. These features have been proved on classifiers such as SVM and Adaboost for the classification of reactions to autobiographical stimuli such as long and short term memories.}

\clearpage 


\setstretch{1.5} 

\acknowledgements{\addtocontents{toc}{\vspace{1em}} 

First and foremost I want to thank my supervisor Dr Vasileios Argyriou. I offer my sincerest gratitude for his patience, advice and guidance over the last three years. His input on the work for this thesis and my own personal development has been invaluable. I wish to express my thanks to the members of the Reading group. Their support during this journey was so important to my research as was the discussion of new and existing ideas in an open and inviting environment. I specifically wish to mention Dr. Jean-Christophe Nebel, Dr. Francisco Florez Revuelta, Dr. Gordon Hunter and Dr. Dimitrios Makris for their guidance and critical insight. Also I thank my fellow labmates, Dr. Pau Climent Pérez, Dr. Rob Dupré, Ioannis Kazantzidis and Cuc Nguyen, for their guidance throughout my PhD, their advices and for all the fun we have had in the last three years.

My sincere thanks to Kingston University for funding my research. I also wish to extend my thanks to the Kingston University post graduate support team who's provisioning of training and administrative support throughout my time was a huge help, and to the IT support team for their help with vital hardware and software.

Finally to my family; miña nai, pai, irmán, curmán, tía e avoa polo apoio que me deron durante este doutorado, a estancia en Londres e a miña vida en xeral. 
}
\clearpage 


\pagestyle{fancy} 

\lhead{\emph{Contents}} 
\tableofcontents 

\lhead{\emph{List of Figures}} 
\listoffigures 

\lhead{\emph{List of Tables}} 
\listoftables 


\clearpage 

\setstretch{1.5} 

\lhead{\emph{Abbreviations}} 
\listofsymbols{ll} 
{
\textbf{1D} & \textbf{1} \textbf{D}imensions \\
\textbf{2D} & \textbf{2} \textbf{D}imensions \\
\textbf{3D} & \textbf{3} \textbf{D}imensions \\
\textbf{4D} & \textbf{4} \textbf{D}imensions \\
\hypertarget{a_AAM}{\textbf{AAM}} & \textbf{A}ctive \textbf{A}ppearance \textbf{M}odels \\
\hypertarget{a_AD}{\textbf{AD}} & \textbf{A}lzheimer's \textbf{D}isease \\
\hypertarget{a_AI}{\textbf{AI}} & \textbf{A}rtificial \textbf{I}ntelligence \\
\hypertarget{a_AR}{\textbf{AR}} & \textbf{A}ugmented \textbf{R}eality \\
\hypertarget{a_AOR}{\textbf{AOR}} & \textbf{A}bnormal \textbf{O}bject \textbf{R}recognition \\
\hypertarget{a_AUs}{\textbf{AU}} & \textbf{A}ction \textbf{U}nit \\
\hypertarget{a_BCI}{\textbf{BCI}} & \textbf{B}rain \textbf{C}omputer \textbf{I}nterface \\
\hypertarget{a_BCS}{\textbf{BCS}} & \textbf{B}ayesian \textbf{C}ompressed \textbf{S}ensing \\
\hypertarget{a_BDTT}{\textbf{BDTT}} & \textbf{B}ot \textbf{D}octor \textbf{T}uring \textbf{T}est \\
\hypertarget{a_BGCS}{\textbf{BGCS}} & \textbf{B}ayesian \textbf{G}roup-Sparse \textbf{C}ompressed \textbf{S}ensing \\
\hypertarget{a_CNN}{\textbf{CNN}} & \textbf{C}onvolutional \textbf{N}eural \textbf{N}etworks \\
\hypertarget{a_DBN}{\textbf{DBN}} & \textbf{D}eep \textbf{B}elief \textbf{N}etworks \\
\hypertarget{a_DLT}{\textbf{DLT}} & \textbf{D}ichotic \textbf{L}istening \textbf{T}est \\
\hypertarget{a_DLTre}{\textbf{DLT}} & \textbf{D}ichotic \textbf{L}istening \textbf{T}est \textbf{r}ight \textbf{e}ar \\
\hypertarget{a_DROZ}{\textbf{DrOZ}} & \textbf{D}r \textbf{O}\textbf{Z} \\
\hypertarget{a_DTNnp}{\textbf{DTNnp}} & \textbf{D}istance \textbf{T}o \textbf{N}ose (\textbf{n}eutral vs. \textbf{p}eak) \\
\hypertarget{a_ECG}{\textbf{ECG}} & \textbf{E}lectrocardiogram \\
\hypertarget{a_EEG}{\textbf{EEG}} & \textbf{E}lectroencephalogram \\
\hypertarget{a_EMG}{\textbf{EMG}} & \textbf{E}lectromyogram \\
\hypertarget{a_EOG}{\textbf{EOG}} & \textbf{E}lectrooculogram \\
\hypertarget{a_ERD/ERS}{\textbf{ERD/ERS}} & \textbf{E}vent \textbf{R}elated \textbf{D}esynchronization/\textbf{S}ynchronization \\
\hypertarget{a_ERP}{\textbf{ERP}} & \textbf{E}vent \textbf{R}elated \textbf{P}otentials \\
\hypertarget{a_FACS}{\textbf{FACS}} & \textbf{F}acial \textbf{A}ction \textbf{C}oding \textbf{S}ystem \\
\hypertarget{a_FER}{\textbf{FER}} & \textbf{F}acial \textbf{E}xpression \textbf{R}ecognition \\
\hypertarget{a_FERA}{\textbf{FERA}} & \textbf{F}acial \textbf{E}motion \textbf{R}ecognition and \textbf{A}nalysis \\
\hypertarget{a_FR}{\textbf{FR}} & \textbf{F}acial \textbf{R}ecognition \\
\hypertarget{a_HCI}{\textbf{HCI}} & \textbf{H}uman \textbf{C}omputer \textbf{I}nteraction \\
\hypertarget{a_HOG}{\textbf{HOG}} & \textbf{H}istogram of \textbf{O}riented \textbf{G}radients \\
\hypertarget{a_ICA}{\textbf{ICA}} & \textbf{I}ndependent \textbf{C}omponent \textbf{A}nalysis \\
\hypertarget{a_IR}{\textbf{IR}} & \textbf{I}nfra\textbf{r}ed \\
\hypertarget{a_kNN}{\textbf{kNN}} & \textbf{k} \textbf{N}earest \textbf{N}eighbors \\
\hypertarget{a_LDA}{\textbf{LDA}} & \textbf{L}ocal \textbf{D}iscriminant \textbf{A}nalysis \\
\hypertarget{a_LSTM}{\textbf{LSTM}} & \textbf{L}ong \textbf{S}hort \textbf{T}erm \textbf{M}emory \\
\hypertarget{a_MCC}{\textbf{MCC}} & \textbf{M}ulti \textbf{C}lass \textbf{C}lassifier \\
\hypertarget{a_MMSE}{\textbf{MMSE}} & \textbf{M}ini \textbf{M}ental \textbf{S}tate \textbf{E}xamination \\
\hypertarget{a_MRI}{\textbf{MRI}} & \textbf{M}agnetic \textbf{R}esonance \textbf{I}maging \\
\hypertarget{a_OCC}{\textbf{OCC}} & \textbf{O}ne \textbf{C}lass \textbf{C}lassification \\
\hypertarget{a_OCSVM}{\textbf{OCSVM}} & \textbf{O}ne-\textbf{C}lass \textbf{S}upport \textbf{V}ector \textbf{M}achine \\
\hypertarget{a_PCA}{\textbf{PCA}} & \textbf{P}rincipal \textbf{C}omponent \textbf{A}nalysis \\
\hypertarget{a_pHOG}{\textbf{pHOG}} & \textbf{p}yramid \textbf{H}istogram of \textbf{O}riented \textbf{G}radients \\
\hypertarget{a_RBF}{\textbf{RBF}} & \textbf{R}adial \textbf{B}asis \textbf{F}unction \\
\hypertarget{a_ROC}{\textbf{ROC}} & \textbf{R}eceiver \textbf{O}perating \textbf{C}haracteristic \\
\hypertarget{a_ROI}{\textbf{ROI}} & \textbf{R}egions \textbf{O}f \textbf{I}nterest \\
\hypertarget{a_SEMdb}{\textbf{SEMdb}} & \textbf{S}pontaneous \textbf{E}motion \textbf{M}ultimodal \textbf{d}ata\textbf{b}ase \\
\hypertarget{a_SLUMS}{\textbf{SLUMS}} & \textbf{S}aint \textbf{L}ouis \textbf{U}niversity \textbf{M}ental \textbf{S}tatus \\
\hypertarget{a_SVM}{\textbf{SVM}} & \textbf{S}upport \textbf{V}ector \textbf{M}achine \\
\hypertarget{a_tSNE}{\textbf{t-SNE}} & \textbf{t} - distributed \textbf{S}tochastic \textbf{N}eighbor \textbf{E}mbedding \\
\hypertarget{a_VAT}{\textbf{VAT}} & \textbf{V}isual \textbf{A}ssociation \textbf{T}est \\
\hypertarget{a_VEs}{\textbf{VE}} & \textbf{V}irtual \textbf{E}nvironments \\
\hypertarget{a_VOG}{\textbf{VOG}} & \textbf{V}ideo \textbf{O}culo\textbf{g}raphy \\
\hypertarget{a_VOM}{\textbf{VOM}} & \textbf{V}irtual \textbf{O}bjects \textbf{M}emorization \\
\hypertarget{a_VR}{\textbf{VR}} & \textbf{V}irtual \textbf{R}eality \\
\hypertarget{a_VRS}{\textbf{VRS}} & \textbf{V}irtual vs. \textbf{R}eal \textbf{S}ounds \\

}







\clearpage 

\lhead{\emph{Symbols}} 

\listofnomenclature{lll} 
{
$c_k$ & Class \\
$S$ & Similarity function \\

$q$ & Quaternion \\
$\mathcal{H}$ & Hyper Complex numbers \\
$q^H$ & Conjugation of Quaternion \\
$\mathbf{q}$ & Quaternion column vector \\
$\mathbf{q}^T$ & Transpose of $\mathbf{q}$ \\
$\mathbf{q}^H$ & Conjugate Transpose of $\mathbf{q}$ \\
$q_r$ & Quaternion component - EEG Channel F7 \\
$q_i$ & Quaternion component - EEG Channel F8 \\
$q_j$ & Quaternion component - EEG Channel AF3 \\
$q_k$ & Quaternion component - EEG Channel AF4 \\
$\mathbf{S}$ & Quaternion Hermitian matrix \\
$\mathbf{\tilde{S}}$ & Complex Hermitian matrix \\
$\sigma$ & Eigenvalues \\

$l_i$ & Normalized coordinates of the $i_{th}$ Facial landmark according to the nose \\
$L_i$ & Coordinates of the $i_{th}$ Facial landmark \\
$L_n$ & Coordinates of the nose landmark \\
$l_i^{(n)}$ & Landmarks' coordinates on the emotion neutral frame \\
$l_i^{(p)}$ & Landmarks' coordinates on the emotion peak frame \\

$p_{i|j}$ & Conditional probability \\
$q_{ij}$ & Joint probability \\
$D_{KL}$ & Kullback-Leibler Divergence \\
$\sigma_i^2$ & Variance \\

$\alpha$ & Amplification factor \\
$B$ & Broadband temporal bandpass filter \\
$\delta$ & Small motion \\
$I$ & Image intensity \\
$f$ & Motion of the image \\

$C$ & Crease pattern \\
$d_i$ & $i_{th}$ distance to nose feature from facial landmarks \\
$d_{mi}$ & $i_{th}$ distance to nose feature from magnified facial landmarks \\
$d_P$ & Distance between Lang's Polygon nodes \\
$d_T$ & Distance between Shadow Tree nodes \\
$e_i$ & $i_{th}$ edge from Origami crease pattern \\
$f$ & Double cycle polygon that results in Lang's Polygon \\
$g$ & Shrinking function \\
$h_i$ & $i_{th}$ pHOG feature \\
$n_i$ & $i_{th}$ node from Origami crease pattern \\
${n_i}_x$ & Coordinate x of node i \\
${n_i}_y$ & Coordinate y of node i \\
$N_{in}$ & Internal node of the shadow tree \\
$N_{ex}$ & Leave node of the shadow tree \\
$T$ & Shadow Tree \\

$F1$ & F1 score \\
$P$ & Precision \\
$R$ & Recall \\


& & \\ 

}







\clearpage 

\lhead{\emph{Publications}}

\publications{\addtocontents{toc}{\vspace{1em}} 

Some ideas and figures have appeared previously in the following publications

Journals:

\begin{enumerate}[label=$\bullet$]
\item \textbf{J. M. F. Montenegro} and V. Argyriou, "Cognitive evaluation for the diagnosis of Alzheimer's disease based on Turing Test and Virtual Environments", \textit{Physiology \& Behavior}, vol. 173, pp. 42-51, 2017.
\item E. I. Konstantinidis, A. S. Billis, R. Dupre, \textbf{J. M. F. Montenegro}, G. Conti, V. Argyriou, and P. D. Bamidis, "IoT of active and healthy ageing: cases from indoor location analytics in the wild", \textit{Health and Technology}, pp. 1-9, 2016.
\item \textbf{J. M. F. Montenegro} and V. Argyriou, "Virtual environments and cognitive tests for dementia diagnosis", \textit{International Journal of Monitoring and Surveillance Technologies Research (IJMSTR)}, vol. 4, no. 1, pp. 10-23, 2016.
\end{enumerate}

Conferences:

\begin{enumerate}[label=$\bullet$]
\item \textbf{J. M. F. Montenegro} and V. Argyriou, "Gaze estimation using EEG signals for HCI in augmented and virtual reality headsets", \textit{IEEE 23rd International Conference on Pattern Recognition (ICPR)}, pp. 1159-1164, 2016.
\item \textbf{J. M. F. Montenegro}, B. Villarini, A. Gkelias, and V. Argyriou, "Cognitive behaviour analysis based on facial information using depth sensors", \textit{International Conference on Pattern Recognition (ICPR) - International Workshop on Understanding Human Activities through 3D Sensors (UHA3DS)}, pp. 4-8, 2016.
\item \textbf{J. M. F.	Montenegro}, A. Gkelias, and V. Argyriou, "Emotion Understanding Using Multimodal Information Based on Autobiographical Memories for Alzheimer’s Patients", \textit{Asian Conference on Computer Vision Workshops (ACCV-W)}, pp. 252-268, 2016.
\item \textbf{J. M. F.	Montenegro} and V. Argyriou, "Cognitive Tests for the Diagnosis of Alzheimer's disease based on Virtual Environments", \textit{IET International Conference on Technologies for Active and Assisted Living (TechAAL)}, pp. 1-5, 2015.
\item \textbf{J. M. F.	Montenegro} and V. Argyriou, "Diagnosis of Alzheimer's disease based on virtual environments", \textit{IEEE 6th International Conference on Information, Intelligence, Systems and Applications (IISA)}, pp. 1-6, 2015.
\end{enumerate}

}


\mainmatter 

\pagestyle{fancy} 



\chapter{Introduction} 

\label{Chapter1} 

\lhead{Chapter 1. \emph{Introduction}} 

Dementia is a term used to encompass a number of symptoms such as the decline of memory, reasoning, language or perceptual interpretation. Specific combinations of these symptoms are used to define different types of dementia such as Alzheimer's or Parkinson's disease. Alzheimer is a disease that usually affects elder people, with the number of cases increasing over the last decades. The number of people with early onset dementia (people under 65 years old) has also increased in the UK and worldwide~\cite{Prince2014}. Alzheimer's is the most common type; it accounts for 60 to 80 percent of cases, and one of the most noticeable symptoms is the difficulty in learning new information. In addition, when the disease advances, there are other symptoms such as disorientation, mood and behaviour changes; confusion about events, date and place; being suspicious about family, friends and caregivers; and difficulty in speaking, writing and walking~\cite{AlzAsso2017}.

According to the World Health Organization, 1.37\% of the estimated deaths around the world in 2030 will be caused by Alzheimer's disease (AD) and other dementias. People with medium dementia have an average life expectancy of 8 years~\cite{AlzAsso2017}. Currently, when Alzheimer's disease is diagnosed, the neuronal damage is spread enough to make it irreversible~\cite{Tarnanas2014}. When neurons die, the other neurons do not divide and replace them, as other cells do, so the damage cannot be reversed~\cite{AlzAsso2017}. Therefore, it is important to detect dementia at its very early stages in order to reduce the deterioration speed~\cite{Abe2013,Whalley2009}.

\begin{figure}
\centering
  \includegraphics[scale = 0.5]{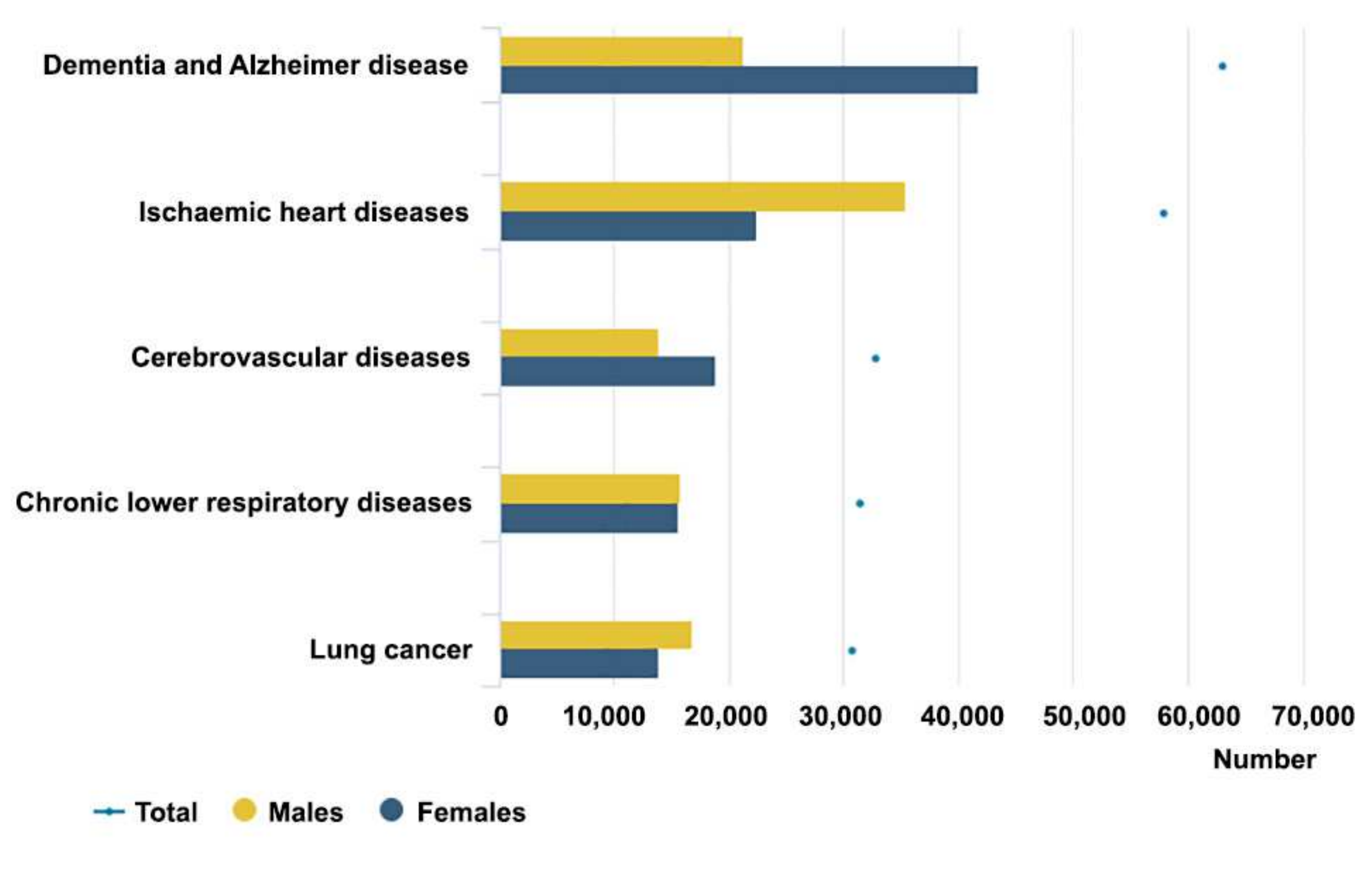}
  \caption{This figure shows the number of deaths in England and Wales during 2016 from top five leading causes. It shows dementia as the major cause of death~\cite{ONS2017}.}
  \label{fig:deathUK}
\end{figure}

The cost of dementia is another issue to take into consideration. The worldwide cost of dementia is \$818 billion, and it will become a trillion by 2018~\cite{Prince2015}. The cost of dementia in the UK estimated in 2013 was 26.3 billion pounds; 4.3 billion were spent on healthcare costs, of which around 85 million were spent on diagnosis~\cite{Prince2014}. Therefore, it is essential to develop affordable diagnosis and support tools to help limit the raising cost associated with dementia. One of the proposed initiatives is focused on the implementation of e-health (the use of Information Communication Technology (ICT)) solutions to reduce the cost and to make the health systems and solutions universally accessible~\cite{Lewis2012}. Arief et al.~\cite{Prince2015} presented the strengths of e-health tools such as the improvement in accessing health-care services by senior citizens, their cost-effectiveness and their efficiency in managing health resources.

Alzheimer's symptoms can be studied to improve the results of previous approaches or create novel and more accurate diagnosis tools relying on new affordable and publicly available technologies. Alzheimer's detection methods are classified into two different categories: invasive and non-invasive. Invasive methods require obtaining data from the interior of the patient's body through procedures such as lumbar puncture or blood extraction. These invasive methods try to define potential biomarkers that prove an accurate indicator of Alzheimer's~\cite{Han2012}. Some biomarkers such as the amount of beta-amyloid and tau in cerebrospinal fluid have been validated as indicators of AD. However, the validated biomarkers tests are complex and expensive therefore current research continues to try and find simpler and more cost effective alternatives~\cite{AlzAsso2017}. In addition, these tests are not always safe and comfortable for the patient, while some of them are unbearably painful. On the other hand, non-invasive tests are harmless and more convenient during the diagnosis process.

\begin{figure}
\centering
  \includegraphics[scale = 0.85]{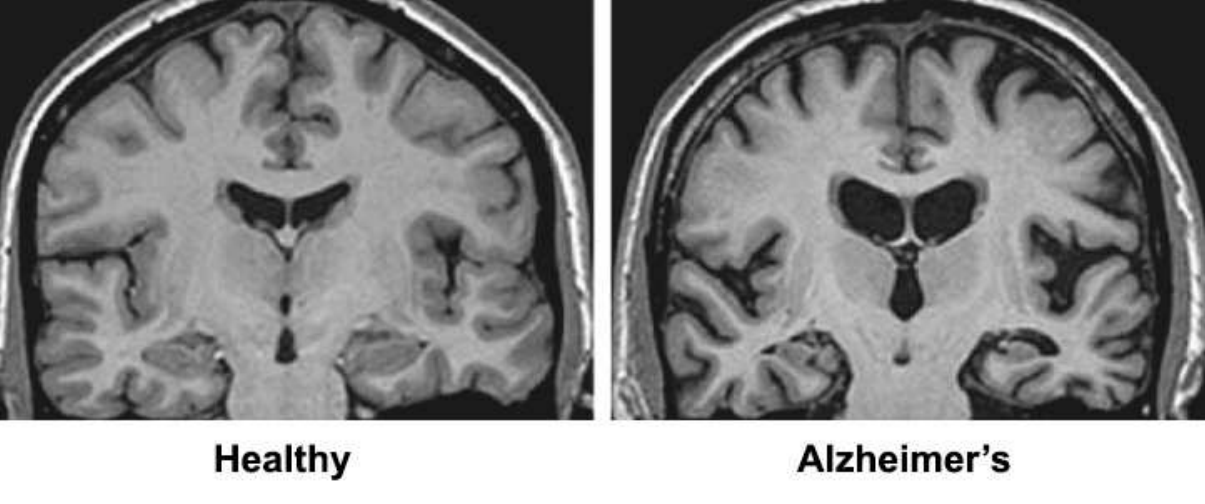}
  \caption{Magnetic Resonance Imaging (MRI) sample of a healthy person's brain (left) and an Alzheimer's patient's one (right).}
  \label{fig:MRI}
\end{figure}

Focusing on non-invasive methods, several approaches are followed in order to detect Alzheimer's disease in its early stages such as neuroimaging, behaviour and emotion analysis or cognitive approaches. Neuroimaging techniques, such as magnetic resonance imaging (MRI) or computed tomography (CT), are used to detect changes in the patients' brains produced by the disease~\cite{AlzAsso2017, Akgul2010}. These techniques compare the head \hyperlink{a_MRI}{MRI} data of the patient with the corresponding data of patients with Alzheimer (see Figure~\ref{fig:MRI}). Nevertheless, since acquisition of these images usually involves the use of medical equipment that is not easily accessible and the process is unpleasant (claustrophobic and noisy), these methods are not suitable for testing large groups of people due to cost and time. They are not recommended either for mental disease patients due to safety and discomfort reasons.

Behaviour analysis approaches try to detect abnormal reactions to common circumstances or problems in daily living activities. Some methods require the installation of sensors in the patient's house in order to identify certain events such as forgetting to turn the gas off~\cite{Abe2013}. One problem with this approach is that it requires the consent of the patient to install the sensors. Moreover, these methods' results are not precise enough (less than 75\% detection rates) to provide an accurate technique for dementia diagnosis at early stages.

Other behaviour related approaches will include emotion analysis. As stated in~\cite{APA2013}, one of the Alzheimer's symptoms is the decline of social cognition, therefore many proposals are focused on patient's capability to recognise emotions~\cite{Sapey-Triomphe15,VandenStock15}. On the other hand, other approaches analyse patient's reactions to specific stimuli. Due to the importance of emotion recognition in different disciplines, such as neuroscience or psychology, several methods have been proposed for human emotion recognition~\cite{Zeng2009}. Several proposals try to analyse those reactions/emotions using different information, such as Electroencephalogram (EEG), eye tracking data, audio or facial gestures. Some dementias such as Lewy Body dementia result in a lack of facial expression, therefore facial expression analysis would be inconceivable. However, these approaches could prove particularly useful in AD where cognitive decline is associated with an increased facial expressiveness~\cite{Seidl12}.

Eye tracking approaches are modelled on how doctors usually perform physical examinations. Eye movement and reflexes are monitored as part of the patient’s mental state evaluation~\cite{AlzAsso2017} since visuospatial functions and visual processes go into decline due to Alzheimer~\cite{Pereiraetal2014}. Therefore, some of the visual methods to evaluate the mental state of a patient include the measure of the reaction time to certain stimuli, the assessment of attention and the evaluation of patient’s visual memory~\cite{Pereiraetal2014,Crutcheretal2009,Rosler2000}.

Although there is merit in using all the previous non-invasive methods, cognitive tests remain the most commonly used in diagnosing \hyperlink{a_AD}{AD}. They are question/task/problem based and are intended to measure patients' cognition, which can be defined as the use of information that has been previously collected by a person to make behavioural decisions~\cite{Rowe2014}. Some of the most used cognitive tests are the Mini mental state examination (MMSE)~\cite{Fountoulakisetal2000} and the clock-drawing test~\cite{Vyhnalek2017}. Cognitive (or aptitude) tests have high accuracy when it comes to Alzheimer's detection. Although, one of their weaknesses is the evaluation of brain's capacity to compensate brain damage (cognitive reserve)~\cite{Tarnanas2014,Whalley2009}. Another well-known problem is the adaptability of the tests according to the patient's IQ, since most of the tasks that integrate an Alzheimer's detection cognitive test usually are too simple to evaluate intelligent patients. Therefore, it is necessary to create tests where the results are not correlated with the IQ of the patient. The use of computerised tests helps to create intelligence adaptable cognitive tests~\cite{Wild2008,Tarnanas2014}. Moreover, based on the available technology, it is possible to design new types and more effective tests such as virtual environments (VEs). Virtual environments provide additional advantages to cognitive tests, since it is possible to immerse the patient in a controlled situation~\cite{Taekman2010,Tarnanas2013,Weibel2011,Garcia2015}.

Parsons et al.~\cite{Parsons2011,ParsonsCourt2011,Parsons2009} have demonstrated the ecological validity of \hyperlink{a_VEs}{VEs} and their benefits for neurocognitive assessment, such as the precision of data retrieved by the computer and the better control of the environment. They also prove the advantages of using virtual reality (VR) over the traditional pen and paper methods.

When it comes to the symptoms of Alzheimer's, the American Psychiatric Association has categorised in~\cite{APA2013} the cognitive domains affected by Alzheimer's disease: complex attention, executive function, learning and memory, language, perceptual-motor and social cognition. The existing tests try to examine specific cognitive areas. For example, the \hyperlink{a_MMSE}{MMSE} studied different cognitive areas such as learning and memory, complex attention and the language cognitive domains.
Analysing the cognitive domains affected by Alzheimer's, a relation can be found with the areas that Alan Turing uses to define intelligence. Alan Turing proposed in 1950 a test related with human and computer cognition. It tests the degree of intelligence of a machine in relation with its ability to impersonate
a human~\cite{French2012}. In order for a computer to be considered intelligent, it should possess the following capabilities~\cite{Russel1995}: natural language processing, knowledge representation, automated reasoning and machine learning. When comparing these capabilities with the correspondent cognitive domains it is possible to consider the use of the existent Turing Test for evaluating human cognition. Furthermore, according to Warwick et al.~\cite{Warwick2014}, the success of the Turing test does not depend only in the quality of the computer's AI; it also depends on the intelligence of the human that is judging the machine. Therefore, if you reverse this test, it is possible to obtain a cognitive test that checks the degree of intelligence and can detect cognitive impairment in humans.


\section{Aims and Objectives}

Some issues were outlined in previous section such as the importance of early \hyperlink{a_AD}{AD} diagnosis and the huge cost of this disease. Taking them into consideration the general aim of this work is described as the improvement of Alzheimer's screening methods for early diagnosis, i.e., the improvement of diagnosis/screening accuracy, the reduction of the associated cost and the increase of the overall accessibility to the end users.

In order to reach these goals the objectives below were set.

\begin{itemize}
\item Propose novel non-invasive cognitive Alzheimer's Screening tests for early diagnosis taking advantage of Virtual Environments.
\item Propose a solution to reduce the effects of cognitive reserve during cognitive tests.
\item Reduce the cost and increase the accessibility of the screening tools using the latest Virtual Reality technology.
\item Create a novel AD dataset that contains multimodal data of reactions elicited from autobiographical memories. These reactions and the absence of them will contribute extra information to issue more robust \hyperlink{a_AD}{AD} screening feedback.
\item Propose novel features for emotion classification focusing on micro-expressions triggered by autobiographical memories.
\end{itemize}


\section{Contributions to Knowledge}

The first contribution is the creation of novel non-invasive Alzheimer's early detection screening tests based on \hyperlink{a_VEs}{VEs} \cite{Montenegro2017,Montenegro2016,Montenegro2015,MontenegroIISA2015}. The proposed and developed tests are the first Alzheimer's detection screening tests that provide a full immersion in Virtual Environments resulting in the patients being more able to focus on the tasks, while additional information related to the environment-patient interaction is obtained. Also, it is possible to adapt some of the tasks to the patient's IQ level during the test, avoiding the ceiling effect and cognitive reserve. Ceiling effect occur when the test results are positively affected by personal characteristics independent of cognitive functioning, being the high IQ one of the most common causes~\cite{Franco2010}. Furthermore, the proposed tests do not consider only memory tasks but introduce novel approaches based on the concept of differentiating between reality and a virtual world; and on a new cognitive test that aims to identify normal and abnormal events or objects present in the virtual scene. Also, novel checks are proposed based on a reversed Turing test, where the patients' cognition is evaluated according to their ability in distinguishing absurd from correct information or a machine from a human \cite{Montenegro2017}.

The second contribution is the creation of a novel multimodal database focused on Alzheimer's disease \cite{MontenegroGaze2016,MontenegroFace2016,MontenegroEmo2016}. It has been proved that for AD patients, semantic, autobiographical and implicit memory are better preserved than recent memory; therefore our work is based on the subjects' autobiographical memory~\cite{Han14,Irish11,APA2013}. Thus a novel dataset was created based on these symptoms providing multimodal data such as \hyperlink{a_EEG}{EEG} signal, eye tracking data and RGB, Infrared (IR) and Depth video data. The Spontaneous Emotion Multimodal database (SEMdb) was created in order to register spontaneous emotions of the participants triggered by autobiographical stimuli. These recorded emotions are supposed to be different between Alzheimer's patients and healthy participants. The database only contains information recorded from healthy subjects so it is possible to infer what could be described as 'normal' human behaviour, thus any abnormality would indicate a cognitive impairment.

The purpose of the last two contributions is the introduction of emotion recognition techniques for the detection of early dementia symptoms. These proposals are presented as extra assessment tools to be used separately from the \hyperlink{a_VR}{VR} tests in order to further evaluate participants cognition. Both contributions present novel features for the classification of emotions, including the emotions related with autobiographical stimuli from \hyperlink{a_SEMdb}{SEMdb}. The third contribution presents novel \hyperlink{a_EEG}{EEG} features \cite{MontenegroGaze2016,MontenegroEmo2016}. Quaternion principal component analysis (Quaternion PCA) is applied to the \hyperlink{a_EEG}{EEG} signal recorded from four frontal \hyperlink{a_EEG}{EEG} sensors (See Section \ref{QuaternionPCA}). The main idea is the extraction of novel features from a reduced number of sensors trying to provide similar emotion classification results to the more expensive, but state-of-the-art methods. A reduced number of sensors would enable an easy implementation of the sensors on Virtual Reality headsets.

Finally, the last contribution introduces novel facial features. Crease patterns similar to the origami ones are created from facial landmarks extracted from RGB facial images. These novel features provide a resistant to noise and partial occlusions novel representation of the facial expressions. They are also utilised with machine learning techniques to improve the performance of emotion classifiers. These last two contributions analyse the different emotions generated on a healthy control group. They create a model that describes the emotions of the healthy population. Thus, every reaction detected out of these models could be considered as a possible sign of dementia. In this case, as an external camera is used to record the subjects' face, it is not possible a combination with current full immersive \hyperlink{a_VR}{VR} technologies.


\section{Structure of the Thesis}

The remainder of this thesis is organised as follows: Chapter~\ref{Chapter2} presents an overview of the literature and background information. Initially an analysis of state-of-the-art cognitive tests for dementia detection is presented, focusing on methods that use Virtual Reality technologies. Afterwards, behaviour and emotion recognition methods are analysed. Firstly, a summary of the \hyperlink{a_EEG}{EEG} methods will be provided. Secondly, facial expression based approaches will be analysed. Finally, an overview of machine learning concepts will be presented and databases for \hyperlink{a_AD}{AD} and emotion analysis will be summarised.

The novel Alzheimer's screening cognitive tests on Virtual Reality are outlined in Chapter~\ref{Chapter3}. Four novel tests demonstrate the Virtual Environment versatility. These tests are compared against computerised state-of-the-art cognitive tests proving better performance for \hyperlink{a_AD}{AD} early screening. This chapter also explains the advantages of the technology used in terms of affordability and accessibility. Finally, the novel tests performance and the participants' assessment of the test are used to analyse the validity of Virtual Environments for \hyperlink{a_AD}{AD} early screening.

Chapter~\ref{Chapter4} introduces the novel Spontaneous Expression Multimodal database. This is the first spontaneous micro expressions due to emotions dataset created recording multimodal data of participants watching autobiographical and memory related stimuli. After \hyperlink{a_SEMdb}{SEMdb} is presented, the usage of \hyperlink{a_EEG}{EEG} data to detect the gaze position on the monitor using the novel Quaternion \hyperlink{a_PCA}{PCA} \hyperlink{a_EEG}{EEG} features is presented. Gaze location can be used to improve interaction on Virtual Environments and it can also be used to evaluate \hyperlink{a_AD}{AD} patients' attention. Tracking gaze using \hyperlink{a_EEG}{EEG} sensors will reduce the size of the current eye tracking devices and it will enable their integration into virtual reality devices. Finally, the novel \hyperlink{a_EEG}{EEG} features are also used for emotion recognition, creating a model that allows the classification of autobiographical memories in healthy participants.

Novel facial features based on origami crease patterns are presented in Chapter~\ref{Chapter5}. The creation of the origami features from facial landmarks is explained in detail. These novel features are tested for the classification of emotions using two datasets: the CK+ \ref{db_CK} and the \hyperlink{a_SEMdb}{SEMdb}. They are compared against state-of-the-art emotion classification approaches. In addition, noise and partial occlusion proof of concept experiments are provided to demonstrate the strengths of the novel features.

Finally in Chapter~\ref{Chapter6} conclusions and intended future work is outlined



\chapter{Literature Review} 

\label{Chapter2} 

\lhead{Chapter 2. \emph{Literature Review}} 

This chapter contains relevant work related to areas such as Alzheimer's Disease screening cognitive tests in Virtual Environments and emotion analysis since they are core elements in this research. Therefore, the most state-of-the-art methods and popular databases are analysed. In addition, other relevant concepts, metrics, methodologies and techniques are addressed in accordance with the work to be presented in subsequent chapters.

\section{Dementia Screening and Diagnosis Tests}
Apart from the invasive/non-invasive classification, Alzheimer's detection methods (see Chapter \ref{Chapter1}) can also be classified as non-cognitive and cognitive tests. Cognitive tests encompass the methods that assess the patients' cognition; these procedures are non-invasive as well as easy to implement. On the other hand, non-cognitive tests include all other methods, irrespective of whether they are invasive or non-invasive, used to detect and diagnose dementia.

\section{Non-Cognitive Tests}

Regarding non-cognitive approaches, in \cite{AlzAsso2017,Han2012} some of the methods used to detect dementia are based on defining potential biomarkers that are acquired through invasive techniques, such as cerebrospinal fluid tau protein and beta-amyloid peptide. Lau et al. proposed in \cite{Lauetal2015} salivary trehalose as a reliable Alzheimer detector biomarker obtained through non-invasive methods, since taking samples from inside the mouth might not be considered body invasion. By using their own cell-based biosensor to measure salivary sugar they managed to accurately classify \hyperlink{a_AD}{AD} patients, Parkinson’s disease patients and healthy groups. Nevertheless, these biomarkers have not been validated yet on large groups. Moreover, salivary studies for diagnosis of diseases are on their early stages and most of these studies are limited by the lack of further research~\cite{Zhang2016}.

Magnetic Resonance Imaging is another non-cognitive technique commonly used for Alzheimer detection that provides accurate results since it enables to visualise the deterioration of the brain. In the works presented in \cite{Unayetal2011,Akguletal2010} the head \hyperlink{a_MRI}{MRI} data of the patient is compared with the corresponding data of patients with Alzheimer's by using machine learning techniques. Agüera‐Ortiz et al.~\cite{AgueraOrtiz2017} present a study using different \hyperlink{a_MRI}{MRI} techniques, such as Diffusion Tensor Imaging and Fluid-attenuated Inversion Recovery, in order to correlate the apathy of \hyperlink{a_AD}{AD} patients with changes in white and grey matter. As a result of their study they found evidence of correlation between deficit of thinking and bilateral damage in the corpus callosum and the internal capsule. Furthermore, they suggest there is a relation between right-sided damage in the brain and apathy, which they associate with the purpose of the right hemisphere to initiate behaviour and monitor the environment. However, acquisition of magnetic resonance images involves the use of expensive, non-easily accessible and uncomfortable medical equipment that is not recommendable for mental disease patients.

Non-cognitive approaches, more or less invasive, are usually uncomfortable, expensive and time-consuming. Therefore, they are not the most convenient for \hyperlink{a_AD}{AD} patients. Due to these drawbacks, our work will focus on cognitive approaches.

\section{Cognitive Tests, Virtual Environments and Sensors}
In Chapter \ref{Chapter1} 'cognition' was defined as the use of the information that has been previously collected by a person to make behavioural decisions~\cite{Rowe2014}. \label{CogDef} However, 'cognition' is such a wide term that it will be not necessary to provide a specific definition~\cite{Allen2017}. Cognition encompasses several domains but this research focuses on those affected by Alzheimer's disease. The American Psychiatric Association defined six neurocognitive domains impaired by \hyperlink{a_AD}{AD}.

\begin{itemize}
\item Complex attention: sustained, divided and selective attention, i.e., the capacity of maintaining attention to a stimulus, focusing on two stimuli at the same time or keeping attention to one specific stimulus while others are interfering.
\item Executive function: planning, decision-making, working memory, responding to feedback, overriding habits and mental flexibility.
\item Learning and memory: immediate, recent and very long-term memory. Recent and very long-term memory include autobiographical memory.
\item Language: understanding and expression.
\item Perceptual-motor: recognition of figures by shape or colour (including face recognition) or hand-eye coordination.
\item Social cognition: ability to identify other people's emotions.
\end{itemize}

This section include the literature approaches that evaluate these domains for \hyperlink{a_AD}{AD} screening. Different approaches for early \hyperlink{a_AD}{AD} screening are analysed starting from the simplest and most used pen and paper assessments. Moreover, literature that demonstrate the advantages of computerized tests is analysed, such as the automatic correction of the patients' educational level bias and the wide range of cognitive domains that could be evaluated in a controlled virtual environment. Finally, approaches that analyse \hyperlink{a_AD}{AD} patients' behaviour using sensors are also presented due to their potential to evaluate \hyperlink{a_AD}{AD} patients continuously without the requirement of specific tests, just analysing daily life activities.

When it comes to cognitive methods that challenge participants' cognition for early detection, the most popular ones are those based on problem-solving tasks and questions, in order to detect a cognitive impairment. Such methods are the most commonly used by doctors. Cordell et al. present in \cite{Cordelletal2013} a comparative study of these methods, including the \hyperlink{a_MMSE}{MMSE}, Mini-Cog test or Saint Louis University Mental Status (SLUMS) \cite{CruzOliveretal2014}.The Mini Mental State Examination is one of the most well-known and used tests \cite{Fountoulakisetal2000,Mitchell2017}. The \hyperlink{a_MMSE}{MMSE} consists of twenty tasks that cover ten domains: orientation, registration, attention, calculation, recall, naming, repetition, comprehension, writing and construction. Amongst these domains, orientation and attention are those which count the most. Mitchell~\cite{Mitchell2017} concludes that the \hyperlink{a_MMSE}{MMSE} is adequate for screening Alzheimer's disease in advanced stages but not Mild Cognitive Impairment (MCI). He also describes some of the limitations included on the \hyperlink{a_MMSE}{MMSE} and other cognitive tests, such as the ceiling effect, since the \hyperlink{a_MMSE}{MMSE} is not complex enough for high IQ patients~\cite{Fountoulakisetal2000,Mitchell2017}. Mini-Cog test is better than the \hyperlink{a_MMSE}{MMSE} for early \hyperlink{a_AD}{AD} detection~\cite{Yang2016}. The test is composed of two tasks that take around three minutes. The first one is a three-word recall task whereas the second requires the participant to draw a clock~\cite{Chan2014}. Each correct recalled word from the first task provides one point and the correct drawing of the clock is one point. Cognitive impairment is diagnosed if either the first task score is 0 or the score of the first task is 1 or 2 and the second task is 0. \hyperlink{a_SLUMS}{SLUMS} is other cognitive screening pen and paper test that is composed of eleven tasks and a maximum score of thirty~\cite{Szczesniak2016}. It takes approximately seven minutes and assesses orientation, memory, reasoning and executive functioning amongst others. Szcześniak et al.~\cite{Szczesniak2016} demonstrate the validity of \hyperlink{a_SLUMS}{SLUMS} examination for Alzheimer's disease screening. Its results are correlated with other examinations such as the \hyperlink{a_MMSE}{MMSE}. In addition, its specificity and sensitivity is superior to the \hyperlink{a_MMSE}{MMSE} for Mild Cognitive Impairment.

Other examples of cognitive tests that directly challenge cognition involve visual or auditory tasks. The memorization of interacting objects in the Visual Association Test (VAT) \cite{Lindeboometal2002} requires the visualization of images. First, six images that contain two related objects or animals are shown to the participants. Afterwards, a new set of six images (where only one of the objects/animals of the previous set appears) is shown and the participants have to recall the missing object/animal. The Dichotic Listening test (DLT) \cite{Ducheketal2005} is used to detect the prevalence of the right ear when sounds are memorised by Alzheimer's patients. With this aim, two one-digit numbers are played on each ear of the participant at the same time, e.g., digit 1 on the right ear and digit 8 on the left ear. This procedure is repeated six times using different numbers and at the end the participants have to recall all the numbers. Healthy participants usually remember digits from both ears equally whereas right ear dominance is detected in \hyperlink{a_AD}{AD} patients. This test also assesses memory impairments. The results provided by these tests show high levels of accuracy and specificity. Nevertheless, it should be noted that these tests do not provide variations to reduce the ceiling effect.

There are methods focused on visual impairments since visuospatial functions and visual processes decline due to Alzheimer \cite{Quentaletal2013,Pereiraetal2014}. These methods evaluate the mental state of patients through the measurement of their reaction time in response to certain stimuli, the assessment of attention or the evaluation of the patient’s visual memory \cite{Pereiraetal2014,Crutcheretal2009}. The assessment of visual impairments is usually analysed by a doctor or by using eye tracking techniques. Eye tracking research is separated into two main areas: eye detection and gaze tracking. Eye detection tries to locate the eyes of a human shown in an image or a video sequence, whereas gaze tracking estimates where a person is looking in 3D space. Gaze is tracked using a device that analyses eye movements (eye tracker), estimating the eyes position and tracking their movement over time to determine the 3D line of sight.

There are different methods for eye detection and gaze tracking that can be classified into two categories: sensor and computer vision based techniques~\cite{AlRahayfehetal2013,Gredebacketal2015,Majarantaetal2014}. Sensor based approaches measure electric potentials utilising the electrooculogram (EOG) or the electroencephalogram. The eye acts as a dipole, considering the cornea as the positive pole and the retina as the negative one. With regard to the \hyperlink{a_EOG}{EOG} sensors, they are located around the eyes, with the electric potential field to be steady when the eye is at its normal state. If the retina approaches one sensor and the cornea the opposite one, a change in the \hyperlink{a_EOG}{EOG} signal is produced, which is used to track the eye movements~\cite{Manabeetal2015,Hladek2017,Steinhausenetal2014}. \hyperlink{a_EEG}{EEG} sensors are placed on the head scalp and they record brain signals and artifacts, such as \hyperlink{a_EOG}{EOG} and other muscle movements~\cite{Muller-Putzetal2015}. Most \hyperlink{a_EEG}{EEG} studies that are not focused on gaze tracking, try to remove these artifacts, in order to work only with brain signals. On the other hand, current methods for eye tracking use \hyperlink{a_EEG}{EEG} signals and their \hyperlink{a_EOG}{EOG} artifacts to detect saccade and pursue movements. These methods use techniques such as Independent Component Analysis (ICA) to remove or extract artifacts, extract certain features, (e.g. amplitude, spectral power) and use them with classifiers such as k-Nearest Neighbour (kNN),~\cite{Tomi2017,Samadi2014}.

Computer vision approaches (video oculography, VOG) use cameras to detect and track eyes over time. Using the obtained information about the eyes’ location and the head pose, the gaze direction is estimated. Eye detection is influenced by the available eye model, the illumination conditions, viewing angle and several other parameters,~\cite{Hansenetal2010}. Other techniques are based on their geometric and photometric properties: shape-based, feature-based, appearance-based and hybrid. Some of these techniques employ active \hyperlink{a_IR}{IR} illumination, since it improves pupil detection~\cite{Stengeletal2015}, with gaze tracking techniques to relate image data and gaze direction. Additionally, other eye movements that can be detected are the fixations and saccades. These techniques require head and pupil position estimation in order to track gaze accurately. Most of these approaches require hardware configurations to obtain head pose invariance
with most of them to be feature or appearance based,~\cite{Hansenetal2005,Woodetal2014,Yangetal2014,Schneideretal2014,Suganoetal2014,Plopskietal2015,Laietal2015}. Since \hyperlink{a_VOG}{VOG} devices are less invasive than sensor based, these methods are more suitable to analyse gaze related impairments, such as attention deficits, in \hyperlink{a_AD}{AD} patients.

Pereira et al. \cite{Pereiraetal2014} analysed different methods that use eye movement to determine visual impairments in Alzheimer patients. These methods compared eye movements of healthy people and Alzheimer patients in terms of fixation duration, refixations or saccade orientation. As a result, most of the studies revealed an increment of saccades, defects in fixations and slow pursuit movements. Nevertheless, as suggested by Pereira et al. in \cite{Pereiraetal2014}, these studies fail to consider attentional impairments as a multi-domain concept and their results still need to be corroborated by future studies.

\begin{figure}
\centering
  \includegraphics[scale = 0.8]{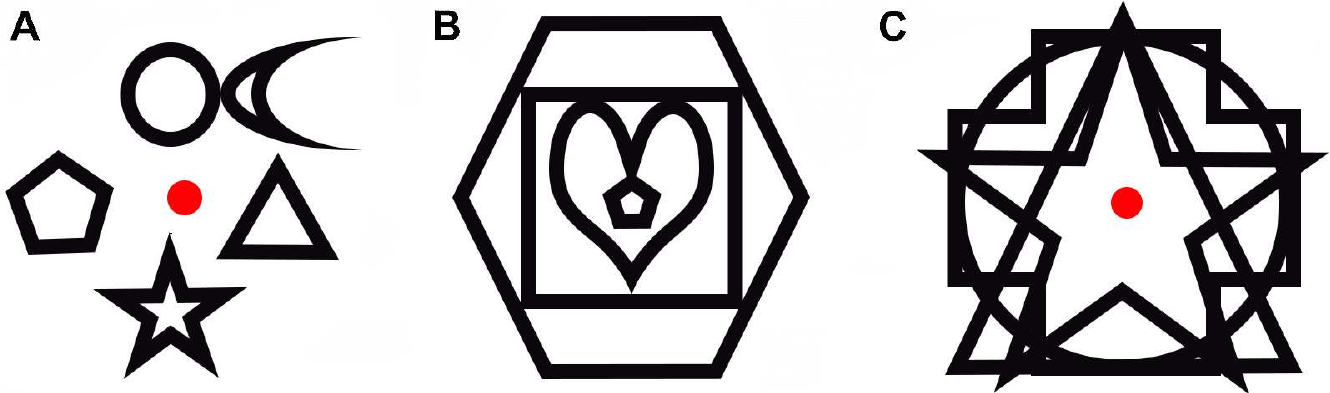}
  \caption{Example of the shapes presentation on each image during simultanagnosia assessment~\cite{Ruiz2017}: a) adjacent, b) embedded, and c) overlapping.}
  \label{fig:simultanagnosia}
\end{figure}

Ruiz el al~\cite{Ruiz2017} relate simultaneous object perception deficits to a visual processing speed impairment. In their research, they assess simultanagnosia, i.e., the inability to perceive more than one object at a time, and visual attention. In order to evaluate simultanagnosia, they use three tests, each one with their own specific tasks. The first test shows images of objects drawn with black lines on a white background. These objects are shown either individually or overlapping. During this test, both the identification speed and the number of errors are analysed. The second test includes dot counting, position discrimination and number location. Results depends on the number of correct answers, which are counted. As for the third test, it analyses the capacity to perceive different shapes drawn with black lines on a white background. The shapes are shown in four levels of difficulty: alone, adjacent, embedded and overlapping (see Figure~\ref{fig:simultanagnosia}). During this test the percentage of errors is analysed. Visual attention is assessed by asking the participant to recall as many letters as possible from an array of letters that is briefly presented. The outcome of these tasks allows to assess visual processing speed and visual short-term memory storage. The overall outcome of these tests proved that the existence of simultaneous object perception deficits in early \hyperlink{a_AD}{AD} patients is linked to impaired visual processing speed.

Fraser et al.~\cite{Fraser2017} classify \hyperlink{a_MCI}{MCI} from healthy participants by using gaze information extracted during two trials: reading of paragraphs silently and aloud. They use as features the information they obtain from saccades and fixations and they combine this data with information about the words which are used in the text, i.e. the duration of a fixation in a word whose number of repetitions in the text is low. Firstly they analyse the results of each trial independently, achieving better accuracy using a Naïve Bayes classifier with the gaze features of the first trial. Finally, they combine both trials achieving the best classification (86\%) using the Naïve Bayes classifier with the combination of words and gaze features. Despite their high accuracy, these results have not been validated yet in large groups. They also present the reading process differences between healthy and \hyperlink{a_MCI}{MCI} patients. Healthy participants read the text from start to finish whereas \hyperlink{a_MCI}{MCI} participants tend to skip words and come back to them later. In addition, this difference is more evident when reading silently.

\subsection{Virtual Environments in Alzheimer Disease screening}

Recently, due to the new advances in computer technologies, Virtual Environments have started to become part of medical tests and rehabilitation therapies. Many approaches have been introduced to validate the use of virtual reality for neurocognitive assessment. In \cite{Parsons2008} the authors evaluate learning and memory cognitive abilities on healthy controls while navigating through a virtual environment using a head-mounted \hyperlink{a_VR}{VR} device. The first step of this experiment consists of showing ten objects that the participant has to memorise. Once memorised, they navigate through five areas of a Virtual Reality city where they have to find the previously memorised objects. At the end of the experiment, the participants are asked to recall all the objects and the areas where they have found them. Results are then compared with traditional neurocognitive methods, proving the correlation between them and validating the use of Virtual Environments for memory and learning assessment. The work presented in~\cite{Parsons2009} assesses complex attention to healthy subjects (both military and civilian) and examines the ecological validity of virtual reality when using head-mounted \hyperlink{a_VR}{VR} devices in comparison to less immersive conditions. Participants are immersed in three low and three high intensity scenarios and record psychophysiological information such as startle eye blink amplitude and heart rate. In those scenarios they also evaluate participants' attention by showing a four-digit number in a central position during the first three scenarios and in a random position in the rest. Their results supported that a high level of immersion evokes a stronger physiological reaction than under low immersion conditions.

Cushman et al. in \cite{Cushman2008} also present a comparative study between a navigation system on \hyperlink{a_VEs}{VEs} and the real world while evaluating Alzheimer's participants' navigation aptitudes and proving the ecological validity of \hyperlink{a_VR}{VR} environments. The navigational assessment test is divided into eight subtests. Firstly, a route demonstration is given in both real and virtual worlds; then the subtests start. These subtests include route recall from ten landmarks, free recall of ten objects seen on the route, pointing to the location of ten objects from a certain point of the route, route-drawing on a map of the environment, route landmarks recall, photograph recognition, photograph location and video location and direction. The results of the tests in both real and virtual scenarios were similar, therefore validating the use of Virtual Scenarios for navigation deficit assessment and screening of early \hyperlink{a_AD}{AD}.

In \cite{Parsons2011}, Parson et al. analyse the appropriate way to create neurocognitive interfaces in \hyperlink{a_VEs}{VEs}. In this work they consider the possible loss of the experimental control related to \hyperlink{a_VEs}{VE} realism. The authors asseverate that an increase of \hyperlink{a_VEs}{VE} realism means a loss in control of the experiment since some of the experiment parameters could not be manipulated while keeping realism, therefore a balance between realism and control should be found in relation with the test type. Another interesting concept is psychophysiological computing; this term refers to a technique which consist of monitoring people with a computer in order to create user-tailored applications. Currently, most of the psychophysiological evaluation is obtained after the \hyperlink{a_VR}{VR} experience, but new advances on Brain Computer Interface (BCI) will allow an automated and real time evaluation.

The work presented by Tarnanas et al. in \cite{Tarnanas2013,Tarnanas2014} demonstrates that the use of \hyperlink{a_VEs}{VEs} is beneficial when it comes to early dementia detection. It is possible to improve the results from previous cognitive tests, since floor and ceiling effects are reduced, resulting in the creation of tests that can be adapted to the patients' IQ. Floor and ceiling effect occur when the test results are negatively or positively affected by personal characteristics independent of cognitive functioning, such as low or high educational level \cite{Franco2010}. Furthermore, it is possible to increase immersion in the task that is in progress. Tarnanas et al. use large screens to display the environment as well as depth sensors to recognise the gestures of the patient's body. This system requires the patient to move to the location where the equipment needed to carry out this test is available, due to the fact that its components are not portable or cost-effective, what makes it unsuitable for e-health applications.

\begin{figure}
\centering
  \includegraphics[scale = 2.3]{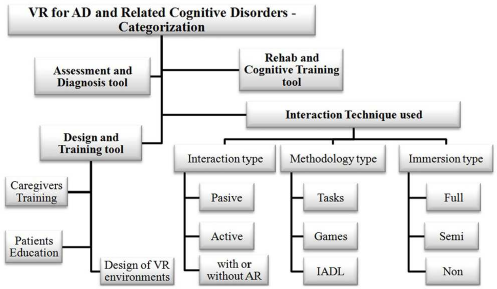}
  \caption{Garcia et al.'s~\cite{Garcia2015} categorization of applications for Alzheimer's Disease using Virtual Reality.}
  \label{fig:ADapplications}
\end{figure}

The study conducted by Garc\'ia-Betances et al. in \cite{Garcia2015} shows the advantages of virtual environments for Alzheimer's disease. In their work they analyse the wide range of applications for \hyperlink{a_AD}{AD} that can be created using \hyperlink{a_VR}{VR} (see Figure~\ref{fig:ADapplications}) and point out the lack of immersion or interaction in most of the current virtual reality applications. They also mention that most of the approaches should be more affordable and accessible for home and nursing environments.

Vallejo et al.~\cite{Vallejo2017} evaluate an \hyperlink{a_AD}{AD} assessment tool based on serious games. This application evaluates several cognitive functions by using Virtual Environments on a PC monitor. Therefore, this is not considered full immersion. The researchers evaluate the performance of both healthy and \hyperlink{a_AD}{AD} participants while these are doing everyday activities. The activities they carry out are all related to shopping and cooking. First, they have to follow a series of arrows in order to get to the market and then they need to remember the way back. They will have previously memorised three ingredients that they need to buy in the market. Finally, when they are back home they have to cook and set the table for two. The cognitive functions evaluated by this task include episodic memory, visuo-spatial orientation, recognition, spatial memory, executive function and attention. As a result of their study, Vallejo et al. found considerable differences between the healthy and the \hyperlink{a_AD}{AD} groups in the amount of time they needed to fulfil the tasks as well as in their rate of success. This survey also highlights the potential usefulness of those errors committed by \hyperlink{a_AD}{AD} patients during task-performance. Finally, the researchers conclude that serious games are a valid tool for cognitive impairment evaluation even though it can be biased, as these activities are sometimes gender-oriented or culture-oriented.

Serino et al.~\cite{Serino2017} propose a training program using Virtual Environments on a monitor for \hyperlink{a_AD}{AD} patients. Their program trains people to navigate in a virtual city providing an interactive aerial view. Participants are in the middle of the city and have to find objects and memorise their location (up to a maximum of three), one at a time (they cannot look for a new object until they have found the previous one). In the second stage, participants are in the city again, but in a different place, and they have to find the locations of the objects that have been previously found in stage one. To analyse cognitive capacity improvement, participants are tested using the \hyperlink{a_MMSE}{MMSE} and neuropsychological battery that include test such as verbal fluency or attentional matrices test. The researchers compared the evolution of patients that underwent a traditional recovery method and patients that had gone through their training program. Results revealed that the participants who had been trained on the \hyperlink{a_VR}{VR} program had improved more in terms of attentional abilities and spatial memory than those \hyperlink{a_AD}{AD} patients trained with traditional methods. Serino et al. consider that this is partly due to the use of the aerial views. Their results also suggest an improvement in executive functioning, however, due to the limited number of participants in the survey, further clinical trials are required to corroborate this.

\subsection{Sensors based behaviour analysis for Alzheimer's Disease screening}

The methods that assess cognition impairments in daily life activities usually require the use of external devices during a certain period of time to analyse patients' behaviour \cite{LopezdeIpinaetal2012, Aztiriaetal2013}. For example, Aztiria et al. \cite{Aztiriaetal2013} placed a sensor on the patient’s foot in order to analyse their gait (step length and step height), since it reflects patients’ level of dementia. However, the gait measurement method still has some drawbacks; for instance, the patient has to wear a device for long periods of time and the results obtained, despite being promising, are still not useful, since they have not been tested in Alzheimer patients. Abe et al. present in \cite{Abe2013} another detection method which uses sensors in the patients' homes to identify certain events. This non-cognitive approach is non-invasive and it requires only the consent of the patient to install the sensors. The main issue with these methods is that the obtained results are not precise enough (detection rates under 75\%) to provide an accurate technique for dementia diagnosis at early stages. Ishii et al.~\cite{Ishii2016} propose a sensor-based system to automatically determine suspicion of dementia. They use a series of sensors which are connected to a cloud where the data is analysed and compared with participants' personal behaviour information in order to provide a diagnosis. In order to analyse behaviour, they check memory loss in situations such as forgetting to turn a faucet off or to take a shower, sleep disorders and wandering with an average accuracy above 80\%. They evaluated this system in different scenarios but not in \hyperlink{a_AD}{AD} patients. Chong et al.~\cite{Chong2017} also propose an automated system for \hyperlink{a_AD}{AD} diagnosis based on sensors. They analyse the participants' behaviour by studying the daily routine activities they carry out at home. They conclude that the system for automated screening diagnosis is promising but the quality of the sensors should be improved and some personalised considerations should be taken since sometimes the patients' mobility is reduced by other reasons than \hyperlink{a_AD}{AD}. Varatharajan et al.~\cite{Varatharajan2017} introduce a method for \hyperlink{a_AD}{AD} detection using wearable devices. They analyse participants' gait by using some special motion detection devices which subjects wear on their feet. These devices collect a huge amount of data that describes the participants' gait such as number of steps, walking speed, stride length or cadence. They identified abrupt changes in foot movement patterns using the middle level cross identification function. Due to the difference in walking speed over time of the participants, the authors propose the use of dynamic time warping to align the signal in time and to classify \hyperlink{a_AD}{AD} and healthy participants' gait. They compare the results in terms of sensitivity and specificity with other classifiers such as Support Vector Machine (SVM), \hyperlink{a_kNN}{kNN} and inertial navigation algorithm (INA). Their outcome validates the use of dynamic time warping for gait classification to diagnose \hyperlink{a_AD}{AD}.


\section{Emotion Analysis for Cognitive Impairment Detection}
Emotion recognition is being widely studied due to its importance as feedback in marketing areas or Human Computer Interaction (HCI). For example, the analysis of emotions will contribute to improve the user's experience when navigating in virtual environments. In addition, participants' spontaneous reactions to controlled stimuli can help to evaluate their mental conditions. Different data modalities have been used for the analysis of human emotions such as facial images, audio or \hyperlink{a_EEG}{EEG}. Amongst these modalities, facial expressions have been the most widely studied. Nevertheless, according to the purpose of the research, a particular type of data may be more adequate than the others. In this section, \hyperlink{a_EEG}{EEG} and facial expression based approaches for emotion recognition will be analysed having into account their high performance for emotion classification as well as their adaptability to \hyperlink{a_VR}{VR} technologies.

\subsection{Brain Computer Interface Approaches}

Brain Computer Interfaces (BCI) is described as the direct communication channel between the brain and an external device~\cite{Anupama2012}. Electroencephalography (EEG) is the most common method to provide this communication channel. \hyperlink{a_EEG}{EEG} based emotion recognition is a less common approach since most researchers use facial or speech data as a source of information for emotion detection. Considering that, amongst other problems, these sources are easy to fake~\cite{Lokannavar15}, \hyperlink{a_EEG}{EEG} provides an extra source that solves problems such as falseness, illumination or speech impairment. On the other hand, \hyperlink{a_EEG}{EEG} signal deals with other challenges, for example, noise and biological and non-biological artifacts~\cite{Soleymani16,Muller-Putzetal2015}, such as electrooculogram (EOG), electromyogram (EMG) and electrocardiogram (ECG). Nevertheless, these biological artifacts are also affected by emotions and are expected to provide extra information to the \hyperlink{a_EEG}{EEG} signal for emotion recognition~\cite{Soleymani16}.

\begin{figure}
\centering
  \includegraphics[scale = 1]{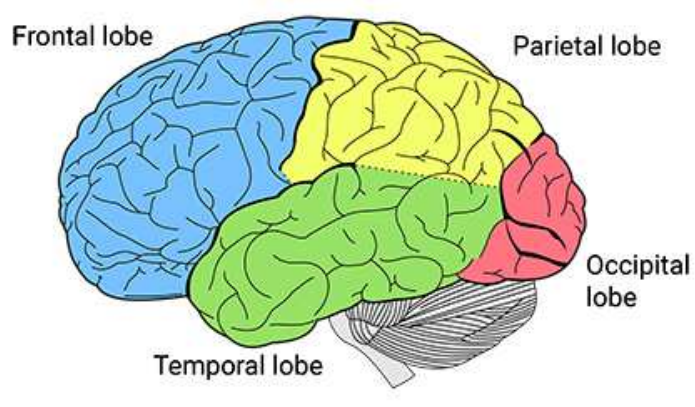}
  \caption{Brain lobes. Emotional activity is located in the frontal lobe and the frontal portion of the temporal lobe of the brain.}
  \label{fig:brain}
\end{figure}

\hyperlink{a_EEG}{EEG} is a medical technique that reads scalp electrical activity. It measures voltage fluctuations that are between 10 and 100 microvolts in a typical adult~\cite{Alarcao2017}. The recorded \hyperlink{a_EEG}{EEG} signal can be divided into frequency bands: delta (1-4Hz), theta (4-7Hz), alpha (8-13Hz), beta (13-30Hz) and gamma (30-100Hz). Frequencies between 4 and 40Hz are generally used to extract emotion-related features~\cite{Preethi2014}. \hyperlink{a_EEG}{EEG} sensors are usually positioned around the scalp in order to receive signals from the whole brain. When it comes to emotions, the brain activity correlated with emotions is focused on the frontal lobe and the frontal portion of the temporal lobe (see Figure~\ref{fig:brain}).

The analysis of the \hyperlink{a_EEG}{EEG} signal usually follows three paradigms: Sensory Evoked Potentials (SEP), Event-Related Potentials (ERP) and Event-Related Desynchronization/Synchronization (ERD/ERS)~\cite{Alarcao2017}. The first paradigm analyses the electrical potential generated after the occurrence of a stimulus: auditory (AEP), visual (VEP) or somatosensory (SsEP). \hyperlink{a_ERP}{ERP} is a measurement of the potential after a short stimulus. One of the most commonly used potentials is P300 for attention and working memory. This is a positive potential that appears 300ms after the stimulus is presented. These potentials can appear at different times and also have negative values such as P100 or N200. \hyperlink{a_ERD/ERS}{ERD/ERS} are used to measure reaction to affective communication. This paradigm analyses changes in the potential within specific frequency bands.

Before working with potentials and extracting relevant features for emotion classification purposes, the \hyperlink{a_EEG}{EEG} signal is preprocessed. In many research works the signal is filtered between 4 and 45 Hz, since emotion-related data is contained in those frequencies and the other frequencies will introduce noise. In addition, some approaches remove artifacts from eyes and other biological signals by using methods such as Independent Component Analysis (ICA) or Blind Source Separation (BSS). Once the \hyperlink{a_EEG}{EEG} signal has been preprocessed, different features are extracted, many of which are based on the aforementioned potentials~\cite{Alarcao2017}. Two types of descriptors can be used for \hyperlink{a_EEG}{EEG} signal analysis: simple descriptors (such as frequency or amplitude) and more complex ones such as asymmetry metrics, time/frequency analysis, topographic mapping, coherence analysis and covariation measures. These descriptors are used depending on the area of study; for example, asymmetry metrics are usually applied to cognitive neuroscience~\cite{Muller-Putzetal2015}. In particular, asymmetric hemispheric differences are used for emotion recognition~\cite{Sohaib13,Petrantonakis10,Alarcao2017}. Amongst all the methods which are used to extract features, some of the most popular ones are Short Time Fourier Transform (STFT), statistical based features such as the mean of the raw signal, and Wavelet Transform (WT)~\cite{Preethi2014,Alarcao2017}. The most relevant research on emotion recognition using \hyperlink{a_EEG}{EEG} signal will be analysed in the following paragraphs.

Lokannavar et al. proposed in~\cite{Lokannavar15} a method to classify four basic emotions: happiness, relax, sadness and fear. They achieved a classification accuracy of 89\% using features that had been previously extracted with the help of Fast Fourier Transform (FFT) and Auto Regression and \hyperlink{a_SVM}{SVM} as a classifier. Nevertheless, they do not provide any information about the data utilised during the experiments. Vijayan et al.~\cite{Vijayan2015} propose a similar approach using DEAP dataset \ref{db_DEAP} to classify four basic emotions: happiness, excitement, sadness and hatred. The method they suggest processes the 128Hz \hyperlink{a_EEG}{EEG} signal obtained from thirty-two electrodes. Firstly, they filter the signal from the 4-45Hz band to keep the information related with emotions and filter 50Hz frequency to remove noise. Afterwards, they use a five-level Wavelet Decomposition to separate the signal in the frequency bands and keep only the Gamma band (30-45Hz). From this band they measure Shannon Entropy and use it as a feature as well as auto-regressive modelling to obtain the final features that are passed to a \hyperlink{a_SVM}{SVM} classifier. As a result, they obtained satisfactory classification accuracy for four emotions.

Atkinson et al.~\cite{Atkinson2016} present an emotion classification approach in an attempt to improve the emotion recognition performance. The \hyperlink{a_EEG}{EEG} data is taken from the DEAP dataset \ref{db_DEAP} and it contains 512 samples per second. Therefore, they start the preprocessing stage by reducing the sample rate to 128Hz. They only work with fourteen relevant channels out of the thirty-two included in the database. The last preprocessing steps include removal of eye artifacts and filtering between 4 and 45Hz. They extract different features from each preprocessed channel such as statistical features, band power from band frequencies, Hjorth parameters and fractal dimension. Genetic Algorithm (GA) and minimum Redundancy Maximum Relevance (mRMR) are compared for feature selection and their selections are used on a \hyperlink{a_SVM}{SVM} classifier to classify six classes on the Valence-Arousal dimensions. As a result, they demonstrate that minimum Redundancy Maximum Relevance dimensionality reduction outperforms Genetic Algorithm for emotion classification purposes.

Zheng et al.~\cite{Zheng2017} analyse \hyperlink{a_EEG}{EEG} signal in order to find stable patterns that define positive, neutral and negative emotions. They created a database (SEED) that contains spontaneous reactions to film clips, which have been previously selected as a trigger for positive, negative and neutral emotions. Firstly, they start analysing the emotions classification performance of different features and classifiers. They conclude that Differential Entropy (DE) features and Graph regularised Extreme Learning Machine (GELM) provide the best results for classifying four classes of Valence-Arousal using DEAP \ref{db_DEAP} and SEED databases. Afterwards, they try to identify stable patterns that describe the three levels of emotion by using the best features and classifier. They demonstrated that positive emotions activate more lateral temporal areas of the brain, neutral emotions have higher alpha responses at parietal and occipital areas, and negative emotions have higher delta responses at parietal and occipital and higher gamma at prefrontal areas.

In general, \hyperlink{a_EEG}{EEG} approaches prove very convenient, since they enable to avoid some of the problems encountered when other data modalities are used, such as noise in audio and illumination in visual data. In addition, \hyperlink{a_EEG}{EEG} signal could also add extra information such as navigational that will boost \hyperlink{a_HCI}{HCI}. Nevertheless, despite most of the previous approaches present promising emotion classification results, these results are still inferior to those obtained when using visual modalities. This is probably due to the fact that visual modalities have been more widely researched than \hyperlink{a_EEG}{EEG}. Consequently, more research on \hyperlink{a_EEG}{EEG} processing should be conducted.


\subsection{Facial Expression Recognition Approaches}

\begin{figure}
\centering
  \includegraphics[scale = 0.9]{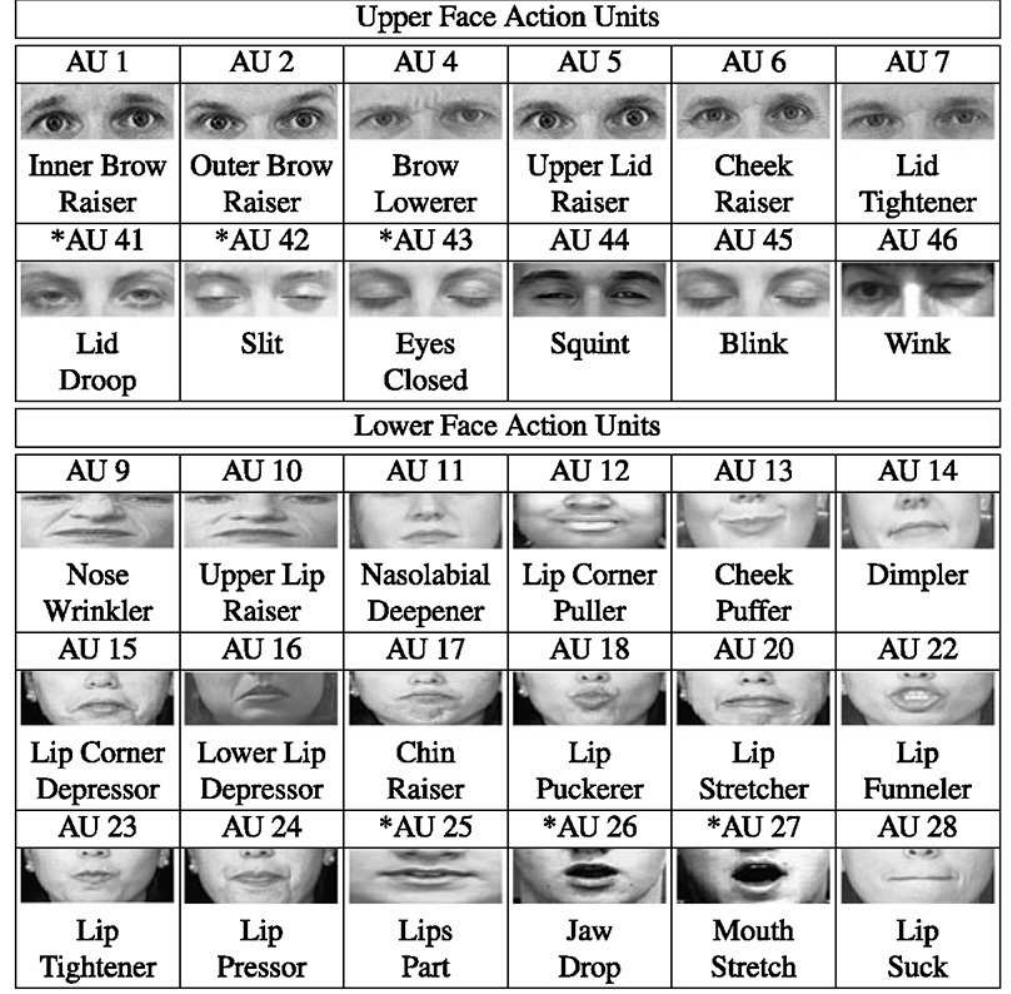}
  \caption{Example of Action Units.}
  \label{fig:AUs}
\end{figure}

Social cognition is one of the cognitive domains affected by Alzheimer. Associated symptoms include changes in behaviour or attitude, a decrease empathy or less ability to recognise facial expressions. Most research works focus on analysing if \hyperlink{a_AD}{AD} patients are able to recognise facial expressions, but none of them attempts to study the patients' facial expressions in order to detect sudden mood changes or how they react to different stimuli. Images and video sequences of faces are highly utilised as a source for emotion recognition. The study of facial expression was part of various disciplines since the Aristotelian era but it was in 1978 when the first automatic recognition study appeared~\cite{Sariyanidi15,Bettadapura12}. It is not clear yet which are the specific pointers required to interpret facial expressions. One of the most used systems for facial expression definition is the Facial Action Coding System (FACS)~\cite{Ekman78}. It defines basic human facial emotions such as happiness, sadness, surprise, fear, anger or disgust; where each of these emotions is described as a combination of Action Units (AUs)(see Figure~\ref{fig:AUs}). Each \hyperlink{a_AUs}{AU} corresponds to a facial configuration. On the other hand, some different approaches try to classify those basic emotions directly, emotion dimensions such as Valence-Arousal, or other type of reactions or emotions that are not amongst the common emotions. Some of the most popular emotion dimensions include arousal, valence, power and expectation. \label{emoDim} These dimensions can be explained as the interval between two states so the approaches provide a measure of the position in that interval. For example, arousal represents the interval between active and inactive or explained in basic emotions, sadness opposed to fear and anger; and valence is the interval between pleasant and unpleasant~\cite{Fontaine2013}.

Methods for facial emotion recognition can be classified according to the approaches applied to the recognition stages: registration, feature selection, dimensionality reduction and classification/recognition~\cite{Sariyanidi15,Bettadapura12}.

Three different approaches can be used for face registration: whole face, parts or points registration. These registration approaches are usually based on Active Appearance Models (AAM)~\cite{Cootes2001,Baltru16}; a method that matches a statistical model of the face to images in order to extract face landmarks and specific face areas. Whole face approaches get the features from the whole face. Littlewort et al.~\cite{Littlewort2009} get image-based features of the whole face, such as Gabor Wavelets, in order to detect \hyperlink{a_AUs}{AUs} for pain recognition. They recorded participants in two different situations: faking pain and experiencing real pain. Firstly, they apply their method to detect pain and afterwards they try to distinguish fake from real pain. Their method starts detecting faces from videos using Viola-Jones \cite{Viola2001,Soni2017} and Procrustes alignment \cite{Gower1975}. A bank of Gabor Wavelet filters eight orientations and nine spatial frequencies and the output magnitudes are used as features. Linear \hyperlink{a_SVM}{SVM} classifiers (See Section \ref{SVM}) were trained for twenty \hyperlink{a_AUs}{AUs}, eighteen for individual \hyperlink{a_AUs}{AUs} and two for combinations of them. The results showed that the detection performance is higher for posed pain. In order to distinguish posed from non-posed pain, a second classifier was trained using the twenty \hyperlink{a_AUs}{AUs} output from a full minute video. They propose two methods for this classification. The first one extracts statistical features such as median or maximum from five hundred frames windows and uses a non-linear \hyperlink{a_SVM}{SVM} for classification. The second method filters at eight frequencies and generates a histogram of the intensities that have been used as features to train a Gaussian \hyperlink{a_SVM}{SVM}. The latter method provided the best discrimination accuracy.

Long et al.~\cite{Long2016} propose an unsupervised basic emotion classification method for video data. Using data from CK database \ref{db_CK} they process six video frames from each subject and emotion. They start detecting and aligning the faces on each frame. The resulting data is divided into 3D spatiotemporal normalized cubes. The sparse coding algorithm is then applied to extract spatiotemporal features. Afterwards, spatiotemporal pyramid max pooling is applied to model and reduce the spatiotemporal features. This method, often compared with others such as Gabor Wavelets or \hyperlink{a_ICA}{ICA}, proves to be the most accurate of them all, as it reaches 81\% accuracy.

\begin{figure}
\centering
  \includegraphics[scale = 1]{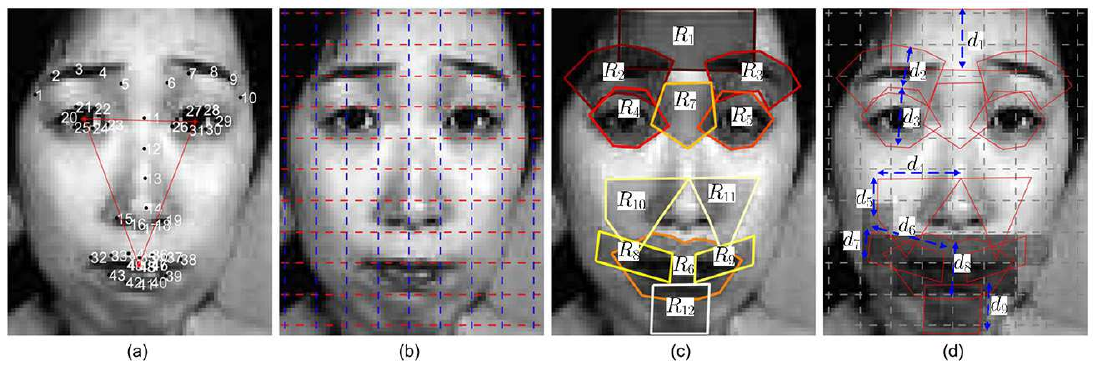}
  \caption{Eighty patches and twelve most important regions \cite{Xie2017}.}
  \label{fig:XIE}
\end{figure}

Face parts approaches focus on those face areas that provide the highest amount of information about face expressions, such as the eyebrows and the mouth. Xie et al.~\cite{Xie2017} propose a face parts approach for classifying seven basic emotions. They start by aligning the images and normalizing their illumination. The images are reduced to $84x68$ and divided into eighty patches, of which the most important twelve regions are selected (see Figure \ref{fig:XIE}). This regions are selected for being the most correlated with expression changes \cite{Ekman78}. The researchers connect twenty \hyperlink{a_AUs}{AUs} to the correspondent patches, ending up with seven areas (eyes, brow, mouth, forehead, nose root, nasolabial and chin). In addition, each of these areas is assigned a label of the basic emotions that can be associated to it. Afterwards, Gabor filters are applied in order to extract surface features from each area. They use a two-stage method. In the first stage, areas features are used in order to classify, with the aid of \hyperlink{a_SVM}{SVM}, the seven basic emotions. Moreover, those three emotions which are more likely to be represented by each image are selected (triplets). During the second stage the \hyperlink{a_AUs}{AUs} and the active areas are weighted for each emotion triplet using the conditional probability matrix. Afterwards, they are optimised with the aid of multi-task sparse learning. Finally, triplets, optimised \hyperlink{a_AUs}{AUs} and active areas are used as input to a weighted \hyperlink{a_SVM}{SVM} classifier that will choose the final expression label. CK+ \ref{db_CK}, JAFFE \ref{db_JAFFE} and SFEW2 \cite{Dhall2011} databases were used for experimental testing, whereas TFEID \cite{LiFen2007}, YALE \cite{Georghiades1997} and EURECOM \cite{Min2014} were reserved for generalization testing. Results show a general improvement of the classification performance when optimization algorithms are applied and detect that weight optimization is counter-productive when the number of training samples is small.

Points based approaches use fiducial points for shape representation. Michel et al.~\cite{Michel03} use a tracker to get twenty-two fiducial points and calculate the distance of each point between a neutral and a peak frame. These distances are used as features of a \hyperlink{a_SVM}{SVM} algorithm in order to classify emotions. Neutral and peak frames are automatically detected when the motion of the points is almost zero. Valstar et al. use Particle Filtering Likelihoods~\cite{Valstar05} in order to extract twenty fiducial points, but they still have to select the initial position of these points manually. These points are normalized by respecting a neutral point (tip of the nose) and a scale transformation is also applied. The distances between certain points are used as features to recognise specific \hyperlink{a_AUs}{AUs} using \hyperlink{a_SVM}{SVM}. Pan et al.~\cite{Pan2016} present a method for classifying six basic emotions thanks to facial landmarks. They extract key facial points using Constrained Local Model (CLM). Finally, they apply Generalized Procrustes Analysis and Coherent Point Drift (CPD) for normalization. They compare this normalization with mean-based and median-based normalizations on CK+ \ref{db_CK}, BU-3DFE \cite{Yin2006} and BU-4DFE \cite{Yin2008} databases. Their results show that this proposal is only better when databases do not include all six basic emotions. Nguyen et al.~\cite{Nguyen2017} propose an approach to classify three types of emotions: negative, blank and positive. They start detecting the faces and extracting sixty-eight facial landmarks. These landmarks are then normalized on rotation and scale. Seventy-nine geometrical features are extracted using the landmarks. Amongst those features we find eccentricity, differential, linear or trigonometric features such as distances and angles between landmarks. Finally, \hyperlink{a_SVM}{SVM} with Radial Basis Function (RBF) kernel is used to classify positive, negative and neutral emotions.

Nicolle et al.~\cite{Nicolle2012} propose a method for emotion recognition (valence, arousal, expectancy and power) using a combination of whole face, face parts, points and audio features. They start using Viola-jones to crop the face area on each image. The points-related features are obtained by using the Point Distribution Model to extract a temporal signal that contains information about head position and facial deformations. The global features are obtained warping the faces into a mean model and using \hyperlink{a_PCA}{PCA} to project them into the principal modes of appearance. To obtain the local/parts features, the face images are divided into patches from interesting areas such as mouth, eyes or eyebrows. \hyperlink{a_PCA}{PCA} is also used to compute temporal signals of the evolution of the local appearance. Afterwards, for each of them, the log-magnitude Fourier spectra are applied in order to obtain dynamic information, that jointly with the mean, standard deviation, global energy and the first and second order spectral moments are used as features. Energy, spectrum and voice-related features are extracted from audio signal. All the features are normalized by subject. They calculate the correlation between these features and their correspondent label and use \hyperlink{a_kNN}{kNN} clustering to select the most representative samples of each label. Finally, they use kernel regression and linear regression for prediction. Results reveal their performance is better than Support Vector Regression on SEMAINE database \ref{db_SEM} for the four emotion classification problem.

Datta el al.~\cite{Datta2017} introduce a new method for classifying six basic emotions. This method combines whole face and point based features. Using CK+ database \ref{db_CK}, Viola-Jones and Active Shapes Model (ASM) are used to extract seventeen facial points. The angles between points are calculated in neutral and peak expression frames and the difference between the peak and neutral angles is used as a geometrical feature vector. The whole face features are extracted from the peak frames. The images are aligned and cropped to resolution of $120x120$ and are then divided into nine blocks. Local Binary Pattern (LBP) is then applied to each block and its output is concatenated to form the Texture-Based feature vector. Each feature vector is tested independently and concatenated using \hyperlink{a_SVM}{SVM} and \hyperlink{a_kNN}{kNN} classifiers. \hyperlink{a_SVM}{SVM} and \hyperlink{a_kNN}{kNN} facial emotion recognition accuracy shows different results. While \hyperlink{a_SVM}{SVM} shows that texture features are the most appropriate when used individually, \hyperlink{a_kNN}{kNN} classifiers obtain better results with the geometric ones. In addition, \hyperlink{a_kNN}{SVM} classifiers prove higher accuracy when used with combined features, whereas \hyperlink{a_kNN}{kNN} classifier's performance is still better when used with geometric features. Datta et al. attribute these results to a lower \hyperlink{a_kNN}{kNN}'s performance when the features dimension is higher.

Those methods can also be organised according to feature representation, where methods can be divided into spatial and spatio-temporal approaches. Spatial approaches include shape representations, histogram of gradients (HOG), low-level histograms or Gabor representations, amongst others. Sariyanidi et al. presented in~\cite{Sariyanidi13} a low-level histogram representation using local Zernike moments for emotion recognition based on \hyperlink{a_kNN}{kNN} and \hyperlink{a_SVM}{SVM} classifiers. Candra et al.~\cite{Candra2016} use a variation of \hyperlink{a_HOG}{HOG} features on a \hyperlink{a_SVM}{SVM} classifier for the classification of seven basic emotions. On the other hand, spatio-temporal approaches obtain features from a range of frames within a temporal window, detecting more efficiently those emotions that cannot be easily differentiated in spatial approaches. Zhao et al.~\cite{Zhao2009} proposed a method that uses spatio-temporal local binary patterns as features and \hyperlink{a_SVM}{SVM} for classifying facial expressions. The method explained in the previous paragraph~\cite{Nicolle2012} is a spatio-temporal approach that uses dynamic cues for classifying emotions dimensions such as expectancy and power. Yan H. propose in~\cite{Yan2017} a multi-metric learning approach to improve the classification of multi-features, in this case, visual and audio features.

When it comes to dimensionality reduction techniques and classification methods, those most commonly used are Principal Component Analyses (PCA)~\cite{Khan2016}. Other popular dimensionality reduction techniques on facial emotion recognition approaches are Local Discriminant Analysis (LDA), Local Fisher Discriminant Analysis (LFDA) and t-distributed Stochastic Neighbor Embedding (t-SNE). Yi et al.~\cite{Yi2013} use \hyperlink{a_tSNE}{t-SNE} and Adaboost to classify seven basic emotions on JAFFE database \ref{db_JAFFE}. They crop the face areas, align the images and resize all of them to $64x64$. Afterwards, they transform each image into 4096 row features. \hyperlink{a_tSNE}{T-SNE} is applied to reduce the dimensionality of these features. Their best facial emotion recognition rate is obtained when the number of dimensions is seventy-five and the number of AdaBoost iterations is a thousand. They also prove that \hyperlink{a_tSNE}{t-SNE} dimensionality reduction provides better results than no reduction or other methods such as \hyperlink{a_PCA}{PCA} or \hyperlink{a_LDA}{LDA}. In addition, some approaches i.e. those presented in~\cite{Xie2016} and~\cite{Sun2017}, propose their dimensionality reduction techniques to improve facial expression recognition methods.

Xie et al.~\cite{Xie2016} propose a dimensionality reduction method for multi-view data. Their technique, called Multi-view Exclusive Unsupervised Dimensionality Reduction, combines multi-view features and reduces their dimensionality. Furthermore, weights are associated to each view in order each view takes part differently in the selected features. Local Binary Pattern on Three Orthogonal Planes (LBP-TOP), \hyperlink{a_HOG}{HOG}, Histogram of Optical Flow (HOF) and Motion Binary Histograms (MBH) are used as input features to a \hyperlink{a_kNN}{kNN} classifier. In order to prove their technique they carry out a test with two facial expression datasets and compare its performance with other common methods that take into account different views such as distributed \hyperlink{a_PCA}{PCA} or distributed spectral embedding. Their results show that a higher degree of classification accuracy is obtained when using methods that combine the features information of the different views. Besides, this system outperforms the rest in their best dimensions.

Sun et al.~\cite{Sun2017} state that dimensionality reduction techniques are aimed at minimising within-class distances and that there is a problem associated to most of these techniques: the influence on the optimization function when the within-class distances are very large. They propose the Enhanced Relevance Feedback (ERF) technique, which is based, as its name suggests, on Relevance Feedback, an algorithm that maximises across-classes distances, minimising the within-class distances and preserving multimedia data distribution. They use a penalty item to this algorithm (the sum of within-class distances) and extract the density information to weight the penalty influence. In addition, they have proposed a classification method based on the Cognitive Gravity Model to leverage the use of the density information, the Enhanced Cognitive Gravity Model (ECGM). In order to demonstrate the validity of their proposal, they use pyramid Histogram of Oriented Gradients (pHOG) as features and compare their facial expression recognition performance on CK+ \ref{db_CK} and JAFFE \ref{db_JAFFE} databases with other dimensionality reduction techniques and other classification methods. Their method obtains better accuracy results for facial emotion classification than the others.

Finally, if organised by classification technique, it can be stated that most of the facial expression recognition methods utilise \hyperlink{a_SVM}{SVM} with different kernels. The most commonly used kernels are Gaussian (\hyperlink{a_RBF}{RBF}) and linear~\cite{Michel03,Valstar05,Zhao2009,Sariyanidi13,Candra2016,Nguyen2017,Xie2017}. Wei et al.~\cite{Wei2016} propose a weighted Gaussian \hyperlink{a_SVM}{SVM} for emotion recognition and compare it with standard Gaussian \hyperlink{a_SVM}{SVM}. Using CK+ dataset \ref{db_CK} they extract fifty-nine facial points from eyebrows, eyes, nose and mouth. Depending on the rigidity of the area, a certain weight is associated to each of them, having into account that the non-rigid region has the biggest impact. Finally, weighted features are used in weighted \hyperlink{a_SVM}{SVM} and \hyperlink{a_SVM}{SVM} classifiers to classify seven basic emotions. The recognition rate of their proposed version of \hyperlink{a_SVM}{SVM} outperformed the standard \hyperlink{a_SVM}{SVM}. The second most used classification algorithm is \hyperlink{a_kNN}{kNN}~\cite{Nicolle2012,Sariyanidi13,Xie2017}. Other approaches use boosting algorithms~\cite{Yi2013} or propose new ones based on those already existing. For instance, Song et al. in~\cite{Song2015} introduce a version of the Bayesian Compressed Sensing (BCS) classifier and Sun et al. in~\cite{Sun2017} present a classification method based on Relevance Feedback algorithm.

All facial emotion recognition approaches face the same challenges such as head-pose and illumination variations, registration errors, occlusions and identity bias. Some of these problems are not included in most of the available databases therefore some of them may not work properly on real conditions. The datasets available for emotion recognition focus on different applications. According to the purpose of the final application and the way it will be executed some of the previous challenges can be relaxed. For example, medical examination applications could be performed in a lab or doctor office where the illumination of the room and the position of the patient is settled beforehand.

Several promising automatic facial expression recognition (FER) approaches exist for emotion classification. Most of them reach a \hyperlink{a_FER}{FER} rate over 90\%. Nevertheless, the results obtained by many of the aforementioned approaches are validated in posed databases or spontaneous databases recorded in controlled environments. Therefore, their emotion classification results cannot be generalised to other datasets and certainly will not work properly when tested in the wild. In order to improve emotion recognition performance, new approaches combine different data modalities (such as \hyperlink{a_EEG}{EEG}, images or audio)~\cite{Nicolle2012,Soleymani16}, propose new features that describe each class more distinctly~\cite{Long2016}, improve classification algorithms~\cite{Wei2016} or, apply deep learning techniques.


\subsubsection{Deep Learning Approaches for Emotion Classification}
The latest methods for facial emotion recognition use deep learning architectures for feature extraction and emotion classification~\cite{Corneanu2016}. Deep learning techniques provide multi-layered hierarchical representations of the facial expressions from large unlabelled datasets~\cite{Sariyanidi15}. Some of the most frequently used deep learning architectures for facial emotion classification are Convolutional Neural Networks (CNN) and Deep Belief Networks (DBN).

Uddin et al.~\cite{Uddin2017} present novel features from RGB and depth videos for the classification of six basic emotions using \hyperlink{a_DBN}{DBN}. They created a depth database containing 40 ten-frame videos. The proposed feature is based on Local Directional Pattern (LDP). It is called Local Directional Position Pattern (LDPP) and it adds to the local directional information of each pixel the top directional strength position with their strength sign. This enables the differentiation of edge pixels with bright and dark areas on their opposite sides. Afterwards, \hyperlink{a_PCA}{PCA} is applied for dimensionality reduction and Generalized Discriminant Analysis (GDA) for feature robustness. The resulting features are used to train a \hyperlink{a_DBN}{DBN} which is made out of three hidden layers of eighty, sixty and twenty neurons. With the aid of both RGB and depth data, they compare their method to other feature extraction techniques such as \hyperlink{a_PCA}{PCA}-\hyperlink{a_LDA}{LDA} and to other classification methods such as Hidden Markov Models (HMM). Their results demonstrate that Local Directional Position Pattern-\hyperlink{a_PCA}{PCA}-General Discriminant Analysis features are more robust and that \hyperlink{a_DBN}{DBN} provided better \hyperlink{a_FER}{FER} performance. In addition, their proposal focused on depth data, illumination variation tolerance and the protection of the subjects' identity. Results reveal that depth \hyperlink{a_FER}{FER}'s rates are better than RGB's.

\begin{figure}
\centering
  \includegraphics[scale = 0.4]{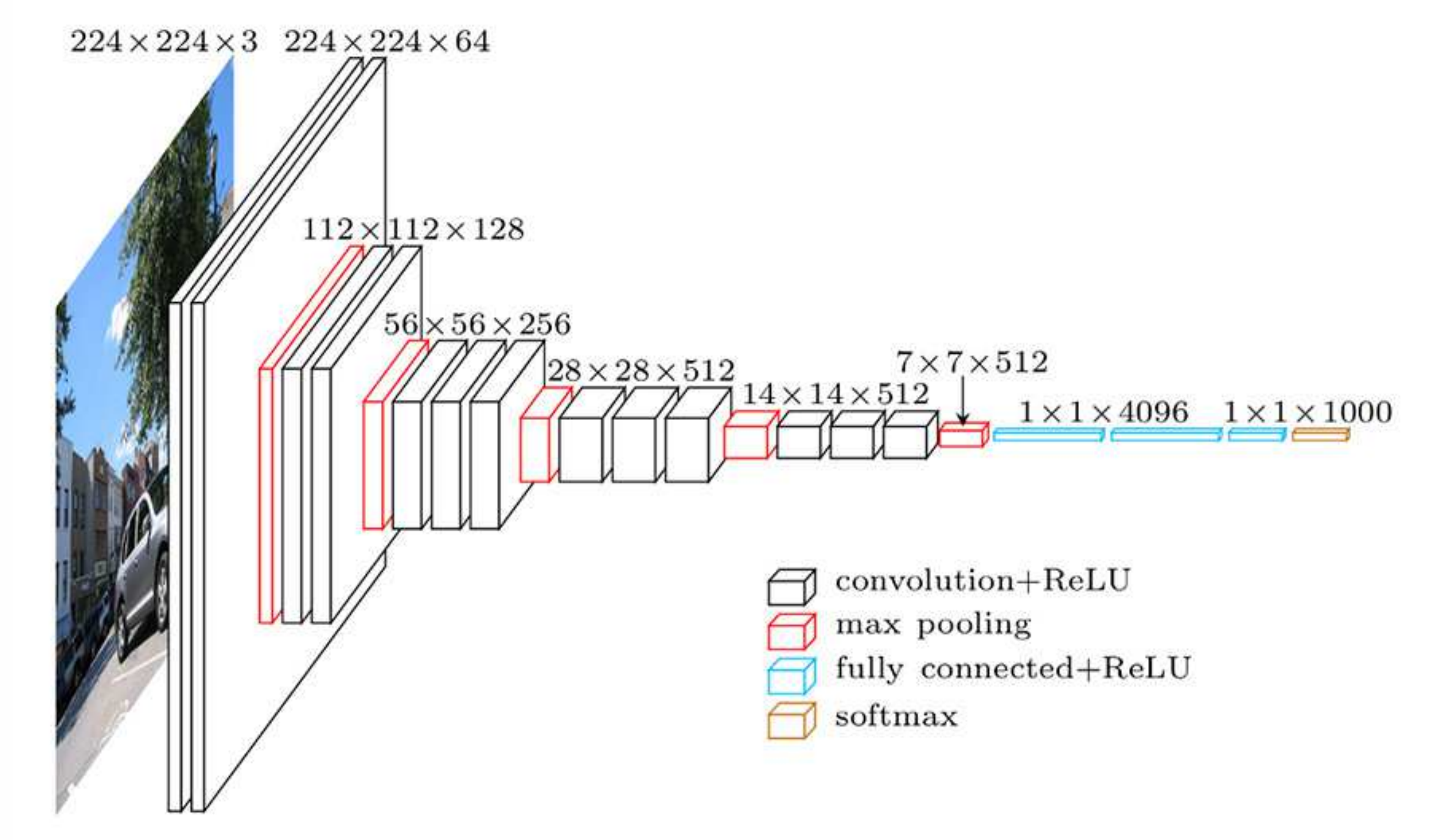}
  \caption{VGG-16 network architecture.}
  \label{fig:vgg}
\end{figure}

Due to the lack of trained neural networks for facial expression recognition and since deep learning approaches require very large datasets that are currently non-existent for facial expression recognition, it is common to fine-tune a purpose-related network that is already trained. Facial recognition (FR) networks are the chosen ones for \hyperlink{a_FER}{FER}. Nevertheless, Ding et al.~\cite{Ding2017} outline a problem when this strategy is used. There may be information related to subject identity after fine-tuning the \hyperlink{a_FR}{FR} network. Therefore, they propose a new distribution function to model the high level neurons of the expression recognition network that will increase the capture of facial expression semantics. Their training method consists of two stages. In the first stage, it is necessary to train a network formed of convolutional, non-linear activation function (ReLU) and pooling layers (the expression network). VGG-16 (see Figure~\ref{fig:vgg}) is used as a \hyperlink{a_FR}{FR} network with FaceNet \cite{Schroff2015} weights. The \hyperlink{a_FR}{FR} net is frozen, with the only exception of layer fc8, and it can be found that its Pooling layer 5 is the best one to provide supervision and help to learn better the high-level expression semantics better from the \hyperlink{a_FR}{FR} net. During the second training stage, a fully connected layer is added to the expression network and it is trained using the weights obtained from stage one. This method has been tested in CK+ \ref{db_CK}, Oulu-CAS VIS \ref{db_CASIA}, TFD \cite{Susskind2010} and SFEW \cite{Dhall2011} databases. Viola-Jones and IntraFace~\cite{Torre2015} are used for face and landmark detection. Faces are then cropped, normalized and resized, and data augmentation is utilised. Finally, the researchers compare their method to some of the state-of-the-art ones for the classification of basic emotions, with positive results.

Deep learning methods for facial emotion recognition can also be separated in spatial and spatio-temporal approaches. For example, Gudi et al.~\cite{Gudi2015} propose a spatial approach for the detection of \hyperlink{a_AUs}{AUs} occurrences and intensity using a seven-layer \hyperlink{a_CNN}{CNN} on frontal cropped faces, as well as BP4D \ref{db_BP} and SEMAINE \ref{db_SEM} databases. It should be noted that the former contains \hyperlink{a_AUs}{AU} and intensity annotation, whereas the latter only includes \hyperlink{a_AUs}{AU}. First, preprocessing is applied to the databases' images. Faces are detected and cropped, then aligned. Afterwards, the grayscale contrast is normalized and data augmentation is utilised. The image size used as input of the \hyperlink{a_CNN}{CNN} is $48x48$ grayscale pixels. The researchers suggested two \hyperlink{a_CNN}{CNNs} composed of Input-Convolutional-Pooling-Convolutional-Convolutional-Fully connected-Output layers. The output layer used with BP4D returns eleven neurons and it is trained with \hyperlink{a_AUs}{AUs} intensity information whereas the SEMAINE one returns six neurons. Unfortunately, results were not as satisfactory as expected, probably because of the small size of the training data, according to the researchers themselves.

Greche et al.~\cite{Greche2017} introduce a facial expression classification approach using distances between facial landmarks as features and a neural network as a classifier. They create their own database recording a hundred and twenty-one 3D facial points of ten subjects displaying six emotions. A comparison between Euclidean distance features and Manhattan distance ones is carried out with the aid of a neural network with one hidden layer. Additionally, they analyse the best number of neurons per layer required to obtain the best performance. As a result, they have obtained perfect results using twenty neurons and Euclidean distances whereas fifty-eight neurons were required for Manhattan distance.

Chu et al.~\cite{Chu2017} propose a spatio-temporal approach using a hybrid network for \hyperlink{a_AUs}{AUs} classification. This hybrid network combines a \hyperlink{a_CNN}{CNN} for spatial feature extraction and a Long Short Term Memory Network (LSTM) for temporal features. The authors state that handcrafted features such as \hyperlink{a_HOG}{HOG} contain bias information, which results in them not being specific enough to describe an \hyperlink{a_AUs}{AU}. Therefore, they propose the use of a \hyperlink{a_CNN}{CNN} based on AlexNet \cite{Krizhevsky2012} to extract features from the images included in the databases. In addition, they suggest that temporal information is also necessary due to the ambiguous and dynamic nature of \hyperlink{a_AUs}{AUs}. In order to capture this information, they propose the use of LSTM to encode crucial information during the transition of two frames. Finally, two fully connected layers are added on top of both \hyperlink{a_CNN}{CNN} and \hyperlink{a_LSTM}{LSTM} in order to unify spatial and temporal information. BP4D \ref{db_BP} and GFT \cite{Girard2017} databases were used in these experiments. During the preprocessing stage, facial detection and data augmentation where applied. Their research outcome shows that their proposed features are less biased than others such as Gabor or \hyperlink{a_HOG}{HOG}. They also compare the F1 scores \ref{eq=f1} obtained to classify twelve \hyperlink{a_AUs}{AUs} using their approach and other deep learning and not deep learning methods. The F1 scores with ten fold cross-validation showed the superiority of deep learning approaches and the suitability of this method, as it outperforms the other deep learning approach, which is based on AlexNet, using only spatial features.

Kim et al.~\cite{Kim2017} introduce a similar approach to Chu et al.'s, claiming that it will solve interpersonal variation problems such as expression intensity variations. From MMI \ref{db_MMI} and CASMEII \ref{db_CASMEII} datasets faces are cropped and aligned using facial landmarks. For each dataset, they manually labelled the onset, onset-apex, apex, apex-offset and offset frames for each expression since they suggest that extracting features from the different phases of expression will reduce the effect of intensity variations. Once all the images have been labelled, data augmentation is applied. Using five images per emotion, one per emotion state label, the \hyperlink{a_CNN}{CNN} is trained. The resulting model is subsequently trained using all the frames to extract the spatial features. Those are used as input to \hyperlink{a_LSTM}{LSTM} to learn the temporal features. The authors also compare their method to deep learning and not deep learning approaches, as well as to their own spatial approach. Results revealed that their method outperforms the other methods for the classification of basic emotions.

Deep learning approaches have been proved reliable and very accurate for many applications. Indeed, in terms of facial expression recognition, these approaches have improved \hyperlink{a_FER}{FER}'s performance using databases. Currently, approaches fine-tune similar networks such as face recognition ones in order to reuse some of the trained facial characteristics and to add specific features from facial expressions, obtaining \hyperlink{a_FER}{FER} rates around 95\%. Nevertheless, most deep learning approaches still need to be validated with data in the wild. In order to achieve better emotion classification results with spontaneous emotions recorded in non-controlled environments, the amount of data should be increased since deep learning performance improves when a higher amount of data is used for training.

\section{Overview of Related Machine Learning Concepts}

Machine learning is a part of artificial intelligence whose aim is to develop techniques that enable computers to learn. As described by Rogers et al.~\cite{Rogers2016}, it aims to find relations between input variables and associated responses that allow the prediction of responses to new input variables. Data from patients' tests and tasks or medical data and their corresponding classes, such as healthy or non-healthy subject, could be utilised by machine learning techniques to find relations between data and labels. Therefore, these relations could be used to classify new data into their corresponding categories. Machine learning techniques can be divided into two groups: supervised and unsupervised. The aim of the latter is to learn about input data, its structure and distribution. Therefore, only input data is necessary to extract information such as related groups of data or the rules that define large portions of data. As for supervised techniques, they focus on those tasks in which is necessary to find the relation between input and output data. Therefore, examples of input and output data are required to extract those relations. The relations that link the input and the responses make up the machine learning model. The performance of the models, which is verified by checking if correct responses are provided when new input variables come into play, is evaluated in order to determine the validity of the machine learning technique. Supervised learning techniques can also be divided into two groups depending on the problem to be solved: classification and regression. The difference between them is that classification returns labels whereas regression returns continuous values. For emotion recognition and cognitive impairment screening purposes, many approaches use classification techniques which can also be split into three groups depending on the number of labels to be classified: one-class classification, binary classification and multiclass classification. Their applications will be analysed in the following subsections.

\subsection{Binary/MultiClass Classification} \label{SVM}
The most frequently used classification types are binary and multiclass. The simplest classification problems are the binary, i.e. classifying data into two categories. One of the most popular binary classifiers is Support Vector Machine (SVM). \hyperlink{a_SVM}{SVM} was proposed for the first time in 1995 by Cortes et al.~\cite{Cortes1995}. It is a binary classifier that uses data which has been previously labelled into two categories to help create a model that enables us to include and classify new data into these categories. The standard \hyperlink{a_SVM}{SVM} is the linear kernel \hyperlink{a_SVM}{SVM}, which from the training data generates an optimal hyperplane that separates the classes. Different changes can be made to this kernel to make non-linear data classification possible. The most widespread one is the Radial Basis Function (RBF or Gaussian). This kernel creates non-linear combinations of the features to move them onto a higher dimensional space where they are linearly separable.

Multiclass problems require separating the given data into more than two categories. This problem can be solved either using multiple binary classifiers or a multiclass one. Two strategies can be applied when using binary classifiers: one versus one (OVO) or one versus all (OVA)~\cite{Krawczyk2017}. One versus one uses one classifier per each possible pair of classes, whereas one versus all uses one classifier per class and it compares each class with an aggregation of the remainder. \hyperlink{a_SVM}{SVM} has also been modified in order to create a multiclass \hyperlink{a_SVM}{SVM} by combining the multiple binary optimization problems into one single optimization function~\cite{He2012}. Nevertheless, one versus one and one versus all strategies are the most commonly used due to the large computational complexity of the multiclass \hyperlink{a_SVM}{SVM}. 


\subsection{One-Class Classification}

One-Class Classification (OCC) aim is to identify one type of objects and distinguish it from others when the data is very imbalanced (with training data mainly from one class). The data from this class is used as positive samples. There might also be data from other classes but, due to factors such as cost and ethics, it is difficult to obtain. This data will be treated as outliers. \hyperlink{a_OCC}{OCC} methods can be organised into three families~\cite{Krawczyk2017}. The first family is composed by density-estimation methods. These need a high amount of training data and they are not robust to outliers inside the training data. The second group deals with clustering-based methods. These consider the structure of the data and they are robust to outliers, but they require training data to represent the whole class. Finally, the third family includes methods that create a boundary around the desired class and the other ones. Their performance and robustness to outliers depend on the setting of this boundary. Rodionova et al.~\cite{Rodionova2016} mention another \hyperlink{a_OCC}{OCC} methods classification depending on the use of outliers: compliant and rigorous. The latter gather all the \hyperlink{a_OCC}{OCC} classifiers that only use positive data for training whereas the former add information given from some outliers to the model. When the data from the classes is overlapped it is generally more convenient to use compliant approaches.

One of the most appropriate applications of \hyperlink{a_OCC}{OCC} is the automatic diagnosis of diseases~\cite{Khan2014}, since sometimes it is difficult to obtain information from certain diseases due to the limited number of cases, to ethics or cost. Lopez et al.~\cite{Lopez2016} compare \hyperlink{a_OCC}{OCC} and multiclass classifiers for the classification of audio signal for early detection of Alzheimer. Speech-related Alzheimer's symptoms include aphasia (speaking and understanding deficits) and Emotional Response problems. The authors have created a database called AZTIAHO that contains video recordings from fifty healthy and twenty \hyperlink{a_AD}{AD} participants. They use a Multi-Layer Perceptron (MLP) for the \hyperlink{a_OCC}{OCC} and the multiclass classifier (MCC). \hyperlink{a_OCC}{OCC} creates a model that represents the healthy group, whereas \hyperlink{a_MCC}{MCC} models two classes, healthy and \hyperlink{a_AD}{AD}. They outcome of their experiments demonstrates \hyperlink{a_OCC}{OCC} outperforms the multiclass when \hyperlink{a_AD}{AD} data is scarce.

Das et al.~\cite{Das2016} propose the use of \hyperlink{a_OCC}{OCC} for the detection of errors in daily activities with the main purpose of helping people with dementia. They suggest the use of \hyperlink{a_OCC}{OCC} classifiers with data extracted from sensors located in the patient's home in order to provide help to both caregivers and patients when undertaking daily activities. Since it is impractical to define all the possible errors that could occur during those activities, the researchers use the data from participants who completed the activities without making errors to train an \hyperlink{a_OCC}{OCC} and the data with error is used for testing. Different features such as number of sensors, event pause or event time probability are used to train the \hyperlink{a_OCC}{OCC}. One Class Support Vector Machine (OCSVM) is the classifier used. \hyperlink{a_OCSVM}{OCSVM} is a boundary method that finds a hyperplane that separates the positive class from the others. In addition, after the outliers from the healthy model are found, other two classifiers are proposed for error classification, one \hyperlink{a_OCC}{OCC} and one \hyperlink{a_MCC}{MCC}. The evaluation of the \hyperlink{a_OCSVM}{OCSVM} shows a feeble performance since many false positives are detected by the classifier. The authors attribute this to inaccurate error annotation. They also compare the use of \hyperlink{a_OCSVM}{OCSVM}-\hyperlink{a_OCSVM}{OCSVM} and \hyperlink{a_OCSVM}{OCSVM}-\hyperlink{a_MCC}{MCC} to detect errors and classify them; their classification results are similar.


\section{Databases}

In~\cite{Bell2015}, Bell et al. identified 117 datasets related to dementia studies. Most of them contain clinical results from \hyperlink{a_AD}{AD} patients such as blood test results, Magnetic Resonance Image scans or cognitive test results (the \hyperlink{a_MMSE}{MMSE}, primarily). For example, the Longitudinal Aging Study Amsterdam (LASA)~\cite{Hoogendijk2016} is a dataset, updated periodically, that contains information such as emotional and cognitive interview-based surveys. It is based on a study of the ageing process experienced by the Dutch population and it contains information about cognitive impairments. AlzGene database~\cite{Bertram2007} contains genotype data pinpointing potential \hyperlink{a_AD}{AD} susceptible genes. ADNI database \cite{Mueller2005} is a longitudinal study that includes clinical data such as cognitive test results and biomarkers, Magnetic Resonance images, positron emission tomography (PET) data, genetic data and biospecimen data such as blood, urine or cerebrospinal fluid. BRAINnet~\cite{Koslow2013} is one of the largest Alzheimer's databases containing general and cognitive information from interviews and cognitive testing applications, \hyperlink{a_EEG}{EEG}, \hyperlink{a_ERP}{ERPs}, autonomic arousal measures, \hyperlink{a_MRI}{MRI} and genomics.

The availability of \hyperlink{a_AD}{AD} databases containing recordings and labels of patients' emotions is limited. Emotion-related proposals study the capacity of \hyperlink{a_AD}{AD} patients to recognise other people's emotions, but not their own. Therefore, in order to analyse patients' emotions, general emotion recognition techniques and databases are considered.

\subsection{Emotion Analysis Databases}
Several databases have been created for emotion analysis purposes. Due to the important role played by the face in emotion recognition, many of these databases contain images of video recordings of facial images. Moreover, other popular databases focus on \hyperlink{a_EEG}{EEG}, audio or contain multimodal information. Emotion analysis databases can be grouped into two categories: posed or non-posed databases. Posed databases usually allow an accurate classification of emotions but they do not represent emotions in real circumstances; therefore, their popularity has been decreasing in the last few years. Another essential difference between databases is recording circumstances. Databases are usually recorded in controlled environments, i.e., controlled background, illumination and participants' movement (head-pose variations). Consequently, results are not reliable in naturalistic conditions. On the other hand, some applications do require a controlled environment; therefore, some datasets might be more suitable for certain applications without requiring naturalistic conditions.

\label{db_JAFFE} The following paragraphs will analyse those emotion databases of more relevance for this research. The Japanese Female Facial Expression database (JAFFE)~\cite{Lyons1998} is a posed dataset that contains facial images of ten Japanese female participants representing seven basic emotions. It includes 213 static greyscale images of women representing neutral, sadness, surprise, happiness, fear, anger and disgust emotions.

\begin{figure}
\centering
  \includegraphics[scale = 0.7]{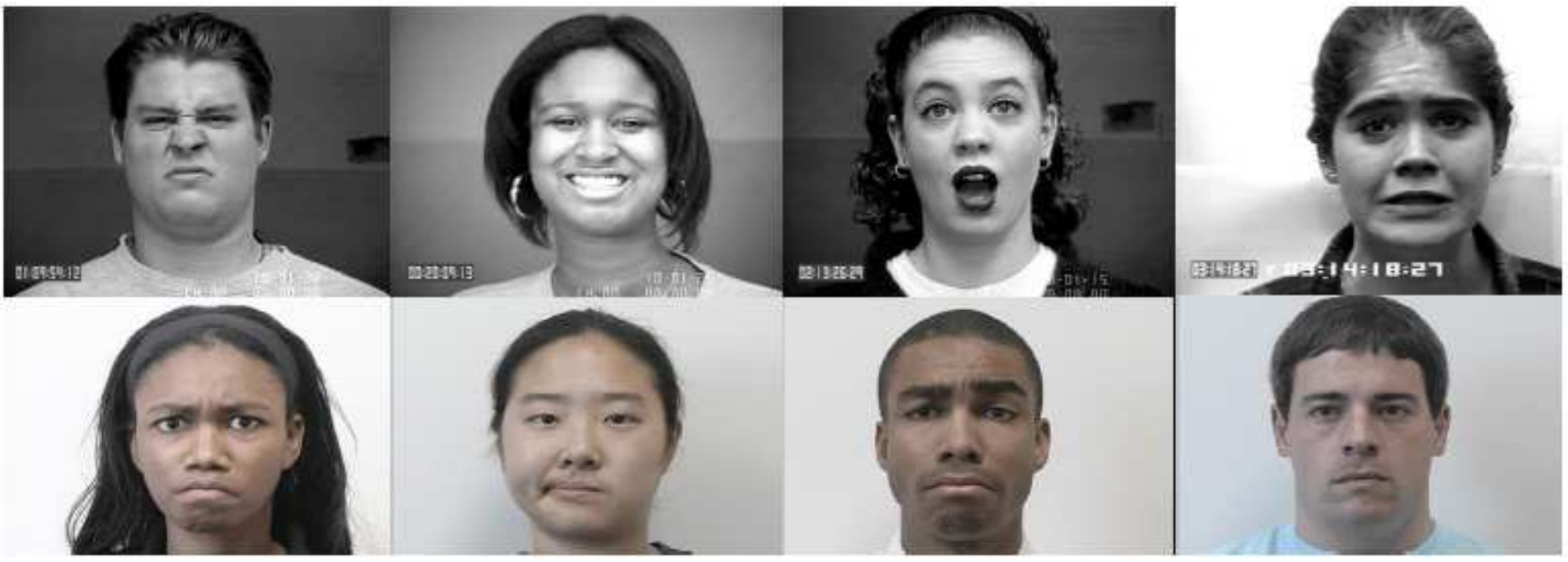}
  \caption{Sample images from CK+ database. The top ones have been taken from CK database~\cite{Lucey2010}.}
  \label{fig:ck+}
\end{figure}

\label{db_CK} Cohn-Kanade (CK) database~\cite{Kanade00} is widely used for emotion classification. CK contains sequences of images of posed expressions from ninety-seven people performing seven basic emotions (neutral, sadness, surprise, happiness, fear, anger and disgust). It provides emotion and action unit labels of the peak expression. As this is a posed database, the authors have additionally created CK+ database~\cite{Lucey2010}, which not only includes posed facial images, but also non-posed. This database contains 593 sequences with their corresponding \hyperlink{a_AUs}{AU} labels, 327 of which also contain labels of the seven basic emotions (see Figure~\ref{fig:ck+}).

\label{db_MMI} MMI database~\cite{Pantic2005} is an ongoing facial expression database. Occasionally, new labelled visual data is added to the database. When it was first presented, this database already included more than 1,500 samples of static images and image sequences, a figure which has grown to currently reach 2,900. Emotion image sequences show facial expression from neutral to peak, and then to neutral again. Therefore, this database includes the onset, apex and offset of emotions. It contains labels of \hyperlink{a_AU}{AUs} and seven basic emotions.

\label{db_FERA} GEMEP-FERA is the database used for the first Facial Emotion Recognition and Analysis (FERA) challenge~\cite{Valstar2011}. It uses part of the Geneva Multimodal Emotion Portrayal (GEMEP) dataset~\cite{Banziger2010}. GEMEP contains over 7,000 audiovisual videos of eighteen posed emotions. Part of this dataset was labelled for AU and emotion detection for \hyperlink{a_FERA}{FERA} challenge: Five emotions and twelve \hyperlink{a_AU}{AUs}.

\label{db_DEAP} The database for emotion analysis using physiological signals (DEAP) provides \hyperlink{a_EEG}{EEG}, face recordings and physiological data of participants while they watch musical videos just for the analysis of human emotional states~\cite{Koelstra12}. Videos showing the faces of thirty-two participants and \hyperlink{a_EEG}{EEG} signal from thrity-two sensors with 512 framerate were recorded for this database. Physiological signals such as blood volume pressure, Galvanic Skin response, breathing pattern, skin temperature, \hyperlink{a_EOG}{EOG} and \hyperlink{a_EMG}{EMG} were recorded from twenty-two, out of the thirty-two participants. The labels of this database classify the frames by levels of valence, arousal, liking and dominance.

\label{db_SEM} SEMAINE database aims to provide voice and facial information to study the behaviour of subjects interacting with human and virtual avatars~\cite{McKeown12}. One hundred fifty people were recorded while taking part in controlled conversations that are thought to evoke happiness, anger, gloom and sensibility. A colour and a RGB camera were used to record frontal face videos and an additional greyscale was used to capture profile views. The audio was recorded using two microphones, one at the front and the other one on the participants' head. Annotators were asked to label the clips considering five dimensions: valence, activation, power, expectation and intensity. In the second place, they labelled seven basic emotions (fear, anger, happiness, sadness, disgust, contempt and amusement) and other information such as data validity. \hyperlink{a_AU}{AUs} from 181 frames were also labelled.

\label{db_MAH} MAHNOB-HCI database was created for the study of those emotions experienced by humans when they are watching multimedia. This database provides a wide range of data such as audio, a RGB video and five monochrome videos of the face, \hyperlink{a_EEG}{EEG}, \hyperlink{a_ECG}{ECG}, respiration amplitude, skin temperature and eye-gaze data~\cite{Soleymani12}. Thirty participants were recorded while watching videos and images and they were also asked to label each fragment on valence and arousal dimensions intensity.

\label{db_DISFA} Denver Intensity of Spontaneous Facial Actions (DISFA) records facial images of twenty-seven participants while they are watching four-minute videos specifically designed to elicit emotions~\cite{Mavadati2013}. It also contains manual annotation of twelve \hyperlink{a_AU}{AUs} intensity, from 0 to 5, on each video frame, where 0 represents 'not present' and 5 means 'maximum intensity'. It also contains the facial landmarks of each frame, which were extracted using active appearance models. The authors have also released an extended version, DISFA+~\cite{Mavadati2016}, in order to provide data to compare posed and spontaneous emotions. They selected nine participants from DISFA and recorded them while imitating thirty facial actions and twelve facial expressions including the twelve \hyperlink{a_AU}{AUs} covered by DISFA.

\begin{figure}[t!]
\centering
  \includegraphics[scale = 0.45]{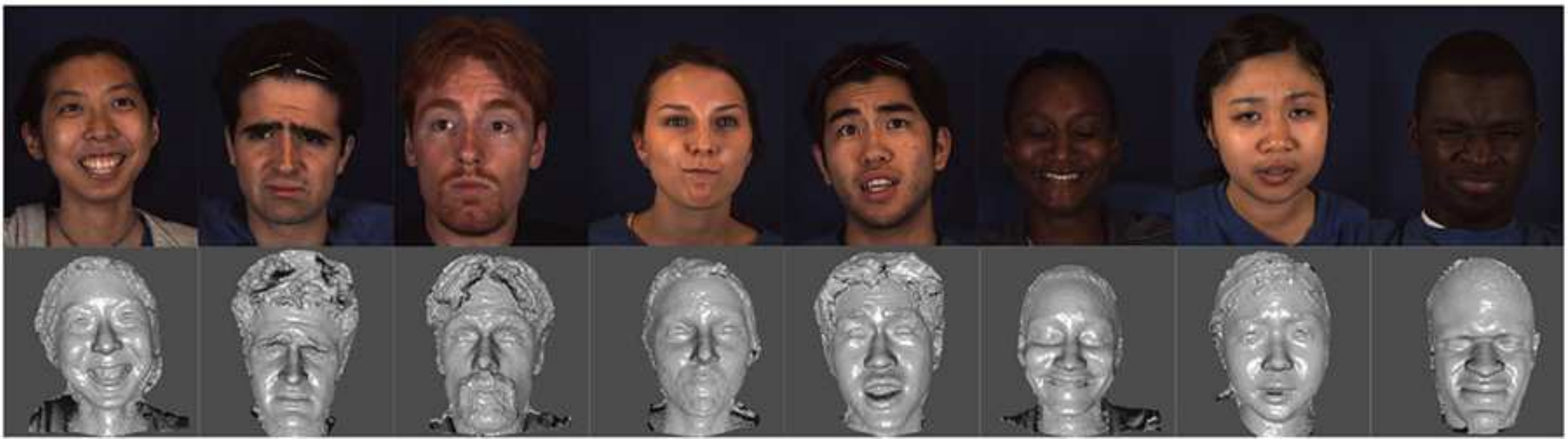}
  \caption{Eight sample emotional expressions from the different activities included in BP4D~\cite{Zhang2014}.}
  \label{fig:BD4P}
\end{figure}

\label{db_CASMEII} CASMEII is a dataset that studies facial micro-expressions for security and medical applications, what requires the use of cameras with a higher framerate and spatial resolution~\cite{Yan14}. Thirty-five participants were recorded in an environment where illumination conditions were strictly controlled while watching high emotional valence videos. In order to elicit micro-expressions, eighteen participants were asked to keep neutral faces whereas the other seventeen were asked to try to refrain facial movements as soon as they realise they were going to produce a facial expression. Annotators excluded irrelevant facial movements and selected sequences of frames with micro-expressions whose length is less than one second. Once the clips were selected they used \hyperlink{a_AU}{AUs} to label each frame. Finally, the database contains 247 micro-expression samples from twenty-six participants with five emotions labels (happiness, disgust, surprise, repression or others), the \hyperlink{a_AU}{AUs} involved and the onset and offset frames.

\label{db_CASIA} Oulu-CASIA NIR \& VIS~\cite{Zhao2011} contains videos recorded under three different illumination conditions: normal, weak and dark illumination. Eighty participants posed six basic emotions (happiness, sadness, surprise, anger, fear and disgust). Near Infrared and Visible light systems were used for recording.

\label{db_BP} Binghamton-Pittsburgh 4D Spontaneous Facial Expression database (BP4D)~\cite{Zhang2014} captured 3D dynamic facial information from forty-one participants while they were undertaking different activities, including an interview, video-clip watching, a surprising noise, an improvisation game, a physical threat, a cold pressor, insults and unpleasant smells. The recording devices were two stereo cameras for 3D information, one texture camera and a standard RGB camera to capture video and audio. The aim of this survey was to trigger a range of emotions such as happiness, amusement, sadness, surprise or startle, embarrassment, fear or nervousness, physical pain, anger or upset, and disgust. The data included in the database is the 3D models of the face and the 2D textures of each frame (see Figure~\ref{fig:BD4P}). Each frame has \hyperlink{a_AU}{AU} labels annotated manually and 2D/3D facial landmarks and head pose information are also included.


\section{Conclusion}

Early \hyperlink{a_AD}{AD} screening and diagnosis methods have been analysed in this Chapter. They have been categorised in non-cognitive, cognitive, virtual reality based and behaviour analysis through sensors approaches. Non-cognitive methods such as \hyperlink{a_MRI}{MRI} have proved accurate and reliable for early detection but expensive and usually uncomfortable. Cognitive approaches on the other hand are usually affordable and easy to implement but they can only be used for screening since their \hyperlink{a_AD}{AD} detection accuracy still can be improved. They have also issues such as the physical limitations to test diverse \hyperlink{a_AD}{AD} symptoms and they are biased by the different IQ of the participants. The computerized \hyperlink{a_VR}{VR} tests offer solutions to the IQ bias and the physical limitations, providing a wide range of tasks that can be used to evaluate different cognitive domains. State-of-the-art \hyperlink{a_VR}{VR} methods are able to improve the cognitive tests \hyperlink{a_AD}{AD} detection accuracy and demonstrate the validity of virtual environments for early \hyperlink{a_AD}{AD} screening. However, they are usually focused on the evaluation of specific cognitive domains without taking advantage of the \hyperlink{a_VR}{VR} potential. When it comes to sensor based approaches, they were included due to their capability to assess patients' impairments without undergoing a test, since they analyse behaviour in a daily life context. Nevertheless, current methods are still not able to reach the accuracy of cognitive test.

Our research propose the study of early AD screening through the evaluation of emotions, since \hyperlink{a_AD}{AD} patients' social cognition is impaired but they are still able to express emotions. There is not research that analyses \hyperlink{a_AD}{AD} patients' emotions triggered by specific stimulus related to \hyperlink{a_AD}{AD} symptoms, that we are aware of, therefore, general methods for emotion recognition are analysed. Approaches that uses \hyperlink{a_EEG}{EEG} and facial features for emotion recognition were studied. The first ones were selected due to the possibility to be integrated in \hyperlink{a_VR}{VR} glasses and they \hyperlink{a_EEG}{EEG} are not as affected by external factors such as facial and voice based approaches. Their detection accuracies are promising but their are still inferior to other data modalities. On the other hand, facial features were selected since state-of-the-art methods show they provide the best emotion detection accuracies, therefore they are the most usually utilised for emotion recognition. These methods proved very accurate for controlled and posed databases but they can be improved for the detection of emotion in the wild.


\chapter{Cognitive Evaluation for the Diagnosis of Alzheimer's Disease based on Turing Test and Virtual Environments} 

\label{Chapter3} 

\lhead{Chapter 3. \emph{Cognitive Evaluation for the Diagnosis of Alzheimer's Disease based on Turing Test and Virtual Environments}} 


\section{Introduction}

Alzheimer's diagnosis requires careful evaluation. Currently, the process carried out by doctors includes medical history analysis, mental status and mood evaluation and blood tests or brain imaging~\cite{AlzAsso2017}. Any of those procedures are just screening tools that help to obtain an indicator of possible mental deterioration and to recommend further assessment if necessary. Many people think that having memory problems is a clear indicator of Alzheimer's but this memory deterioration could also be caused by other health problems such as depression. Therefore, a deep analysis of the problem needs to be assessed by professionals taking into account various Alzheimer's symptoms.

Alzheimer's patients' cognition is clearly damaged since the beginning of the disease. Cognition can be defined as the use of the information that has been previously collected by a person to make behavioural decisions~\cite{Rowe2014} but it is such a complex term that it is difficult to provide a definition that covers all the affected domains \ref{CogDef}. Therefore, since many cognitive symptoms are easily detectable, several screening tests were created to challenge patients' cognition in early stages of dementia. Cognitive tests are the most utilised mental status screening tests for early Alzheimer's diagnosis. Since most of these tests can be performed with only pen and paper and they can be completed in less than fifteen minutes, they can be applied immediately without any cost. Other reasons why they are so widely used by doctors are their high accuracy and the fact that they are not invasive. Other non-cognitive methods can be slightly more accurate for Alzheimer's detection but they usually require invasive or uncomfortable procedures.

\begin{figure}[!t]
 \centering
 \includegraphics[scale=0.6]{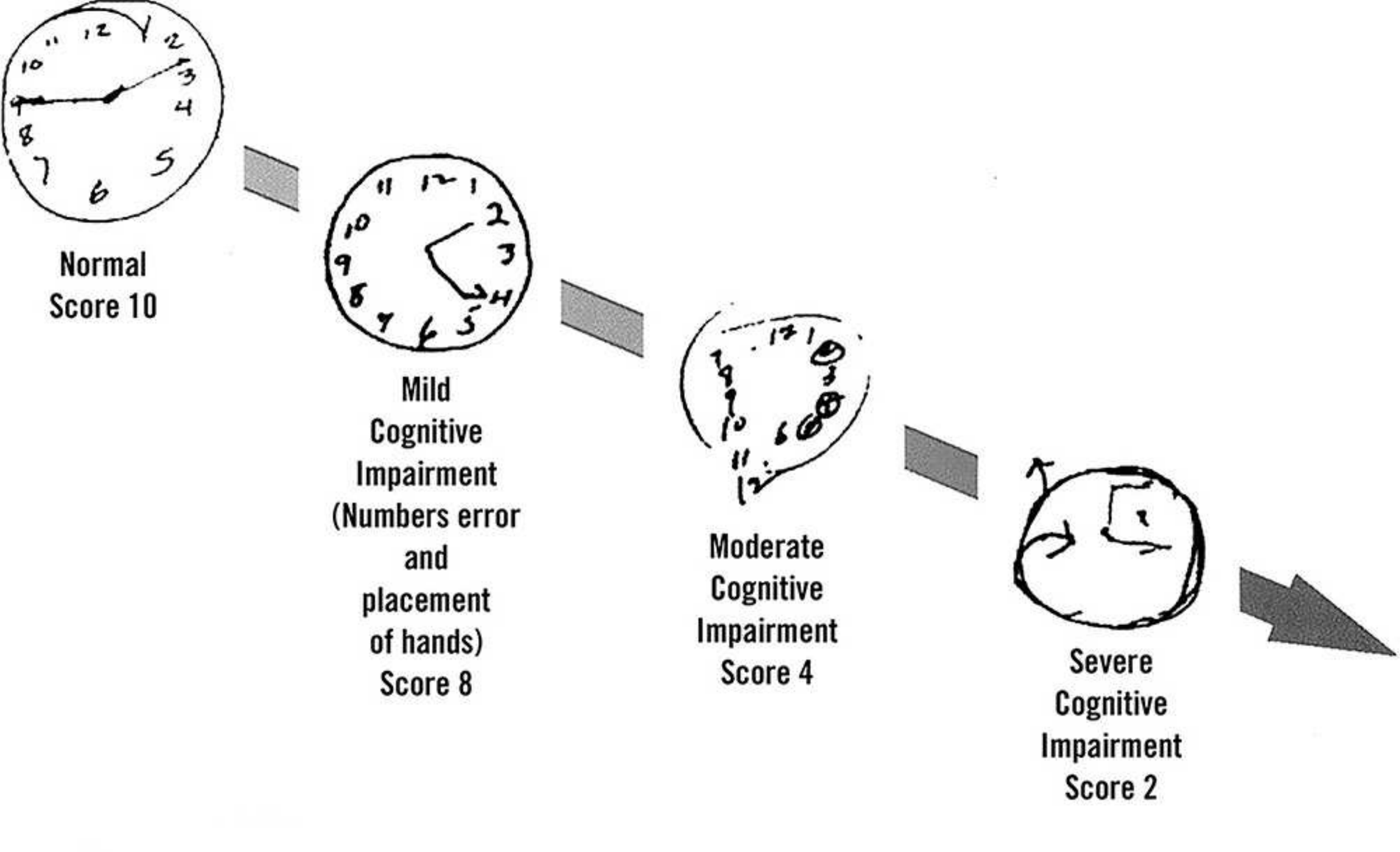}
 \caption{Clock Drawing Test for Alzheimer's detection. The patient is asked to draw a clock at a specific time. The subject is evaluated accordingly to the number and type of errors~\cite{Sunderland1989}.}
 \label{fig:Figure0}
 \end{figure}

Most of these pen and paper tests, such as the Mini Mental State Examination (MMSE), the clock-drawing test (see Figure~\ref{fig:Figure0}) or the Saint Louis Mental Status Examination (SLUMS), evaluate diverse mental capacities such as patients' temporal orientation, concentration or memory. These tests are score based examinations where the diagnosis is given by scoring thresholds. The problem with these tests is their lack of adaptability to the patient. It has been demonstrated that distinct tasks are necessary according to the patients' IQ. Aforementioned tests usually contain easy tasks that are too simple for people with high IQ. Cognitive reserve, or the capacity of the brain to compensate brain damage, explains why people with higher IQ perform better in those cognitive tests. Thus, taking cognitive reserve into account could improve early Alzheimer's detection~\cite{Tucker2011}.

\begin{figure}[!t]
 \centering
 \includegraphics[scale=0.6]{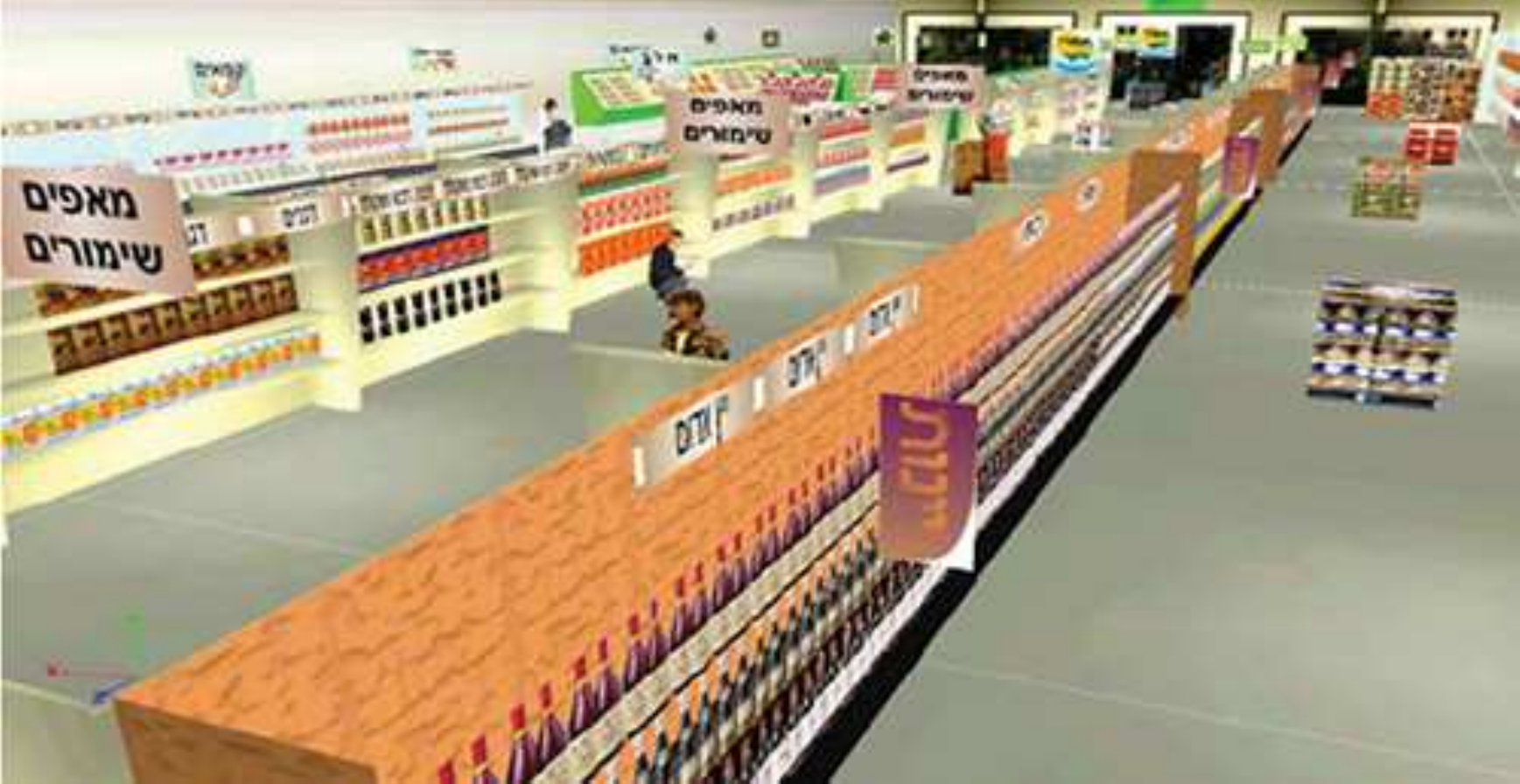}
 \caption{The VAP Supermarket~\cite{Werner2009} was one of the first tools for mild cognitive assessment. The participants have to buy some items on a virtual supermarket displayed on a PC monitor. The whole purchase task is analysed in order to detect executive functioning deficits.}
 \label{fig:Figure01}
 \end{figure}

Computerization of cognitive tests has allowed to take into account cognitive reserve. Computer tools can analyse the data and react accordingly, providing the adaptability missed in pen and paper tests. In addition, Virtual Environments have thrust the creation of cognitive tasks. \hyperlink{a_VEs}{VEs} provide a safe environment where any kind of test can be generated. Therefore, a wider range of cognitive tasks can be created out of real world conditions. First \hyperlink{a_VEs}{VEs} tests were utilised in average size monitors and the graphics related technology did not allow the creation of realistic environments (see Figure~\ref{fig:Figure01}). Most recently, Virtual Reality technology provided immersive \hyperlink{a_VR}{VR} glasses and powerful graphic resources being possible to create ultra-realistic virtual scenarios. This glasses technology appeared in the market at a reasonable price so any medical centre could afford one. Currently, this \hyperlink{a_VEs}{VEs} technology is accessible from many mobile phones and it will be part of most of them in the near future. As a result, \hyperlink{a_VR}{VR} Alzheimer's screening tools could be widely accessible allowing the patient to be tested from everywhere, at low cost and directly connected with a doctor.

As such, within this chapter the following contributions are presented: the design of four cognitive screening tests for early Alzheimer's Disease detection taking advantage of \hyperlink{a_VEs}{VEs} adaptability and versatility. The tests variate according to the educational level of the participant in order to reduce the cognitive reserve effect, creating more difficult tasks when necessary. Some of the created tests would not be easily reproducible in natural conditions. \hyperlink{a_VEs}{VEs} allowed the use of certain objects that would be unlikely to obtain in real world and the creation of unnatural behaviours. These tests have been developed for \hyperlink{a_VR}{VR} glasses and mobile phone \hyperlink{a_VR}{VR} in order to make them widely accessible. The screening tests were evaluated on Alzheimer's and healthy groups and their results were evaluated to validate the diagnosis results. Each participant evaluated the tool providing information about its convenience and comfort. Additionally, three state-of-the-art tests have also been computerised in order to be able to provide a comparative study.


\section{Related Work}

Virtual environments have been used for different mental health issues such as phobias, autism or stress; focussing mainly on assessment and treatment~\cite{Haniff2014,Freeman2017}. These methods have proved that virtual environments are suitable for mental health issues since they provide a sense of realism in safe conditions~\cite{Haniff2014}. Due to the improvement on Virtual Reality technology over the last years, the amount of research done on virtual environments has been increased and the immersion experience has been improved significantly~\cite{ParsonsCarlew2017}.

Despite the increasing number of research for mental health, the amount of dementia research on Virtual Environments is still reduced. This research includes some approaches focused on improving Alzheimer's patients' daily life. For example, Ferreira et al.~\cite{Ferreira2017} have proved on Virtual Environments that auditory assistance can enhance the patients' task performance. Other approaches try to improve the Alzheimer's disease detection rate. Werner et al. present in~\cite{Werner2009} a virtual supermarket on a monitor where the participants have to purchase seven items. During the test all the actions required to complete the task are analysed. As a result, Werner et al. proved that their method allows to discriminate between Mild Cognitive Impairment participants and a control group. Sauzéon et al.~\cite{Sauzeon2016} projected a \hyperlink{a_VEs}{VE} of an apartment on a wide screen where the participants have to remember a number of objects and their location. The results of this experiment identified age-related differences and the effects of Alzheimer’s disease on cognitive functioning.

In the work presented by Tarnanas et al. in~\cite{Tarnanas2013,Tarnanas2014}, they used large screens to display the environment and some depth sensors to recognise the patients’ gestures. The participants were immersed in a virtual apartment where they confronted a fire evacuation drill that required to solve multiple tasks, some of them simultaneously. Their work demonstrated that the use of \hyperlink{a_VEs}{VE} is beneficial when it comes to early dementia detection and that it is possible to increase the immersion of the patient in the task that is in progress.

Previous approaches showed different tasks and situations that can be implemented in Virtual Environment and that can be used to challenge participants' cognition. Nevertheless, the immersion is not complete. The usage of big size monitors/projections improves the engagement on the tasks but it still can be improved. Moreover, some of them require the patients to move to the location where the facilities were available, since the required components are not portable or cost-effective. Current technologies such as virtual reality glasses are affordable and they will improve the immersion on the tasks considerably.


\section{Methodology} \label{CH3method}

\begin{figure}[!t]
 \centering
 \includegraphics[scale=1]{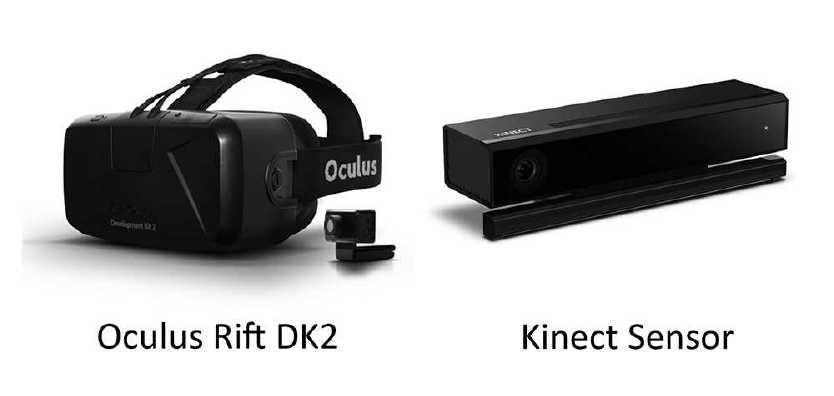}
 \caption{Virtual Reality glasses (Oculus Rift or Samsung Gear VR) and Depth sensor (Microsoft Kinect 2) are the devices required to perform the tests.}
 \label{fig:Figure1}
 \end{figure}

\begin{figure}[!t]
  \centering
  \includegraphics[scale=1]{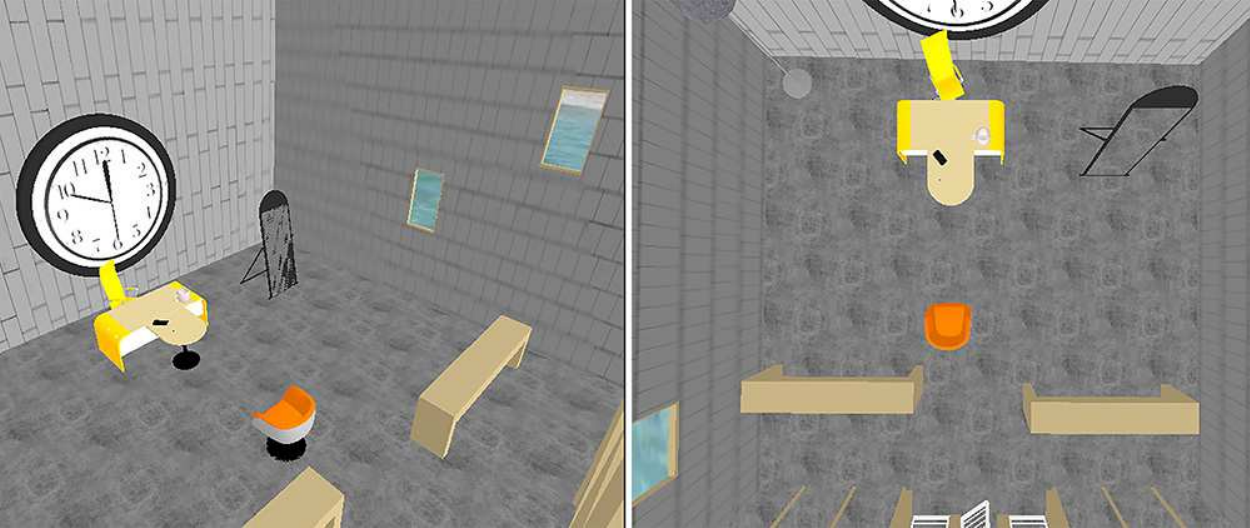}
    \caption{Virtual Reality Room. Corner and top down view}
  \label{fig:Figure2}
 \end{figure}

The proposed methods for Alzheimer's detection were developed to be conducted in Virtual Environments. Thus, \hyperlink{a_VR}{VR} glasses are used in order to maximise the patients' immersion in the \hyperlink{a_VEs}{VE} and a depth camera is utilised to track the patients' movements and animate their avatars in order to increase the immersion (see figure~\ref{fig:Figure1}). These tools are affordable and they can be reused by other patients, resulting to a low cost Alzheimer’s detection application. The main scene was designed such as to recreate a doctor's office or any other real room that the patient can recognise (office, living room), see figure~\ref{fig:Figure2}. Then, the tests were designed taking advantage of the virtual environment versatility, such as the absence of physical limits.

\begin{figure}[!t]
 \centering
  \includegraphics[scale=1]{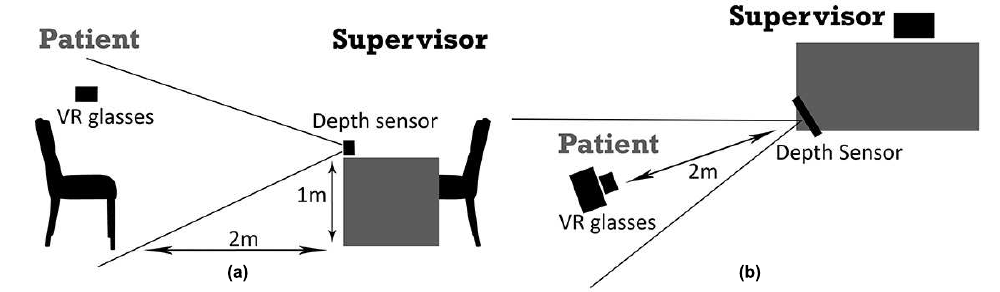}
    \caption{(a) Top down view showing the position of the devices, and the location of the patient and the supervisor. Around the patient should be enough space to avoid collisions with any real objects. (b) Side view showing the position of the devices, and the location of the patient and the supervisor. The depth sensor should be located approximately one meter above the floor and the patient has to be approximately two meters away from the depth sensor, in order to be in the view range.}
  \label{fig:Figure3}
 \end{figure}

\begin{figure}[!t]
 \centering
  \includegraphics[scale=.9]{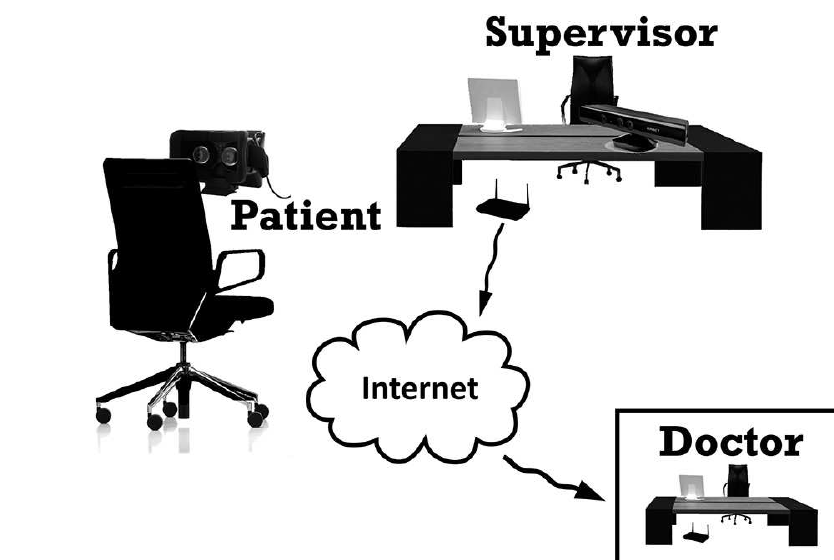}
    \caption{Early detection Test setup. The patient is sat in front of the depth sensor, wearing the Virtual Reality glasses. The supervisor/doctor monitors the test from the computer locally or remotely.}
  \label{fig:Figure4}
 \end{figure}

Regarding the real environment, since Alzheimer's patients are mainly elder people; all tasks are performed seated on a chair requiring only minimal movements. The patient is sitting on a comfortable chair about two meters in front of the depth sensor and is helped to place on the Virtual Reality glasses (see figures~\ref{fig:Figure3} and~\ref{fig:Figure4}). Once the patient feels comfortable with the glasses, we proceed with the tests. The doctor (or test supervisor) has to be in front of the computer locally or remotely, in order to control the software application and is advised to take notes during the process typing details that may be useful later, since it is possible to obtain in that way extra information about the patient's behaviour. Despite most of the tests are automatic and the patient only has to follow the instructions, some actions, such as starting a new task or playing sounds, requires the doctor's input in order to provide enough time to the patient to perform the actual tasks or to add a degree of randomness and adaptability.

All the proposed tests should be performed under the supervision of a healthy person following the instructions on the screen or the test's information sheets. Despite the fact that the test could be monitored by any person without mental impairments, it is recommendable that the supervisor during the tests to be a doctor, either locality or remotely, since additional information may be obtained during the whole process. In this chapter, four tests are introduced: Virtual Objects Memorization, Abnormal Objects Recognition, Virtual vs. Real Sounds and Bot Doctor Turing Tests.

\subsection{Virtual Objects Memorization Test (VOM)}
\hyperlink{a_VOM}{VOM} is a test focused on memory and its main objective is the analysis of patients’ learning and memory cognitive domain but it could also be used to get information about perceptual-motor impairments\cite{APA2013, Montenegro2015, Montenegro2016, Montenegro2017, MontenegroIISA2015}. It is split into three steps.

\begin{itemize}
\item The first step of this memory test is the recognition of six different virtual objects. The objects are commonly used during everyday activities making them easily identified and recognised. The objective of this first task is to check if the patient has problems in recognising quotidian objects (see figure~\ref{fig:Figure5}). One by one the objects are shown in front of them together with three options/names. If they select the correct name on their first attempt they get one point. They will not move to the next object until they get the correct name so they could always identify each object. Once all the objects are recognised, a virtual monitor displays a picture of all of them in specific positions over a table. The positions of some of these objects are the ones where they normally are located on a desk, e.g., the monitor on the back of the table and the keyboard in front of it. Then, the patient is asked to memorise the objects but nothing is mentioned about their position.

\item During the second step, the subject is asked about the objects' locations, in order to check the visual and the associative memory of the patient. In more details, the desk is shown in front of the patient and a number appears in the position where an object was located, while a list of the six possible objects is provided (see figure~\ref{fig:Figure6}). The subjects are asked about the object that was located in the position indicated by the number and they have to select the name of the correct one. The fact that the objects were located in logical positions should help the user to select the right answers.

 \begin{figure}[!t]
 \centering
  \includegraphics[scale=.9]{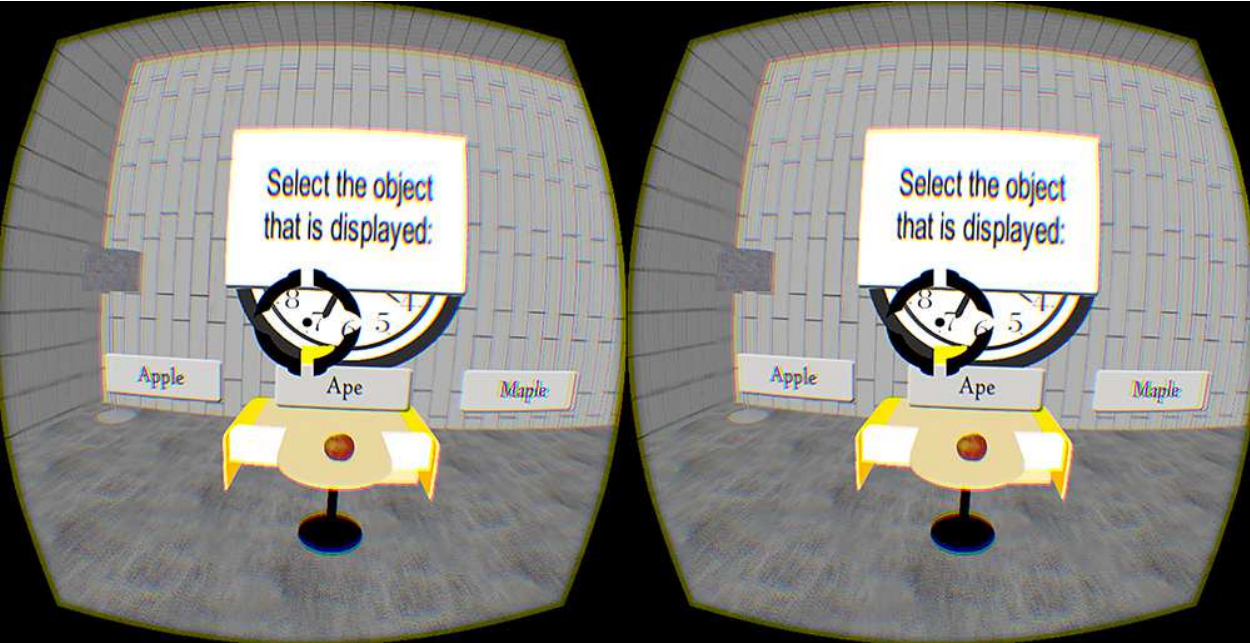}
    \caption{Virtual Object Memorization tasks. Object Recognition task where the patient has to identify and select the correct name of the objects on a table.}
  \label{fig:Figure5}
 \end{figure}

\item Finally, in the last task, pairs of interacting objects are placed in the virtual room and have to be located and memorised by the patient. Then, one of the objects is shown and the subject has to recall its interacting pair and their functionality. Once the third stage is completed, there is an extra task associated with the first step, where the patient is asked to recall the objects that were shown. The aim of the last two tasks is to evaluate the patient's short-term memory incorporating interacting objects and their functionalities. So, the novel part here is focused mainly on the combination of interacting objects with their functionalities.
\end{itemize}

Also, it should be mentioned that in order to proceed through the test, the subject has to select amongst different options. This selection is performed moving the head to aim towards the desired option using the direction of view (scope), as we can see in Figures~\ref{fig:Figure5} and~\ref{fig:Figure6}. Therefore, the patients have to move slightly their head to fulfil the tasks. This interaction could be improved and be made more realistic, for example, by tracking participants' gaze or gestures or voice recognition. However, since we aimed for reduced movements (due to reduced mobility of some elder people) and since by the time of the development of the tool the technology available was not yet reliable enough to be used or it was difficult to combine with \hyperlink{a_VR}{VR} headsets, the head interaction was kept. In next chapter (Chapter~\ref{Chapter4}), a method for gaze tracking using a reduced amount of \hyperlink{a_EEG}{EEG} sensors that could facilitate this \hyperlink{a_HCI}{HCI} is proposed.

This test, if all tasks are completed successfully, provides in total 21 points. The points are associated with the recognition of the objects, the recall of their position and the recall of the pairs of objects themselves and their functionality.

 \begin{figure}[!t]
   \centering
    \begin{subfigure}[b]{0.48\textwidth}
        \includegraphics[scale=.6]{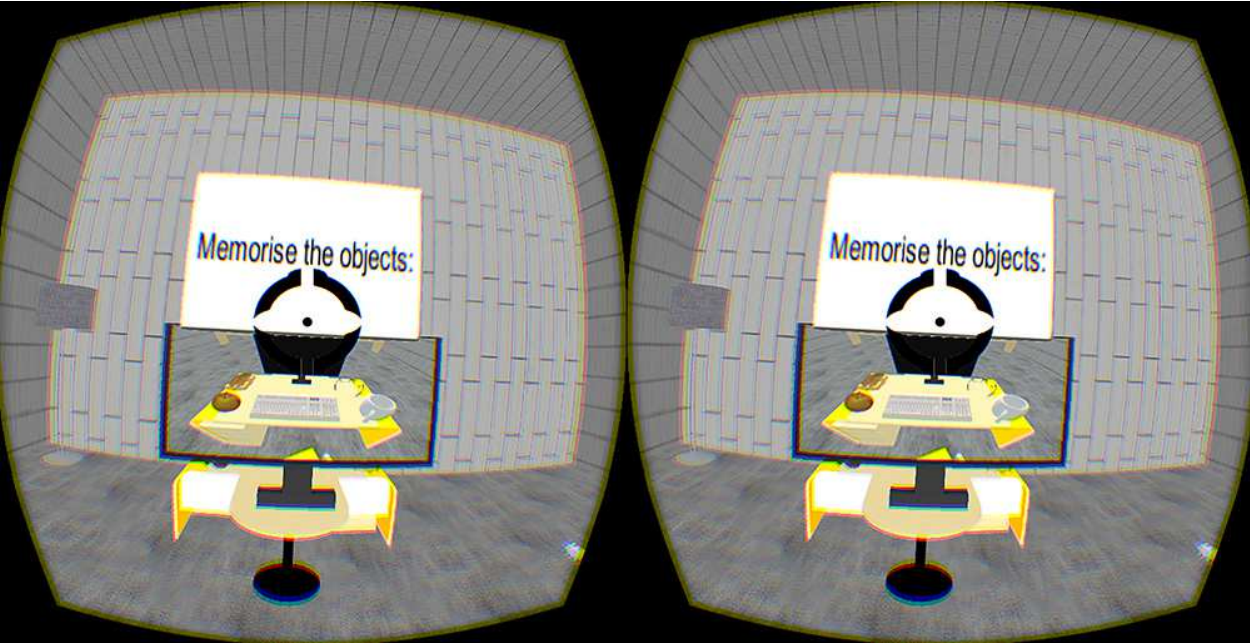}
        \caption{}
    \end{subfigure} \ \ \  %
    \begin{subfigure}[b]{0.48\textwidth}
        \includegraphics[scale=1]{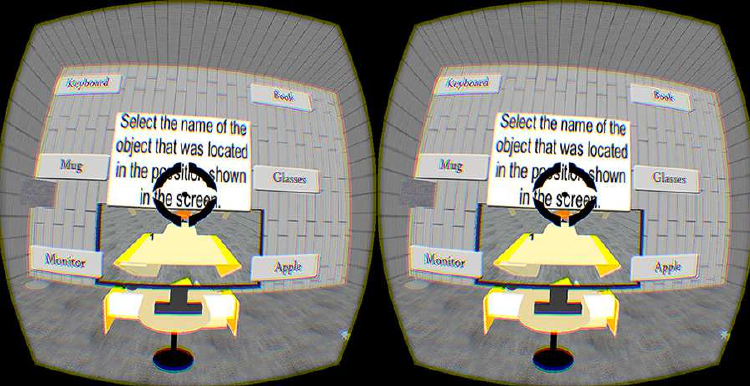}
        \caption{}
    \end{subfigure}\\
    \centering
    \begin{subfigure}[b]{0.5\textwidth}
        \includegraphics[scale=.3]{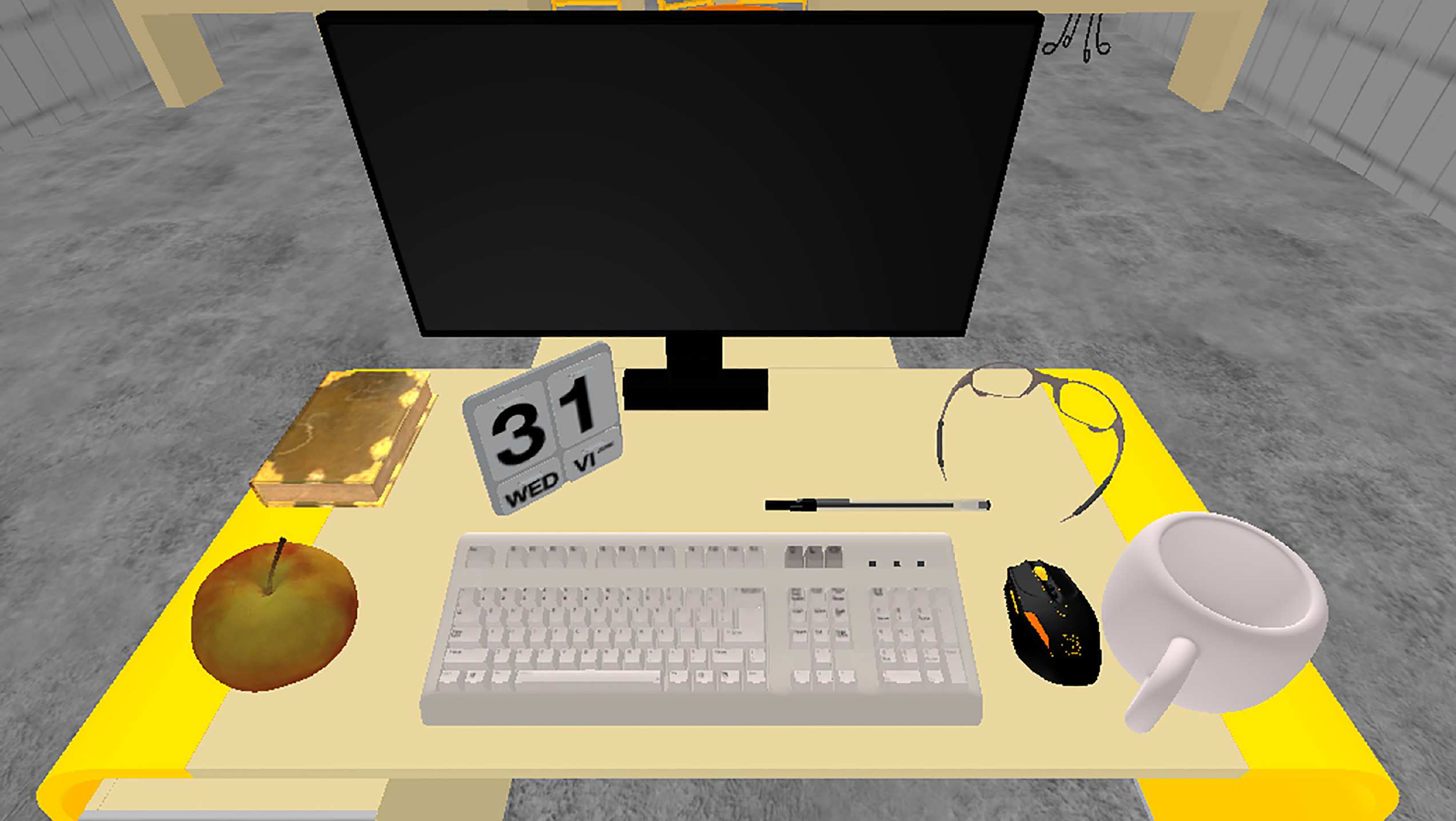}
        \caption{}
    \end{subfigure}%
    \caption{Virtual Object Memorization tasks. (a) The objects in their respective location are shown in order to be memorised by the patient. (b) Objects' location task where the patient has to select the name of the object that was shown on the screen. (c) A close view of the objects and their locations.}
  \label{fig:Figure6}
 \end{figure}

\subsection{Abnormal Objects Recognition Test (AOR)} \label{AORsection}
\hyperlink{a_AOR}{AOR} is a test that evaluates if the patient is able to discern if something is abnormal or not within the virtual environment and it is divided into three separate stages \cite{Montenegro2015, Montenegro2016, Montenegro2017, MontenegroIISA2015}. During the first stage, the patient is asked to detect any abnormalities present in the room in order to analyse their perceptual-motor cognitive skills \cite{APA2013}. Table~\ref{tabAOR} describes the abnormalities that can be find in the room. The participant is expected to find all the abnormalities in the room and describe each abnormality they find. Each abnormality correctly detected is scored with one point. The abnormalities search challenges participants attention and evaluates their semantic memory.

\begin{figure}[!t]
 \centering
  \includegraphics[width=.65\columnwidth]{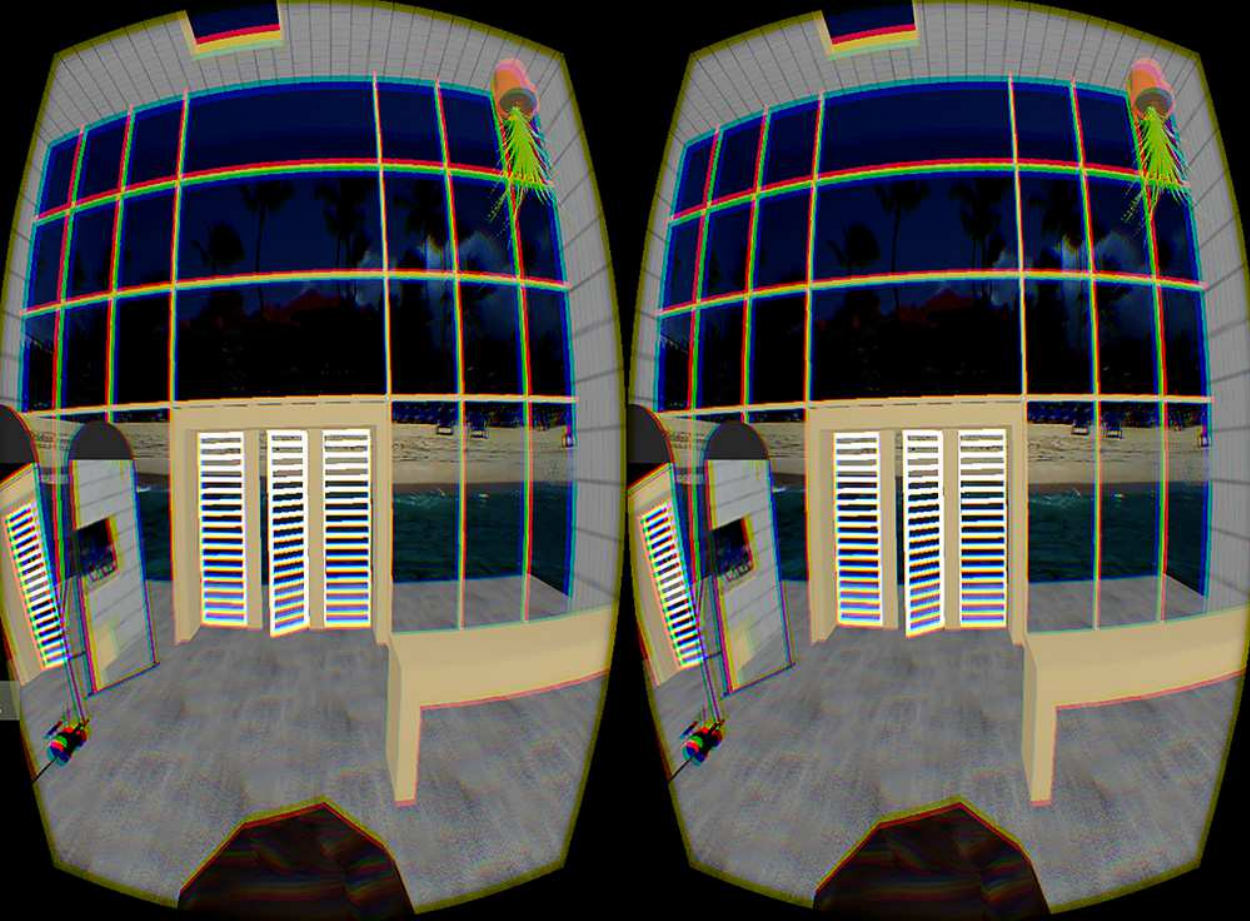}
    \caption{Abnormal Objects Recognition test preview. Some abnormalities such as the upside-down pot and the toy car going through the mirror can be observed in this figure.}
  \label{fig:Figure7}
 \end{figure}

The subject in the second stage is requested to read the clock that is located on one of the walls in the room. Shortly afterwards the patient is asked if the illumination (i.e. ambient light) of the room is in accordance with the time, (see Figure~\ref{fig:Figure7}).

\setlength{\tabcolsep}{4pt}
\begin{table*}[!b]
\caption{\label{tabAOR} Abnormalities in the room ordered by identification difficulty}
    \centering
        \begin{tabular}{r|rrrrr}
        \hline
        \noalign{\smallskip}
        \textbf{Object} & \textbf{Abnormality}\\
        \noalign{\smallskip}
        \hline
        \noalign{\smallskip}
        Chair & Wanders around the room\\
        \noalign{\smallskip}
        \hline
        \noalign{\smallskip}
        Mug   & Spins\\
        \noalign{\smallskip}
        \hline
        \noalign{\smallskip}
        \multirow{ 2}{*}{Car}   & Goes through the wall\\
              & Change shape when goes through the mirror\\
        \noalign{\smallskip}
        \hline
        \noalign{\smallskip}
        Potted plant    & Upside down\\
        \noalign{\smallskip}
        \hline
        \noalign{\smallskip}
        Mirror    & Does not reflect\\
        \noalign{\smallskip}
        \hline
        \end{tabular}
\end{table*}
\setlength{\tabcolsep}{1.4pt}

Finally, the objective of the third stage is to evaluate the coordination of the patient when the physics of a mirror reflection are changed. An avatar, whose movements are linked to the patients’ movements, is displayed in front of them. Therefore, it will look like a mirror; however, the movements of the avatar are reversed. Once the patients detect how the avatar moves, they are asked to perform specific movements with the avatar (see Table~\ref{tab2}). This task allows the study of perceptual-motor and executive function impairment \cite{APA2013}.

This test can provide a maximum of 10 points. Seven points are associated with the correct detection of the abnormalities and three are associated with the correct movement of the avatar.

\setlength{\tabcolsep}{4pt}
\begin{table*}[!t]
\caption{\label{tab2} Movements that the patient has to perform to move the avatar and their evaluation}
    \centering
        \begin{tabular}{r|r}
        \hline
        \noalign{\smallskip}
        \textbf{Task} & \textbf{Answer and Evaluation}\\
        \noalign{\smallskip}
        \hline
        \noalign{\smallskip}
        Raise your right hand & \multirow{ 2}{*}{No = 1 point}\\
        Is the avatar behaving as a mirror? & \\
        \noalign{\smallskip}
        \hline
        \noalign{\smallskip}
        Move the body of the avatar towards the mug & \multirow{ 2}{*}{First attempt = 1 point}\\
        (there is a mug on the right of the avatar) & \\
        \noalign{\smallskip}
        \hline
        \noalign{\smallskip}
        Grab the mug   & First attempt = 1 point\\
        \noalign{\smallskip}
        \hline
        \end{tabular}
\end{table*}
\setlength{\tabcolsep}{1.4pt}

 \begin{figure}[!t]
 \centering
  \includegraphics[width=.7\columnwidth]{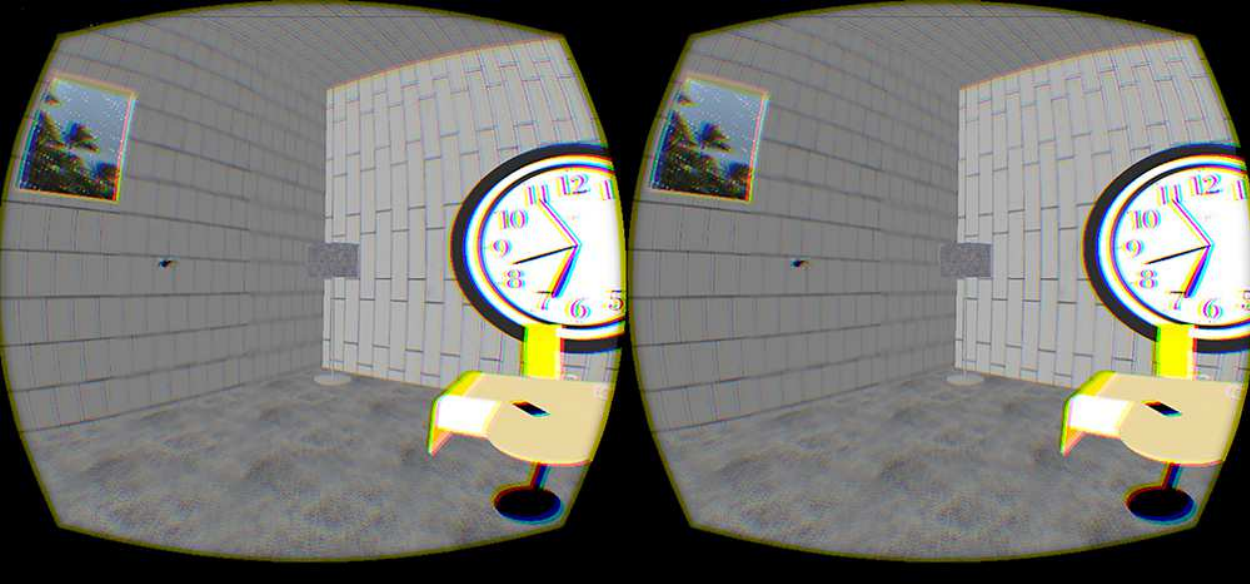}
    \caption{Virtual vs Real Sounds test preview. Some of the visual clues such as the fly, the clock and the mobile phone appear on this figure.}
  \label{fig:Figure8}
 \end{figure}

\subsection{Virtual vs Real Sounds Test (VRS)}
The main objective of \hyperlink{a_VRS}{VRS} test is the detection and recognition of objects/events through audio and to check if the patient is able to discern between real and virtual sounds \cite{Montenegro2015, Montenegro2017, Montenegro2016}. As a result, this test analyses patients' mental and cognitive flexibility, their complex attention and memory \cite{APA2013}.

Different sounds are played during the test, such as a clock ticking, a fly, rain at the background, etc., (see Figure~\ref{fig:Figure8}). The patient has to recognise and name all the sounds that they listen during the test. In addition, the virtual environment will provide some visual clues to help them with the identification. Once the subjects have identified all the sounds, they are asked about the origin of the sounds. During this stage, the source of some sounds is coming either from the virtual environment or the real world reproduced by the supervisor. This second part requires the patient to recall all the sounds that were played and to identify which ones were reproduced by the supervisor. This test gives a maximum score of 6 points. One point is obtained for each object that is recognised and categorised properly as real or virtual.

\subsection{Bot-Doctor Turing Test (BDTT)}
\begin{figure}[!t]
 \centering
  \includegraphics[width=.65\columnwidth]{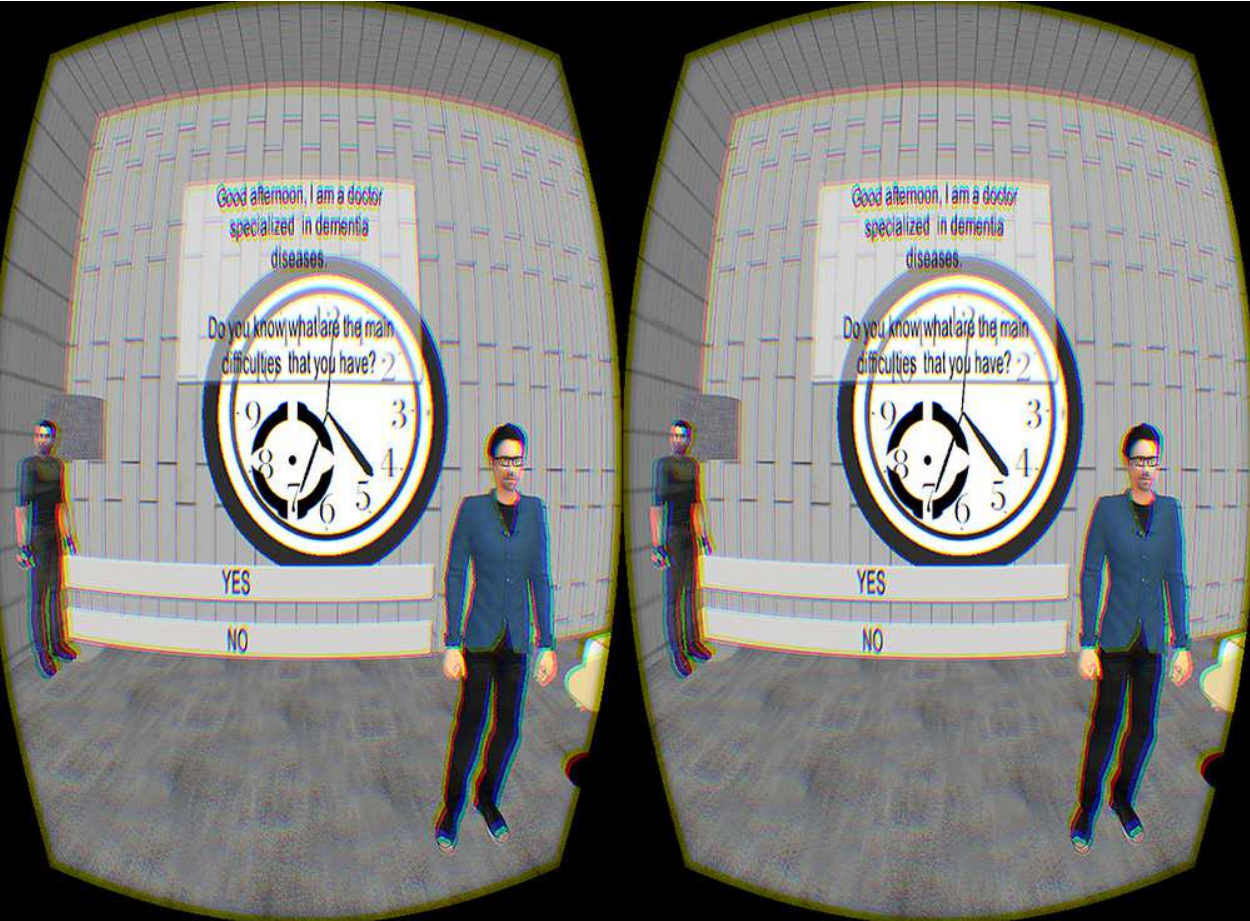}
    \caption{Selection Based - Bot Doctor Turing Test preview. The two avatars represent a doctor. The closest one is the one asking the questions in the screen, once it finishes, the other avatar will move to the position of this avatar and it will ask its questions. The participant has to read the information above and choose one of the options below moving the scope over the desired choice using head movements.}
  \label{fig:Figure9}
\end{figure}

The proposed Turing test is based on the ability of a patient to evaluate if a computer is able to impersonate a human \cite{Montenegro2017}. In order to perform the test, a human and a computer are located in a room, while the patient is in another location trying to communicate with the human and the computer in the first room without knowing who is responding, (see Figure~\ref{fig:Figure9}). The patient in the second room asks questions aiming to distinguish who is the human and who is the computer \cite{French2012}. We propose two different versions of the Alzheimer Turing test to evaluate patients' executive function, such as the ability to hold information and manipulate it and assess their process of making decisions \cite{APA2013}. Only the first one was implemented. The second one could not be implemented since by the time the tool was developed the accessible chatbots technology required was not good enough, but it was also proposed since it represents better the inverse Turing test idea.

\subsubsection{Selection Based - Bot-Doctor Turing Test}
In this test two virtual doctors have been designed to provide answers about a specific topic to the patient based on a simple Artificial Intelligence (AI) architecture. The conversation is formatted using a tree-structured state machine and is based on the same discussion and interaction system used in classic adventure video games. In more details, the discussion system provides a set of questions to the patients and based on their choice the \hyperlink{a_AI}{AI}-Bot-Doctors will provide an answer. The patient is informed that will have two discussions with two different Bot-Doctors and each \hyperlink{a_AI}{AI} system provides related but different answers. One of them will provide logical and correct answers, whereas the other will provide incorrect and sometimes absurd responses.

The whole discussion takes place in a virtual room, and two different human like avatars are used for the Bot-Doctors. The patients are immersed using the virtual reality glasses in the Virtual Environment, allowing them to see the avatars of the virtual doctors. Different variations of this test can be designed depending on the avatars used to represent the \hyperlink{a_AI}{AI}-Bots. In more details the Doctors' avatar can have either human or robotic appearance. Therefore, there can be two human avatars, two robotic avatars or a mixture of them. The objective of using non-human-like avatars is to check if the patient will have a different interaction if the avatar that provides the information does not look as a human. So, with this approach we can evaluate if the visual characteristics of the avatars can affect the provided answers. The test terminates either when the patients decide that they have asked enough questions to the avatars or if all the provided questions have been answered. This test gives in total 3 points. One point if the answer is that both doctors provided real answers and three if they distinguish them correctly.

\subsubsection{Script Based – Bot-Doctor Turing Test}
The second Alzheimer Turing test that is suggested is script based using chatbots. Chatbots are defined as computer programs that can maintain a conversation with a person \cite{Shieber2011}. ELIZA, IBM's Watson and Apple's Siri are some of the most famous chatbots. ELIZA was the first chatbot and was created by Joseph Weizenbaum; IBM's Watson is a chatbot that won in Jeopardy TV show in 2011; and Apple's Siri is the iPhone's personal assistant application. Also, many open source chatbots are available such as ALICE that could be used in this Bot-Doctor Turing test, but they were not good enough by the development time of the tool.

In order to interact with chatbots, people start a chat with the \hyperlink{a_AI}{AI}, they write the messages on the computer chat screen and the \hyperlink{a_AI}{AI} or human will engage in conversation. If voice recognition and synthesis are supported then this option can be selected. The proposed test uses a chatbot and a human to have a discussion with the patient. Therefore, all the entities that are involved in this test are the same as the used in a Turing Test, two humans and one computer. The patients are asked to chat for five minutes with each entity and, at the end, they are asked to decide which entity is a human. The intelligence of the chatbot is settled in accordance with patient's IQ in order to avoid the ceiling effect. It is expected that Alzheimer's patients fail distinguishing a simple \hyperlink{a_AI}{AI} based chatbot from a human. Regarding the scoring system, this test provides exactly the same points as the previous one.


\section{Results}

\subsection{Participants}
The twenty participants of this test are aged between 23 and 82 years old including both healthy people and Alzheimer's patients diagnosed with mild dementia less than a year ago. Furthermore, Alzheimer's patients in advanced conditions were tested but, the process was not possible to be completed, they could not focus on the tasks and they could only wear the \hyperlink{a_VR}{VR} glasses for a short time. This does not affect our proposal since our aim is the creation of a model that represents healthy people. Moreover, in case outliers (people with cognitive impairments) were needed for testing, only early \hyperlink{a_AD}{AD} would be tested.

The participants were recruited taking into account their sex, age and educational background. The recruitment aim was to gather a compensated gender and educational background and spread in age population. These decisions were taken in order to be able to analyse a possible bias due to these elements. Moreover, it was possible to select people from different countries, allowing us to detect any cultural related problem during the test. When it comes to the \hyperlink{a_AD}{AD} participants, the required ethics to work with them were submitted but rejected, therefore due to time limitations, the \hyperlink{a_AD}{AD} patients that took part in the experiment were Spanish \hyperlink{a_AD}{AD} patients diagnosed with mild and severe \hyperlink{a_AD}{AD}, whom were tested in Spain. The addition of the \hyperlink{a_AD}{AD} participants do not contribute to validate the screening tool, but it helped to check that the results obtained were coherent and if the virtual reality tool is adequate for this type of patients. Amongst the selected subjects half were male and the other half female of whom eleven had college or higher educational background and nine did not reach high school. None of the Alzheimer's participants' education level reached high school. The participants' nationalities included French, Spanish, Vietnamese and Greek so the non-English speakers were tested in their own language. It has to be mentioned that one of the patients was illiterate so the test was performed orally.

\subsection{Data}
The main data collected during the experiment are the individual scores for each test. As it was explained in the Methodology section \ref{CH3method}, each test has a maximum score and each task provides a different number of points. In more details, regarding the recognition and memory tasks, each recalled and recognised object scores 1 point.

\begin{figure}[!t]
   \centering
    \begin{subfigure}[b]{0.5\textwidth}
        \includegraphics[scale=.8]{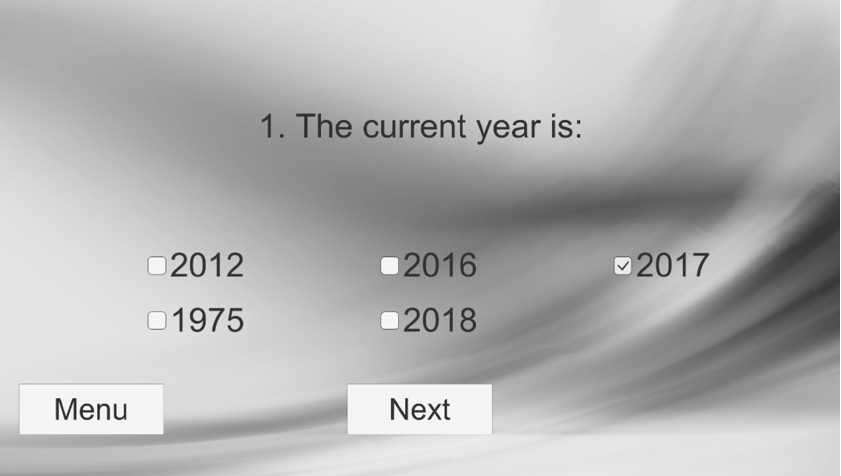}
    \caption{}
    \end{subfigure}%
    \begin{subfigure}[b]{0.5\textwidth}
        \includegraphics[scale=.8]{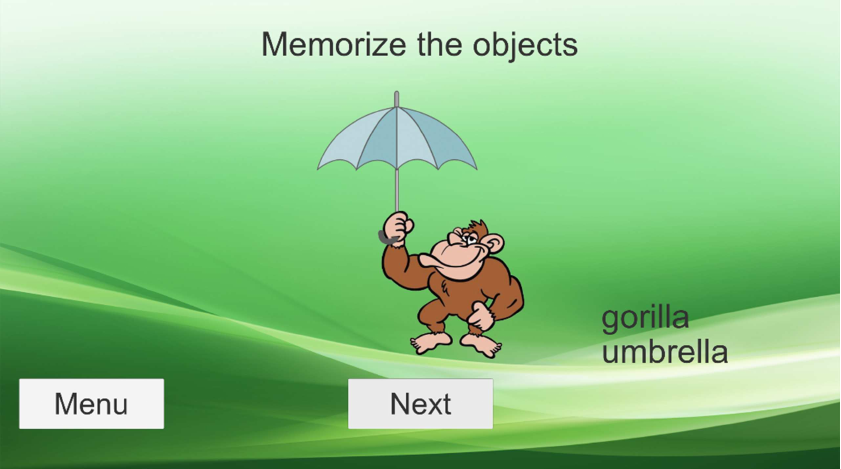}
    \caption{}
    \end{subfigure}\\%
    \begin{subfigure}[b]{0.5\textwidth}
        \includegraphics[scale=.8]{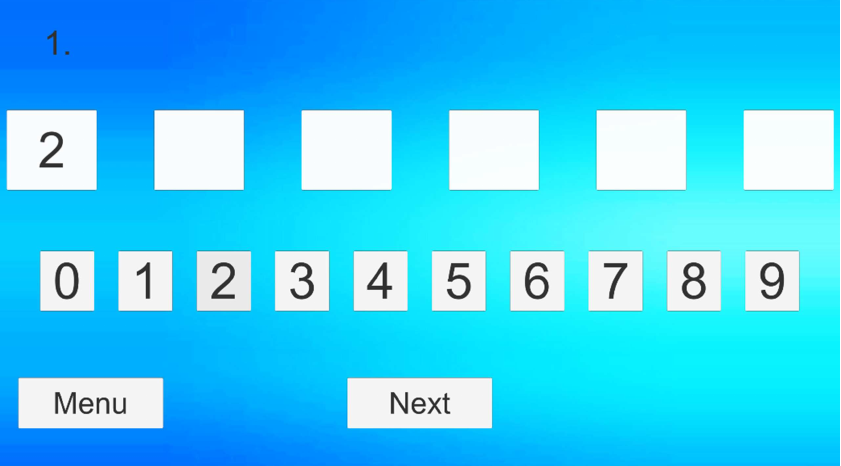}
    \caption{}
    \end{subfigure}%
    \caption{State-of-the-art test on its computerised version. a) Doctor Oz Memory Quiz. b) Visual Association Test. c) Dichotic Listening Test}
  \label{fig:Figure12}
 \end{figure}

In addition, patients have been also tested using other well-known state-of-the-art Alzheimer screening tests, in order to perform a comparative study and allow the evaluation of the proposed novel non-invasive diagnosis screening tests. Breaks were made between tests to avoid participants getting tired and reduce bias effects due to fatigue. All the tests were computerised so the patients have to use a computer through the whole process retaining also the same conditions among all tests. The state-of-the-art screening tests used in our comparative study are Dr. Oz Memory Quiz \cite{CruzOliveretal2014}; Visual Association Test \cite{Lindeboometal2002}; and Dichotic Listening Test \cite{Ducheketal2005} (see Figure~\ref{fig:Figure12}).

The Dr. Oz Memory Quiz is an Alzheimer's detection screening method that is based on \hyperlink{a_SLUMS}{SLUMS} test \cite{CruzOliveretal2014}. It is formed by seventeen questions and tasks that score 1 point each.

In the case of Visual Association Test \cite{Lindeboometal2002}, six pairs of objects are shown consecutively, and after the patients memorise them, one object of each pair is shown and they have to recall its couple. This test gives a maximum score of 6 points, one per recalled couple.

During the Dichotic Listening Test, the patients have to memorise six digits played to each ear separately and the process is repeated eight times \cite{Ducheketal2005}. As the results are stored by ear, the maximum score is 24 points per ear. This test is based on the fact that the difference of the recalled digits between the left and right ear in a healthy person is minimal compared with the predominance of the right ear for patients with Alzheimer's. Therefore, the data collected from this test is the difference of the left ear score minus the right ear score. The maximum score for each ear (24) is added in order to obtain positive values. Finally, the maximum possible value is delimited to 24, since left predominance or both ear equivalence are considered as healthy results (DLTre - Dichotic Listening Test right ear). In this way, the results decrease when the right ear predominance is high. Therefore, these results and the ones obtained by the other tests can be compared. Additionally, the number of digits recalled are also collected. This value will represent the memorization capacity.

\subsection{Experimental Procedure}
This section explains the process the participants went through since they entered the testing room until they left it. Firstly, the general purpose of the test, the technology they were going to use and a summary of the type of tasks they would have to go through were explained and the Instructions sheet was provided. They were asked to read the Instructions (see Appendix \ref{AppendixA} - Instructions \ref{AppAinstructions}), they were informed that the supervisor of the experiment would guide them through all the process and once they finished reading, the experiment started.

The participants were first evaluated using the state-of-the-art-tests. These required them to sit in front of a laptop and to follow the instructions on the screen. In case the participant were not familiarized with mouse and keyboard usage, they were helped to select the options (only the eldest participant required help). Headphones were provided for the \hyperlink{a_DLT}{DLT} test. They were allowed to ask any doubt about the tool at any time and to rest between tests as long as they required.

The second part of the experiment involved the novel virtual reality tests, therefore the subjects were asked to sit in a location in front of the Kinect sensor where they would not collide with any object in the room and far enough from the sensor to able to detected them. The instructions of the tests were reminded and they were helped to put on the \hyperlink{a_VR}{VR} glasses before each test. The glasses were removed at the end of each test so the subjects could rest before the next one. The first time they enter the virtual environment, they are allowed to practice their interaction with the interface. They have time to get used to navigate moving the head and selecting objects by fixing their sight. The interaction is very simple and their ability to using the devices is not evaluated so there should not be any bias due to different level of expertise when using technology.

The supervisor annotated the participants answers, their doubts and any extra information about participants behaviour and reactions during each test. At the end of the last test, the participants were asked about their impression of the tool and the Interview sheet was provided (see Appendix \ref{AppendixA} - Interview \ref{Interview}).

\subsection{Evaluation}

The scores obtained by the participants during all the test are presented in Table \ref{tab:Ch3Results}. This section will evaluate these results, analysing their correlation between tests and with participants' data (health status, educational background, gender and age).

\setlength{\tabcolsep}{2pt}
\begin{table*}[!t]
\caption{\label{tab:Ch3Results} Scores obtained for each test}
    \centering
      \begin{threeparttable}
        \begin{tabular}{rrrrr|cccc|cccc}
        \hline
        \noalign{\smallskip}
        \textbf{Age} & \textbf{Country} & \textbf{Sex} & \textbf{St} & \multicolumn{1}{c|}{\textbf{E.B.}} & \hyperlink{a_DROZ}{\textbf{DrOz}} & \hyperlink{a_VAT}{\textbf{VAT}} & \multicolumn{2}{c|}{\hyperlink{a_DLT}{\textbf{DLT}}} & \hyperlink{a_VOM}{\textbf{VOM}} & \hyperlink{a_AOR}{\textbf{AOR}} & \hyperlink{a_VRS}{\textbf{VRS}} & \hyperlink{a_DBTT}{\textbf{DBTT}}\\
        \noalign{\smallskip}
        \hline
        \noalign{\smallskip}
        & &  &  &  &  &  & \multicolumn{1}{|c}{\textbf{Sco}} & \multicolumn{1}{c|}{\textbf{FRE}} & \\
        \noalign{\smallskip}
        \hline
        \noalign{\smallskip}
        23 & Spain & F & H & O-HS & 15 & 6 & 48 & 6 & 21 & 8 & 5 & 1\\
        24 & Spain & M & H & O-HS & 16 & 6 & 47 & 6 & 21 & 9 & 5 & 3\\
        24 & Spain & M & H & U-HS & 16 & 6 & 47 & 4 & 18 & 8 & 5 & 1\\
        25 & Greece & M & H & O-HS & 15 & 6 & 42 & 3 & 21 & 10 & 6 & 3\\
        26 & Greece & M & H & O-HS & 16 & 6 & 36 & 6 & 21 & 9 & 5 & 1\\
        27 & Vietnam & F & H & O-HS & 14 & 3 & 39 & 3 & 19 & 9 & 5 & 3\\
        29 & Greece & F & H & O-HS & 15 & 6 & 45 & 3 & 20 & 8 & 6 & 3\\
        30 & Spain & F & H & O-HS & 16 & 6 & 45 & 3 & 19 & 9 & 5 & 3\\
        44 & Spain & M & H & O-HS & 17 & 6 & 48 & 4 & 21 & 10 & 5 & 3\\
        45 & France & M & H & O-HS & 17 & 6 & 46 & 5 & 20 & 9 & 6 & 3\\
        46 & UK & M & H & O-HS & 16 & 5 & 40 & 5 & 21 & 10 & 4 & 3\\
        48 & Spain & F & H & O-HS & 16 & 3 & 43 & 6 & 18 & 8 & 5 & 3\\
        53 & Spain & F & H & U-HS & 16 & 2 & 42 & 5 & 20 & 8 & 5 & 1\\
        53 & Spain & F & H & U-HS & 15 & 6 & 44 & 6 & 16 & 9 & 5 & 3\\
        54 & Spain & F & H & U-HS & 16 & 6 & 41 & 3 & 19 & 6 & 5 & 3\\
        56 & Spain & M& H & U-HS & 13 & 6 & 42 & 4 & 21 & 8 & 5 & 1\\
        58 & Spain & F & H & U-HS & 11 & 6 & 37 & 4 & 20 & 6 & 5 & 1\\
        60 & Spain & M & H & U-HS & 15 & 6 & 40 & 5 & 19 & 8 & 5 & 0\\
        78 & Spain & M & mAD & U-HS & 9 & 0 & 21 & 6 & 5 & 2 & 2 & 0\\
        82 & Spain & F & H & U-HS & 16 & 5 & 39 & 3 & 14 & 5 & 3 & 3\\
        \noalign{\smallskip}
        \hline
        \end{tabular}
        \begin{tablenotes}
        \item[1] The top row contains Age, Country, Sex, St: participants' health State, E.B: Educational Background, the three state-of-the-art tests and the proposed tests
        \item[2] The second row contains the two scores obtained during the DLT test. Sco: total number of digits recalled with both ears and FRE: score of First digits recalled with the Right Ear.
        \item[3] The values of the third column means M:Male and F:Female
        \item[4] The values of the forth column means H:Healthy and mAD: mild Alzheimer's Disease
        \item[5] The values of the fifth column means O-HS: Over High School and U-HS: Under High School
       \end{tablenotes}
      \end{threeparttable}
\end{table*}
\setlength{\tabcolsep}{1.4pt}


The mean normalized results for each test in relation to the subject's health status are shown in Table~\ref{tab3}. It is observed that healthy patients have achieved results close to the maximum score of each test. For example, the normalized averaged score of the Abnormal Objects Recognition test, for healthy subjects, is 0.826316, which is close to the maximum score of 1. On the other hand, the Alzheimer's patients have obtained significantly lower scores in comparison with the healthy participants. For instance, healthy subjects have obtained a mean of 0.924821, whereas the Alzheimer's subjects scored 0.238100 in the Virtual Objects Memorization test. This separation between results is also apparent in the other state-of-the-art tests, such as the relation 0.986842-0.375000 (Healthy – Alzheimer) on the Dichotic Listening test (\hyperlink{a_DLTre}{DLTre}). Table~\ref{tab4} demonstrates that the p-value of the novel tests proves that their results are correlated with the healthy and Alzheimer's status of the patients. When it comes to sensitivity and specificity in relation to the detection of healthy patients, the results are perfect. This suggests that more subjects need to be tested in order to corroborate the results of the novel Alzheimer's screening tests.

 \begin{figure}[!t]
 \centering
  \includegraphics[width=.9\columnwidth]{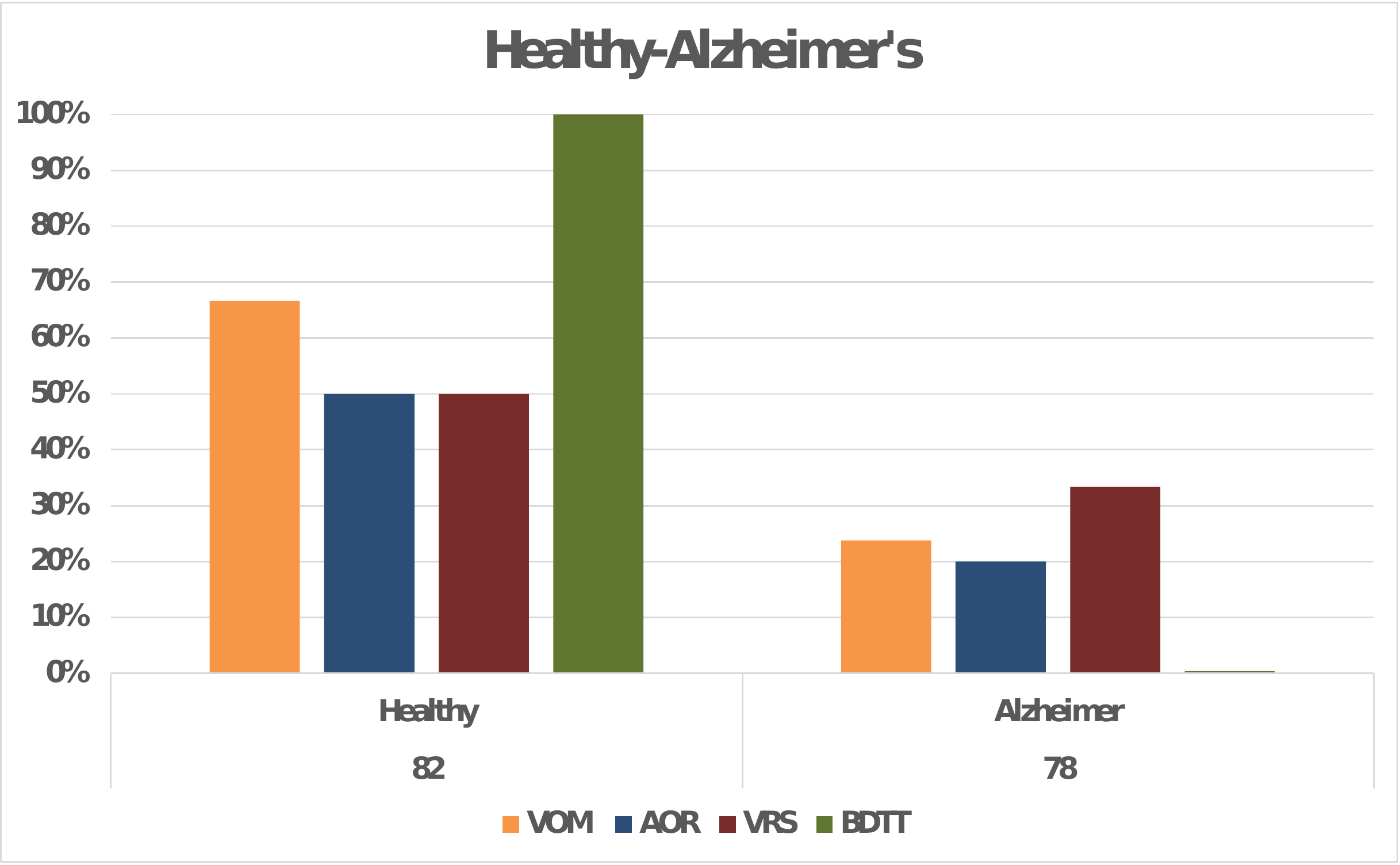}
    \caption{Worst healthy control result vs. best Alzheimer's result comparison. The scores of each test are show as percentages where the 100\% correspond to the maximum result possible of each test.}
  \label{fig:Figure11}
 \end{figure}

Figure~\ref{fig:Figure11} compares the results from the healthy participant that obtained the worst results and the Alzheimer's participant's ones. This comparison is shown to prove there is a considerable difference between the results obtained by a healthy and a mild \hyperlink{a_AD}{AD} patient. Nevertheless, since there is not enough \hyperlink{a_AD}{AD} participants to corroborate these results, they should be validated.

\setlength{\tabcolsep}{4pt}
\begin{table*}[!t]
\caption{\label{tab3} Normalised mean value of the tests regarding the Healthy and Alzheimer's patients}
    \centering
        \begin{tabular}{r|rrrrrrrr}
        \hline
        \noalign{\smallskip}
             & \textbf{Age} & \hyperlink{a_DROZ}{\textbf{DrOz}} & \hyperlink{a_VAT}{\textbf{VAT}} & \hyperlink{a_DLT}{\textbf{DLT}} & \hyperlink{a_DLTre}{\textbf{DLTre}}\\
        \noalign{\smallskip}
        \hline
        \noalign{\smallskip}
        \textbf{Healthy} & 42.47 & .900953 &	.894732 & .889253 &	.986842\\
        \noalign{\smallskip}
        \hline
        \noalign{\smallskip}
        \textbf{Alzheimer} & 78.00 & .529400 & .000000 & .437500 & .375000\\
        \noalign{\smallskip}
        \hline
        \hline
        \noalign{\smallskip}
             & \textbf{Age} & \hyperlink{a_VOM}{\textbf{VOM}} & \hyperlink{a_AOR}{\textbf{AOR}} & \hyperlink{a_VRS}{\textbf{VRS}} & \hyperlink{a_BDTT}{\textbf{BDTT}}\\
        \noalign{\smallskip}
        \hline
        \noalign{\smallskip}
        \textbf{Healthy} & 42.47 & .924821 & .826316 & .833311 & .736832\\
        \noalign{\smallskip}
        \hline
        \noalign{\smallskip}
        \textbf{Alzheimer} & 78.00 & .238100 & .200000 & .333300 & .385600\\ %
        \noalign{\smallskip}
        \hline
        \end{tabular}
\end{table*}
\setlength{\tabcolsep}{1.4pt}

Table~\ref{tab4} shows the correlation between the novel tests and the participant's health status, age, educational background and gender. It demonstrates that most of the results of the tests are not correlated with the age and the educational level of the patients, since the $p$-values do not reject the $Null$ hypothesis. If it is considered that the values are correlated when the $p$-value is under 0.01, then all the test show no correlation with the participants' parameters excepting the \hyperlink{a_AOR}{AOR} test. The novel test \hyperlink{a_AOR}{AOR} results a $p$-value of 0.003 (age) and 0.004 (educational level), indicating a relationship with those factors (age and educational background). Therefore, tasks that are more difficult can be added for patients with high educational background in order to eliminate this correlation. When it comes to the relationship between the results and the gender of the participants, the results of the male participants and the females' ones are not correlated (see Table~\ref{tab4}).

\setlength{\tabcolsep}{4pt}
\begin{table*}[!t]
\caption{\label{tab4}T-test for each test according to the healthy and Alzheimer's cases. The p-value (Sig. (2-tailed)) is less than 0.01 so the Null Hypothesis is rejected}
    \centering
        \begin{tabular}{r|rrrrr}
        \cline{2-6}
        \noalign{\smallskip}
            & & \multicolumn{4}{c}{Sig.(2-tailed)}\\
        \noalign{\smallskip}
        \cline{2-6}
        \noalign{\smallskip}
            &     & \textbf{Dementia} & \textbf{Age} & \textbf{Educational} & \textbf{Gender}\\
            &     & &  & \textbf{Level} & \\
        \noalign{\smallskip}
        \hline
        \noalign{\smallskip}
        \hyperlink{a_DROZ}{\textbf{DrOz}} & e.v.a\textsuperscript{a} & .000 &	.033 & .067 & 1.000\\
         & e.v.n.a\textsuperscript{b} & . & .088 & .104 & 1.000\\
        \noalign{\smallskip}
        \hline
        \noalign{\smallskip}
        \hyperlink{a_VAT}{\textbf{VAT}} & e.v.a & .001 & .325 & .462 & .615\\
         & e.v.n.a & . & .395 & .492 & .615\\
        \noalign{\smallskip}
        \hline
        \noalign{\smallskip}
        \hyperlink{a_DLT}{\textbf{DLT}} & e.v.a & .000 & .038 & .111 & .616\\
         & e.v.n.a & . & .076 & .142 & .619\\
        \noalign{\smallskip}
        \hline
        \noalign{\smallskip}
        \hyperlink{a_DLTre}{\textbf{DLTre}} & e.v.a & .000 & .198 & .262 & .400\\
         & e.v.n.a & . & .314 & .326 & .410\\
        \noalign{\smallskip}
        \hline
        \noalign{\smallskip}
        \hyperlink{a_VOM}{\textbf{VOM}} & e.v.a & .000 & .052 & .045 & .908\\
         & e.v.n.a & . & .128 & .084 & .909\\
        \noalign{\smallskip}
        \hline
        \noalign{\smallskip}
        \hyperlink{a_AOR}{\textbf{AOR}} & e.v.a & .000 & .003 & .004 & .433\\
         & e.v.n.a & . & .020 & .013 & .435\\
        \noalign{\smallskip}
        \hline
        \noalign{\smallskip}
        \hyperlink{a_VRS}{\textbf{VRS}} & e.v.a & .000 & .061 & .078 & .818\\
         & e.v.n.a & . & .113 & .104 & .818\\
        \noalign{\smallskip}
        \hline
        \noalign{\smallskip}
        \hyperlink{a_BDTT}{\textbf{BDTT}} & e.v.a & .000 &	.131 & .041 & .438\\
             & e.v.n.a & .  & .178   & .060 & .445\\
        \noalign{\smallskip}
        \hline
        \multicolumn{6}{r}{\textsuperscript{a}\footnotesize{Equal Variances Assumed}}\\
        \multicolumn{6}{r}{\textsuperscript{b}\footnotesize{Equal Variances Not Assumed}}
        \end{tabular}
\end{table*}
\setlength{\tabcolsep}{1.4pt}

The correlation between the \hyperlink{a_DROZ}{DrOZ}, \hyperlink{a_VAT}{VAT} and \hyperlink{a_DLT}{DLT} tests, and the novel ones is shown in Table~\ref{tab5}, where the Pearsons's correlation has been calculated, including the correlation coefficient and the $p$-value. The Pearsons's correlation show values between -1 and 1, been 1 fully correlated, 0 no correlation and -1 inverse correlation.  Most of the novel tests result high correlation values with most of the state-of-the-art,  excepting the \hyperlink{a_BDTT}{BDTT}. The Pearsons correlation value of the tests are between $0.832$ for \hyperlink{a_VOM}{VOM} -\hyperlink{a_DLTre}{DLTre}  and $0.486$ for \hyperlink{a_VRS}{VRS}-\hyperlink{a_DROZ}{DrOZ} and all the p-values are significant below $0.05$. However, the \hyperlink{a_BDTT}{BDTT} test Pearsons maximum correlation value is $0.388$ with \hyperlink{a_DROZ}{DrOZ} but the p-values are above $0.1$ so the correlation is not significant. This table shows that there is no correlation between state-of-the-art tests results and the \hyperlink{a_BDTT}{BDTT}'s ones, therefore it is not useful for the assessment of cognitive impairments. This can be attributed to the few amount of points of the test (three points) since the difference between a positive or negative assessment is one point score.

\setlength{\tabcolsep}{4pt}
\begin{table*}[!t]
\caption{\label{tab5}Correlation between state-of-the-art tests and novel ones}
    \centering
        \begin{tabular}{r|rrrrr}
        \cline{2-6}
        \noalign{\smallskip}
            &     & \hyperlink{a_DROZ}{\textbf{DrOz}} & \hyperlink{a_VAT}{\textbf{VAT}} & \hyperlink{a_DLT}{\textbf{DLT}} & \hyperlink{a_DLTre}{\textbf{DLTre}}\\
        \noalign{\smallskip}
        \hline
        \noalign{\smallskip}
        \hyperlink{a_VOM}{\textbf{VOM}} & Pearsons correlation & .589** & .666** & .741** & .832**\\
         & Sig. (2-tailed) & .006 & .001 & .000 & .000\\
        \noalign{\smallskip}
        \hline
        \noalign{\smallskip}
        \hyperlink{a_AOR}{\textbf{AOR}} & Pearsons correlation & .676** & .526* & .723** & .724**\\
         & Sig. (2-tailed) & .001 & .017 & .000 & .000\\
        \noalign{\smallskip}
        \hline
        \noalign{\smallskip}
        \hyperlink{a_VRS}{\textbf{VRS}} & Pearsons correlation & .486* & .602** & .720** & .688**\\
         & Sig. (2-tailed) & .030 & .005 & .000 & .001\\
        \noalign{\smallskip}
        \hline
        \noalign{\smallskip}
        \hyperlink{a_BDTT}{\textbf{BDTT}} & Pearsons correlation & .388 & -.060 & .228 & .249\\
             & Sig. (2-tailed) & .100 & .807 & .347 & .304\\
        \noalign{\smallskip}
        \hline
        \multicolumn{6}{r}{\textsuperscript{**}\footnotesize{Correlation is significant at the 0.01 level (2-tailed)}}\\
        \multicolumn{6}{r}{\textsuperscript{*}\footnotesize{Correlation is significant at the 0.05 level (2-tailed)}}
        \end{tabular}
\end{table*}
\setlength{\tabcolsep}{1.4pt}

As a summary, proposed tests' correlation with state-of-the-art tests and their lack of bias due to participants' age, educational background and gender, prove novel virtual test as valid early \hyperlink{a_AD}{AD} screening tools. On the other hand, due to the few amount of participants, these values are not conclusive and need to be validated. Since the purpose of the tests is to be able to identify healthy participants and discard as cognitive impairment any result out of the healthy model, there is not need to add more \hyperlink{a_AD}{AD} participants. However, the number of healthy participants should increase significantly to create an accurate and valid model.

In addition to the numerical results, some notes were collected during the testing process that could be used by the doctors to make a decision. The most interesting information collected on these notes is the lack of attention of Alzheimer's patients due to the animated virtual environment. For example, \hyperlink{a_VOM}{VOM} test contains some animated objects close to the questions' area that divert the attention of the patients, making them to forget about the task in progress. This reveals a problem in the Complex Attention cognitive domain.

\subsection{Qualitative Evaluation of the Proposed System}
At the end of the process, subjects were asked to evaluate the application. They had to fill a form evaluating the quality of the application in terms of interaction; simplicity of instructions/processes and the comfort of the \hyperlink{a_VR}{VR} glasses. Additionally, participants had the option to write any additional comments about the proposed cognitive tests (see Appendix \ref{AppendixA} - Interview \ref{Interview}).

Table~\ref{tab6} shows the overall evaluation of the proposed cognitive tests provided by the participants. The interaction with the application (i.e. keyboard, mouse, depth sensor and virtual reality glasses), the quality of the instructions (i.e. written and/or oral provided by the supervisor) and the comfort of the \hyperlink{a_VR}{VR} glasses were evaluated from 0 (really bad) to 5 points (really good). The overall evaluation obtained is positive, since the average result in all these categories is above 3 points, with elder people being more enthusiastic with new technologies, probably due to the completely new to them features and capabilities.

\setlength{\tabcolsep}{4pt}
\begin{table*}[!t]
\caption{\label{tab6}Qualitative evaluation of the proposed cognitive tests by the participants. Four characteristics were evaluated from 0 (Very bad) to 5 (Very good)}
    \centering
        \begin{tabular}{r|rrrrr}
        \hline
        \noalign{\smallskip}
        \textbf{Type} & \textbf{Mouse Inter} & \textbf{Kinect Inter} & \textbf{Instructions} & \textbf{VR Glasses}\\
        \noalign{\smallskip}
        \hline
        \noalign{\smallskip}
        \textbf{ $\mathbf{>=50}$} & 4.17(0.983) & 4.00(1.155) & 4.43(0.787) & 3.86(0.9)\\
        \noalign{\smallskip}
        \hline
        \noalign{\smallskip}
        \textbf{$\mathbf{<50}$} & 4.33(0.778) & 3.33(0.985) & 4.17(0.835) & 3.50(1.0)\\
        \noalign{\smallskip}
        \hline
        \end{tabular}
\end{table*}
\setlength{\tabcolsep}{1.4pt}

The average evaluation of the proposed approach was positive. Most of the participants evaluated the application as motivating and interesting; and the interaction and instructions as simple and easy to understand. On the other hand, there were some important comments that we took into consideration for possible tool upgrades.

\begin{itemize}
\item The glasses were not comfortable at the end of the test for some participants.
\item The colors of the environment were not natural or comfortable.
\item Some models quality was not good to be easily identifiable.
\item The quality of some images should be improved since they produce eye fatigue.
\item The graphical interface should be more attractive.
\end{itemize}

\subsection{Tool Upgrade}

\begin{figure}[!t]
   \centering
    \begin{subfigure}[b]{0.49\textwidth}
        \includegraphics[scale=.205]{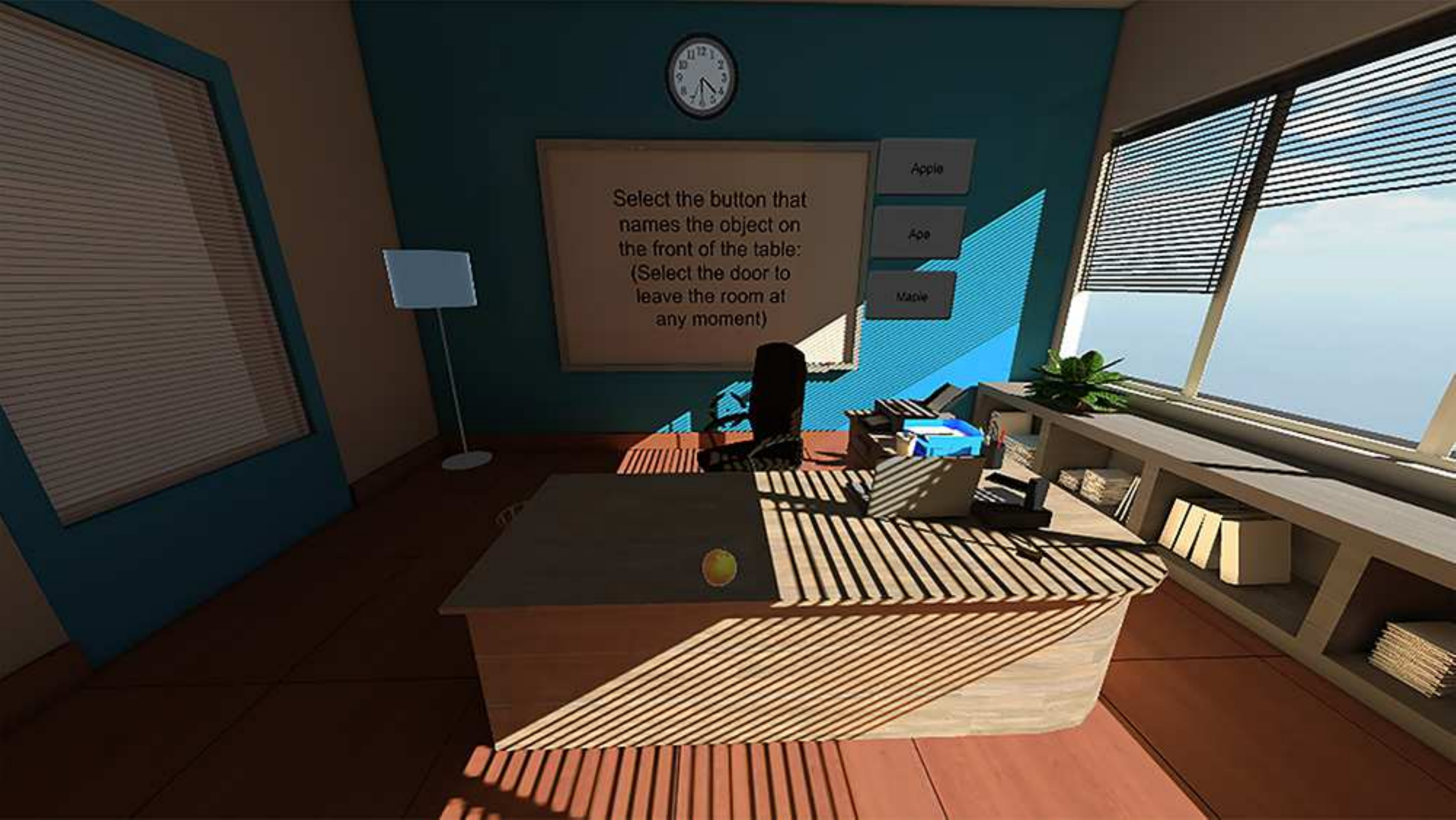}

    \end{subfigure}%
    \begin{subfigure}[b]{0.49\textwidth}
        \includegraphics[scale=.08]{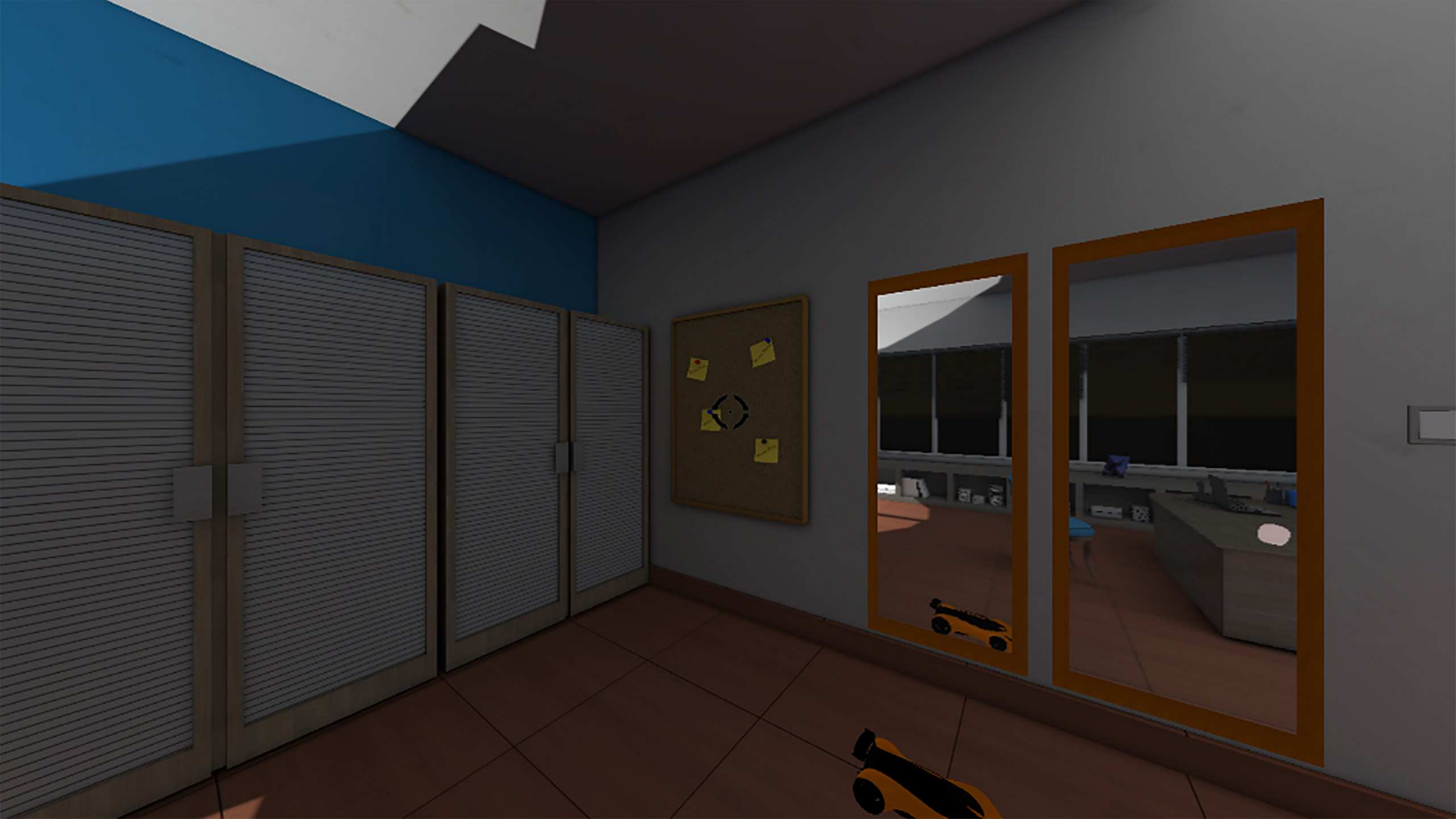}

    \end{subfigure}\\%
    \begin{subfigure}[b]{0.49\textwidth}
        \includegraphics[scale=.205]{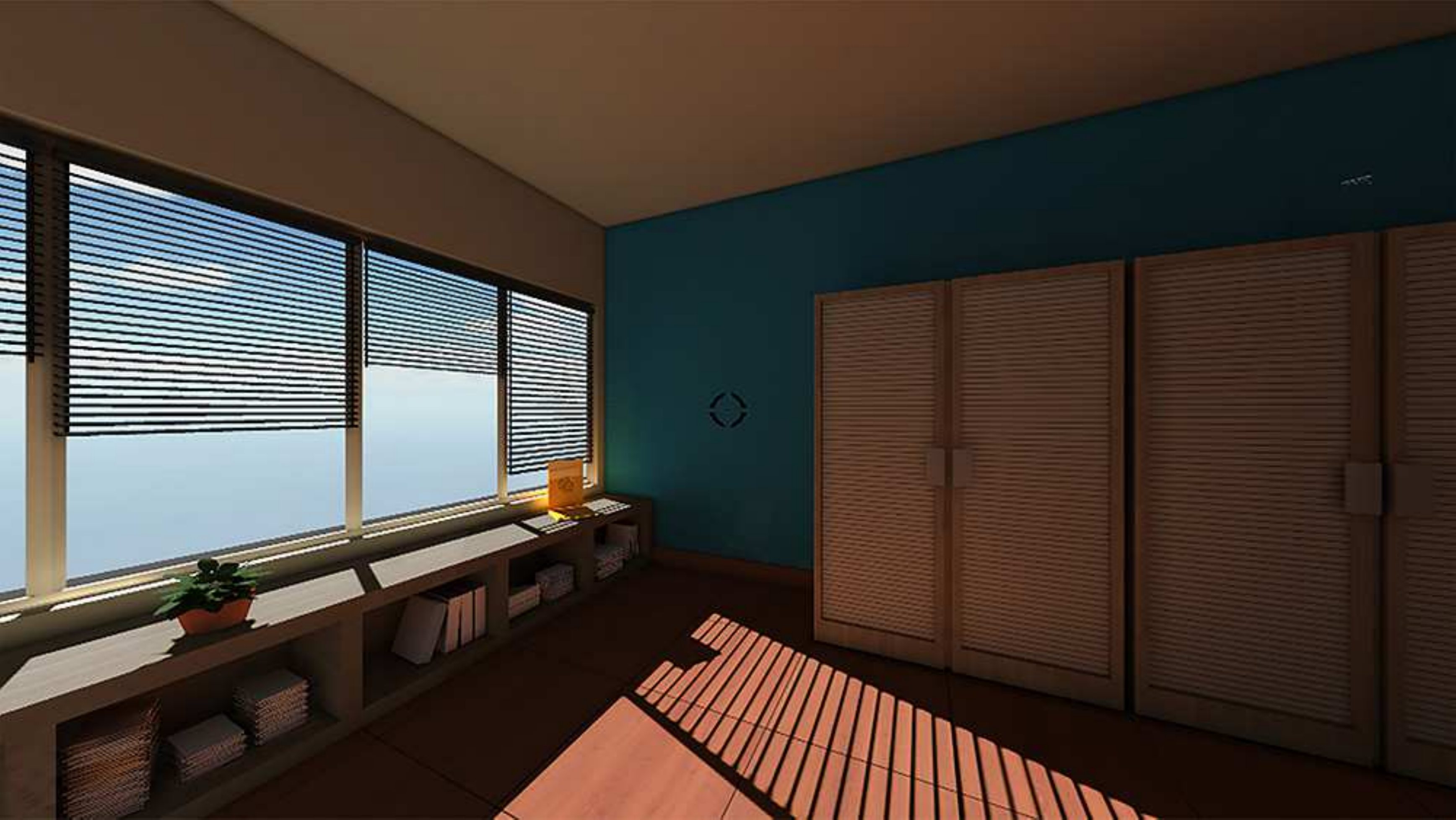}

    \end{subfigure}%
    \begin{subfigure}[b]{0.49\textwidth}
        \includegraphics[scale=.08]{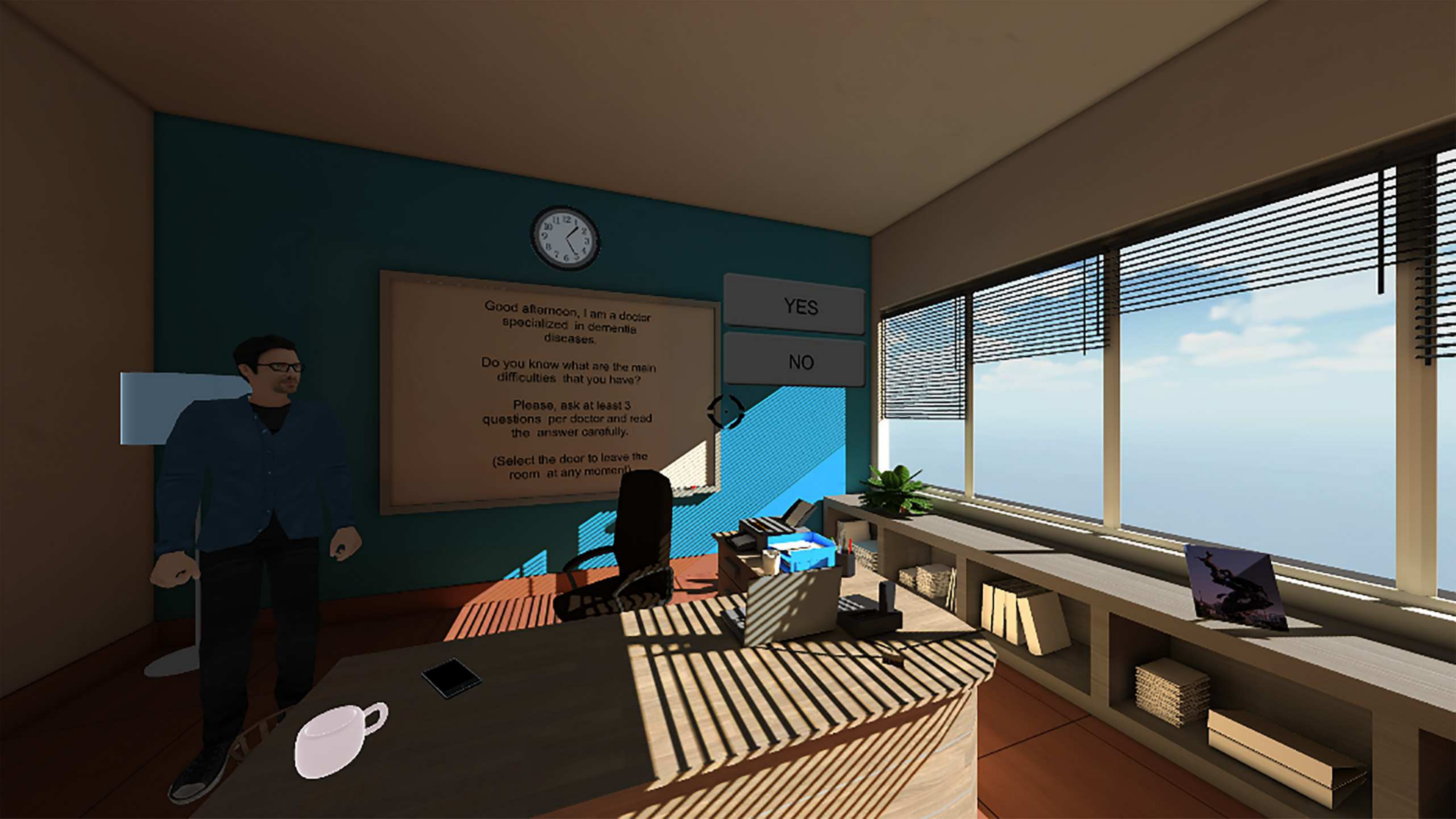}

    \end{subfigure}%
    \caption{Tool upgrade. The graphics have been improved and the Virtual Environment of the office is more realistic.}
  \label{fig:Figure13}
 \end{figure}

In order to make the tool available to the general public and taking into account the participants and the results evaluation, the tool has been upgraded. The novel version of the tool was redesigned to be supported by mobile phones headset and the latest version of the \hyperlink{a_VR}{VR} glasses. Since latest devices provide buttons to interact with the \hyperlink{a_VE}{VE}, it allowed us to remove the selection timer making the navigation more dynamic. The graphics have been improved and the room where the test takes place is more realistic (see Figure~\ref{fig:Figure13}). Some tasks have been removed from the mobile phone version to facilitate the usage of the tool everywhere. Therefore, the avatar movement using the Kinect from \hyperlink{a_AOR}{AOR} test and the real sound from \hyperlink{a_VRS}{VRS} test have been removed. These tasks are still available in the \hyperlink{a_VR}{VR} headset version.

\setlength{\tabcolsep}{4pt}
\setlength{\tabcolsep}{1.4pt}

Cognitive reserve has also been taken into account, the screening tool has been adapted to participants' IQ using their educational background as reference. Therefore, the tasks and evaluation system of three out of four tests have been updated. \hyperlink{a_VOM}{VOM} test displays three more items for memorisation for high educational background participants: a calendar, a pen and a PC mouse. \hyperlink{a_AOR}{AOR}'s most difficult tasks, mirror reflection and daylight relation to clock time, are only required for high IQ subjects. Furthermore, an extra abnormality is added for those participants: the 30th of February is shown in a calendar. \hyperlink{a_VRS}{VRS}'s mobile phone sound has been removed from low and normal educational background subjects so they do not have to memorise it nor have to identify that the sound comes from the real world.

Finally, a database has been created for the storage of the tests' results without storing any personal information. The stored results in this database could be a precious asset for future research.


\section{Discussion}

Alzheimer is a disease without foreseen cure, that is affecting more people every year, which related research and associated services (care) cost are increasing rapidly. Our objective is to provide affordable approaches that would help improving the current cognitive based detection systems.

Virtual Environments start to be part of medical treatments and diagnosis methods. They have been proved to be considerably useful since they provide secure and controlled environments to apply cognitive tests and help to increase the effectiveness of the related methodologies. These facts have been further validated with the proposed novel tests, since the results obtained show that it is possible to create an accurate e-health screening Alzheimer's diagnostic system. In this work, new methods for detection of Alzheimer's disease based on Virtual Environments were introduced. The proposed tests are focused on the evaluation of memory loss related to common objects, recent events, the diagnosis of problems in expressing and understanding language; and the ability to recognise abnormalities and to differentiate between virtual worlds and reality. In addition, two novel tests related with Alan Turing Imitation Game were proposed, where the human's intelligence is evaluated instead of the machine's one. The proposed approaches were evaluated in a comparative study with other well-known state-of-the-art cognitive tests. Finally, the obtained results indicate that the proposed methodologies and tests can provide accurate indications of the presence of Alzheimer's disease. They also show the virtual environments' potential to further improve \hyperlink{a_AD}{AD} diagnosis methods by providing tests more adaptive to patients. However, the implemented Turing's related test did not provide enough evidence to demonstrate its validity. Moreover, the number of the participants needs to be increased in order to prove the screening tool is able to generate a valid healthy participants model.

\hyperlink{a_HCI}{HCI} in Virtual Environments still can provide more realism. Currently, \hyperlink{a_AD}{HCI} in Virtual Reality devices is based on controllers and head movement, while a more realistic interaction would consider eye tracking and voice, gestures or touch interactions. These types of interactions were not implemented since the technology available during the development of the tools did not allow their implementation. AD patients are usually elder people not used to new technologies, therefore, a simpler and more realistic \hyperlink{a_HCI}{HCI} will simplify the fulfilment of the \hyperlink{a_AD}{AD} cognitive impairment screening tests. Touch interaction techniques have not achieve yet reliable interaction levels and voice interaction performance deteriorates on noisy scenarios. There are some reliable eye tracking techniques but they are not easy to combine with \hyperlink{a_VR}{VR} headsets. On the other hand, eye tracking could be adapted to \hyperlink{a_VR}{VR} devices by using \hyperlink{a_EEG}{EEG} signal. Using a reduced number of sensors, they could be easily added to the \hyperlink{a_VR}{VR} headsets. Since current \hyperlink{a_EEG}{EEG} eye tracking approaches use around fourteen to thirty-two sensors, they should be optimise to fit within the \hyperlink{a_VR}{VR} glasses.

In addition, the tests' capabilities could also be improved by adding extra information about participants' physical and emotional state. For example by analysing participants' reactions/emotions during their performance. Emotions classification can be used as a mechanism to detect dementia. For example, \hyperlink{a_AD}{AD} patients usually have difficulty recalling recent events, therefore, lack of reaction when recent autobiographical memories are triggered by visual stimuli will suggest an impairment in the learning and memory cognitive domain. The analysis of \hyperlink{a_AD}{AD} patients' emotions triggered by autobiographical memories have not been done previously, therefore, state-of-the-art emotion classification approaches should be used in order to be able to classify these emotions. For example, most approaches analyse facial expressions so there would be a wide amount of methods to classify emotions. Furthermore, the number of \hyperlink{a_EEG}{EEG} approaches is also considerable and they will allow an easy combination with \hyperlink{a_VR}{VR} technologies.




\chapter{Classification of emotions triggered by autobiographical memories with EEG and facial expression recognition}

\label{Chapter4} 

\lhead{Chapter 4. \emph{EEG Approach for Classification of Emotions Triggered by Autobiographical Memories}} 


\section{Introduction}
The American Psychiatric Association mention in~\cite{APA2013} the lack of social cognition as one of the cognitive domains affected by Alzheimer's disease, where \hyperlink{a_AD}{AD} patients are not able to read other people's emotions. Many projects are focused on studying dementia patients' capability to recognise emotions~\cite{Sapey-Triomphe15,VandenStock15} whereas a minority tries to analyse patients' facial expressions to specific stimuli. Since \hyperlink{a_AD}{AD} patients' facial expression is increased~\cite{Seidl12} the lack of them in presence of specific stimuli can also be a good indicator of cognitive deterioration.

In literature, several approaches for automatic emotion recognition are focused on the variety of human interaction capabilities or biological data. For example, the study of speech and other acoustic cues in~\cite{Weninger15}, body movements in~\cite{Chowdhuri16}, Electroencephalogram in~\cite{Lokannavar15,Valenzietal2014}, facial expressions or combinations of previous ones such as speech and facial expressions in~\cite{Nicolaou11} or \hyperlink{a_EEG}{EEG} and facial expressions in~\cite{Soleymani16}. Due to the fact that our purpose is the study of \hyperlink{a_AD}{AD} patients' emotions, and since their mobility or speech could be limited depending on their stage, \hyperlink{a_EEG}{EEG} and facial emotion detection would be more convenient. This chapter focuses on the automatic detection of emotions triggered by \hyperlink{a_AD}{AD} related stimuli for \hyperlink{a_AD}{AD} early detection using a novel \hyperlink{a_EEG}{EEG} approach. This approach aims to improve the emotion recognition accuracy using a reduced number of \hyperlink{a_EEG}{EEG} sensors so it could be easily integrated with \hyperlink{a_VR}{VR} technology. The evaluation of cognitive impairments on Virtual Environments have proved reliable for \hyperlink{a_AD}{AD} screening. Virtual environments have demonstrated they provide realistic scenarios to perform a wide variety of tasks \cite{Garcia2015,Parsons2011,Tarnanas2014,Vallejo2017,Serino2017,Berg2017}. Current Virtual Reality technologies allow a complete immersion but the \hyperlink{a_HCI}{HCI} could be improved by adding other technologies such as \hyperlink{a_EEG}{EEG} sensors~\cite{Frey2017}. \hyperlink{a_EEG}{EEG} signal can be processed for different purposes that will improve the virtual reality experience such as gaze tracking, objects selection or emotional feedback~\cite{Pike2016}. In addition, the information about the emotional state of the participant can also be used for cognitive impairments diagnosis.

In order to analyse \hyperlink{a_AD}{AD} patients' emotions impairments, it is required a dataset that contains recorded images or videos of people reacting to specific stimulus that would trigger different emotions to healthy and \hyperlink{a_AD}{AD} patients. Since it was not possible to find a database with this kind of information, a novel dataset was created for Alzheimer's Detection. This database contains recording of subjects while performing three tasks focused on different \hyperlink{a_AD}{AD} cognitive impairments. The main novelty of this database is the evaluation of subjects' autobiographical memory since it has been proved that for \hyperlink{a_AD}{AD} patients, semantic, autobiographical and implicit memory are more preserved than recent memory~\cite{Han14,Irish11,APA2013}. The dataset provides multimodal data such as \hyperlink{a_EEG}{EEG}, eye-gaze data and RGB, \hyperlink{a_IR}{IR} and Depth of participant's while watching stimuli on a monitor (see Figure~\ref{fig:dataClass}). Another novelty of the database is it contains recordings of the participants' reactions when specific images related and unrelated to their personal life are shown. Due to the difficulty obtaining data from \hyperlink{a_AD}{AD} patients our database only contains recordings from healthy participants. Since it is easier to obtain data from healthy subjects it is common there is always high imbalance between healthy and \hyperlink{a_AD}{AD} data. This imbalance problem is usually solved using a One Class Classification approach~\cite{Das2016,Lopezde2016}. Our case could be treated as a rigorous \hyperlink{a_OCC}{OCC} problem where only the positive set of examples (the healthy participants' data) is used as training data~\cite{Rodionova2016}. Therefore, using the healthy data it is possible the creation of healthy emotion models so any reaction detected out of the model is considered as a cognition impairment.

Our study analyses expected emotions~\cite{Soleymani16}, thus our classification is based on the expected emotions according to the images displayed during the test rather than on the emotions felt (represented on the captured video). This work investigates healthy people to analyse the differences and level of the emotional inputs generated from the available image classes. Three different approaches were presented to classify emotions. Firstly, \hyperlink{a_EEG}{EEG} based approach using only four sensors was presented and compared with state-of-the-art methods. Currently many approaches that process \hyperlink{a_EEG}{EEG} signal for emotion recognition or \hyperlink{a_BCI}{BCI} use headsets composed of fourteen, thirty-two or sixty-four electrodes/sensors. The proposed method is focused only on the four frontal sensors since our aim is the reduction of the amount of sensors in order to improve the usability of virtual reality for \hyperlink{a_AD}{AD} screening. At the same time, since literature shows facial emotions recognition techniques usually achieve better emotion detection accuracies, a facial emotion recognition approach based on facial landmarks to be compared with the \hyperlink{a_EEG}{EEG} one is presented. Finally, since facial approach probed better, a second version of the facial landmarks approach is presented using depth data. All these approaches are performed using advanced techniques for dimensionality reduction, i.e., \hyperlink{a_tSNE}{t-SNE}, \hyperlink{a_PCA}{PCA} as input to supervised learning approaches.

Furthermore, before the previous approaches are presented, an approach that uses the \hyperlink{a_EEG}{EEG} signals to detect a coarse gaze direction instead of eye trackers is introduced as a proof of concept. The need of cameras and light sources limits the applicability eye trackers on full immersive Augmented Reality (AR)/\hyperlink{a_VR}{VR} headsets. Therefore their combination with full immersive dementia diagnosis virtual environment systems is not practical due to the size and the amount of the acquisition devices required to detect eyes and gaze position. Due to this, this approach is based on a reduced number of \hyperlink{a_EEG}{EEG} sensors that could be easily added to the \hyperlink{a_VR}{VR} headset.

Eye movements are essential in order to gather information and move through the visual world. They are linked to personality and reflect cognitive processes and human emotional states. Eye-gaze tracking is the process of measuring either the point of gaze or the eye motion in relation to the head. Nowadays, there are many applications of eye and gaze tracking, including human-computer interaction, supporting disabled people, alerting drivers, diagnosing visual disorders, marketing and the understanding of human mental state. It can also facilitate the navigation and improve the quality of experience on Virtual and Augmented Reality applications significantly. These advantages applied on Virtual Reality medical examinations, such as \hyperlink{a_AD}{AD} screening, will help the patients to focus on the tasks reducing the possibility of errors due to a difficult \hyperlink{a_HCI}{HCI}. The coarse gaze direction estimation can improve the overall \hyperlink{a_HCI}{HCI} in \hyperlink{a_AR}{AR}/\hyperlink{a_VR}{VR} devices significantly allowing interaction with the eyes especially for item or menu interactions, making it easy for \hyperlink{a_AD}{AD} patient to interact on the Virtual Environments. A realistic and easy interaction helps to reduce a plausible bias on the results due to a difficult \hyperlink{a_HCI}{HCI} during the tasks.

This gaze tracking approach is also based on supervised learning techniques introducing novel feature descriptors based on a simplified quaternion representation, which is later used in the proposed \hyperlink{a_EEG}{EEG} emotion recognition approach. In the proposed descriptor, the data provided by the four frontal \hyperlink{a_EEG}{EEG} sensors are combined into a single quaternion, and used as input to machine learning techniques trying to identify the areas where the eyes are focused on. This would allow 2D and 3D navigation using the gaze position instead of controllers or head movements, providing the capability to select menu options or objects easily.

As such, within this chapter the following contributions are introduced: the novel spontaneous emotion multimodal database (SEMdb) based on autobiographical memories is propounded. Before focusing on the main application of the database, autobiographical emotion analysis for \hyperlink{a_AD}{AD} screening, a method for eye tracking detection using novel features optimised for improving \hyperlink{a_HCI}{HCI} on \hyperlink{a_AD}{AD} screening virtual reality systems is demonstrated. During this demonstration the novel Quaternion \hyperlink{a_PCA}{PCA} \hyperlink{a_EEG}{EEG} features are introduced for the first time. Afterwards, two emotion classification approaches using features from different data modalities are explained and their validity is probed against state-of-the-art methods. Firstly, the novel Quaternion \hyperlink{a_PCA}{PCA} \hyperlink{a_EEG}{EEG} and the RGB features are compared against state-of-the-art RGB and \hyperlink{a_EEG}{EEG} features. Finally, Depth and \hyperlink{a_EEG}{EEG} features whose dimensionality is reduced using \hyperlink{a_tSNE}{t-SNE} are compared against the state-of-the-art facial features. Both last contributions have proved that novel features provide better emotion classification performance than the correspondent state-of-the-art. These approaches focused only on healthy participants in order to generate a model that describes emotions from healthy participants, being able to identify cognitive impairments related with \hyperlink{a_AD}{AD} when the emotions are out of the model.


\section{Related Work}

Virtual Environments improves the performance and provides a wider range of tasks for the screening of Alzheimer's disease as demonstrated in Chapter~\ref{Chapter3}. \hyperlink{a_BCI}{BCI} interaction can contribute significantly in the performance of these Virtual Reality technologies when it comes to direct interaction (object selections, navigation) and indirect such as collecting users feedback to adapt to each individual. Pike et al.~\cite{Pike2016} endorse the use of Brain Computer Interaction combined with Virtual Reality. They support the use of \hyperlink{a_BCI}{BCI} technologies, such as \hyperlink{a_EEG}{EEG} sensors, to take measures that enable to have a good \hyperlink{a_VR}{VR} experience. Most of these are usually measured once the \hyperlink{a_VR}{VR} experience is finished, such as NASA TLX \cite{Hart1988} or Workload Profile \cite{Stanton2005}. According to Pike et al., \hyperlink{a_BCI}{BCI} will generate feedback continuously providing a more objective, automated, continuous and quantitative evaluation. They also suggest the use of a reduced number of sensors in the Pre-Frontal Cortex area (area of the brain located on the forehead) since activities such as working memory, emotions, decision-making or mental workload occur there. The measurement of previous activities and perception and attention are currently well supported by \hyperlink{a_BCI}{BCI} technologies. Pike et al. mention these measures in order to use them to design a good \hyperlink{a_VR}{VR} experience or to provide feedback on task based training. In addition, these measures could be useful in many areas such as gaming, where \hyperlink{a_BCI}{BCI} techniques could be used for interaction purposes or to adapt the game to participants mental state. The use of users' mental state such as boredom, anxiety or flow (game immersion) in combination with predictive analysis could be used to optimise players' gaming experience \cite{Mallapragada2018}. Moreover, as we have introduced in previous chapters, most of these measures can also be used to provide \hyperlink{a_AD}{AD} diagnosis.

The neurocognitive domains affected by \hyperlink{a_AD}{AD} according to~\cite{APA2013} include attention, perception, memory and emotions impairments. The emotions impairment is related with the \hyperlink{a_AD}{AD} patient's capability to identify emotions on other people. We propose the analysis of memory impairments through emotions. There is not research applied to the study of emotions elicited by recent and distant autobiographical memories, therefore, \hyperlink{a_EEG}{EEG} and facial techniques used for basic emotion recognition are used. Amongst the wide research techniques we selected two approaches, one for each type of data. Michel et al. in~\cite{Michel03} uses CK \ref{db_CK} facial expression database to classify six basic emotions. They use a facial point tracker to get twenty-two fiducial points from the face images and calculate the distance of each point between a neutral face and a peak expression frame of each emotion. These distances are used as features of a SVM algorithm in order to classify the emotions. Neutral and peak frames are automatically detected from each sequence when the motion of the points is almost zero. This approach obtains good classification results but it is tested on a posed database.~\cite{Soleymani16} uses a combined \hyperlink{a_EEG}{EEG} and facial approach using multimodal data of spontaneous emotions provided by MAHNOB-HCI database \ref{db_MAH}. Their work combines \hyperlink{a_EEG}{EEG} data and facial RGB data to classify basic emotions. The \hyperlink{a_EEG}{EEG} signal was captured using thirty-two sensors and the power spectral density was extracted from overlapping one second windows. The facial approach extracts forty-nine fiducial points and calculates the distance from thirty-eight of these points to a reference point. Finally, they use regression models for emotion detection. As a result, they have obtained better results using the facial data and conclude that the good performance of the \hyperlink{a_EEG}{EEG} results are due to the facial artifacts present in the \hyperlink{a_EEG}{EEG} signal. Other methodologies and databases are analysed in Chapter~\ref{Chapter2}.

Apart from feedback about user's mental state, \hyperlink{a_BCI}{BCI} is also used for navigating and selection purposes on Virtual Environments. Kerous et al.~\cite{Kerous2016} summarise the techniques used to navigate and select objects in a Virtual Environment through \hyperlink{a_BCI}{BCI} techniques. They separate these techniques in two: the ones based on evoked potentials and the oscillatory \hyperlink{a_EEG}{EEG} activity ones. Amongst the evoked potentials they focus on two. Visual-Evoked Potential is observed in \hyperlink{a_EEG}{EEG} signal when a visual stimulus flashing at specific frequency is observed. P300 is a potential that appears 300ms after a stimulus different from others is detected. Oscillatory \hyperlink{a_EEG}{EEG} activity study all the brain activity, usually analysing the different wave bands. In order to find objects gaze tracking is more realistic, Chun et al.~\cite{Chun2016} propose a combination of \hyperlink{a_BCI}{BCI} and Eye tracking technologies for object control in \hyperlink{a_VE}{VE}. In their work, eye movements are used as cursor in the \hyperlink{a_VE}{VE} and \hyperlink{a_BCI}{BCI} is used for selection. They train two classes using the \hyperlink{a_EEG}{EEG} signal: concentration or not-concentration. A custom eye tracker device is attached to the \hyperlink{a_VR}{VR} glasses and the \hyperlink{a_BCI}{BCI} headset is worn independently. Since \hyperlink{a_BCI}{BCI} technologies also have provided promising results on gaze tracking, the eye tracking device could be removed. Tomi et al.~\cite{Tomi2017} propose a system for eye position estimation using \hyperlink{a_EEG}{EEG} signal. Their design is divided in two sections: calibration and \hyperlink{a_EEG}{EEG} signal processing. The calibration stage describes the new coordinates system, translating the eye movement inside the \hyperlink{a_VR}{VR} glasses into the \hyperlink{a_VR}{VR} world. The \hyperlink{a_EEG}{EEG} signal processing stage measures the eye position in relation to the visual field using evoked potentials. During this stage they filter the signal to reduce noise and use Independent Component Analysis to separate artifacts. Afterwards, they extract spectral power and amplitude related features for classification on a \hyperlink{a_kNN}{kNN} classifier. Eye tracking not only can help with navigation on \hyperlink{a_VE}{VE}; \hyperlink{a_AD}{AD} can be screened by analysing visual related impairments. Lutz et al.~\cite{Lutz2017} defend the use of eye tracking in \hyperlink{a_VR}{VR} therapy, especially the use of it for real time attention assessment.


\section{Methodology}

\subsection{Spontaneous Emotions Multimodal Database (SEMdb)}
\hyperlink{a_SEMdb}{SEMdb} is a multimodal dataset for spontaneous emotional reaction recognition that contains multimodal information of nine healthy participants: five females and four males between 27 to 60 years old with different educational backgrounds taken while completing cognitive/visual tests \cite{MontenegroGaze2016, MontenegroFace2016, MontenegroEmo2016}. The general purpose of this dataset is \hyperlink{a_AD}{AD} screening. Due to the difficulty in obtaining data from \hyperlink{a_AD}{AD} patients, this dataset contains only healthy information that can be used to create single models using One Class Classification techniques. These techniques will classify any outlier from the healthy model as a possible cognitive impairment helpful for \hyperlink{a_AD}{AD} diagnosis. Khan et al.~\cite{Khan2014} supported the use of \hyperlink{a_OCC}{OCC} for disease diagnosis since low false positive are required and techniques such as One Class Support Vector Machine usually provide high true positive rates.

The participants were asked to complete three tasks: gaze calibration, numeral search and autobiographical tasks.

\begin{itemize}
\item The Gaze Calibration Task requires participants to fix their gaze to specific areas in a monitor during a specific amount of time \cite{MontenegroGaze2016}. A white rectangle on a black background is shown in nine different positions during five seconds each (see Figure~\ref{fig:sectionsSEMdb}). The outcome of this task can be utilised for gaze tracking application. Our main purpose with this test is to have gaze guided data in order to correlate the \hyperlink{a_EEG}{EEG} signal with the Eye tracker information. \hyperlink{a_BCI}{BCI} for navigation proved to be very adequate on virtual environments~\cite{Kerous2016}. \hyperlink{a_BCI}{BCI} simplifies the combination of devices required for eye tracking and eye tracking reduces the interaction complexity with Virtual Environments, making it more realistic. In addition, other important applications of eye tracking can be applied for \hyperlink{a_AD}{AD} diagnosis such as the evaluation of \hyperlink{a_AD}{AD} patients sustained attention, that is, the maintenance of attention over time.

\begin{figure}
\centering
\includegraphics[scale=1]{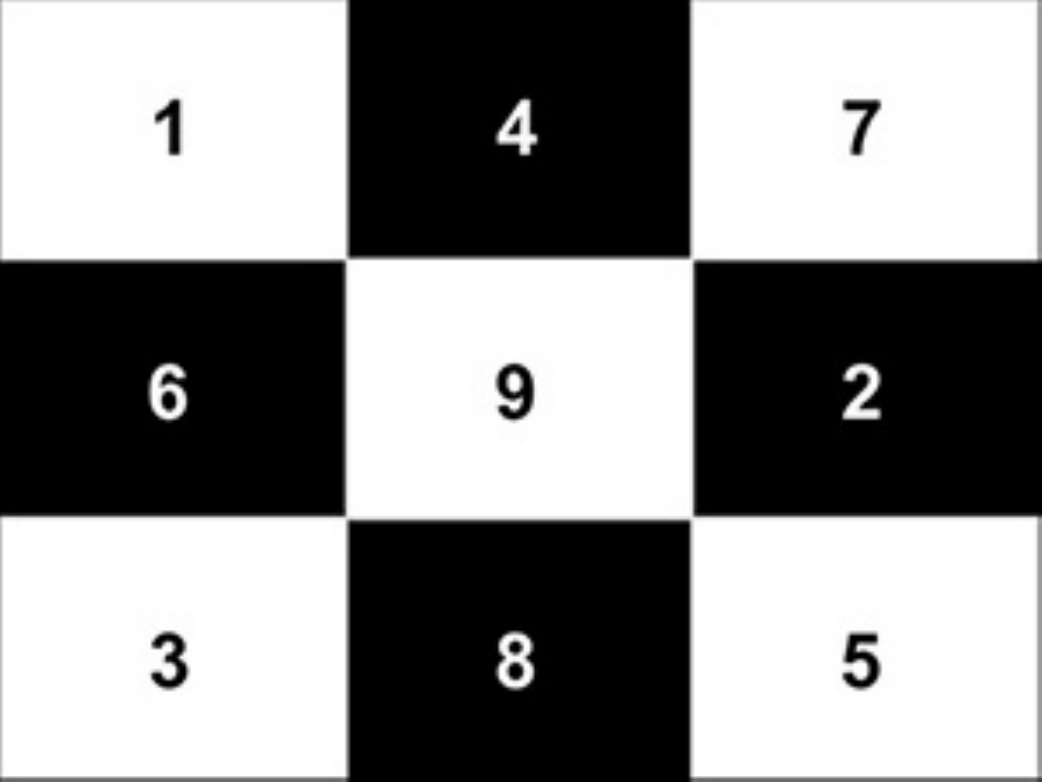}
\caption{
Each white area is shown in these nine positions with the other areas in black
}
\label{fig:sectionsSEMdb}
\end{figure}

\item The Number Search Test is based on an existent dementia detection test that relies on the fact that the number and duration of fixations of dementia patients is higher than healthy participants' during search tasks~\cite{Rosler2000,Pereiraetal2014}. During this test, an image with letters randomly displayed on the screen and one number amongst the letters are shown; the participant has to locate the number and press a key when they find it to proceed to the following search, up to a total of four images (see Figure~\ref{fig:searchSEMdb}). The number of fixations and their duration and the time spent to find the number are recorded by the eye tracker. Additionally, apart from the location of the gaze on the screen, it also records a flag value that indicates when the user's gaze goes over the number area.

\begin{figure}[!t]
   \centering
    \begin{subfigure}[b]{0.34\textwidth}
        \includegraphics[scale=0.1]{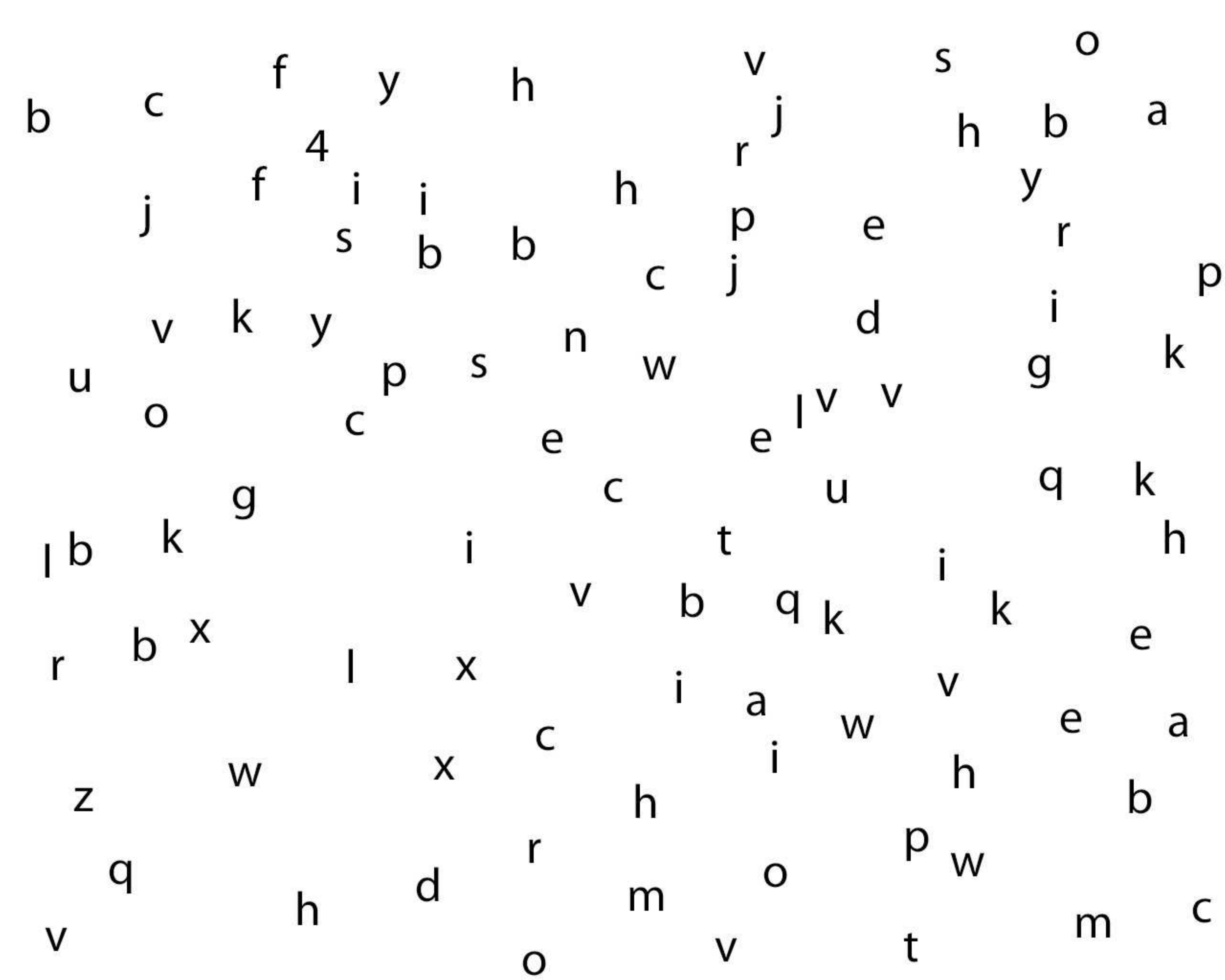}
    \caption{}
    \end{subfigure}%
    \begin{subfigure}[b]{0.34\textwidth}
        \includegraphics[scale=0.1]{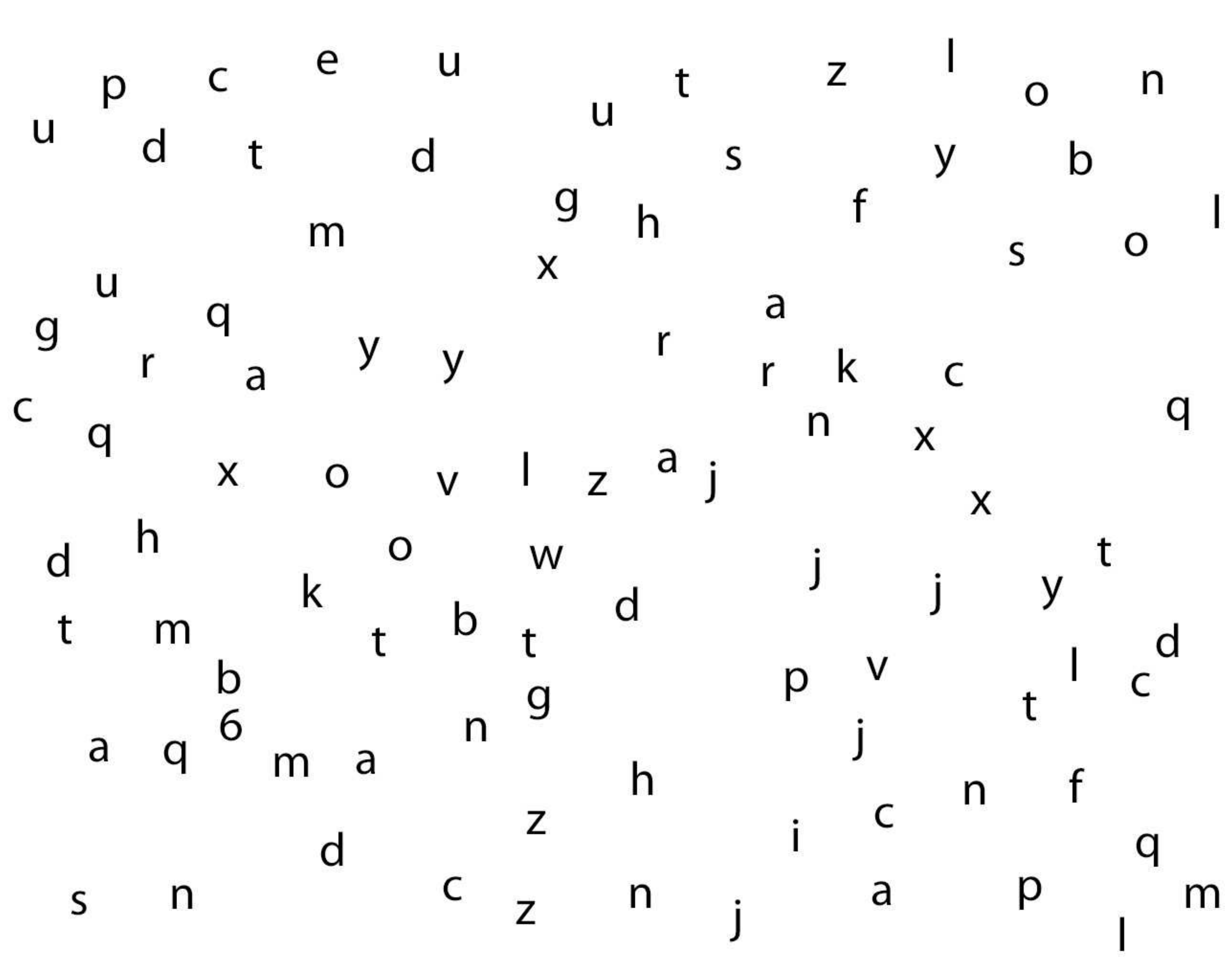}
    \caption{}
    \end{subfigure} \\%
    \begin{subfigure}[b]{0.34\textwidth}
        \includegraphics[scale=0.1]{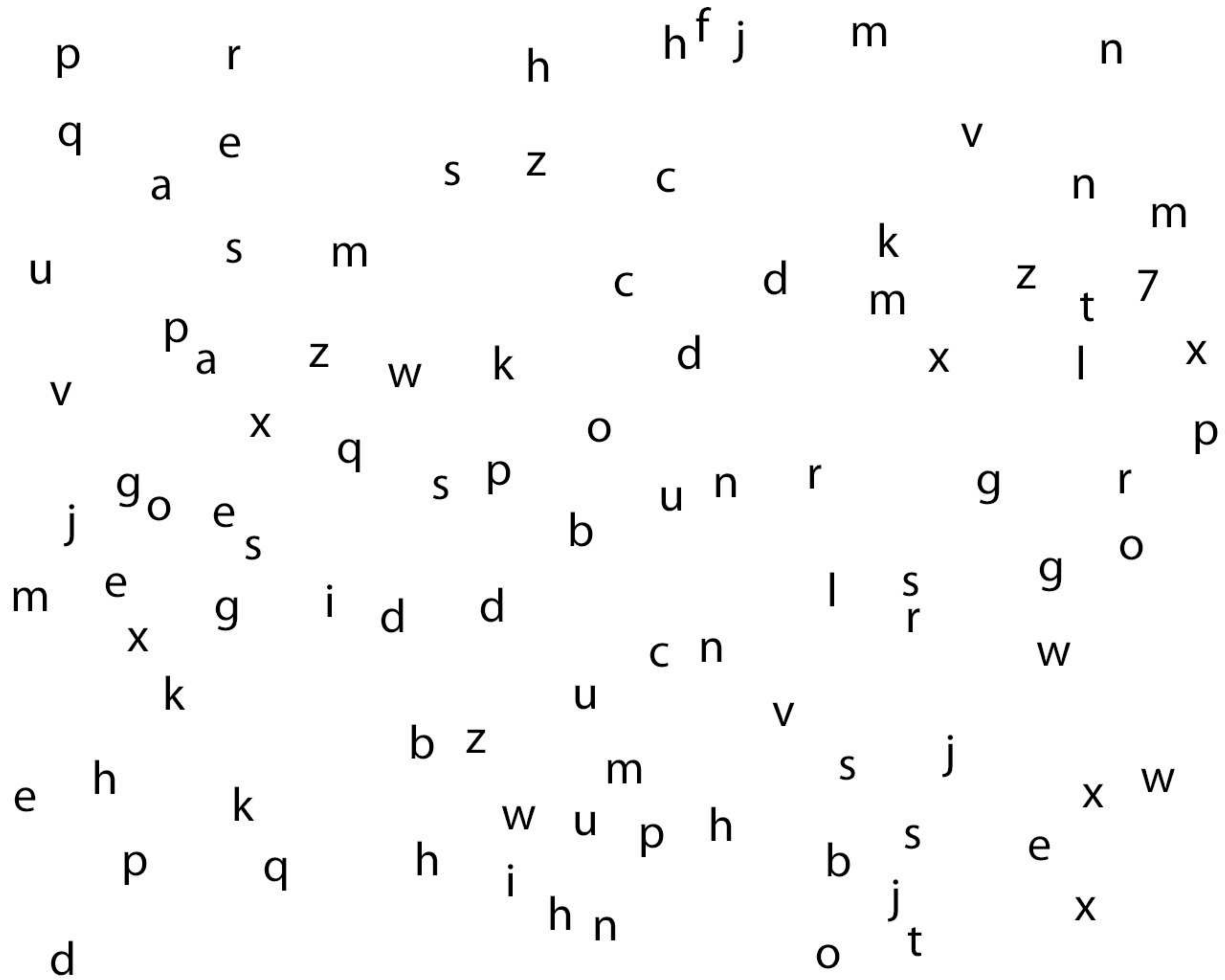}
    \caption{}
    \end{subfigure}%
    \begin{subfigure}[b]{0.34\textwidth}
        \includegraphics[scale=0.1]{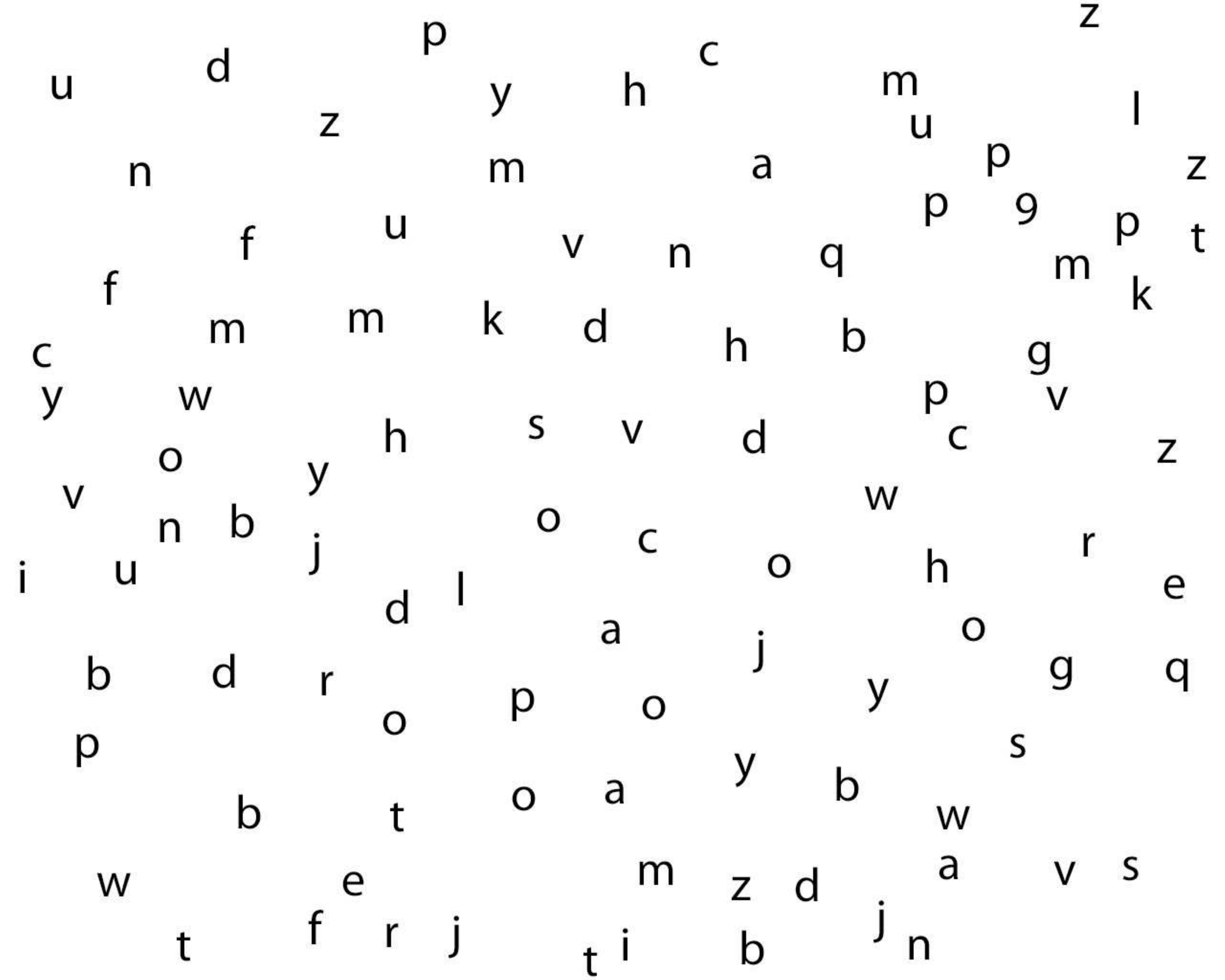}
    \caption{}
    \end{subfigure}%
    \caption{Images shown during the numeral search task. Each image is shown individually, proceeding to the next one when the number is found.}
  \label{fig:searchSEMdb}
\end{figure}

\item \label{autobioEmo} The novelty of \hyperlink{a_SEMdb}{SEMdb} is enclosed in the third task (autobiographical) \cite{MontenegroFace2016, MontenegroEmo2016}. It provides non-posed reactions to autobiographical and non-autobiographical visual stimuli data. The main contribution of \hyperlink{a_SEMdb}{SEMdb} is the use of personalised images for each participant. These images are photos of themselves or their relatives and friends both from the recent and distant past. Moreover, the participants did not know that those images were used so the reactions were genuine. Additionally, images of famous and unknown to the participants' people and places were shown. These stimuli have been added based on the \hyperlink{a_AD}{AD} patients' decline of two main cognitive domains: learning and memory and perceptual motor~\cite{APA2013}. One of the symptoms produced by the learning and memory cognitive domain impairment is the difficulty recalling recent events whereas the distant ones are preserved~\cite{Kirk2017}. \hyperlink{a_AD}{AD} patients suffer from amnesia that affects recent acquired memories. This decline reduces the access to memories that defines their identity \cite{ElHaj2015}. The purpose of the autobiographical reactions is the detection of this lost of recent personal information by comparing them with remote ones. In addition, the data obtained from the non-autobiographical reactions is focused on the lost of well-know general information, such as the recognition of famous people or places, by proving their reactions are the same when new information (unknown information) is provided. Moreover, the perceptual motor cognitive domain impairment produces face recognition difficulties~\cite{Lavallee2016}. Finally, a last series of images that present abnormalities were also shown. Half of this series presents images with optical effects and the other half are formed by autobiographical images that have been modified by adding unreal characters. This last series of stimuli is related with \hyperlink{a_AOR}{AOR} test described in Section \ref{AORsection}. \hyperlink{a_AOR}{AOR} test evaluated the participants' capacity to detect abnormalities. This series of stimuli will provide data to evaluate this capacity and also their reaction to those abnormalities. In more detail we had the following classes of images with the corresponding expected spontaneous emotions or reactions (see Figure~\ref{fig:SEMdbClassIma}). Since ten repetitions have been recorded for five seconds each per participant a total of ninety instances per class are provided, obtaining a 7.5 minutes video.

\begin{enumerate}[label=\alph*)]
\item10 images of distant past faces of the subjects and their relatives.
\item10 images of recent past faces of the subjects and their relatives.
\item10 images of distant past group of relatives, including themselves.
\item10 images of recent past group of relatives, including themselves.
\item10 images of famous people.
\item10 images of unknown to the subject persons.
\item10 images of famous places/objects.
\item10 images of unknown to the subject places/objects.
\item10 images of abnormalities: 5 optical effects and 5 modified autobiographical images.
\end{enumerate}

\end{itemize}

\begin{figure}[!t]
   \centering
    \begin{subfigure}[b]{0.24\textwidth}
        \includegraphics[scale=0.5]{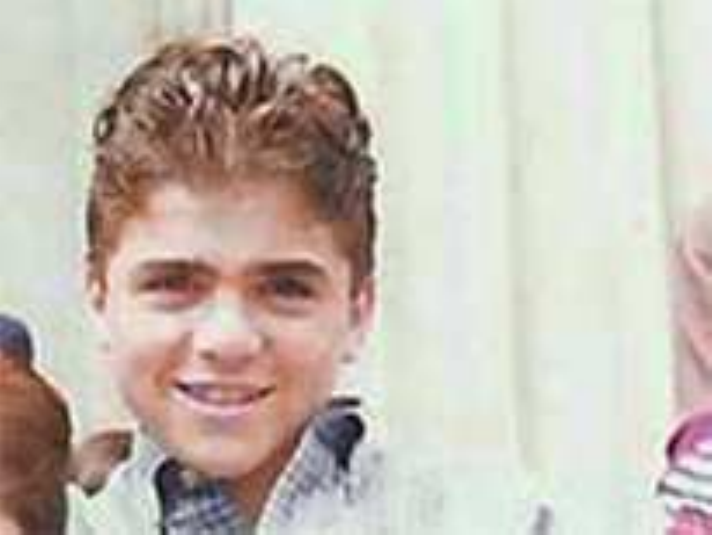}

    \end{subfigure}%
    \begin{subfigure}[b]{0.24\textwidth}
        \includegraphics[scale=0.5]{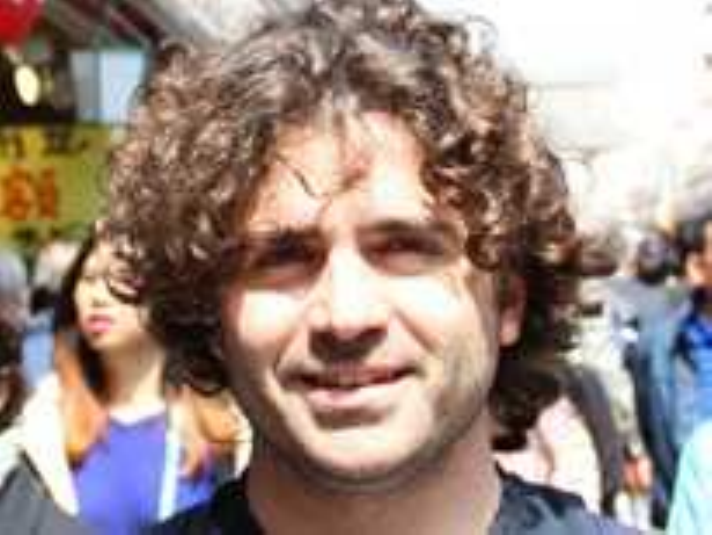}

    \end{subfigure}%
    \begin{subfigure}[b]{0.24\textwidth}
        \includegraphics[scale=0.5]{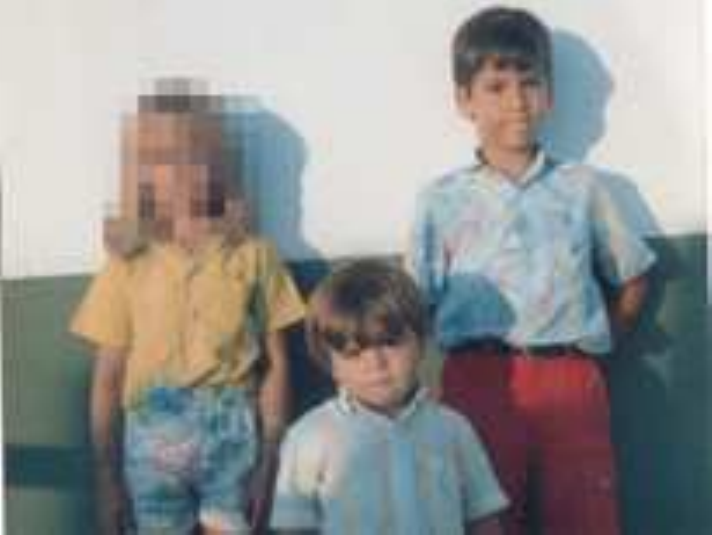}

    \end{subfigure}%
    \begin{subfigure}[b]{0.24\textwidth}
        \includegraphics[scale=0.5]{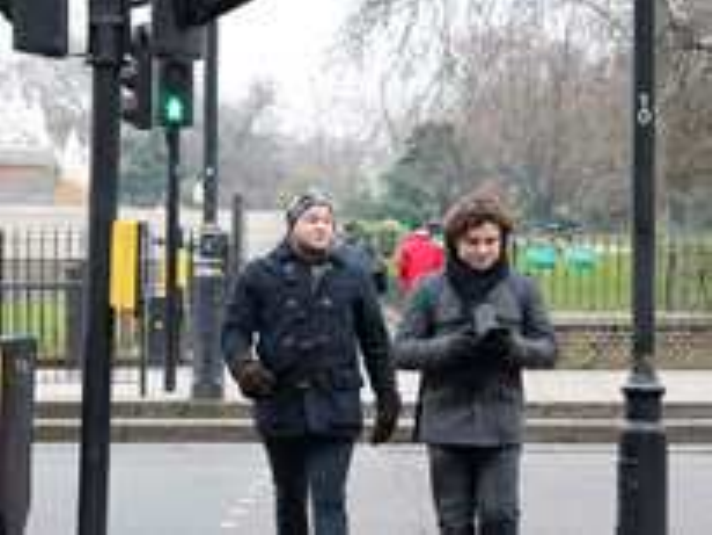}

    \end{subfigure} \\%
    \begin{subfigure}[b]{0.24\textwidth}
        \includegraphics[scale=0.5]{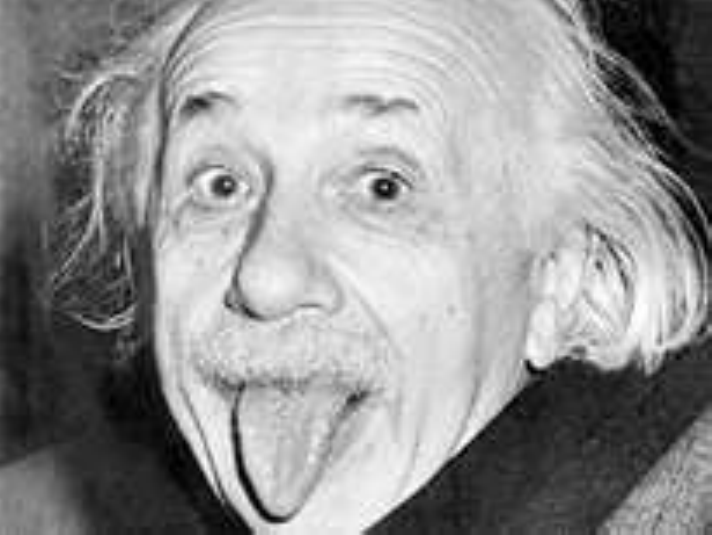}

    \end{subfigure}%
    \begin{subfigure}[b]{0.24\textwidth}
        \includegraphics[scale=0.5]{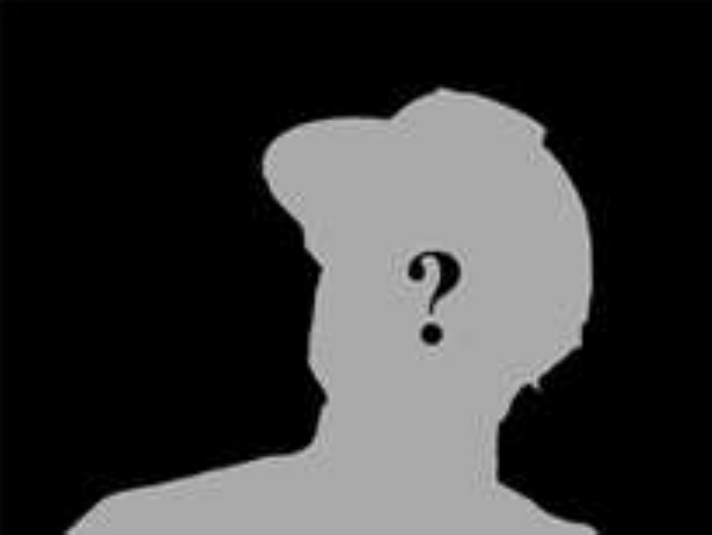}

    \end{subfigure}%
    \begin{subfigure}[b]{0.24\textwidth}
        \includegraphics[scale=0.5]{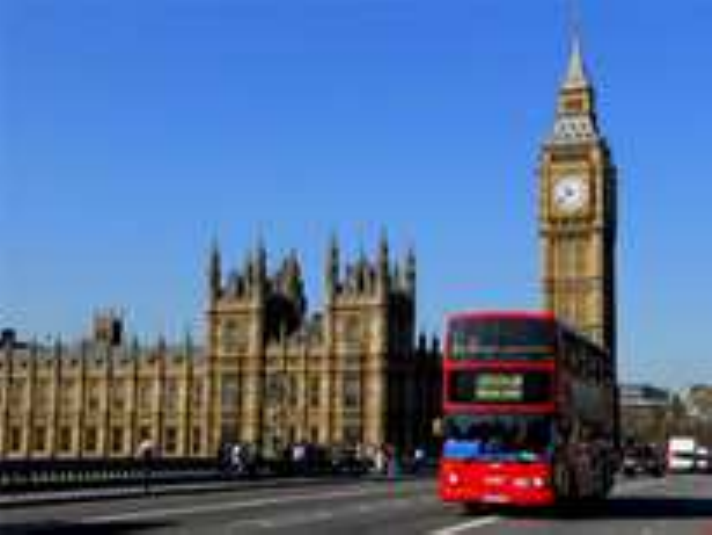}

    \end{subfigure}%
    \begin{subfigure}[b]{0.24\textwidth}
        \includegraphics[scale=0.5]{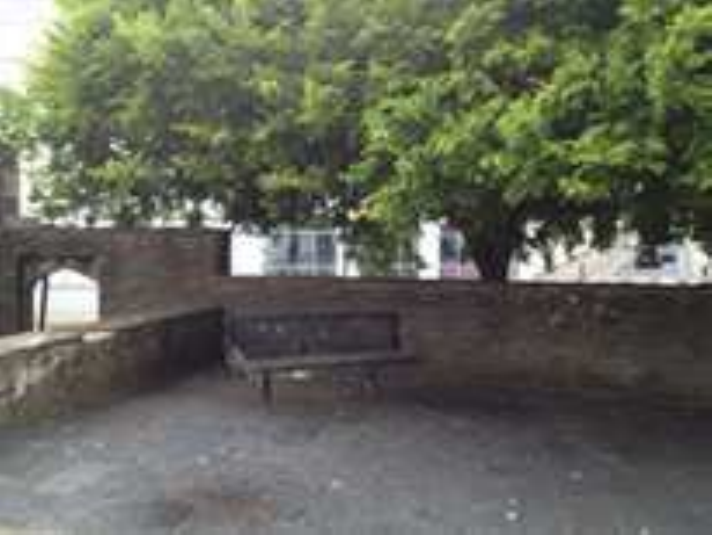}

    \end{subfigure} \\%
    \begin{subfigure}[b]{0.24\textwidth}
        \includegraphics[scale=0.5]{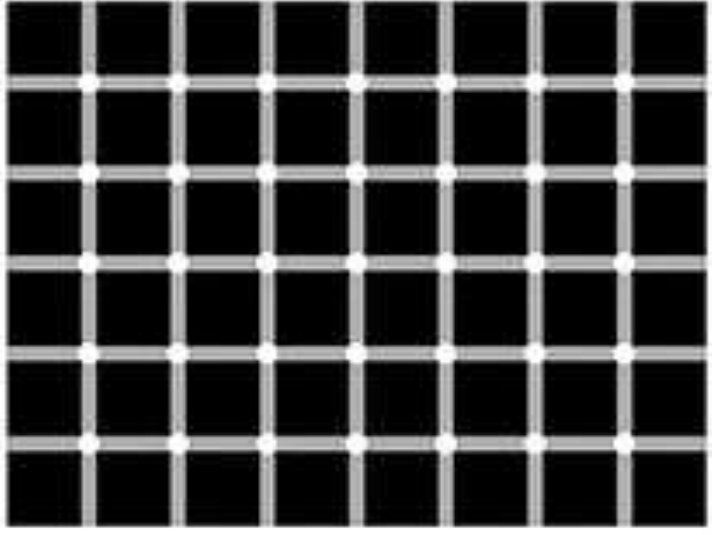}

    \end{subfigure}%
    \begin{subfigure}[b]{0.24\textwidth}
        \includegraphics[scale=0.5]{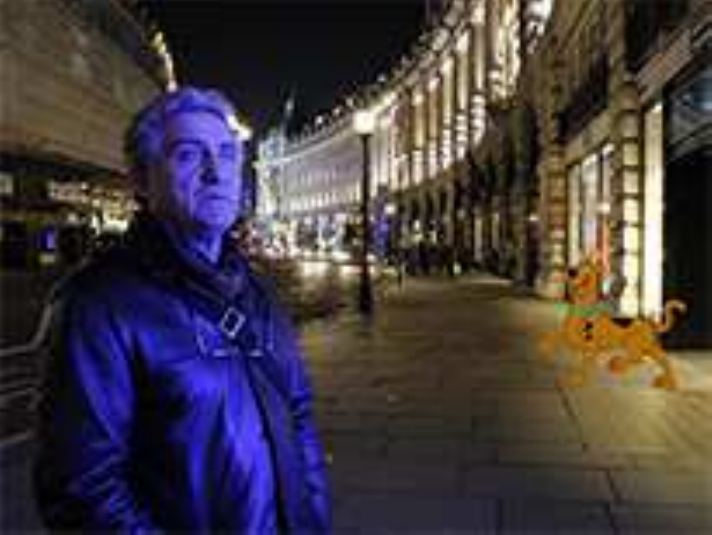}

    \end{subfigure}%
   \caption{Example of each class of images displayed to the subjects during the recording of the database.}
  \label{fig:SEMdbClassIma}
\end{figure}

\begin{figure}
\centering
\includegraphics[width=130mm]{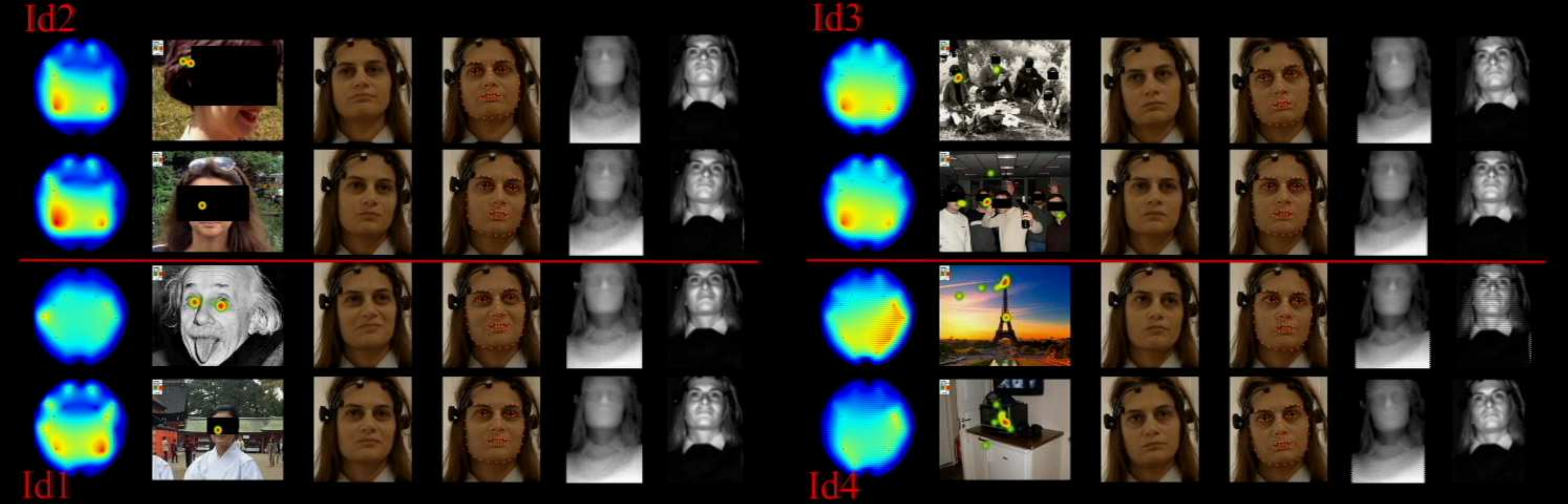}
\caption{
Data modalities contained in the database and the related classes analysed in our approach (see Table~\ref{table:classes} for the emotion definitions). The left figure shows, from top to bottom, images of people from distant vs recent past; and famous vs unknown people. The right figure shows from top to bottom images of groups of people from distant vs recent past; and famous vs unknown places. The different modalities from left to right in each case are EEG, gaze tracked heat map, RGB, facial landmarks, depth and IR.
}
\label{fig:dataClass}
\end{figure}

\begin{figure}
\centering
\includegraphics[width=100mm]{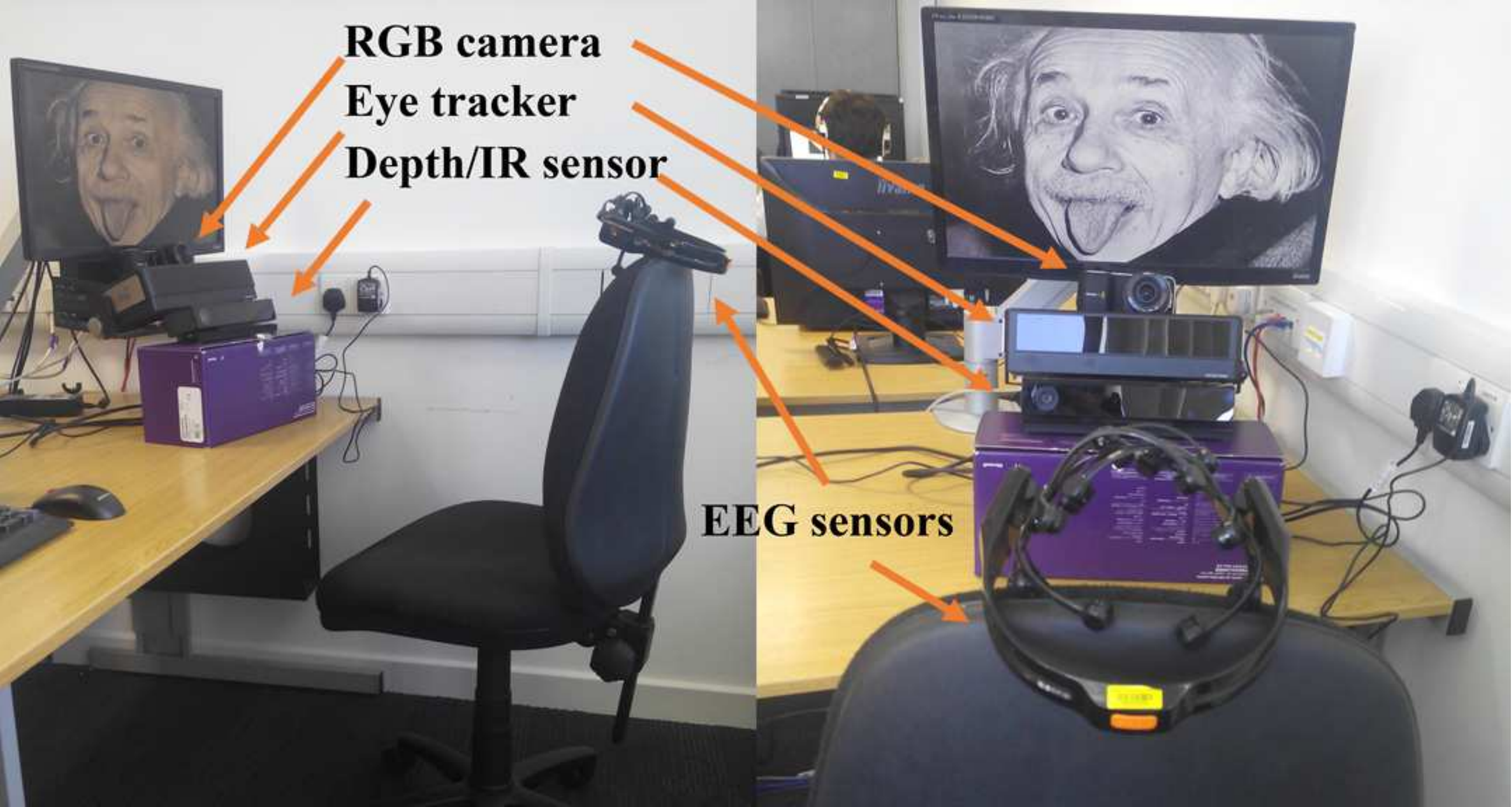}
\caption{
Location of the devices during the recording of the database.
}
\label{fig:office}
\end{figure}

The recorded data is provided in different data modalities: HD RGB, depth and \hyperlink{a_IR}{IR} frames of the face, \hyperlink{a_EEG}{EEG} signal and eye gaze data; which were recorded using four different devices: a 30fps HD RGB camera, \hyperlink{a_IR}{IR}/Depth sensors (Kinect), an eye tracker (Tobii eye tracker) and \hyperlink{a_EEG}{EEG} sensors (Emotiv headset)(see Figure~\ref{fig:dataClass}). The recording of the data has been done in a controlled environment e.g. an office. Before we proceeded with the tests, the participants were requested to read and sign the ethics form (see Appendix~\ref{AppendixB}). The participants were asked to put on the \hyperlink{a_EEG}{EEG} headset and they were seated in a comfortable chair in front of the test screen, the RGB camera, the Kinect sensor and the eye tracker. The height of the chair was adjusted so that the eye tracker would detect their eye movements (see Figure~\ref{fig:office}). Once the eye tracker is detecting the participants' eyes and all the \hyperlink{a_EEG}{EEG} sensors are receiving good quality signal the test begins. The instructions of the tests are provided verbally at the beginning of each test. Furthermore, each test starts showing a red image for synchronization.

\subsubsection{Data Collection}

All the data (HD RGB, depth and \hyperlink{a_IR}{IR} frames of the face, \hyperlink{a_EEG}{EEG} signal and eye gaze data) from the different devices was collected simultaneously using time-stamps for synchronization. During the acquisition process, the subjects were looking at a monitor in front of them, while wearing the \hyperlink{a_EEG}{EEG} headset. The RGB camera, the eye tracker and the depth/\hyperlink{a_IR}{IR} sensor were located under the monitor. \hyperlink{a_EEG}{EEG} raw data was collected from \hyperlink{a_EEG}{EEG} sensors/electrodes placed on the scalp of the participants. The \hyperlink{a_EEG}{EEG} device returns the amplified signal of the fourteen sensors/channels, a value that indicates the quality of the signal, the gyroscope measurements, the battery level and the time when the capturing process started. The HD RGB camera was adjusted to focus on participants faces, recording 1080p resolution videos at 30fps. The eye-tracker provides information that includes the coordinates of the eyes, fixations, saccades, areas of interest and time-stamps. The information about the location of the eyes is provided both in pixel and millimetre coordinates. The areas of interest where created as indicated in Table~\ref{table:ROIs}. The eye tracker was calibrated for each participant before starting the test. The depth and infrared images where recorded using Microsoft Kinect 2. Both these images have a $640x480$ resolution.

\setlength{\tabcolsep}{4pt}
\begin{table}
\begin{center}
\caption{
Regions of Interest (ROIs) annotated during the recording of the tasks. Some tasks included more than one type of ROI. For example, when the faces are shown during the autobiographical task, two ROIs with different labels are used: participant's face and face (not participant's face).
}
\label{table:ROIs}
\centering
\begin{tabular}{l|l|l}
\hline
\noalign{\smallskip}
 Task &  Images &  \hyperlink{a_ROI}{ROIs}\\
\noalign{\smallskip}
\hline
\noalign{\smallskip}
\multirow{2}{*}{\centering Gaze calibration} & \multirow{2}{0.5\linewidth}{9 images with a white rectangles located in different locations} & White rectangle\\
&&\\
\noalign{\smallskip}
\hline
\noalign{\smallskip}
\multirow{2}{*}{\centering Numeral search}  & \multirow{2}{0.5\linewidth}{4 images with one number mixed around letters} & Number\\
&&\\
\noalign{\smallskip}
\hline
\noalign{\smallskip}
\multirow{13}{*}{\centering Autobiographical} & \multirow{2}{0.5\linewidth}{10 images of distant past faces} & Face \\
                 &                                & Participant's face \\
                 \noalign{\smallskip}
                 \cline{2-3}
                 \noalign{\smallskip}
                 & \multirow{2}{0.5\linewidth}{10 images of recent past faces} & Face \\
                 &                                & Participant's face \\
                 \noalign{\smallskip}
                 \cline{2-3}
                 \noalign{\smallskip}
                 & \multirow{1}{0.5\linewidth}{10 images of distant past group of people} & Participant's face \\
                 \noalign{\smallskip}
                 \cline{2-3}
                 \noalign{\smallskip}
                 & \multirow{1}{0.5\linewidth}{10 images of recent past group of people} & Participant's face \\
                 \noalign{\smallskip}
                 \cline{2-3}
                 \noalign{\smallskip}
                 & \multirow{1}{0.5\linewidth}{10 images of famous faces} & Face \\
                 \noalign{\smallskip}
                 \cline{2-3}
                 \noalign{\smallskip}
                 & \multirow{1}{0.5\linewidth}{10 images of unknown faces} & Face \\
                 \noalign{\smallskip}
                 \cline{2-3}
                 \noalign{\smallskip}
                 & \multirow{1}{0.5\linewidth}{10 images of famous places} & None \\
                 \noalign{\smallskip}
                 \cline{2-3}
                 \noalign{\smallskip}
                 & \multirow{1}{0.5\linewidth}{10 images of unknown places} & None \\
                 \noalign{\smallskip}
                 \cline{2-3}
                 \noalign{\smallskip}
                 & \multirow{1}{0.5\linewidth}{5 images of optical effects} & None \\
                 \noalign{\smallskip}
                 \cline{2-3}
                 \noalign{\smallskip}
                 & \multirow{2}{0.5\linewidth}{5 images of modified autobiographical images} & Abnormality \\
                 &                                              & Participant's face \\
\noalign{\smallskip}
\hline
\end{tabular}
\end{center}
\end{table}
\setlength{\tabcolsep}{1.4pt}

\subsection{Gaze Estimation using EEG Signals for HCI in Augmented and Virtual Reality Headsets}

\begin{figure}[!t]
  \centering
  \includegraphics[width=0.9\columnwidth]{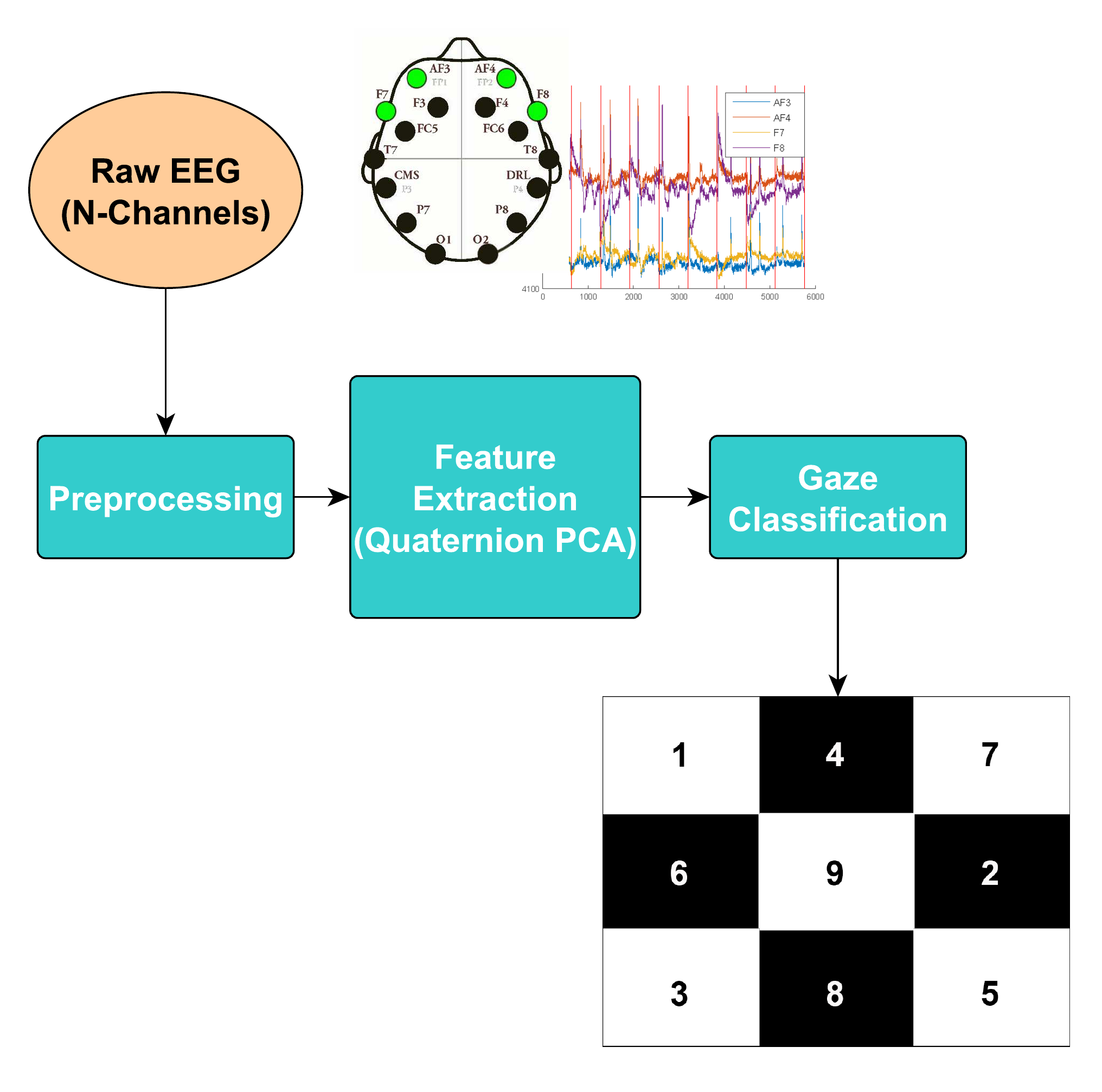}
  \caption{The proposed approach. The raw data is preprocessed, the quaternion PCA features are extracted and feeded to a classifier which will output the correspondent classes labels.
  }\vspace{-4mm} \label{fig:DiagramEye}\vspace{-3mm}
\end{figure}

The proposed method for gaze estimation is based on a supervised learning framework that incorporates a training and a testing stage, (see Figure~\ref{fig:DiagramEye}) \cite{MontenegroGaze2016}. This section analyses the suggested methodology, and provides details for the data acquisition process. Also, all the pre-processing steps are analysed and the proposed novel quaternion based descriptor is presented.

\subsubsection{Data Acquisition}
Initially, raw data was collected from \hyperlink{a_EEG}{EEG} sensors placed on the scalp of the participants. During the acquisition process, the subjects were looking at different locations on a monitor in front of them. Regarding the setup, the subjects were seated in front of an eye tracker and a PC monitor, while wearing the \hyperlink{a_EEG}{EEG} headset. Amongst the data provided by the \hyperlink{a_EEG}{EEG} device, only the amplified signal of the fourteen sensors/channels and the time when the capturing process started were required. When it comes to the eye-tracker information just the coordinates of the eyes, the areas of interest and the time-stamps associated to each image displayed were utilised to create the ground truth.

\subsubsection{Pre-processing}
Since the data is available, the obtained values are normalized for each channel in order to reduce the signal differences between the subjects. The average value is subtracted and the result is divided by their standard deviation. In addition, a median filter is used to reduce the noise produced by the electronic amplifier, the power line interference and any other external interferences.

\subsubsection{Feature Extraction for Gaze Classification - Quaternion Principal Component Analysis} \label{QuaternionPCA}
\begin{figure}[!t]
  \centering
  \includegraphics[scale=.8]{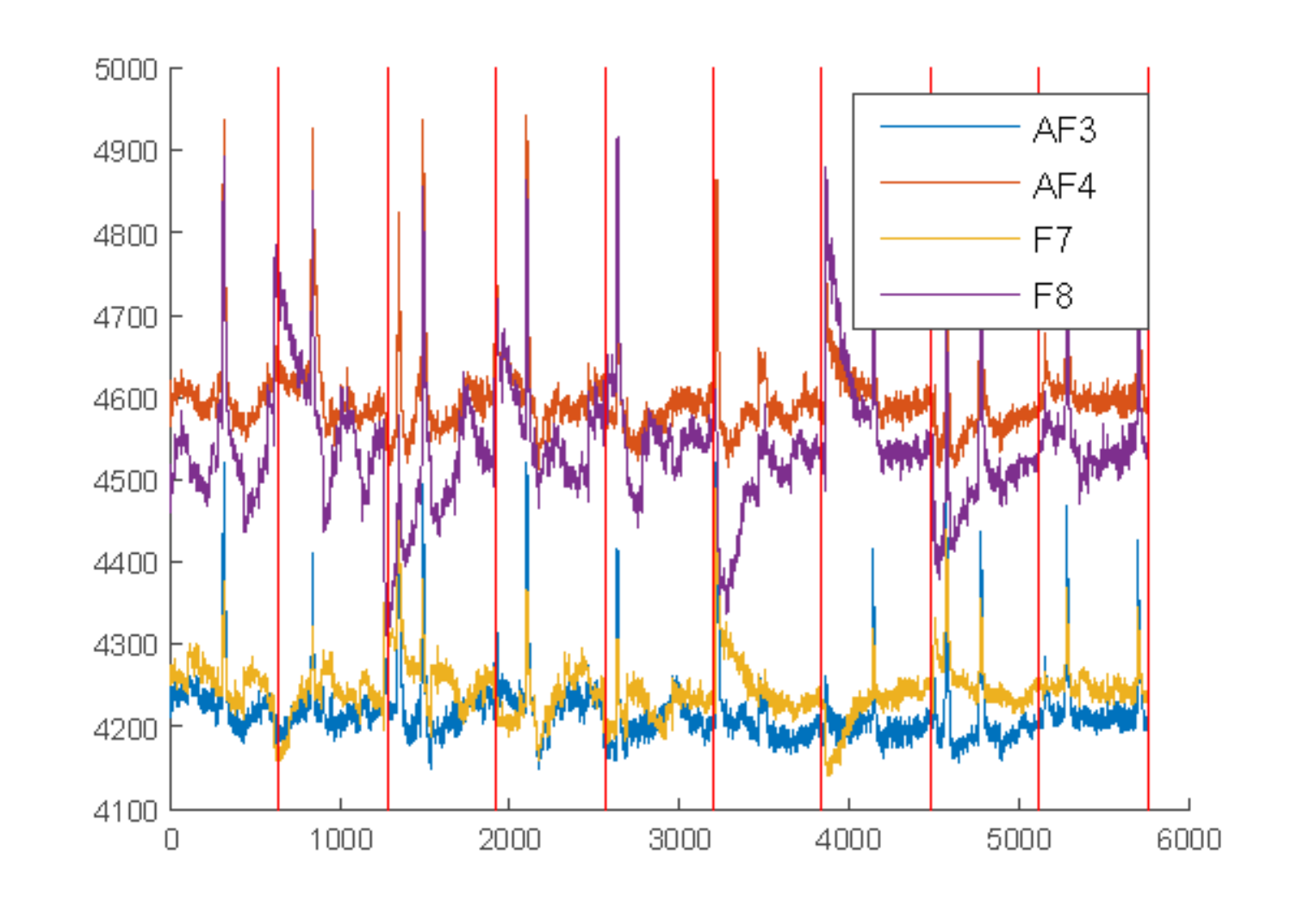}
  \caption{Raw signal example of 4 of the 14 channels (AF3, AF4, F7 and F8). The y axis represents the voltage ($\mu V$) and the x axis the number of samples. Every red line denotes 5 seconds} \label{fig:rawEEGsignal}\vspace{-4mm}
\end{figure}

The training stage is oriented towards the extraction of features from the preprocessed data of the \hyperlink{a_EEG}{EEG} channels (see Figure \ref{fig:rawEEGsignal}) to create a classification model. The selected \hyperlink{a_EEG}{EEG} channels are combined into a novel quaternion representation and \hyperlink{a_PCA}{PCA} is used to reduce the dimensionality of the obtained feature vector \cite{MontenegroGaze2016, MontenegroEmo2016}.

It is true that a vector can be decomposed in linearly independent components, in the sense that they can be combined linearly to reconstruct the original vector. However, depending on the phenomenon that changes the vector, correlation between the components may exist from the statistical point of view. If they are independent our proposed descriptor does not provide any significant advantage, but if there is correlation this is considered. In most of the cases during the feature extraction process complex or hyper-complex features are generated but decomposed to be computed by a classifier. For example, normals and gradients in 2D/3D are features consisting of more than one element and this decomposition can imply a loss of information.

To do so, vectorial features can be represented more precisely using a complex or hyper-complex representation \cite{Adali2011,Li2011}. Since, in our case and many similar scenarios, vectorial features such as a location, speed, gradients or angles, are the primary source of information, a hyper-complex representation of these features is more efficient allowing better correlation between these channels \cite{Adali2011,Li2011,Chai2009}. The proposed method exploits the hyper-complex (quaternion) representation capturing the dependencies within the \hyperlink{a_EEG}{EEG} sensors located on the sides of the head and the ones over the eyes, \cite{Li2009,Bonita2014}.\\

\begin{figure}[!t]
  \centering
  \includegraphics[width=.3\columnwidth]{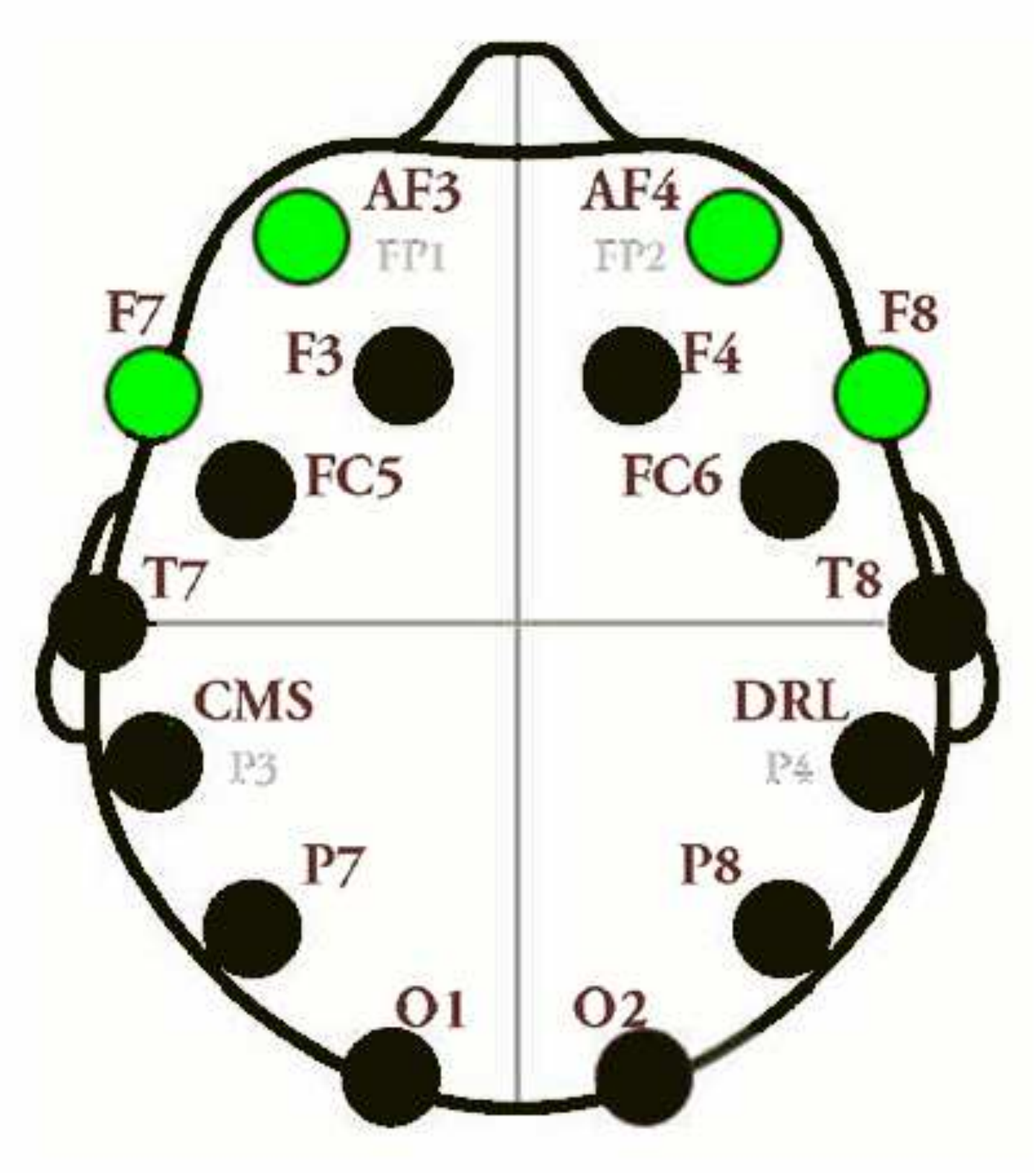}
  \vspace{-3mm}
  \caption{Only 4 of the 14 channels (AF3, AF4, F7 and F8) are used and combined in a single quaternion representation.} \label{fig:emotivsensors}
\end{figure}
\vspace{-4mm}

In order to reduce the number of the selected hyper-complex features without increasing the complexity, quaternion \hyperlink{a_PCA}{PCA} is applied. In more detail, the quaternion representation was introduced in \cite{Bihan2003,Chen2015} as a generalization of the complex numbers. A quaternion $q \in \mathcal{H}$ has four components:
\begin{equation}
q = q_r + q_ii +q_jj + q_k k
\end{equation}
where $q_r,q_i,q_j,q_k \in \Re$  and $i,j,$ and $k$ satisfy
\begin{equation}
\begin{array}{rl}
i^2& = j^2 = k^2=-1,\,\, ij = -ji = k\\
jk & = -kj = i, ki = -ik = j
\end{array}
\end{equation}
Conjugation of quaternions denoted by $H$ is analogous to conjugation of complex numbers elements and is defined as:
\begin{equation}
q^H = q_r - q_i i - q_jj - q_k k.
\end{equation}
The square of the norm of a quartenion is defined as
\begin{equation}
||q||^2 = q_r^2 +q_i^2 + q_j^2 + q_k^2 = q^H q.
\end{equation}
with $(q_1q_2)^H = q_2^H q_1^H$ and the four components $(q_r, q_i, q_j, q_k)$ to correspond to the available four \hyperlink{a_EEG}{EEG} channels (see figure~\ref{fig:emotivsensors}).

Let quaternion column vector $\mathbf{q} = [q_1, \ldots, q_F]^T \in \mathcal{H}^F$ where $T$ denotes simple transposition be the \hyperlink{a_EEG}{EEG} values over time. The conjugate transpose of vector $\mathbf{q}$
is denoted by $\mathbf{q}^H$. There is an isomorphy between a quaternion and a complex $2\times 2$ matrix defined as
\begin{eqnarray}\label{eq:quatIsom}
\textbf{Q} &=& \left[
\begin{array}{cc}
q_r + q_i i  &  q_j + q_k i  \\
-q_j + q_k i &  q_r  - q_i i
\end{array}
\right]
\end{eqnarray}

Let $\mathbf{x}_l$ be the $F$-dimensional vector obtained by writing in lexico graphic ordering and form  $\mathbf{X}=[\mathbf{x}_1|\cdots| \mathbf{x}_N]\in\mathcal{H}^{F\times N}$. Also we denote by ${\mathbf{\bar{x}}} = \frac{1}{N} \sum_{i=1}^N \mathbf{x}_i$ and $\mathbf{\overline{X}}$ the sample mean and the centralised sample matrix $\mathbf{X}$, respectively. A projection vector is denoted by $\mathbf{u}\in \mathcal{H}^{F}$ and by $y_i = \mathbf{u}^H \mathbf{x}_i$ the projection of $\mathbf{x}_i$ onto $\mathbf{u}$. We want to maximise the (sum of the) variances of the data assigned to a particular class (gaze location)
\begin{equation}
\begin{array}{rl}
E(\mathbf{u})& = \sum_{l=1}^N ||y_l - \tilde{m}||^2 = \sum_{l=1}^N ||\mathbf{u}^H (\mathbf{x}_i- \mathbf{m})||^2
\\
                 & = \mathbf{u}^{H} \sum_{l=1}^N (\mathbf{x}_l- \mathbf{m})(\mathbf{x}_l- \mathbf{m})^H\mathbf{u}\\
                 & = \mathbf{u}^{H} \mathbf{S} \mathbf{u}
\end{array}
\end{equation}
where $\mathbf{S} = \mathbf{\bar{X}}\mathbf{\bar{X}}^H$. It can be easily proven that matrix $\mathbf{S}$ is a quaternion Hermitian matrix i.e., $S_{ij} = S_{ji}^H$.

In order to find $K$ projections $\mathbf{U}=[\mathbf{u}_1|\ldots|\mathbf{u}_k]\in \mathcal{H}^{F \times K}$ we may generalise $E(\mathbf{U})$:
\begin{equation}
\begin{array}{rl}
 \mathbf{U}_o &=\arg \max_{\mathbf{U}\in \mathbb{H}^{F\times p}}  E(\mathbf{U}) \\
              &=\arg \max_{\mathbf{U}\in \mathbb{H}^{F\times p}}  \mbox{tr}[\mathbf{U}^H \mathbf{S} \mathbf{U}] \\
\mbox{s.t.}&\,\,\mathbf{U}^H \mathbf{U} = \mathbf{I}.
\end{array}
\end{equation}
We aim at solving the above noted problem by using the isomorphic complex form that can be reformulated as
\begin{equation}
\begin{array}{rl}
 \mathbf{\tilde{U}}_o &=\arg \max_{\mathbf{\tilde{U}}}  \mbox{tr}[\mathbf{\tilde{U}}^H \mathbf{\tilde{S}}
 \mathbf{\tilde{U}}] \\
\mbox{s.t.}&\,\,\mathbf{\tilde{U}}^H \mathbf{\tilde{U}} = \mathbf{I}.
\end{array}
\end{equation}

Since $\mathbf{S}$ is a quaternion Hermitian matrix, $\mathbf{\tilde{S}}$ is a complex Hermitian.
Also, given that $\mathbf{\tilde{S}}$ is a positive semidefinite Hermitian matrix (i.e., it only has non-negative eigenvalues) the solution $\mathbf{\tilde{U}}_0$ is given by the $p$ eigenvectors of $\mathbf{\tilde{S}}$ that correspond to $p$ largest eigenvalues. We want an efficient algorithm for performing eigen-analysis to $\mathbf{\tilde{S}}$, which is a complex $2F\times 2F$ matrix and can be written
as $\mathbf{\tilde{S}} = \mathbf{\tilde{X}} \mathbf{\tilde{X}}^H$ where $\mathbf{\tilde{X}}\in\mathbb{C}^{2n\times F}$ and needs $O((2F)^3)$ time.

In general, a given quaternion Hermitian matrix $\mathbf{A}$ has $n$ nonnegative real eigenvalues (due to the non-commutative multiplication property of quaternions, two kinds of its eigenvalue exists; in this paper we are only interested in the left eigenvalues) $\mathbf{l} = [\sigma_1, \ldots, \sigma_n]$. Let $\mathbf{\tilde{A}}$ be its complex form

\begin{eqnarray}
\mathbf{\tilde{A}} &=& \left[
\begin{array}{cc}
\mathbf{A}_r + i \mathbf{A}_i &  \mathbf{A}_j + i \mathbf{A}_k \\
-\mathbf{A}_j+ i\mathbf{A}_k  &  \mathbf{A}_r - i \mathbf{A}_i.
\end{array}
\right]\nonumber
\end{eqnarray}

then the eigenvalues of $\mathbf{l}_{2n} = [\sigma_1,\sigma_1,\ldots,\sigma_n,\sigma_n]$. Representing $\mathbf{A} = \mathbf{B}\mathbf{B}^H$, where $\mathbf{B}$ is a quaternion matrix, and considering $\mathbf{\tilde{A}}$ and $\mathbf{\tilde{B}}$ to be the complex forms of matrices $\mathbf{A}$ and $\mathbf{B}$, respectively, then $\mathbf{\tilde{A}}$ will be given by $\mathbf{\tilde{A}} = \mathbf{\tilde{B}}\mathbf{\tilde{B}}^H$. So, based on this analysis, we can write $\mathbf{\tilde{S}} = \mathbf{\bar{X}} \mathbf{\bar{X}}^H$. Also by defining matrices $\mathbf{A}$ and $\mathbf{B}$ so that $\mathbf{A} = \mathbf{\Gamma}\mathbf{\Gamma}^H$ and $\mathbf{B} = \mathbf{\Gamma}^H\mathbf{\Gamma}$ with $\mathbf{\Gamma}\in \mathcal{C}^{m \times r}$, and considering $\mathbf{U}_A$ and $\mathbf{U}_B$ to be the eigenvectors corresponding to the non-zero eigenvalues $\mathbf{\Lambda}_A$ and $\mathbf{\Lambda}_B$ of $\mathbf{A}$ and $\mathbf{B}$, respectively, we finally obtain $\mathbf{\Lambda}_A = \mathbf{\Lambda}_B$ and $\mathbf{U}_A = \mathbf{\Gamma}\mathbf{U}_B \mathbf{\Lambda}_A^{-\frac{1}{2}}$.

Thus, according to the above, in a classification problem such as the current one for gaze location, we may represent the quaternion Hermitian matrix (descriptor) providing a subspace analysis method in the quaternion domain. So, assuming we have a quaternion matrix $P$ with dimension $m\times n$, we consider $n$ to be the total number of the captured data and $m$ the number of the actual hyper-complex features. A quaternion \hyperlink{a_PCA}{PCA} of $P$ as it was analysed above seeks a solution that contains $r$ $(r<m,n)$ linearly independent quaternion eigenvectors in the columns of $Q$ $(m\times r)$ so that $P=QA$ where the rows of $A$ $(r\times n)$ contain the $r$ quaternion principal component series. As a result a solid representation of the selected quaternion features is obtained, while the computational complexity is low.

The dimensionality of the quaternion feature vector was reduced using Quaternion-\hyperlink{a_PCA}{PCA}, and the final model is created using \hyperlink{a_kNN}{kNN} classifiers (see Gaze Classification block on Figure~\ref{fig:DiagramEye}). The \hyperlink{a_kNN}{kNN} algorithm finds the k-nearest neighbors among the training data, and they are used to weight the category candidates \cite{Khasnobish2010}. The performance of this algorithm depends on two factors: the similarity function and the $k$ value (e.g. if $k$ is too large, big classes will overwhelm the small ones). The approaches that can be used are shown in the equations below.
\begin{equation}\label{eq:knn1}
\begin{array}{ll}
y(\mathbf{d}_i)=\arg \max_{\mathbf{k}} \sum_{\mathbf{x}_j\in \mathbf{kNN}}||y(\mathbf{x}_j,\mathbf{c}_k)||
\end{array}
\end{equation}
\begin{equation}\label{eq:knn2}
\begin{array}{ll}
y(\mathbf{d}_i)=\arg \max_{\mathbf{k}} \sum_{\mathbf{x}_j\in \mathbf{kNN}} ||s(\mathbf{d}_i,\mathbf{x}_j)y(\mathbf{x}_j,\mathbf{c}_k)||\\
\end{array}
\end{equation}
where $\mathbf{d}_i$ is the test data, with $\mathbf{x}_j$ to belong in class $\mathbf{c}_k$, and $s(\mathbf{d}_i,\mathbf{x}_j)$ represents the similarity function for $\mathbf{d}_i$ and $\mathbf{x}_j$. Equation \ref{eq:knn1} result is the class with the maximum number of samples in the k neighbor and Equation \ref{eq:knn2} is the class with maximum sum of similarities. Also, the most commonly used similarity functions are Euclidean, City block, Cosine and Correlation distances.

\subsubsection{Testing}

The objective of the testing stage is the classification of new raw data. Therefore, new incoming raw data is preprocessed, transformed into quaternions and the features are extracted using Quaternion-\hyperlink{a_PCA}{PCA}, following the same pre-processing steps as in the training stage. Once the features are extracted, the model created during the training is used to classify the new input data and determine the gaze location.

\subsection{Emotion Understanding using Multimodal Information based on Autobiographical Memories for Alzheimer's Patients}

\begin{figure}
\centering
\includegraphics[width=120mm]{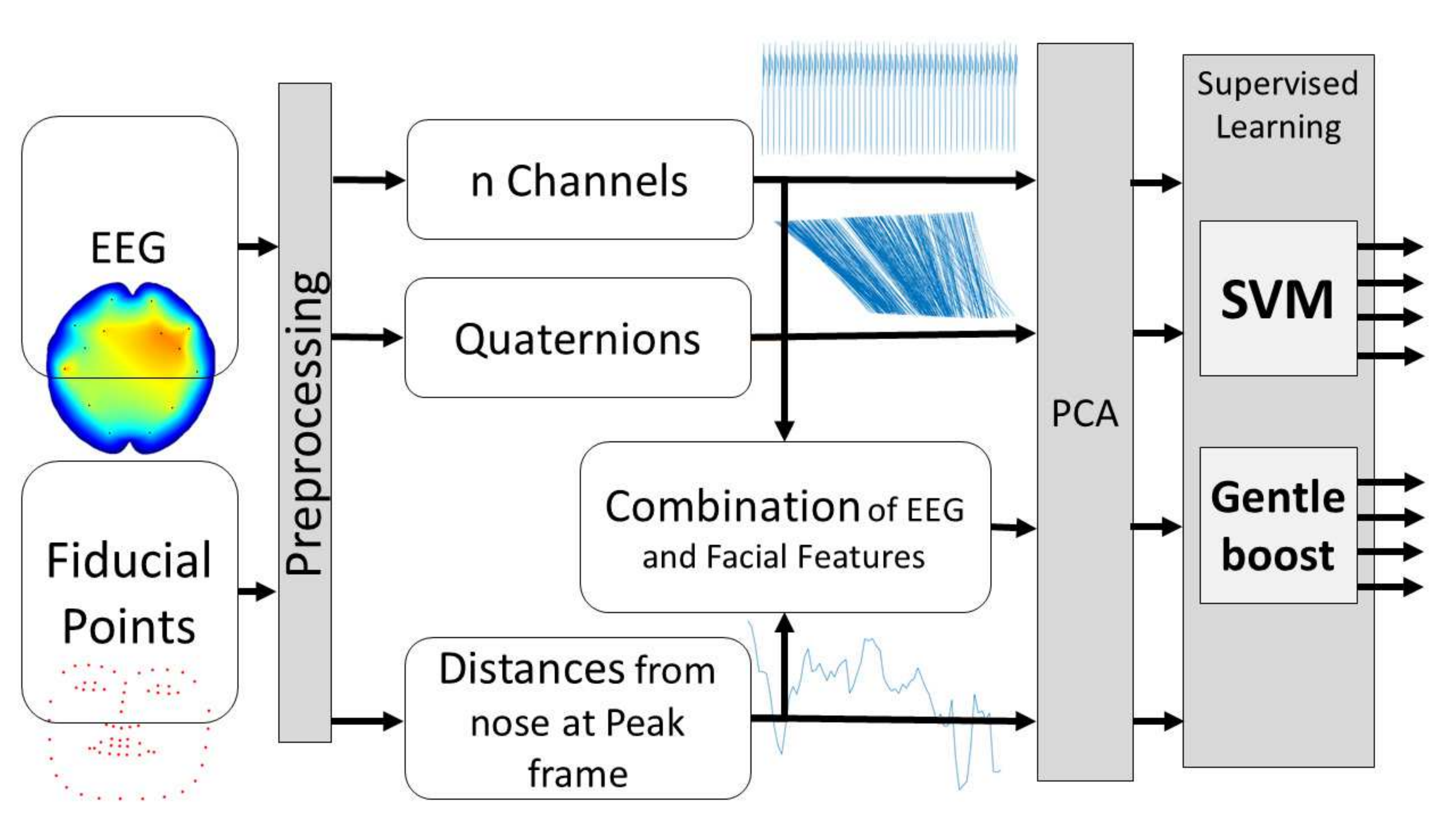}
\caption{
Diagram demonstrating our approach. The two modalities of data used (Fiducial Points, EEG) go through the process independently. The combination of the features concatenates the features extracted from the fiducial points with the EEG features, including all combinations.
}
\label{fig:diagram}
\end{figure}

Our approach intends to classify the reaction of the participants using two data modalities: the \hyperlink{a_EEG}{EEG} data and the fiducial points obtained from the RGB face images \cite{MontenegroEmo2016}. Using each modality and combining them (see Fig.~\ref{fig:diagram}), two binary classifications have been performed trying to recognise spontaneous reactions from distant and recent memories that were triggered during our experiments (see Table~\ref{table:classes}). The main reaction to be detected is the 'positive recognition' reaction versus the 'indifference' reaction. Additionally, a stronger recognition reaction is expected when the participant watches images from the distant past.

\setlength{\tabcolsep}{4pt}
\begin{table}
\begin{center}
\caption{
Classes chosen for recognition and the expected reaction.
}
\label{table:classes}
\begin{tabular}{llll}
\hline\noalign{\smallskip}
Id & Class 1 & Class 2 & Expected emotion\\
\noalign{\smallskip}
\hline
\noalign{\smallskip}
1 & Famous faces & Unknown faces & Recognition vs\\
  &  &  & Neutral reaction\\
2  & Distant past images & Recent past images & Long-term memory \\
   & of the participant  & of the participant & recognition vs short- \\
   & family and friends faces & family and friends faces & term memory recognition\\
3 & Distant past images & Recent past images & Long-term memory\\
  & of group of people & of group of people & recognition vs short-\\
  & including the participant, & including the participant, & term memory recognition\\
  & family and friends faces & family and friends faces & \\

4 & Famous places, & Unknown places & Recognition vs
\\
  & objects/brands & and objects & Neutral reaction
\\
\noalign{\smallskip}
\hline
\end{tabular}
\end{center}
\end{table}
\setlength{\tabcolsep}{1.4pt}

Our approach extracts features from both data modalities: \hyperlink{a_EEG}{EEG} and Facial points. The facial fiducial points were obtained using Baltru et al.'s approach~\cite{Baltru16} from a 30 frame rate video. This approach obtains sixty-eight fiducial points per frame. The coordinates of the fiducial points have been preprocessed normalizing them according to a neutral face point (i.e. nose)~\cite{Valstar05} to obtain rigid head motions invariant features. The \hyperlink{a_EEG}{EEG} data was recorded using an \hyperlink{a_EEG}{EEG} headset (Emotiv Epoc) which collects \hyperlink{a_EEG}{EEG} from fourteen sensors at 128Hz.

\label{distanceTOnose}Once the data was collected and preprocessed, the features were extracted. The spatio-temporal facial features studied were based on Michel et al.'s work~\cite{Michel03}. The distance of each coordinate to the nose point was measured. The first frame of each subject data was considered as the neutral face since at the beginning of the test a neutral pose is expected. Each frame was compared to the neutral face, calculating the frame that varies most from the neutral face. That frame was selected as the peak frame and used as a $p$ points long feature vector. For \hyperlink{a_EEG}{EEG}, three variations of features were analysed: (i) a combination of the fourteen channels, (ii) a combination of the four frontal channels and (iii) novel features combining the four frontal channels into a quaternion representation based on quaternion principal component analysis. Those features were implemented exactly as explained in the previous section.

Alongside the individual features modalities, a combination of the aforementioned \hyperlink{a_EEG}{EEG} and facial features has also been analysed. This combination comprises the attachment of the \hyperlink{a_EEG}{EEG} features vector to the facial one. Once the features are structured properly, dimensionality reduction is applied using \hyperlink{a_PCA}{PCA} and the reduced features are used as input to two supervised learning algorithms: \hyperlink{a_SVM}{SVM} and GentleBoost. A leave-one-out approach is used so the features obtained from $N-1$ of the participants, $N$ being the number of participants, are used for training and the remaining participants' data is used for testing. Moreover, k-fold cross-validation has been applied so the final results are the average of all the folds.

\subsection{Cognitive Behaviour Analysis based on Facial Information using Depth Sensors}
\begin{figure}
\centering
\includegraphics[width=120mm]{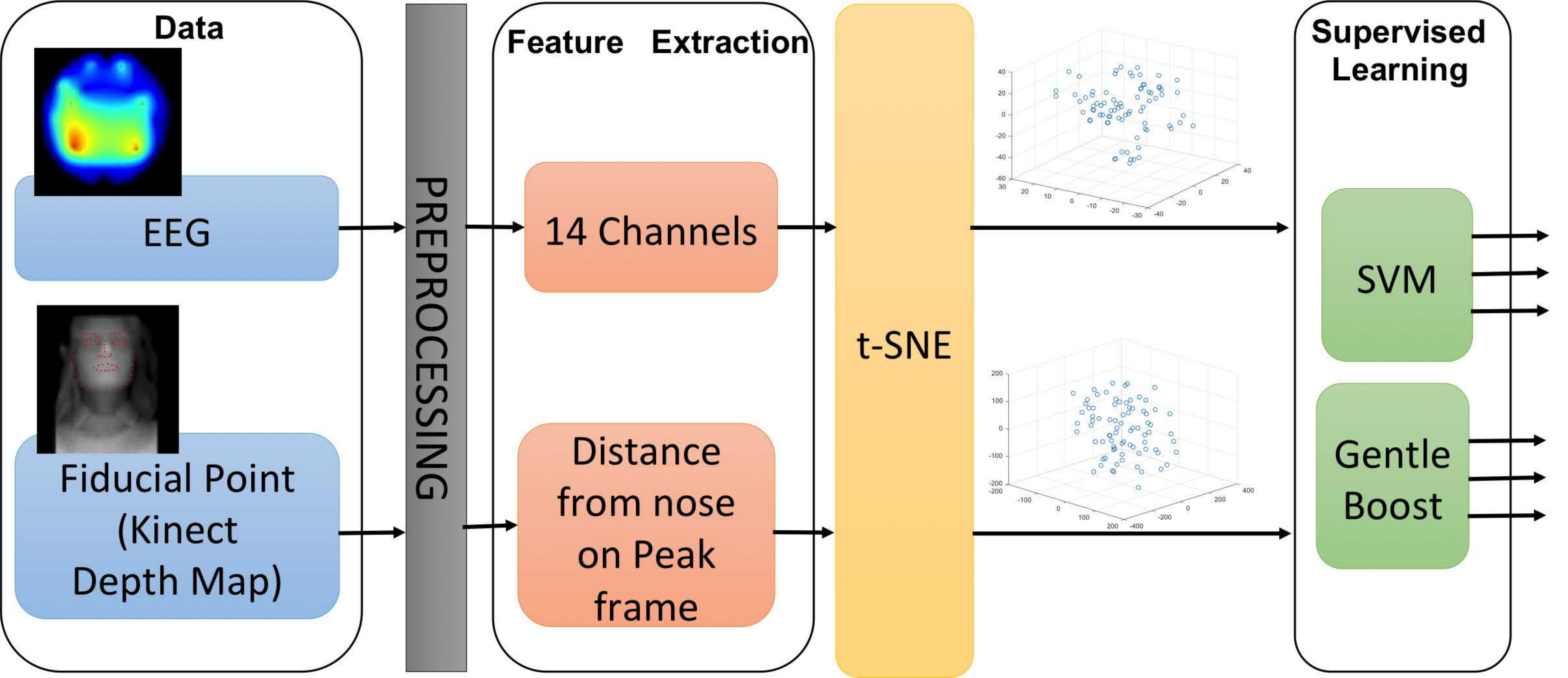}
\caption{
Diagram representing our approach using the two modalities of data (EEG in the upper part of the diagram and Fiducial Points from depth images in the lower part of the diagram). It is possible to go through the process independently or combining the two modalities concatenating the features extracted from the fiducial points with the EEG features.
}
\label{fig:diagramDepth}
\end{figure}

In this approach features are extracted from two data modalities: 3D facial points from depth data acquired using Kinect and \hyperlink{a_EEG}{EEG} signals \cite{MontenegroFace2016}. As shown in~\ref{fig:diagramDepth}, two feature descriptors were used in our classification problem based on the corresponding modalities. The classes for classification are the same as the classified on the previous section.

The fiducial facial points were obtained from a 30 frame rate \hyperlink{a_IR}{IR} video using the methods in \cite{Baltru16,Michel03}, obtaining 68 facial fiducial points per frame. The third coordinate is obtained from the depth data since their correspondence with the \hyperlink{a_IR}{IR} frames is provided. In order to obtain rigid head motions invariant features, the coordinated of the fiducial facial points were normalized according to a neutral face points, which correspond to the nose. After preprocessing the data, the feature vectors were extracted.

We expect that at the beginning of each test, before showing the image, the subject is in a neutral pose. Therefore, we select as the neutral face the one that corresponds to the first frame and we compare each frame to it. The frame that results more distant from the neutral face is selected as the peak frame. The
distance of these points to the nose point were measured in the peak frame and all of them are combined in a single feature vector.

Regarding the \hyperlink{a_EEG}{EEG} signals, they were acquired from fourteen sensors at 128Hz using an \hyperlink{a_EEG}{EEG} Headset. These fourteen channels represent the features that have been obtained from the \hyperlink{a_EEG}{EEG} data. Only the fourteen sensors approach is utilised in this section to show comparative results; since it has already been probed that the facial features provide better classification results we will focus on the facial features comparison. During the preprocessing stage a median filter was applied to remove the noise and the mean was subtracted in each channel.

The binary classification is performed separately using the features extracted from the depth frames and the \hyperlink{a_EEG}{EEG} data. Once the features vectors are structured properly, we apply the \hyperlink{a_tSNE}{t-SNE} method in order to reduce the number of selected descriptors generating a manifold representation.

More speficically, \hyperlink{a_tSNE}{t-SNE}~\cite{Maaten2008} is a non-linear dimensionality reduction technique used to embed high-dimensional data into a low-dimensional space (e.g., two or three dimensions for human-intuitive visualization). Given a set of $N$ high-dimensional faces of people under different illumination conditions (i.e. data-points) $x_1, ..., x_N$, \hyperlink{a_tSNE}{t-SNE} starts by converting the high-dimensional Euclidean distances between data-points ($||x_i-x_j||$) into pairwise similarities given by symmetrized conditional probabilities. In particular, the similarity between data-points $x_i$ and $x_j$ is calculated from~\ref{eq=tsne1} as:

\begin{equation}\label{eq=tsne1}
  p_{ij} = \frac{p_{i|j} + p_{j|i}}{2N}
\end{equation}

where $p_{i|j}$ is the conditional probability that $x_i$ will choose $x_j$ as its neighbour if neighbours were picked in proportion to their probability density under a Gaussian centred at $x_i$ with variance $\sigma_i^2$, given by~\ref{eq=tsne2}:

\begin{equation}\label{eq=tsne2}
  p_{i|j} = \frac{\exp(\frac{-||x_i-x_j||^2}{2\sigma_i^2})}{\sum_{k \neq i} \exp(\frac{-||x_k-x_i||^2}{2\sigma_1^2})}
\end{equation}

In the low-dimensional space the Student-t distribution (with a single degree of freedom: $f(x) = \frac{1}{\pi (1+x^2)}$ ) that has much heavier tails than a Gaussian (in order to allow dissimilar objects to be modelled far apart in the map) is used to convert distances into joint probabilities. Therefore, the joint probabilities $q_{ij}$ for the low-dimensional counterparts $y_i$ and $y_j$ of the high-dimensional points $x_i$ and $x_j$ are given by

\begin{equation}\label{eq=tsne3}
  q_{ij} = \frac{(1+||y_i-y_j||^2)^-1 }{\sum_{k\neq i} (1+||y_k-y_l||^2)^-1)}
\end{equation}

The objective of the embedding is to match these two distributions (i.e.,~\ref{eq=tsne1} and~\ref{eq=tsne2}), as well as possible. This can be achieved by minimising a cost function which is the Kullback-Leibler divergence between the original ($p_{ij}$) and the induced ($q_{ij}$) distributions over neighbours for each object

\begin{equation}\label{eq=tsne4}
  D_{KL}(P||Q) = \sum_{i} \sum_{j} p_{ij} \log \frac{p_{ij}}{q_{ij}}
\end{equation}

The minimization of the cost function is performed using a gradient descent method which have the following simple form:

\begin{equation}\label{eq=tsne5}
  \frac{\partial D_{KL}}{\partial y_i} 4 \sum_{j} \frac{(p_{ij}-q_{ij})(y_i-y_j) }{(1+||y_i-y_j||^2)}
\end{equation}

The reduced feature vectors (manifolds) from the two data modalities represent the input to two supervised learning algorithms: \hyperlink{a_SVM}{SVM} and GentleBoost. Features from eight participants have been used for training, and data from one participant have been used for testing, according to the leave-one-out approach. The final results are the average of all the iterations, since k-folding cross-validation has been applied.


\section{Results}

\subsection{Gaze Estimation using EEG Signals for HCI in Augmented and Virtual Reality Headsets}
\subsubsection{Acquisition Process and Dataset}
Nine participants without severe visual impairment were selected with the purpose to record data. Amongst the population of participants, who are part of \hyperlink{a_SEMdb}{SEMdb}, four wore glasses or contact lenses during the test. Non-intrusive devices for \hyperlink{a_EEG}{EEG} and eye tracker data collection were utilised. The \hyperlink{a_EEG}{EEG} signal was collected using Emotiv EPOC headset. This \hyperlink{a_EEG}{EEG} headset is formed of sixteen sensors; two of them are reference sensors, therefore the brain signal is collected from the remaining fourteen sensors/channels at 128Hz. Nevertheless, since our aim is to combine our method with wearable equipment such as Virtual Reality and Augmented Reality devices, only the information of the four sensors located closer to the eyes are used (see figure~\ref{fig:emotivsensors}). The eye-tracker data is captured using the Tobii Eye tracker, a device located below the monitor and tracks the pupil position returning gaze data at 60Hz. These data include time-stamps of each sample, which allow the synchronization with the \hyperlink{a_EEG}{EEG} signal in order to be used also as ground truth.

The acquisition process itself is separated into three stages: the installation, the actual test, and the data collection. Regarding the installation, the participants are helped to put on the \hyperlink{a_EEG}{EEG} headset until the signal quality is excellent for all the sensors. They are seated about 50cm to 80cm in front of the monitor, on a chair that is lifted to a position where their eyes are detected by the eye-tracker.

Once they are comfortably seated, the calibration of the \hyperlink{a_VOG}{VOG} device is performed and the test process is explained. A white area that changes position over time is shown on the monitor and the subjects are asked to look at that location. The duration of the test is forty-five seconds; nine positions of the white area are shown with a duration of approximately five seconds each. Their \hyperlink{a_EEG}{EEG} signal and gaze data is recorded during that period.

\begin{figure}[!t]
  \centering
  \includegraphics[width=.9\columnwidth]{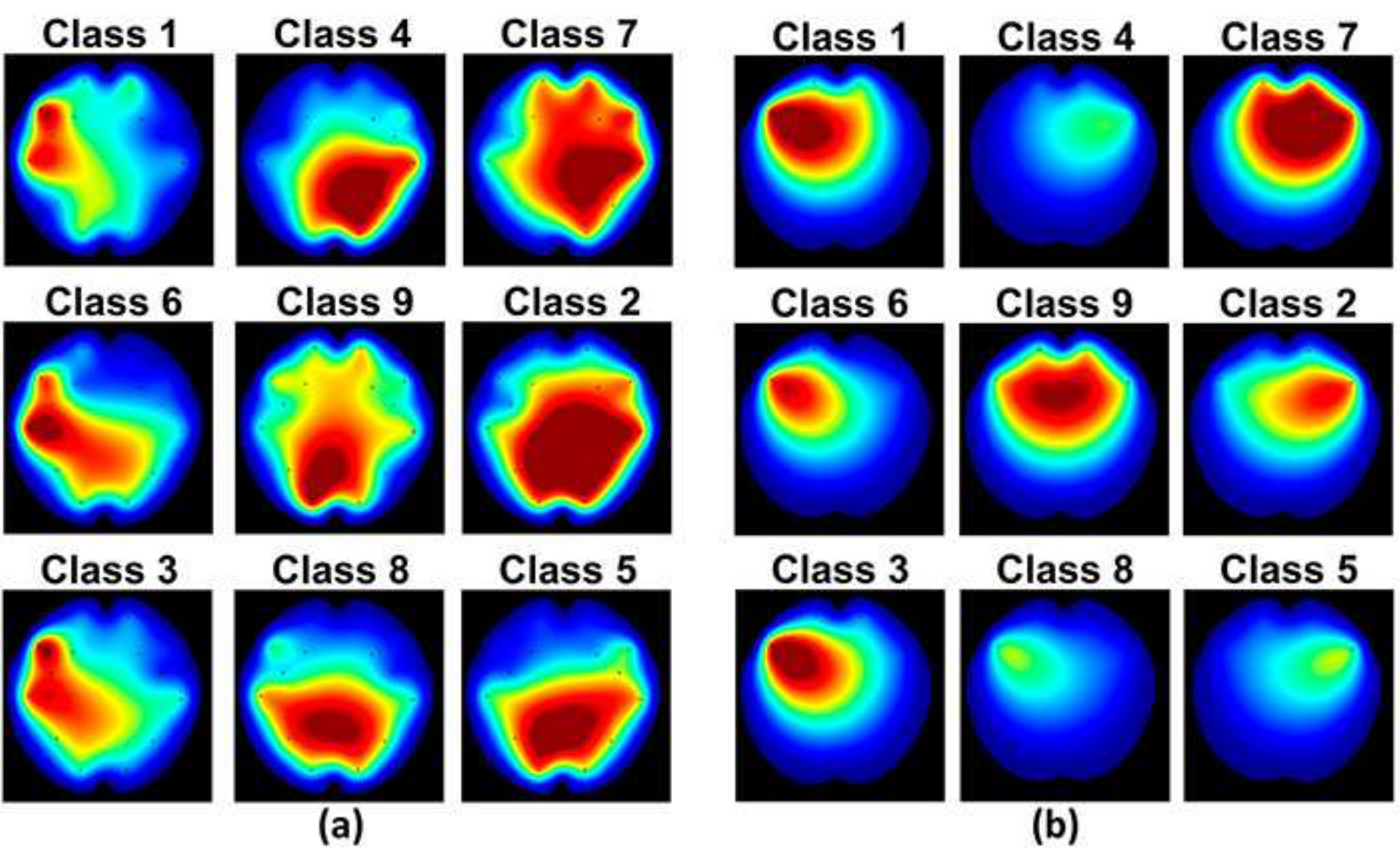}\vspace{-4mm}
  \caption{Examples of the EEG signals for each gaze location using (left) all the fourteen channels and (right) only the four of them. The classes represented in the figures belong to the nine positions shown in Figure \ref{fig:sectionsSEMdb}.
  } \label{fig:colourEEG}\vspace{-4mm}
\end{figure}

The proposed approach uses the four channels/sensors located on the frontal lobe (AF3, AF4, F7 and F8). AF3 and AF4 are affected by vertical eye movements; and F7 (left) and F8 (right) by horizontal eye movements \cite{Suresh2016}, this is, the \hyperlink{a_EEG}{EEG} signal at these sensors present a peak when there is a vertical or horizontal eye movement. These sensors record data at 128 frame rate. Figure~\ref{fig:colourEEG} shows the heatmap of the values obtained by each sensor at a specific time while the participant's gaze was focused on one class/area of the screen (See Figure \ref{fig:sectionsSEMdb}). The image on the left show fourteen sensors values and the image on the right only the aforementioned four.

\subsubsection{\itshape EEG Quaternion PCA Features}
After the data is preprocessed to reduce the noise the proposed novel quaternion descriptors are estimated. The four channels are combined in a single quaternion, with the four components $(q_r, q_i, q_j, q_k)$ to correspond to the sensors located to the sides of the head (F7 and F8) and the ones above the eyes (AF3 and AF4), respectively. Horizontal movements produce opposite-going voltage traces in F7 and F8 (e.g. when the eye is moving to the left F7 amplitude is increased and F8 is decreased), \cite{Jensenetal2002}. The changes in the AF3 and AF4 sensors are proportional, when it comes to vertical eye movements, \cite{GuerreroMosqueraetal2009}. The data is split into two different sets: $67\%$ for training and $33\%$ for testing. The separation of the data is done following the “Leave-persons out” protocol, that is, a set of people is removed to obtain the minimum test set that contains instances of all classes. Since the number of subjects is small, k-fold cross validation is used to limit overfitting problems. The proposed method uses nine-fold cross validation, with the testing data to be taken from a different set of users on each fold. Quaternion \hyperlink{a_PCA}{PCA} is used to reduce the features by selecting the number of dimensions according to the minimum necessary to obtain the best results, in this case the results were optimal when the data was reduced to fifteen dimensions.

\begin{figure}[!t]
  \centering
  \includegraphics[scale=.5]{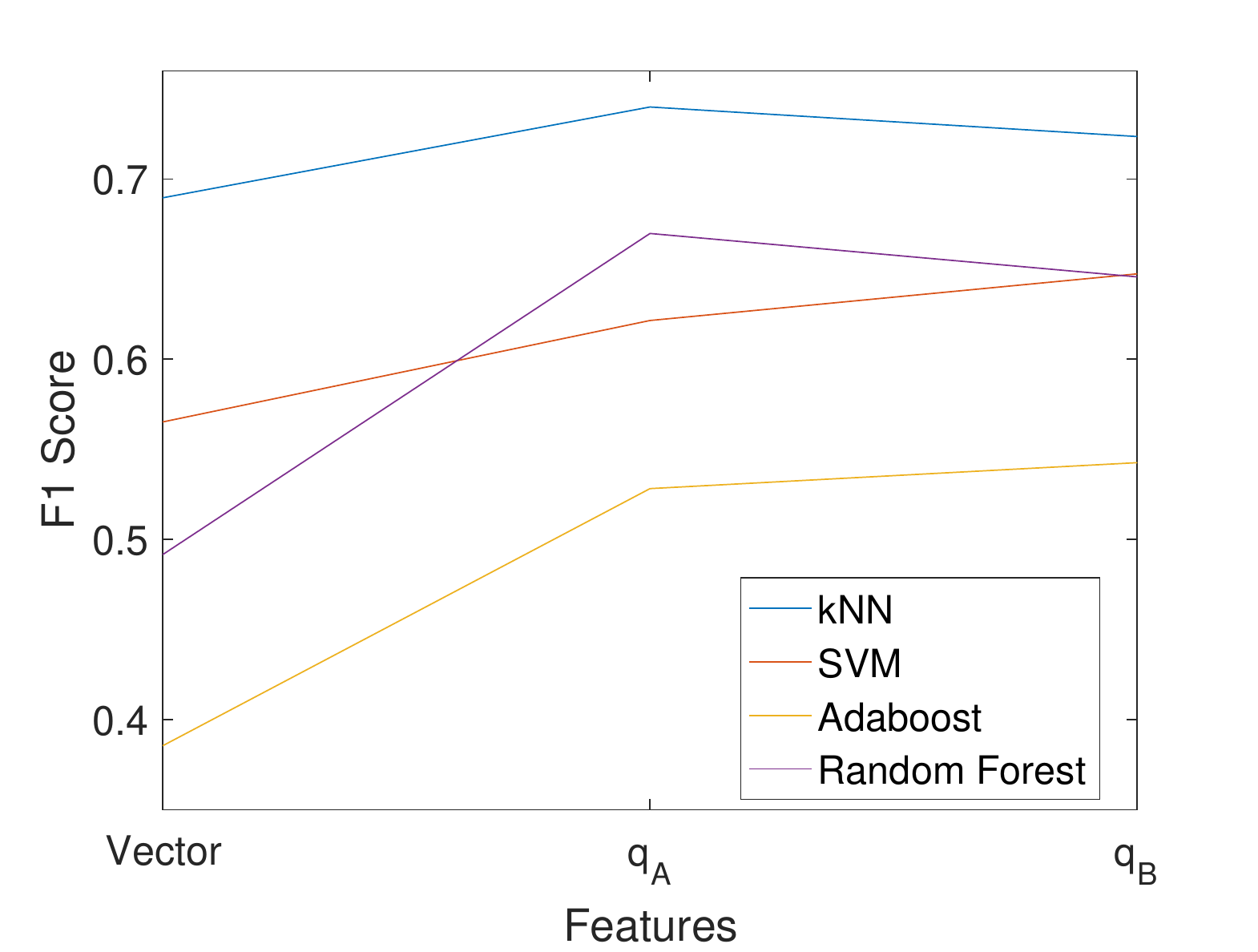}
  \caption{The average F1 scores of the classes.}
  \label{fig:PCAtest}\vspace{-4mm}
\end{figure}

\subsubsection{\itshape Gaze Location Classification F1 Scores}
The classification accuracy of the proposed approach was evaluated using three different metrics: precision (P), recall (R) and F1 score. Precision is the number of true positives (TP) divided by the amount of true positives plus false positives (FP) $P=\frac{TP}{TP+FP}$, and recall is the number of true positives divided by the amount of true positives plus false negatives (FN) $R=\frac{TP}{TP+FN}$, whereas the F1 score is a measure of the tests' accuracy based on the precision and recall and it is defined as $F1=2*\frac{P*R}{P+R}$.

Four different classification methods were used in our comparative study: \hyperlink{a_kNN}{kNN}, \hyperlink{a_SVM}{SVM}, AdaBoost (Adab) and Random Forests (RF). \hyperlink{a_kNN}{kNN} is a multi class classification method that assigns new unclassified samples to the class to which the majority of its $k$ nearest neighbours belong. It assumes that the $k$ nearest neighbors of a test sample are located at roughly the same distance from it (this approach uses cosine distance). \hyperlink{a_SVM}{SVM} classifier is a machine learning algorithm that maps the input features into a higher dimensional feature space. A linear decision surface is then constructed in this high-dimensional-feature space so that the margin between the surface and the nearest point is maximised. Since \hyperlink{a_SVM}{SVM} is a binary classification method, it is transformed to a multiclass classifier according to the one vs all relation. AdaBoost classifier combines weak classifiers to create a strong one. The weak classifier has to solve a sequence of learning problems and they are weighted according to their results. The final strong classifier is a weighted combination of the weak classifiers. AdaBoostM2 is a multiclass AdaBoost version, where each weak learner is associated to one class. Random forests is a machine learning classifier that uses a set of tree predictors and weights their output in order to perform a prediction. A tree predictor is a classifier that recursively partitions a data set into smaller subdivisions according to a set of tests defined on each branch of the tree. At the end of the tree, there are the final nodes that are linked with the label of the classes.

The outcomes presented in table~\ref{tab1} are the averaged results of the nine-fold process. The results shown in the table present the results for each classification method, using the obtained features in a vector representation or the proposed one combined in a single quaternion. With regards to the novel quaternion representation two combinations of the four channels were utilised and evaluated altering channel 2 and 3, resulting in the following cases $q_A=(q_r, q_i, q_j, q_k)$ and $q_B=(q_r, -q_j, q_i, q_k)$. The improvement can be seen by observing the results using the proposed quaternion representation over all the classifiers. Furthermore, in figure~\ref{fig:PCAtest} compares all the classifiers in terms of F1 score over the vector and quaternion representation of the descriptors are compared and the improvement that is obtained using the proposed quaternion representation can be seen.

\setlength{\tabcolsep}{4pt}
\begin{table}
\caption{\label{tab1} Results for all the classifiers.}
  \centering
  \begin{threeparttable}
    \begin{tabular}{rr|rrr}
    \hline\noalign{\smallskip}
    Features & Method & Precision & Recall & F1 \\
    \noalign{\smallskip}
    \hline
    \noalign{\smallskip}
    Vector     & \hyperlink{a_kNN}{kNN}   & 0.7500  & 0.6914 & 0.6896 \\
                 & \hyperlink{a_SVM}{SVM}   & 0.5838 & 0.6049 & 0.5652 \\
                 & Adab & 0.4185 & 0.4568 & 0.3855 \\
             & RF & 0.5594 & 0.5432 & 0.4916 \\
    \noalign{\smallskip}
    \hline
    \noalign{\smallskip}
    $q_A$\tnote{1} & \hyperlink{a_kNN}{kNN}   & 0.7901 & 0.7407 & \textbf{0.7400} \\
                 & \hyperlink{a_SVM}{SVM}   & 0.6321 & 0.6667 & 0.6215 \\
                 & Adab & 0.5691 & 0.5432 & 0.5282 \\
                 & RF & 0.7031 & 0.6790 & \textbf{0.6698} \\
    \noalign{\smallskip}
    \hline
    \noalign{\smallskip}
    $q_B$\tnote{2} &  \hyperlink{a_kNN}{kNN}   & 0.7741 & 0.7284 & 0.7236 \\
                & \hyperlink{a_SVM}{SVM}   & 0.6821 & 0.6667 & \textbf{0.6474} \\
                 & Adab & 0.5679 & 0.5556 & \textbf{0.5426} \\
                 & RF & 0.7080 & 0.6543 & 0.6457 \\
    \noalign{\smallskip}
    \hline
    \end{tabular}
    \begin{tablenotes}
    \item[1] $q_A$: Quaternion $(q_r, q_i, q_j, q_k)$
    \item[2] $q_B$: Quaternion $(q_r,-q_j, q_i, q_k)$
   \end{tablenotes}
  \end{threeparttable}
  \label{tab:addlabel}\vspace{-4mm}
\end{table}
\setlength{\tabcolsep}{1.4pt}

\subsection{Emotion Understanding using Multimodal Information based on Autobiographical Memories for Alzheimer's Patients}
This section shows and analyses the classification results obtained using the \hyperlink{a_EEG}{EEG} and Facial approaches presented in the previous section using \hyperlink{a_SVM}{SVM} and gentleboost classifiers. The results are represented by the F1 score which is a measure of accuracy that takes into account the precision and recall. A leave-one-out approach and a $k$-fold cross validation is applied for all the participants in our database (\url{http://homepages.inf.ed.ac.uk/rbf/CVonline/Imagedbase.htm\#face}). These results are compared with the ones obtained using as features the ones suggested in~\cite{Soleymani16}.

\begin{figure}
\centering
\includegraphics[width=120mm]{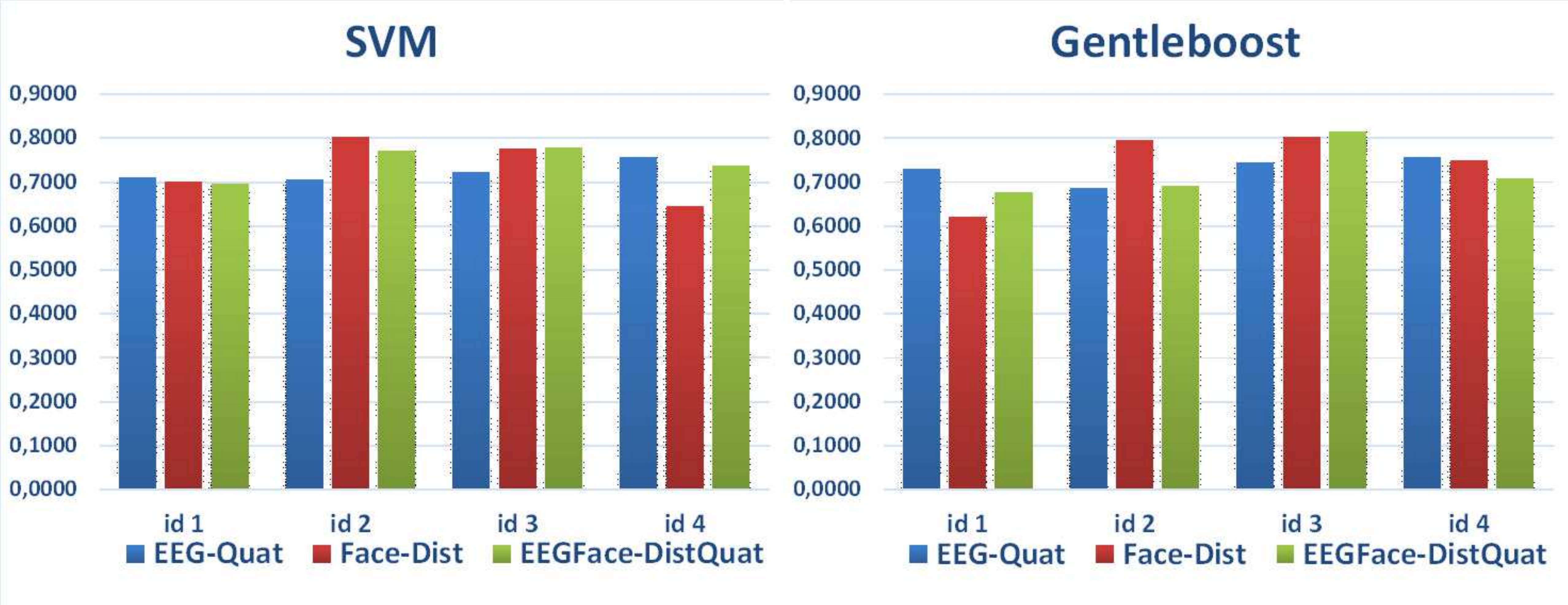}
\caption{
Results of the features that provide the best results for each classification using SVM and Gentleboost classifier: EEG quaternion, Face distance and EEGFace distances plus quaternion.
}
\label{fig:bestResACCV}
\end{figure}

\begin{figure}
\centering
\includegraphics[width=80mm]{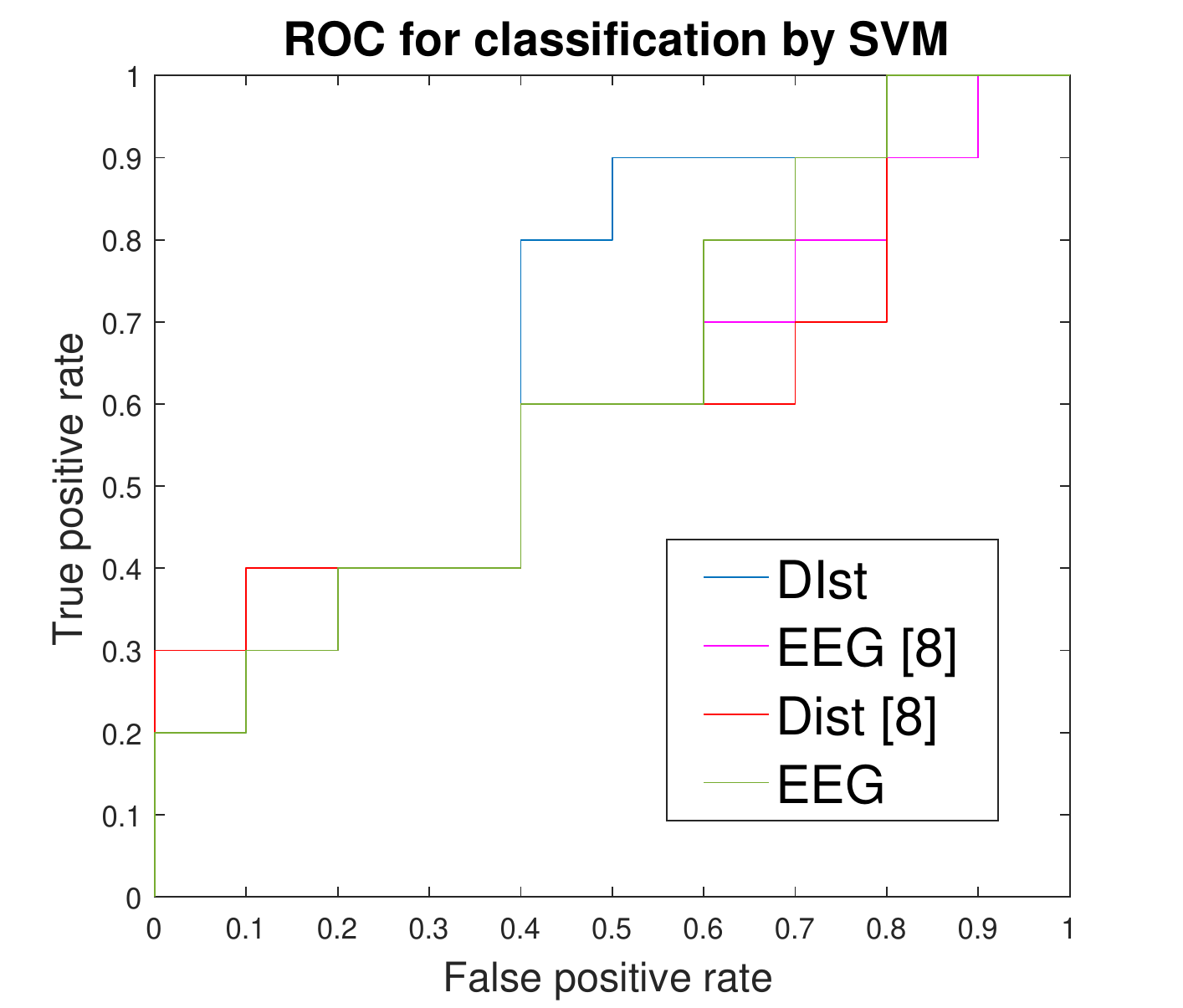}
\caption{ROC curve of the proposed method based on facial features in comparison to the ones proposed by Soleymani et al.~\cite{Soleymani16}.}
\label{fig:ROCaccv}
\end{figure}

Tables~\ref{tab:SVMaccv} and~\ref{tab:boostACCV} show the F1 scores for all the modalities and both classifiers, \hyperlink{a_SVM}{SVM} and gentleboost, respectively. Also, the precision and recall values are shown in table~\ref{tab:precRecACCV}, while an overview of the best outcomes is presented in figure~\ref{fig:bestResACCV}. Furthermore, the \hyperlink{a_ROC}{ROC} curves of the proposed method based on facial features in comparison to the ones proposed by Soleymani is shown in figure~\ref{fig:ROCaccv}. The results of both individual modalities (\hyperlink{a_EEG}{EEG} and facial) are coherent and adequate for the detection of emotions with overall F1 values around 70\%. Comparing both data modalities, facial fiducial landmarks provide slightly better results than \hyperlink{a_EEG}{EEG} signal for both classifiers; and the combination of both modalities only improves slightly the results using the gentleboost classifier. These results are in alignment with the results obtained by~\cite{Soleymani16}. On the other hand, the emotions related with unknown and known people or places have been recognised with higher accuracy using \hyperlink{a_EEG}{EEG} features. We assume that this is due to a minimal difference on facial expressions during the recognition of famous, but not personally related, versus the unknown people or places.

\setlength{\tabcolsep}{4pt}
\begin{table}[ht]
  \centering
  \caption{F1 scores obtained using SVM. See Table~\ref{table:classes} for id definitions.}
  \begin{threeparttable}
    \begin{tabular}{ll|llll|l}
    \hline\noalign{\smallskip}
    \multicolumn{2}{c|}{\hyperlink{a_SVM}{\textbf{SVM}}} & id 1 & id 2 & id 3 & id 4 & Overall\\
    \noalign{\smallskip}
    \hline
    \noalign{\smallskip}
    \multicolumn{6}{c}{\hyperlink{a_EEG}{\textbf{EEG}}} \\
    \noalign{\smallskip}
    \hline
    \noalign{\smallskip}
    \multicolumn{2}{l|}{Soleymani~\cite{Soleymani16}} & 0.6002 & 0.5677 & 0.6194 & 0.7122 & 0.6249\\
    Proposed & 14 Ch\tnote{1} & 0.5972 & 0.6882 & 0.6507 & 0.6704 & 0.6516 \\
    & 4 Ch & 0.6637 & 0.6965 & 0.7177 & 0.6725 & 0.6876 \\
    & Quaternion & \textbf{0.7105} & \textbf{0.7043} & \textbf{0.7225} & \textbf{0.7553} & \textbf{0.7232}\\
    \noalign{\smallskip}
    \hline
    \noalign{\smallskip}
    \multicolumn{6}{c}{\textbf{Face}}\\
    \noalign{\smallskip}
    \hline
    \noalign{\smallskip}
    \multicolumn{2}{l|}{Soleymani~\cite{Soleymani16}} & 0.6235 & 0.6699 & 0.6722 & \textbf{0.6942} & 0.6650\\
    Proposed & Dist\tnote{2} & \textbf{0.6987} & \textbf{0.8028} & \textbf{0.7750} & 0.6438 & \textbf{0.7301}\\
    \noalign{\smallskip}
    \hline
    \noalign{\smallskip}
    \multicolumn{6}{c}{\textbf{EEGFace}}\\
    \noalign{\smallskip}
    \hline
    \noalign{\smallskip}
    \multicolumn{2}{l|}{Soleymani~\cite{Soleymani16}} & 0.6429 & 0.7090 & 0.6502 & 0.6461 & 0.6620 \\
    Proposed & Dist +14 Ch & \textbf{0.6950} & 0.6825 & 0.7452 & 0.7313 & 0.7135 \\
     & Dist +4 Ch & 0.6887 & 0.7470 & \textbf{0.7843} & 0.7319 & 0.7380 \\
     & Dist +Quaternion & \textbf{0.6945} & \textbf{0.7699} & 0.7774 & \textbf{0.7372} & \textbf{0.7448} \\
    \noalign{\smallskip}
    \hline
    \end{tabular}%
    \begin{tablenotes}
    \item[1] X Ch: the EEG method are named according to the number of electrodes/Channels used
    \item[2] Dist is the method based on facial landmarks distances
   \end{tablenotes}
  \end{threeparttable}
  \label{tab:SVMaccv}%
\end{table}%
\setlength{\tabcolsep}{1.4pt}

\setlength{\tabcolsep}{4pt}
\begin{table}
  \centering
  \caption{F1 scores obtained using Gentleboost. See Table~\ref{table:classes} for id definitions.}
  \begin{threeparttable}
    \begin{tabular}{ll|llll|l}
    \hline\noalign{\smallskip}
    \multicolumn{2}{c|}{\textbf{Boost}} & id 1 & id 2 & id 3 & id 4 & Overall\\
    \noalign{\smallskip}
    \hline
    \noalign{\smallskip}
    \multicolumn{6}{c}{\hyperlink{a_EEG}{\textbf{EEG}}}\\
    \noalign{\smallskip}
    \hline
    \noalign{\smallskip}
    \multicolumn{2}{l|}{Soleymani~\cite{Soleymani16}}  & 0.6891 & 0.6843 & 0.6515 & 0.7540 & 0.6947 \\
    Proposed & 14 Ch\tnote{1} & 0.7030 & 0.6508 & 0.6901 & 0.6670 & 0.6777 \\
     & 4 Ch & 0.7061 & 0.6622 & 0.7792 & 0.7035 & 0.7128 \\
     & Quaternion & \textbf{0.7297} & \textbf{0.6861} & \textbf{0.7439} & \textbf{0.7565} & \textbf{0.7291} \\
    \noalign{\smallskip}
    \hline
    \noalign{\smallskip}
    \multicolumn{6}{c}{\textbf{Face}}\\
    \noalign{\smallskip}
    \hline
    \noalign{\smallskip}
    \multicolumn{2}{l|}{Soleymani~\cite{Soleymani16}}  & \textbf{0.7068} & 0.7362 & 0.7295 & 0.6579 & 0.7076\\
    Proposed & Dist\tnote{2} & 0.6200 & \textbf{0.7934} & \textbf{0.8024} & \textbf{0.7481} & \textbf{0.7410}\\
    \noalign{\smallskip}
    \hline
    \noalign{\smallskip}
    \multicolumn{6}{c}{\textbf{EEGFace}}\\
    \noalign{\smallskip}
    \hline
    \noalign{\smallskip}
    \multicolumn{2}{l|}{Soleymani~\cite{Soleymani16}}  & \textbf{0.7146} & 0.7296 & 0.6711 & 0.7412 & 0.7141\\
    Proposed & Dist +14 Ch & 0.6444 & 0.7084 & 0.7841 & 0.7148 & 0.7129 \\
     & Dist +4 Ch & 0.6327 & \textbf{0.7694} & 0.7871 & \textbf{0.7680} & \textbf{0.7393} \\
     & Dist +Quaternion & 0.6753 & 0.6911 & \textbf{0.8145} & 0.7077 & 0.7222\\
    \noalign{\smallskip}
    \hline
    \end{tabular}%
   \begin{tablenotes}
    \item[1] X Ch: the EEG method are named according to the number of electrodes/Channels used
    \item[2] Dist is the method based on facial landmarks distances
   \end{tablenotes}
  \end{threeparttable}
  \label{tab:boostACCV}%
\end{table}%
\setlength{\tabcolsep}{1.4pt}

\setlength{\tabcolsep}{4pt}
\begin{table}
  \centering
  \caption{Best precision and recall values of our approach corresponding to EEG Quaternion, Facial distance and Distance plus Quaternion features in comparison with the ones obtained by~\cite{Soleymani16} features.}
  \begin{threeparttable}
    \begin{tabular}{l|l|llll|l|llll|l}
    \hline\noalign{\smallskip}
    \multicolumn{1}{c}{} & \multicolumn{1}{c|}{} & \multicolumn{5}{c|}{\hyperlink{a_SVM}{\textbf{SVM}}} & \multicolumn{5}{c}{\textbf{Boost}}\\
    \noalign{\smallskip}
    \hline
    \noalign{\smallskip}
    \multicolumn{1}{c}{} & & id 1 & id 2 & id 3 & id 4 & OA & id 1 & id 2 & id 3 & id 4 & OA\tnote{3}\\
    \noalign{\smallskip}
    \hline
    \noalign{\smallskip}
    \multicolumn{1}{c}{} &      & \multicolumn{5}{c|}{\textbf{EEG}} & \multicolumn{5}{c}{\textbf{EEG}}\\
    \noalign{\smallskip}
    \hline
    \noalign{\smallskip}
    Soleymani~\cite{Soleymani16} & P\tnote{1}  & 0.664 & 0.687 & 0.733 & 0.735 & 0.705 & 0.749 & 0.721 & 0.714 & 0.787 & 0.743\\
     & R\tnote{2}  & 0.644 & 0.622 & 0.655 & 0.716 & 0.659 & 0.705 & 0.694 & 0.672 & 0.761 & 0.708 \\ \hline
    Quaternion & P & 0.758 & 0.737 & 0.746 & 0.784 & $\mathbf{0.757}$ & 0.751 & 0.723 & 0.783 & 0.780 & $\mathbf{0.759}$ \\
     & R & 0.722 & 0.711 & 0.727 & 0.761 & $\mathbf{0.730}$ & 0.733 & 0.694 & 0.750 & 0.761 & $\mathbf{0.734}$ \\
    \noalign{\smallskip}
    \hline
    \noalign{\smallskip}
    \multicolumn{1}{c}{} & \textbf{} & \multicolumn{5}{c|}{\textbf{Face}} & \multicolumn{5}{c}{\textbf{Face}}\\
    \noalign{\smallskip}
    \hline
    \noalign{\smallskip}
    Soleymani~\cite{Soleymani16} & P & 0.716 & 0.713 & 0.727 & 0.807 & 0.741 & 0.820 & 0.749 & 0.806 & 0.762 & 0.784\\
     & R & 0.661 & 0.722 & 0.705 & 0.722 & 0.702 & 0.738 & 0.777 & 0.750 & 0.705 & 0.743\\ \hline
    Distances & P & 0.749 & 0.849 & 0.838 & 0.710 & $\mathbf{0.787}$ & 0.684 & 0.823 & 0.860 & 0.786 & $\mathbf{0.788}$ \\
    & R & 0.711 & 0.816 & 0.794 & 0.683 & $\mathbf{0.751}$ & 0.644 & 0.800 & 0.816 & 0.755 & $\mathbf{0.754}$\\
    \noalign{\smallskip}
    \hline
    \noalign{\smallskip}
    \multicolumn{1}{c}{} & & \multicolumn{5}{c|}{\textbf{EEGFace}} & \multicolumn{5}{c}{\textbf{EEGFace}}\\
    \noalign{\smallskip}
    \hline
    \noalign{\smallskip}
    Soleymani~\cite{Soleymani16} & P & 0.797 & 0.730 & 0.708 & 0.711 & 0.737 & 0.751 & 0.742 & 0.790 & 0.808 & $\mathbf{0.773}$\\
     & R & 0.688 & 0.733 & 0.688 & 0.683 & 0.698 & 0.744 & 0.772 & 0.705 & 0.755 & $\mathbf{0.744}$ \\
    \noalign{\smallskip}
    \hline
    \noalign{\smallskip}
    Distances + & P & 0.764 & 0.806 & 0.806 & 0.767 & $\mathbf{0.786}$ & 0.722 & 0.736 & 0.847 & 0.735 & 0.760\\
    Quaternion & R & 0.711 & 0.777 & 0.783 & 0.744 & $\mathbf{0.754}$ & 0.688 & 0.705 & 0.822 & 0.716 & 0.733\\
    \noalign{\smallskip}
    \hline
   \end{tabular}%
   \begin{tablenotes}
    \item[1] P is the Precision
    \item[2] R is the Recall
    \item[3] OA are the OverAll results
   \end{tablenotes}
  \end{threeparttable}
  \label{tab:precRecACCV}%
\end{table}%
\setlength{\tabcolsep}{1.4pt}

In \hyperlink{a_EEG}{EEG}, the use of four channels provides similar results to the fourteen ones. The proposed quaternion based features improves the overall results by more than 1\%. The proposed facial features also provide better F1 scores than the ones used in~\cite{Soleymani16} in most of the classification scenarios. On the other hand, the results of the combined features are not always consistent in terms of which combination is the best one.

\subsection{Cognitive Behaviour Analysis based on Facial Information using Depth Sensors}
This section shows and analyses the recognition results obtained using the facial and \hyperlink{a_EEG}{EEG} features using \hyperlink{a_SVM}{SVM} and gentleboost classifiers. The results are represented by the F1 score.

\begin{figure}[ht]
\centering
\includegraphics[width=80mm]{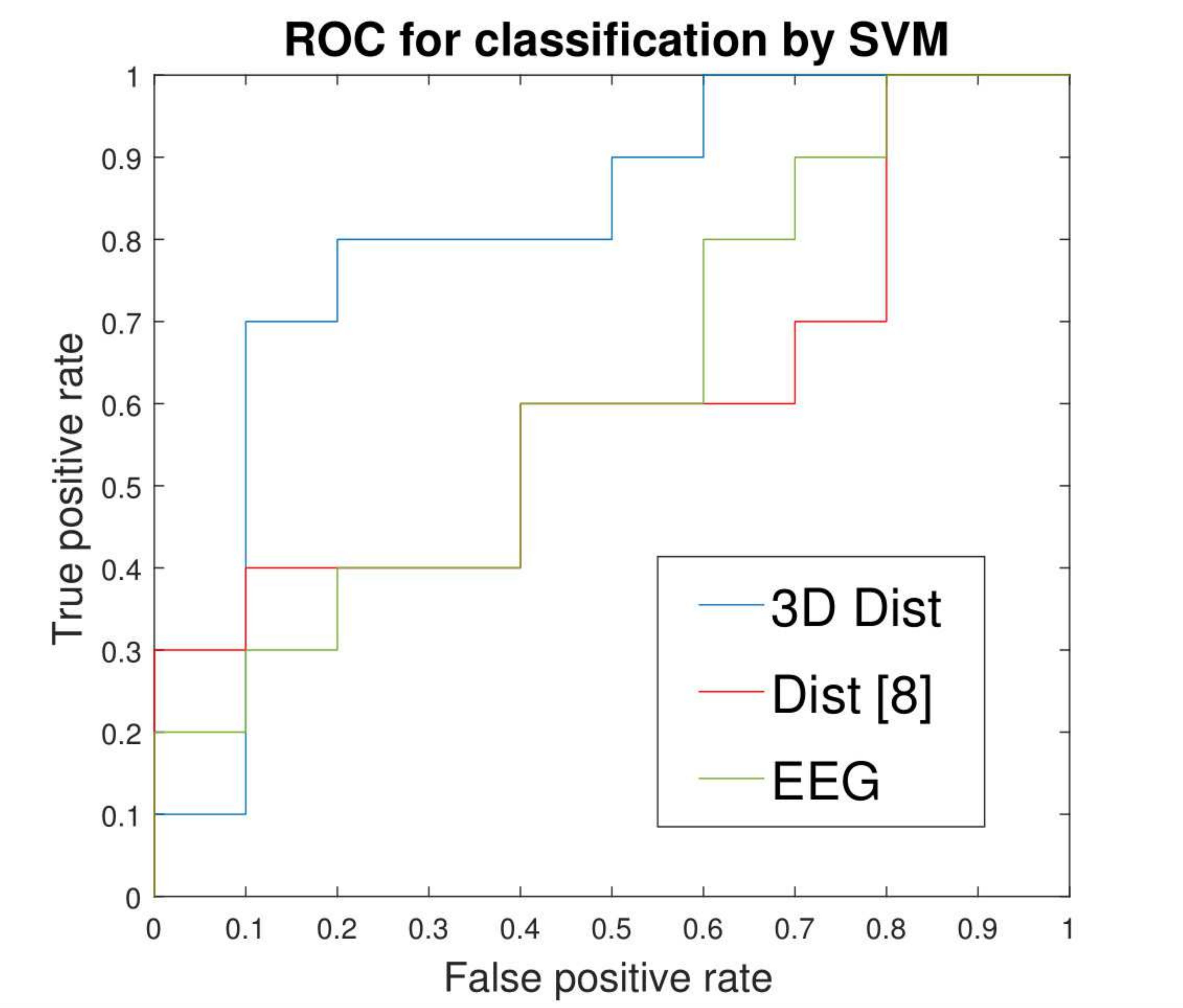}
\caption{ROC curve of the proposed method based on facial features in comparison to the ones proposed by Soleymani et al.~\cite{Soleymani16}.}
\label{fig:ROCdepth}
\end{figure}

Tables~\ref{tab:DepthT2} and~\ref{tab:DepthT3} show the F1 scores for all the modalities and both classifiers, \hyperlink{a_SVM}{SVM} and gentleboost, respectively. Furthermore, the \hyperlink{a_ROC}{ROC} curves of the proposed method based on facial and \hyperlink{a_EEG}{EEG} features in comparison to the ones proposed by Soleymani el al~\cite{Soleymani16} are shown in figure~\ref{fig:ROCdepth}. The results of both individual modalities (\hyperlink{a_EEG}{EEG} and facial) are coherent and adequate for the detection of emotions with overall F1 values above 70\%. Comparing both data modalities, face depth data provide slightly better results than \hyperlink{a_EEG}{EEG} for both classifiers. The classifiers have provided similar results, with \hyperlink{a_SVM}{SVM} to result more accurate estimates in the case of face depth data. The proposed facial features also provide better F1 scores than the state-of-the-art in most classifications. On the other hand, the \hyperlink{a_EEG}{EEG} features are not so consistent and the best results are provided using gentleboost.

\setlength{\tabcolsep}{4pt}
\begin{table}
  \centering
  \caption{F1 Score obtained using SVM classifier. See Table~\ref{table:classes} for id information}
    \begin{tabular}{l|llll|l}
    \hline\noalign{\smallskip}
    \multicolumn{1}{c|}{\hyperlink{a_SVM}{\textbf{SVM}}} & \multicolumn{1}{c}{id 1}& \multicolumn{1}{c}{id 2}& \multicolumn{1}{c}{id 3}& \multicolumn{1}{c}{id 4} & \multicolumn{1}{|c}{Overall}\\
    \noalign{\smallskip}
    \hline
    \noalign{\smallskip}
    EEG & 0.6069 & 0.6854 & 0.7106 & 0.7416 & 0.6861\\
    Face & 0.7035 & 0.7776 & 0.7276 & 0.6001 & $\mathbf{0.7022}$\\
    Soleymani~\cite{Soleymani16} & 0.6235 & 0.6699 & 0.6722 & 0.6942 & 0.6650\\
    \noalign{\smallskip}
    \hline
   \end{tabular}%
  \label{tab:DepthT2}%
\end{table}%
\setlength{\tabcolsep}{1.4pt}

\setlength{\tabcolsep}{4pt}
\begin{table}
  \centering
  \caption{F1 Score obtained using SVM classifier. See Table~\ref{table:classes} for id information}
    \begin{tabular}{l|llll|l}
    \hline\noalign{\smallskip}
    \multicolumn{1}{c|}{\textbf{Boost}} & \multicolumn{1}{c}{id 1}& \multicolumn{1}{c}{id 2}& \multicolumn{1}{c}{id 3}& \multicolumn{1}{c}{id 4} & \multicolumn{1}{|c}{Overall}\\
    \noalign{\smallskip}
    \hline
    \noalign{\smallskip}
    EEG & 0.6307 & 0.6789 & 0.7183 & 0.7338 & 0.6904\\
    Face & 0.6646 & 0.7826 & 0.6871 & 0.7143 & $\mathbf{0.7121}$\\
    Soleymani~\cite{Soleymani16} & 0.7068 & 0.7362 & 0.7295 & 0.6579 & 0.7076\\
    \noalign{\smallskip}
    \hline
   \end{tabular}%
  \label{tab:DepthT3}%
\end{table}%
\setlength{\tabcolsep}{1.4pt}

The classification of famous faces vs unknown faces, that is, the recognition of expected recognition vs neutral emotions, have been recognised with lower accuracy in both cases of facial and \hyperlink{a_EEG}{EEG} data. On the other hand, the classification of distant past faces vs recent past faces and distant past groups vs recent past groups, or the classification long-term vs short-term memory reactions resulted in higher recognition rates when facial features were utilised.

\section{Discussion}

A novel database (\hyperlink{a_SEMdb}{SEMdb}) has been created focusing on natural reactions to specific autobiographical and non-autobiographical stimuli that intend to elicit different emotions. This database provides facial videos and \hyperlink{a_EEG}{EEG} signals, amongst other information, that can be used for emotion recognition and other purposes. This database contains only data from healthy participants. Many automatic disease diagnosis related approaches use rigorous One Class Classification techniques due to the imbalanced nature of the data~\cite{Khan2014}. Rigorous \hyperlink{a_OCC}{OCC} allows the creation of reliable models -healthy people's emotions models in our case- using the data collected from healthy people for training purposes only. Therefore, any outlier of these models will be considered as a sign of a possible impairment. Using this database this chapter shows different usages of its data. Firstly, a proof of concept approach that represents some of the possible usages of the database provides a novel gaze estimation system using \hyperlink{a_EEG}{EEG} signal for selection based tasks and interactions. This approach focused on proposing an eye tracking method that works with a reduced number of sensors so that they are easily added to a Virtual Reality headset used for dementia diagnosis. Since eye movements are important in human computer interaction and Virtual and Augmented reality, this will change the way humans interact with the environment. Eye tracking will allow a more realistic navigation, making it easier for \hyperlink{a_AD}{AD} patients -usually elderly people with motion difficulties and problems to understand new technologies-. Our method is adapted to the requirements in terms of size and weight of the \hyperlink{a_VR}{VR} and \hyperlink{a_AR}{AR} headset devices, therefore it only uses the four \hyperlink{a_EEG}{EEG} sensors located on the front of the head without requiring external cameras. Additionally a novel quaternion feature representation was suggested to classify gaze positions improving the overall accuracy of the available classifiers significantly. This novel feature representation was also used for the main purpose of the database: an approach for classification of natural reactions to specific autobiographical and non-autobiographical stimuli that intend to elicit different emotions. Two different data modalities have been used independently and combined for the classification elicited by memories: \hyperlink{a_EEG}{EEG} and facial data. The overall results demonstrate that facial features outperform the \hyperlink{a_EEG}{EEG} ones for emotion recognition. The novel quaternion \hyperlink{a_EEG}{EEG} and the implemented facial features result in accurate classification rates. In parallel, an experiment used advanced features based on \hyperlink{a_tSNE}{tSNE} manifolds providing an accurate representation of the depth information and the \hyperlink{a_EEG}{EEG} data. Overall, the face depth representation provides more accurate classification rates in comparison to the other descriptors and data modalities utilised in our comparative study.

As the research results support that facial emotion recognition classification performance overcomes the \hyperlink{a_EEG}{EEG} one, a practical integration of these facial approaches. Since full immersive \hyperlink{a_VR}{VR} systems require \hyperlink{a_VR}{VR} headsets that cover the eyes, the integration with those systems is not possible but they could be used as an extra tool to evaluate \hyperlink{a_AD}{AD} cognitive impairments.



\chapter{Novel Origami Features for Facial Emotion Recognition} 

\label{Chapter5} 

\lhead{Chapter 5. \emph{Novel Origami Features for Facial Emotion Recognition}} 


\section{Introduction}

Facial expression and \hyperlink{a_EEG}{EEG} approaches for emotion recognition were compared in Chapter~\ref{Chapter4} where autobiographical emotion classification based on facial approaches have proved more accurate than \hyperlink{a_EEG}{EEG} ones. The better performance of facial features for emotion classification has also been demonstrated by other works such as~\cite{Soleymani16}. Huang et al.'s work~\cite{Huang2017} also shows a better performance of the facial expression approach and introduces two fusion methods for both modalities (\hyperlink{a_EEG}{EEG} and facial features) that increase the emotion recognition accuracies. Previous chapter analysed \hyperlink{a_EEG}{EEG} approaches due to their convenient combination on current virtual reality headsets, which are very useful for dementia screening tests. On the other hand, the use of facial expression techniques on their own for cognitive impairment analysis could be an interesting approach that will reduce the cost even more. In addition, the \hyperlink{a_VR}{VR} glasses or \hyperlink{a_EEG}{EEG} headset will not be necessary, therefore, it would avoid the possibility of elder patients feeling uncomfortable wearing those devices. In this chapter, the proposed methods will focus on facial expression techniques.

The facial expressions are usually linked to human emotions~\cite{Wan2014}. Several techniques have been proposed for facial expressions interpretation since the first automatic recognition study appeared in 1978~\cite{Alpher2015,Bettadapura12}. The Facial Action Coding System~\cite{Ekman78} is a well-known and commonly used system that describes facial expressions as independent configurations of the face, called Action Units. \hyperlink{a_AUs}{AUs} are codes that describe certain facial configurations, e.g. \hyperlink{a_AUs}{AU} 8 describes the fact that the lips are towards each other. The combination of \hyperlink{a_AUs}{AUs} can be used to describe basic emotions such as happiness, sadness, surprise, fear, anger and disgust. Some state-of-the-art methods for emotion classification focuses only on the classification of \hyperlink{a_AUs}{AUs}, only the \hyperlink{a_AUs}{AUs} or they also develop a method to link the \hyperlink{a_AUs}{AUs} with the basic emotions. For example, sadness is a combination of \hyperlink{a_AUs}{AUs} 1, 4 and 15 (inner brow raiser, brow lowerer and lip corner depressor). Other authors analyse images directly in order to detect the basic emotions and other researchers model the affect dimensions \ref{emoDim}.

Facial emotion recognition methods face the same challenges, such as head-pose and illumination variations, registration errors, occlusions and identity bias~\cite{Sariyanidi15}. In addition, some of them use posed emotions data, which is usually less challenging than spontaneous emotions~\cite{Wan2014,Sariyanidi15}. Some of these problems are not included in most of the available databases therefore many of them may not work properly in real conditions. For example, many databases are recorded on illumination controlled environments where the participants are always facing the camera. On the other hand, some facial emotion analysis applications do not require uncontrolled environments, e.g., medical examinations/tests.


The data provided by most of facial emotion recognition datasets is usually utilised to classify the six basic human emotions or the action units that compose those emotions using machine learning algorithms. Amongst the most common classifiers for facial emotion analysis, \hyperlink{a_SVM}{SVM}, boosting techniques and artificial neural networks are the most used~\cite{FERA2017}. The input to these classifiers is a series of features extracted from the available data that will provide distinctive information of the emotions such as facial landmarks, \hyperlink{a_HOG}{HOG}, SIFT descriptors or Gabor Wavelets~\cite{Corneanu2016,Zeng2009}. Facial landmarks are widely used, if not as features for classification they are used for facial alignment and normalisation alongside with other registration methods~\cite{Argyriou2007,Argyriou2011}. Other approaches use a transformation of these landmarks such as CANDIDE grids to represent the face, this is, a polygon face mask. Then, the geometrical displacements of the mesh are used as features~\cite{Kotsia2005,Kotsia2007,Kotsia2008}.

\begin{figure}[ht]
\centering
\includegraphics[scale=0.15]{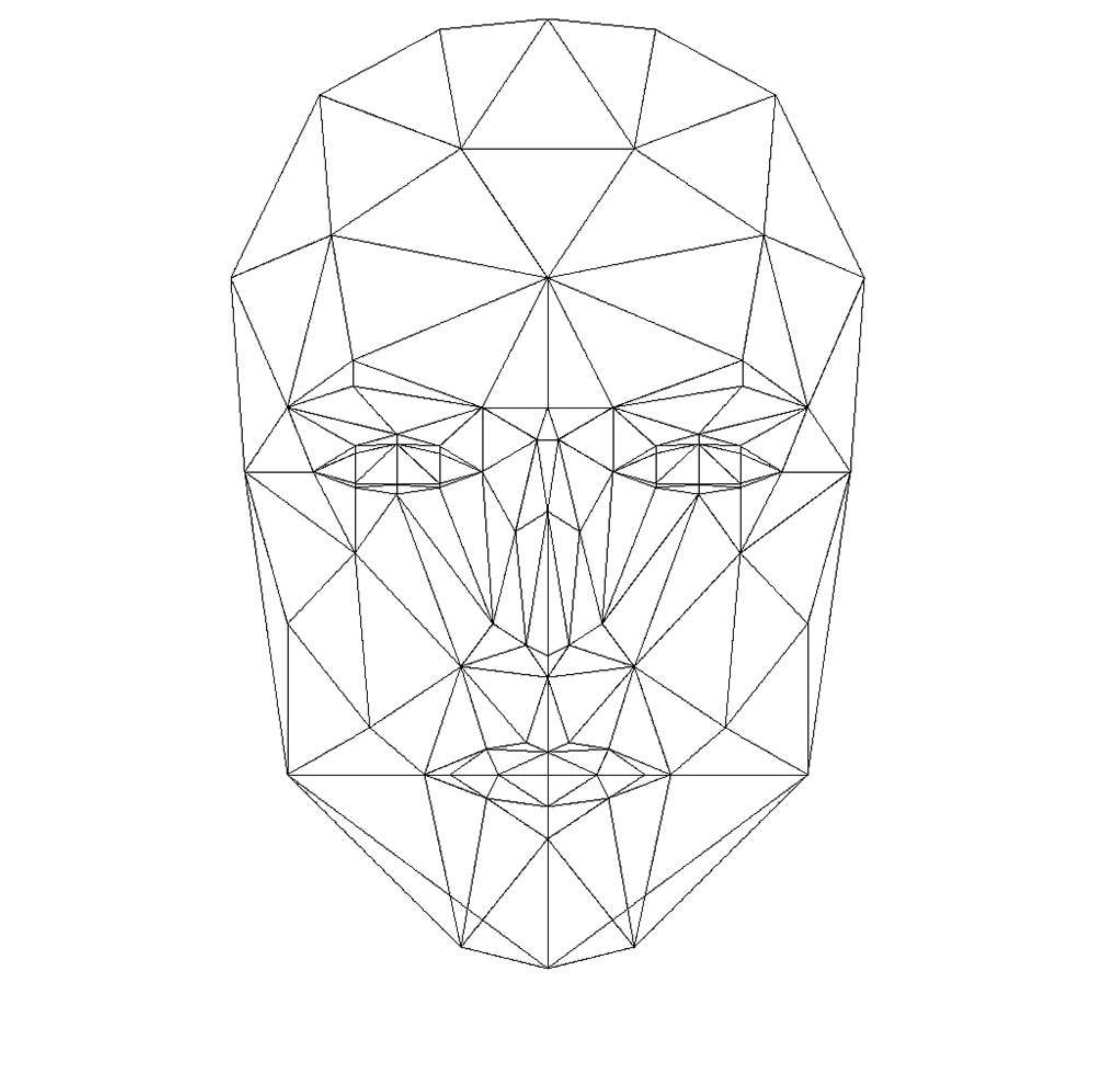}
\caption{CANDIDE grid example.}
\label{fig:candideGrid}
\end{figure}

The purpose of this chapter is the introduction of novel descriptors based on an innovative representation of the face and movement amplifier preprocessing methods for facial emotion classification. It demonstrates the improvement of the classification performance when those are combined with state-of-the-art features. The preprocessing method uses Eulerian magnification in order to magnify the motion of the face movements as presented in~\cite{Wadhwa2016}; which provides a more pronounced vision of the facial expressions. The facial features are based in Lang's Universal Molecule Algorithm~\cite{Bowers2015} resulting in a crease pattern representation of the facial landmarks. These crease pattern facial descriptors improve the performance when classifying \hyperlink{a_AUs}{AUs}, seven basic emotions and autobiographical memories related emotions. They were also proved noise and partial occlusions resistant. This proposal is focused on facial features that require an external camera to track the facial landmarks, therefore it is proposed as an additional test for \hyperlink{a_AD}{AD} screening but not to combine with Virtual Environments.


\section{Related Work}

The approaches that analyse \hyperlink{a_AD}{AD} patients' emotions elicited by autobiographical memories are nonexistent or they cannot be easily found. Consequently, proposals that classify standard facial expression are utilised for comparison. Candra et al.~\cite{Candra2016} propose a facial emotion recognition approach to recognise emotions from patients with psychological problems. These emotions can be used by psychotherapist to provide effective counsel and treatment. Candra et al. use CK+ dataset \ref{db_CK} to classify seven basic emotions. Their approach starts detecting the face, mouth and eyes regions using Viola-Jones algorithm. A Canny edge detector is applied to these regions and \hyperlink{a_HOG}{HOG} features are extracted from the resultant edge images. Finally a \hyperlink{a_SVM}{SVM} classifier with Radial Basis Function (Gaussian) kernel is used for classification. The classification accuracy of their proposed features (edge plus \hyperlink{a_HOG}{HOG}) is high but they do not manage to outperform the \hyperlink{a_HOG}{HOG} ones. Song et al.~\cite{Song2015} present an approach for facial expression classification that performs finely in different databases such as CK+ \ref{db_CK}, G-FERA \ref{db_FERA} and DISFA \ref{db_DISFA}. They present an action unit recognition system based on an improved Bayesian Compressed Sensing (BCS) classifier. Their system uses whole face registration and spatial features representation approach. It extracts features from the peak frame using pyramid histogram of gradients (pHOG) with eight bins on three different pyramid levels. The improved \hyperlink{a_BCS}{BCS} exploits the sparsity and co-occurrence structure via group-wise sparsity inducing priors. They take into account the fact that some \hyperlink{a_AUs}{AUs} usually appear when other \hyperlink{a_AUs}{AUs} are present. Therefore they have created group structures from the co-occurrence of the \hyperlink{a_AUs}{AUs}; and those group structures are used to increase the \hyperlink{a_AU}{AU} recognition performance. Their best classification results are obtained when their method is applied to CK+ database, reaching accuracies of 94\%. CK and its extension CK+~\cite{Lucey2010} are widely used facial expression databases. They are composed of sequences of images of different subjects' posed and non-posed emotions, action units and emotions labels for some of those images and facial landmarks.

Ali et al.~\cite{Ali2016} also use CK \ref{db_CK} posed database to propose a facial emotion recognition approach for partial occluded images. During the first stage of their method they used Viola-Jones to detect the faces region. Afterwards, they used Radon Transform to project the 2D image to 1D. Empirical Mode Decomposition was then used to decompose the 1D projection into a small number of components. Their dimensionality was reduced using \hyperlink{a_PCA}{PCA} and \hyperlink{a_LDA}{LDA} and the output was used as features of a \hyperlink{a_SVM}{SVM} classifier. They analysed four types of occlusions: top, bottom, left and right half of the face. Their results showed good emotion classification accuracy and demonstrated that most of the relevant information for emotion classification is in the lower region.



\section{Methodology}

\begin{figure}[!b]
\centering
  \includegraphics[scale = 0.4]{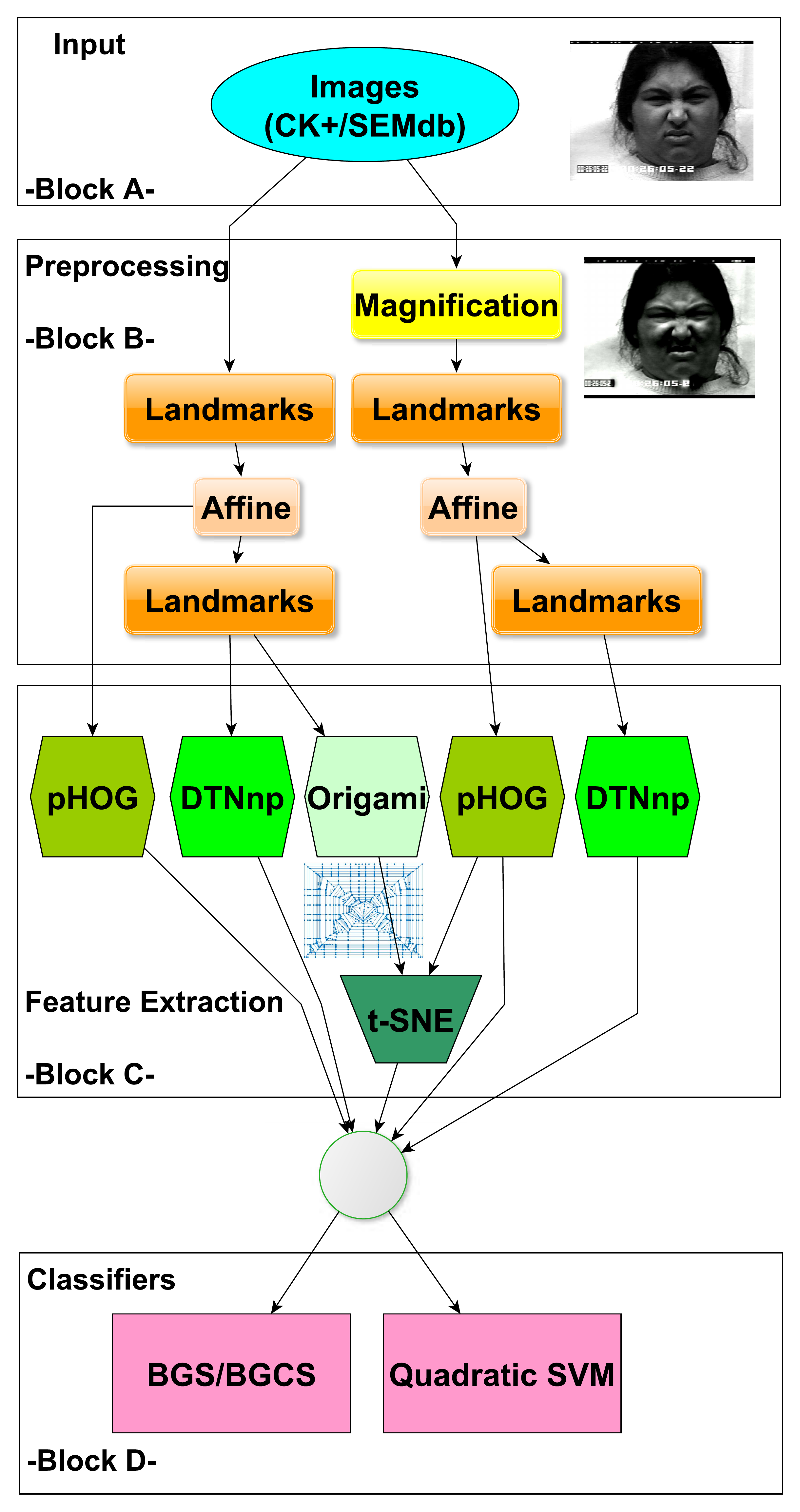}
  \caption{Methodology diagram. It represents the steps through which the data passes for training and testing purposes in our methodology. Each block of the figure are explained in next subsections: block B in \ref{ProcessingPori}, block C in \ref{FeatExtPori} and block D in \ref{ClassiPhase}.}
  \label{fig:diagramOri}
\end{figure}

The proposed methodology comprises a data preprocessing, a feature extraction and a classification phase. In the preprocessing phase, facial landmarks are extracted with and without Eulerian Video Magnification~\cite{Wadhwa2016} and the affine transform is applied for landmark alignment. In the feature extraction phase, the outputs of the affine transform are provided as inputs to a pyramid Histogram of Gradients (pHOG) descriptor. Moreover, new facial landmarks are extracted from the outputs of the affine transform and they are used as inputs to (1) the Lang's Universal Molecule algorithm~\cite{Lang1996} to extract novel origami features and (2) a facial feature displacement (i.e., the Euclidean distance between the nose location in a neutral and a 'peak' frame representative of the expression) descriptor, based on~\cite{Michel03}. Finally, in the classification phase the \hyperlink{a_pHOG}{pHOG} features are used as input to a Bayesian Compressed Sensing (BCS) and a Bayesian Group-Sparse Compressed Sensing (BGCS) classifiers, while the origami and nose-distance features are provided as inputs to a Quadratic \hyperlink{a_SVM}{SVM} classifier. The \hyperlink{a_tSNE}{t-SNE} is applied to the origami features to reduce their dimensionality before they are fed into the \hyperlink{a_SVM}{SVM} classifier. The utilised methodology is summarised in Figure~\ref{fig:diagramOri} and is described in detail in the following subsections.

\subsection{Preprocessing Phase} \label{ProcessingPori}

The two main techniques used in the data preprocessing phase (apart from the facial landmarks extraction) are the affine transform for landmark alignment and the Eulerian Video Magnification (EVM)~\cite{Wadhwa2016}.

\begin{figure}
\centering
  \includegraphics[scale = 0.45]{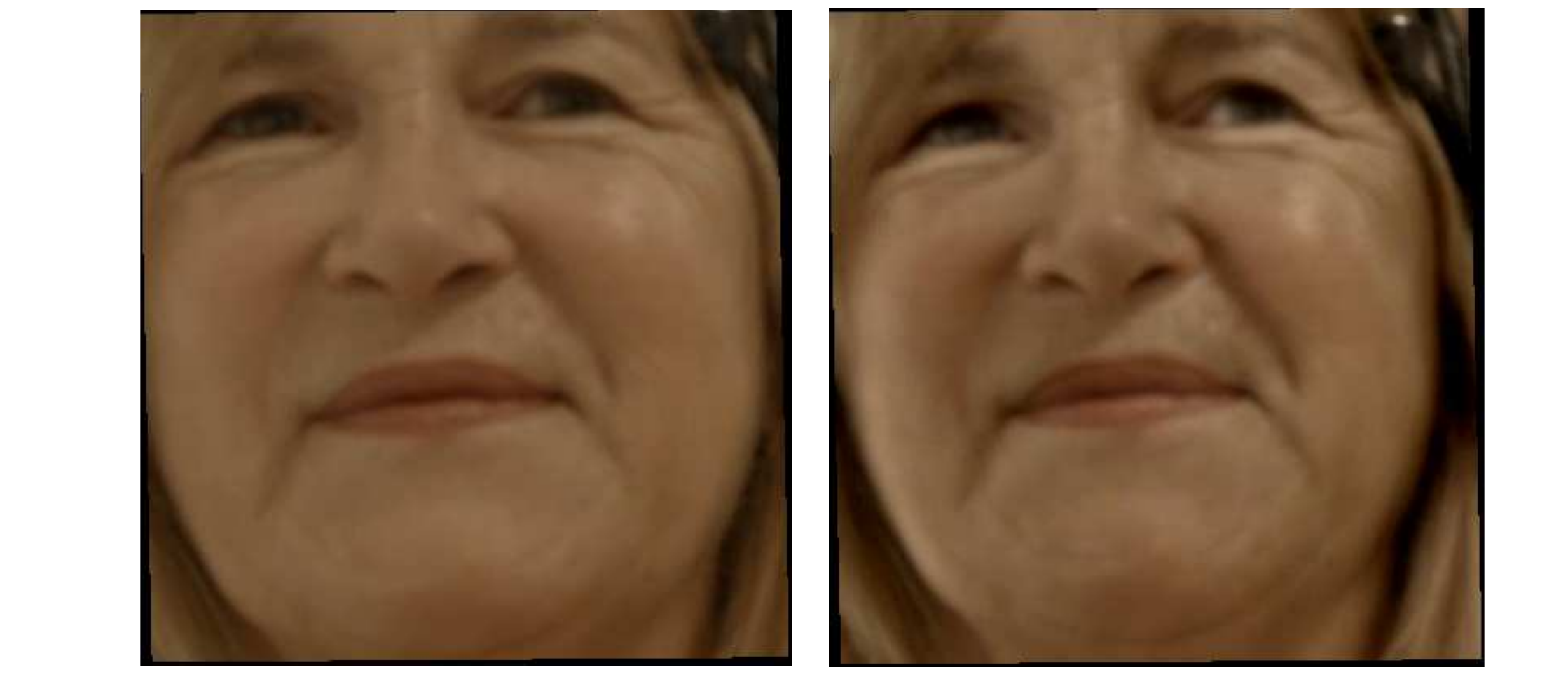}
  \caption{Eulerian Magnification applied to one of SEMdb RGB images where the participant was watching long distance memory autobiographical stimuli.}
  \label{fig:magni}
\end{figure}

The Eulerian Video Magnification is applied in order to augment the motion perception on the images by amplifying the colour variation in a given temporal frequency band at diverse spatial locations (see Figure \ref{fig:magni}). The Eulerian Video Magnification works as follows: By representing the image intensity at position (pixel) $x$ at time $t$ as $I(x,t)$ we can use a displacement function $\delta(t)$ to express any transactional motion in the image as $I(x,t)=f(x+\delta(t))$ where $I(x,0)=f(x)$. The first-order Taylor series expansion is then used to approximate $I(x,t)$ as
\begin{equation}\label{eq=magnificationTaylor}
I(x,t)\approx f(x) +\delta(x)\frac{\partial f(x)}{\partial x}.
\end{equation}
An appropriate broadband temporal bandpass filter is applied to $I(x,t)$ to cancel out $f(x)$ while keeping the $\delta(x)\frac{\partial f(x)}{\partial x}$ part (assuming that the motion $\delta(t)$ is within the bandpass filter)
\begin{equation}\label{eq=magnificationBandPass}
B(x,t)=\delta(x)\frac{\partial f(x)}{\partial x}.
\end{equation}
The output of the filter is amplified by $\alpha$ and added back to $I(x,t)$
\begin{equation}\label{eq=magnification}
I'(x,t)=I(x,t)+\alpha B(x,t)\approx f(x)+(1+\alpha)\delta(x)\frac{\partial f(x)}{\partial x}
\end{equation}
It can be easily observed from Equation (\ref{eq=magnification}) that the motion of the image $f(x)$ has been amplified by $(1+\alpha)$. This practical case (i.e., a small motion $\delta(t)$ that is entirely within the bandpass filter) demonstrates the concept of the Eulerian magnification. The broader applicability of the Eulerian magnification, in videos which includes bigger motions not entirely within the filter, has been demonstrated in~\cite{Wu2012}.

In the preprocessing phase different output sequences of images are generated based on combinations of different processing techniques as it is shown in block B of Figure \ref{fig:diagramOri}.

\textbf{SEC-1}: The first output comprises affine transformation of the original sequence of images which are then rescaled to 120$\times$120 pixels and converted to greyscale. This is given as input to the \hyperlink{a_pHOG}{pHOG} descriptor in the next phase.

\textbf{SEC-2}: The second output comprises Eulerian magnified images which are then affine transformed, rescaled to 120$\times$120 pixels and converted to greyscale. This is also given as input to the \hyperlink{a_pHOG}{pHOG} descriptor in the next phase.

\textbf{SEC-3}: In the third output, the Baltru et al.'s approach \cite{Baltru16} is used before and after the affine transformation to obtain the facial landmarks. Combining these outcomes with 3D face reconstruction methods ~\cite{Argyriou2009,jackson2017vrn} the proposed approach can be extended to support 3D representations. This is provided as input to the facial feature displacement descriptor and our proposed Lang's Universal Molecule based algorithm for novel origami features extraction in the next phase.

\textbf{SEC-4}: In the fourth output, the facial landmarks are obtained from the Eulerian magnified images, the affine transform is applied and the facial landmarks are obtained again. This is given as input to the facial feature displacement descriptor for feature extraction in the next phase.

\subsection{Feature Extraction Phase} \label{FeatExtPori}

Three schemes for features extraction have been applied in this phase to the different preprocessed data as shown in block C of the Figure \ref{fig:diagramOri}, i.e., (1) the pyramid Histogram of Gradients (pHOG), (2) a facial feature displacement descriptor based on~\cite{Michel03} and (3) a new descriptor which uses the Lang's Universal Molecule algorithm~\cite{Lang1996} to extract novel origami features. These schemes will be described in details in the following.

\textbf{\hyperlink{a_pHOG}{pHOG} features extraction}: The magnified affine transformed sequence (i.e., SEC-2) is provided as input to the \hyperlink{a_pHOG}{pHOG} descriptor. More specifically, eight bins on three pyramid levels of the \hyperlink{a_pHOG}{pHOG} are applied in order to obtain a row of $h_m$ features per sequence. This is the first set of novel features proposed in this work. For comparison purposes, the same process is applied to the unmagnified sequence (i.e., SEC-1) and a row of $h$ features per sequence is obtained accordingly.

\textbf{Facial feature displacement}: The magnified facial landmarks (i.e., fourth sequence output of the preprocessing phase) are normalized according to a fiducial face point (i.e., nose) to account for head motion in the video stream. In other words, the nose is used as a reference point, such that the position of all the facial landmarks are independent of the location of the subject's head in the images. If $L_i=[L_{i_x}\quad L_{i_y}]$ are the original image coordinates of the $i$-th landmark, and $L_n=[L_{n_x}\quad L_{n_y}]$ the nose landmark coordinates, the normalized coordinates are given by $l_i=[L_{i_x}-L_{n_x}\quad L_{i_y}-L_{n_y}]$.

The facial displacement features are vectors of landmark displacements between landmarks locations in a neutral and 'peak' frame representative of the expression. The displacement of the $i$-th landmark (i.e., $i$-the vector element) is calculated by the Euclidean distance
\begin{equation}
d(l_i^{(p)},l_i^{(n)})=\sqrt{(l_{i_x}^{(p)}-l_{i_x}^{(n)})^2- (l_{i_y}^{(p)}-l_{i_y}^{(n)})^2}
\end{equation}
between its normalized position in the neutral frame ($l_i^{(n)}$) and the "peak" frame ($l_i^{(p)}$). In the following we will be referring to these features as distance to nose (neutral vs peak) features (DTNnp). The output of the \hyperlink{a_DTNnp}{DTNnp} descriptor are $d_m$ long row vectors per image sequence.

For comparison purposes, the same process is applied to unmagnified facial landmark sequences (i.e., the third sequence of the preprocessing phase) and a row of $d$ features per sequence is obtained accordingly.

\subsubsection{Origami Features}
The origami features are created from the normalized facial landmarks (i.e., SEC-3). The descriptor is based on Robert J. Lang's Universal Molecule algorithm and uses $o$ facial landmarks in order to create an undirected graph of $n$ nodes and $e$ edges representing the facial crease pattern. The $n$ nodes contain the information of the $x$ and $y$ landmark coordinates while the $e$ edges contain the IDs of the two nodes connected by the corresponding edge (which represent the nodes/landmarks relationships). The facial crease pattern creation process is divided into three main steps: shadow tree, Lang's polygon and shrinking.

\begin{figure}
\centering
  \includegraphics[scale = 0.45]{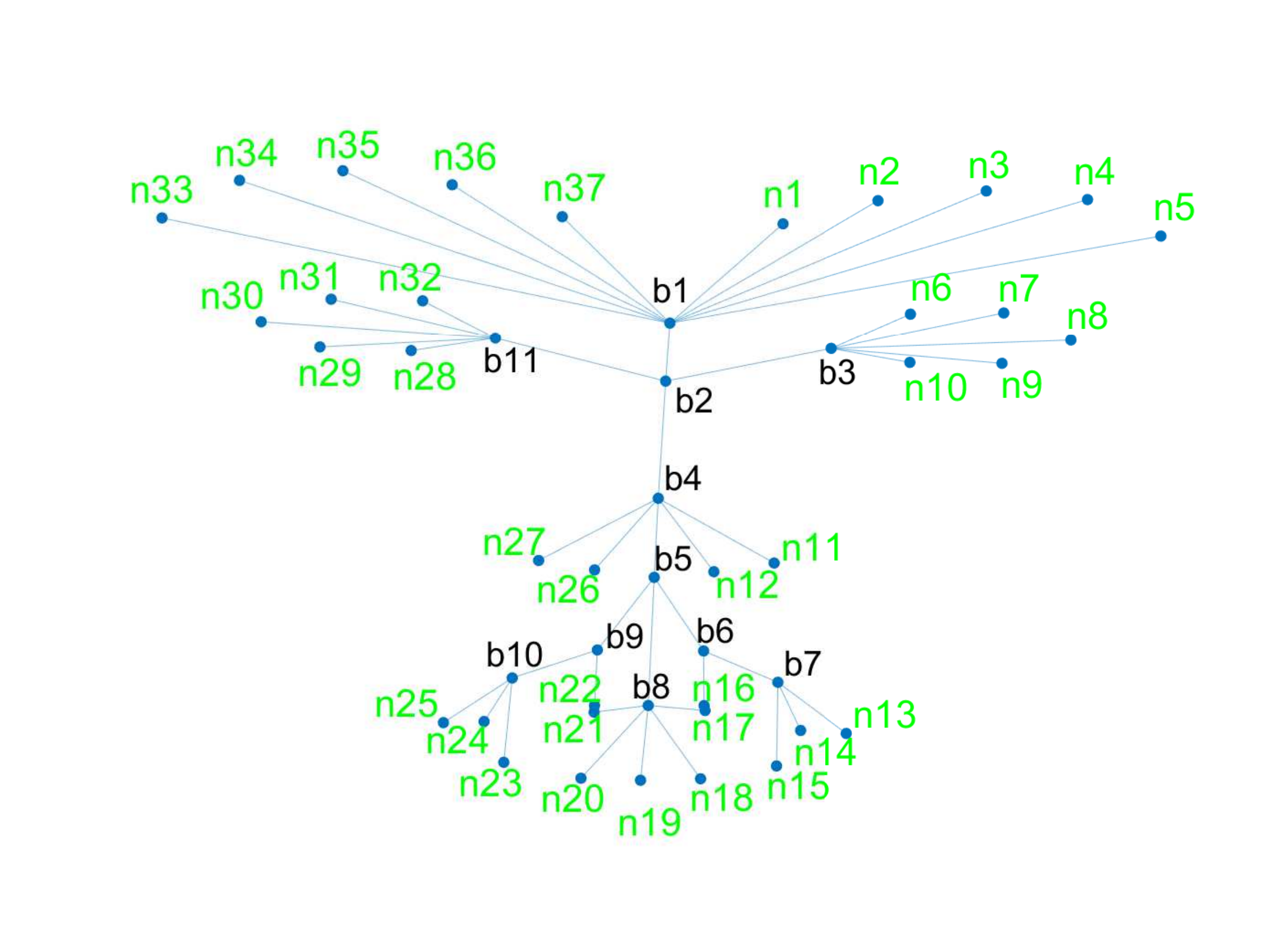}
  \caption{Shadow tree. The facial landmarks are linked in order each side of the tree is symmetric.}
  \label{fig:shadowTree}
\end{figure}

The first step implies the extraction of the flap projection of the face, the shadow tree, from the facial landmarks. This shadow tree or metric tree $(T,d)$ is composed of leaves (external nodes) $N_{ex} = {n_1,...,n_p}$, internal nodes $N_{in} = {b_1,...,b_q}$, edges $E$ and distances $d$ on the edges. This distance is the Euclidean distance between connected landmarks (nodes) through an edge in the tree. It is just a distance measured between each landmark that is going to be used during the Lang's Polygon creation and during the shrinking step as part of the conditions to trigger events. The shadow tree is created as a linked version of 2D facial landmarks of the eyebrows, eyes, nose and mouth (see Figure~\ref{fig:shadowTree}).

\begin{figure}
\centering
  \includegraphics[scale = 0.45]{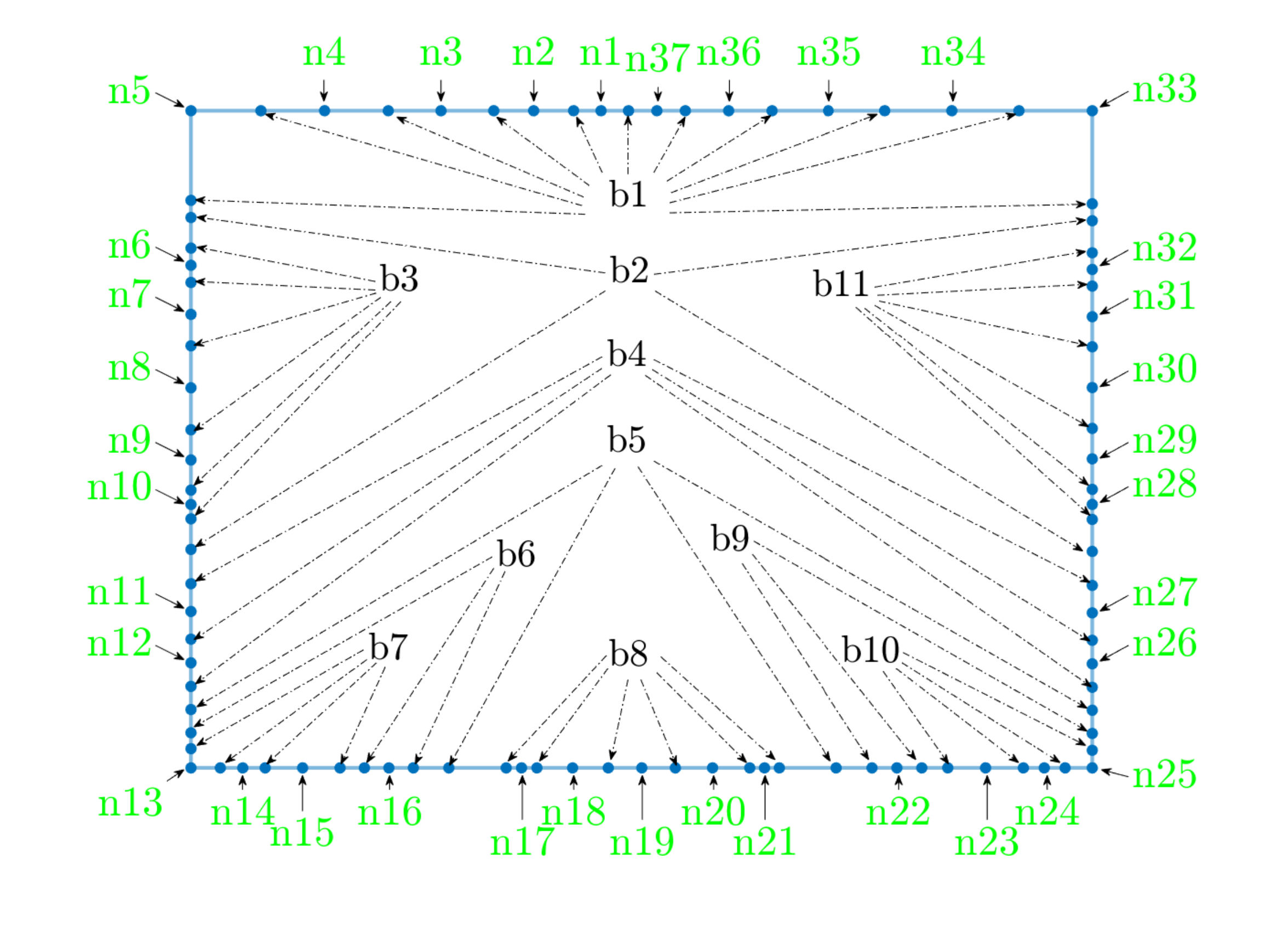}
  \caption{Our Lang's polygon. Based on Lang's polygon rules, a rectangle shaped initial Lang's polygon was created from the shadow tree in order all the faces start with the same initial polygon shape.}
  \label{fig:langPolygonSquare}
\end{figure}

During the second step, a convex doubling cycle polygon~\cite{Bowers2015}, Lang's polygon $L_p$, is created from the shadow tree $(T,d)$, following the procedure to create double cycling polygons $f$. This process, as explained in~\cite{Bowers2015}, consists in walking from leaf node to leaf node of the tree until reaching the initial node. For example, the path to go from $n_1$ to $n_2$ includes the path from $n_1$ to $b_1$ and from $b_1$ to $n_2$, the path from $n_2$ to $n_3$ also requires to pass through $b_1$ and the path from $n_5$ to $n_6$ goes through $b_1$, $b_2$ and $b_3$. In order the resultant polygon is convex, we shaped it as a rectangle (see Figure~\ref{fig:langPolygonSquare}) where the top side contains the eyebrows landmarks, the sides are formed by the eyes and nose landmarks and the mouth landmarks are at the bottom. This Lang's polygon represents the area of the paper that is going to be folded.

The resultant Lang's polygon is a convex polygonal region that has to satisfy the next condition~\cite{Lang2004,Bowers2015}: the distance between the leaf nodes $n_i$ and $n_j$ in the polygon ($d_P$) is equal or greater than the distance of those leaf nodes in the shadow tree ($d_T$). This property is mainly due to the origami properties, since a shadow tree comes from a folded piece of paper, once the paper is unfolded to see the crease pattern, the distances on the unfolded paper $d_P(n_i,n_j)$ are going to be always bigger or equal to the distances in the shadow tree $d_T(n_i,n_j)$ but never smaller.

\begin{equation}\label{eq=oriCondition}
   d_P(n_i,n_j)\geq d_T(n_i,n_j)
\end{equation}

\begin{figure}
\centering
  \includegraphics[scale = 0.55]{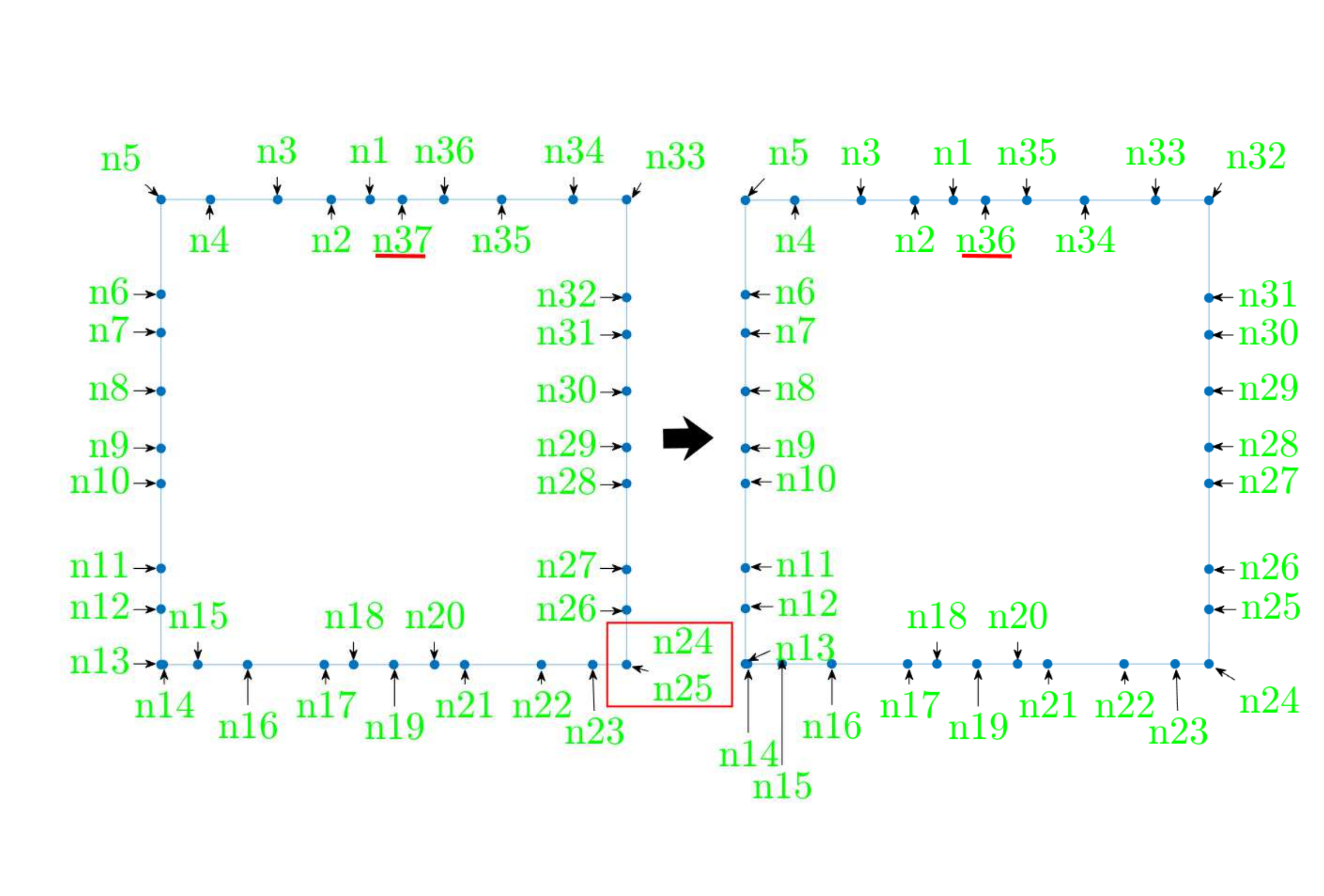}
  \caption{Contraction event. In this case, the distance between nodes 24 and 25 is 0 so one of the nodes is eliminated. Therefore, the number of leaf nodes is reduced to 36. Nodes 13 and 14 are close but not enough to trigger a contraction event.}
  \label{fig:contraction}
\end{figure}

The third step corresponds to the shrinking of the Lang's polygon. All the edges are simultaneously moved towards the centre of the polygon at constant speed until one of two possible events happens: contraction or splitting. The contraction event happens when the points collide in the same position (see Eq.~\ref{eq=contractCond}). In this case, only one point is kept and the shrinking process continues (see Figure~\ref{fig:contraction}).

\begin{equation}\label{eq=contractCond}
   d_P(n_i,n_j)\leq th, \quad \textrm{then} \quad n_i = n_j
\end{equation}

where $j=i+1$ and $th$ is a positive number $\approx0$.

\begin{figure}
\centering
  \includegraphics[scale = 0.55]{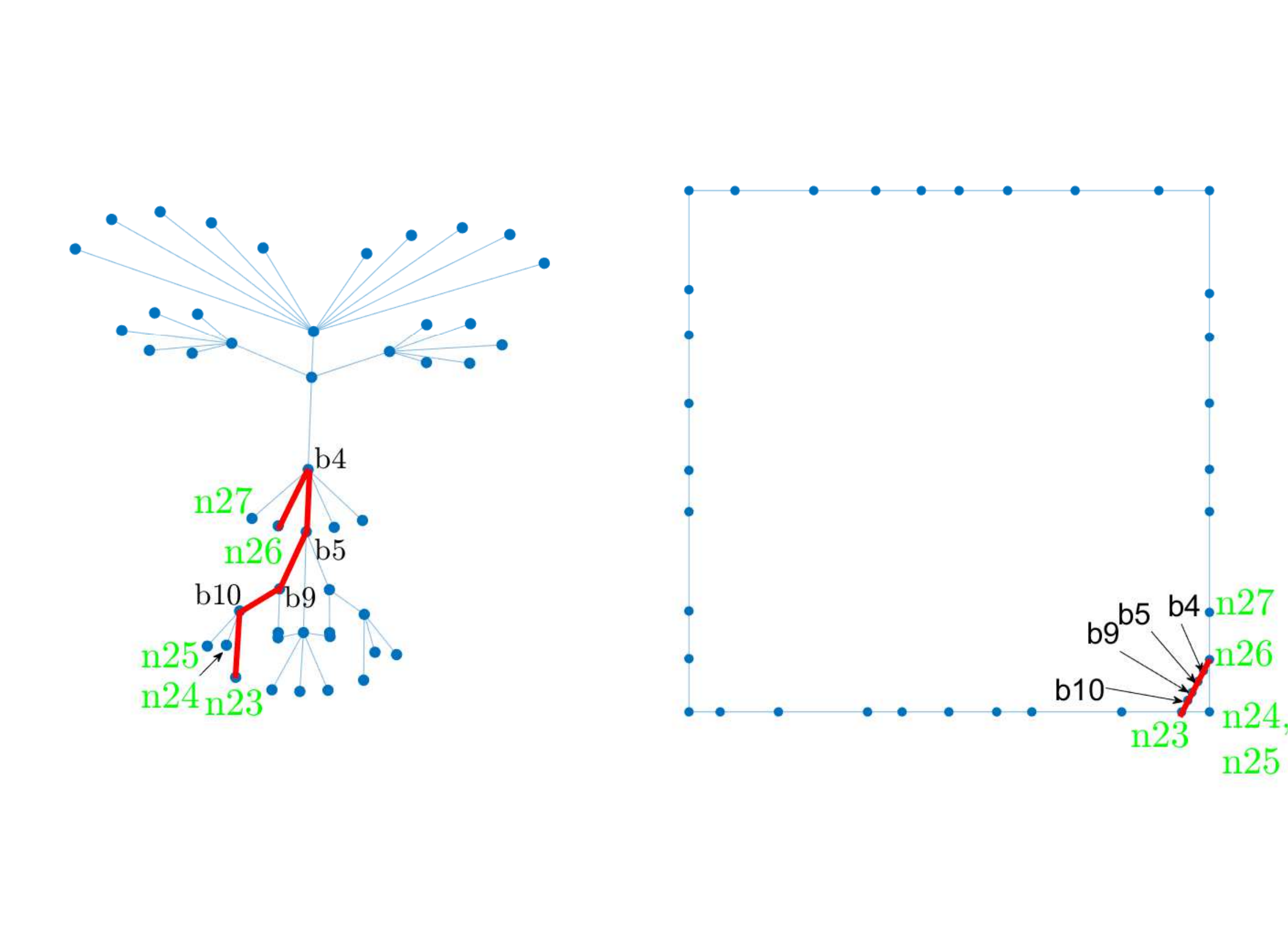}
  \caption{Splitting event. When the distance between two non-consecutive points in Lang's polygon (nodes 23 and 26) is the same as the distance of those two points in the shadow tree, a new edge is created. The intermediate nodes in the tree are also located in the new edge (nodes b4, b5, b9, b10).}
  \label{fig:splitting0}
\end{figure}

\begin{figure}
  \includegraphics[scale = 0.6]{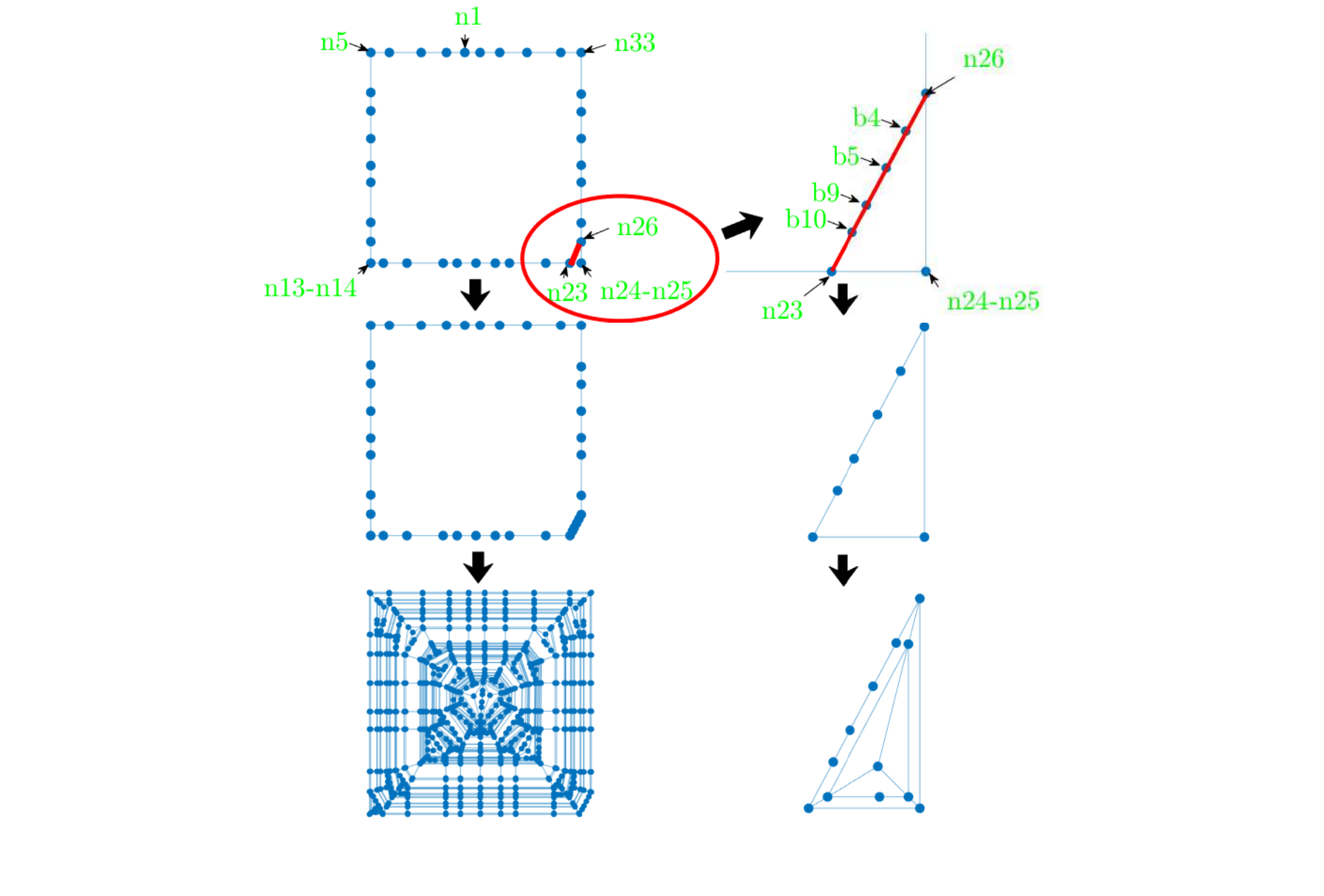}
  \caption{Splitting event. The new edge created due to the splitting event divides the polygon into two new polygons that are shrank independently. Since a new edge was created between nodes 23 and 26, the main polygon was divided in two. Due to this splitting event, the big polygon formed by all nodes except node 24-25 (Node 24-25 is a node formed from the contraction event of the nodes 24 and 25) and the small polygon formed by nodes 23, 24-25 and 26 are processed independently.}
  \label{fig:splitting1}
\end{figure}

The splitting event happens when the distance between two non-consecutive points is equal to their shadow tree distance (see Figure~\ref{fig:splitting0}).

\begin{equation}\label{eq=splitCond}
   d_P(n_i,n_k)\leq d_T(n_i,n_k) + th
\end{equation}

where $k\geq i+1$ and $th$ is a positive number $\approx0$. As a consequence of this event a new edge is created between these points creating two sub-polygons. The shrinking process continues on each polygon separately (see Figure~\ref{fig:splitting1}).

\begin{figure}
\centering
  \includegraphics[scale = 0.4]{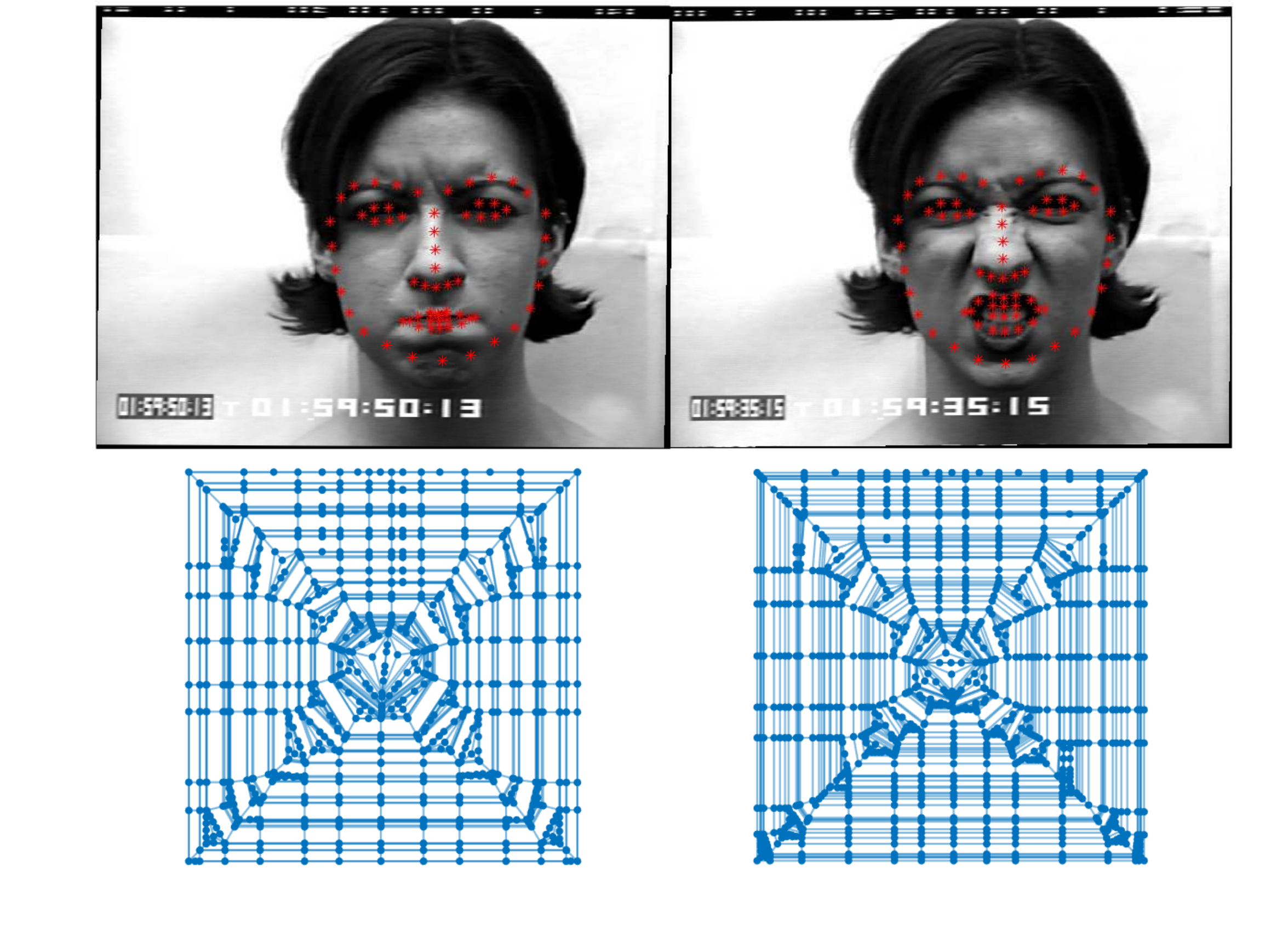}
  \caption{This image shows the crease patterns resultant from the universal molecule algorithm applied to our rectangular Lang's polygon produced from the landmarks of two faces. As the rectangular polygon is shrank the rectangle starts becoming smaller and splitting in the corners when the distance between points is equal to the distances in the shadow tree.}
  \label{fig:creasePattern}
\end{figure}

Finally, the process will finish when all the points converge, creating a crease pattern, $C = g(f({T,d}))$, where $(T,d)$ is the shadow tree, $f(T,d)$ is the Lang Polygon, $C$ is the crease pattern and $f$ and $g$ are the process to create the double cycle polygon and shrinking functions respectively. Due to the structure of our initial Lang's polygon, the final crease patterns will have an structure similar to the shown in Figure~\ref{fig:creasePattern}.

The crease pattern is stored as an undirected graph containing the coordinates of each node and the edges information (IDs of linked nodes). Due to the fact that $x$ and $y$ coordinates and the linked nodes are treated as vectorial features, these can be represented more precisely by a complex or hyper-complex representation, as shown in~\cite{Adali2011}. A vector can be decomposed into linearly independent components, in the sense that they can be combined linearly to reconstruct the original vector. However, depending on the phenomenon that changes the vector, correlation between the components may exist from the statistical point of view. If they are independent our proposed descriptor does not provide any significant advantage, but if there is correlation this is considered. In most of the cases during the feature extraction process complex or hyper-complex features are generated but decomposed to be computed by a classifier~\cite{Adali2011,Li2011}. In our case, the coordinates of the nodes are correlated and also the nodes id that represent the edges. Therefore, the nodes and edges are represented as show in~\ref{eq=nodes} and~\ref{eq=edges}.

\begin{equation}\label{eq=nodes}
   n = {n_i}_x + i* {n_i}_y
\end{equation}

\begin{equation}\label{eq=edges}
   e = {{e_i}\_nodeID_1 + i*{e_i}\_nodeID_2}
\end{equation}

\subsection{Classification Phase} \label{ClassiPhase}

Two state-of-the-art methods have been used in the third and last phase for emotion classification (see block D of Figure \ref{fig:diagramOri}). The first method is based on a Bayesian Compressed Sensing (BCS) classifier, and its improved version for action units detection, a Bayesian Group-Sparse Compressed Sensing (BGCS) classifier, similar to~\cite{Song2015}. The second method is similar to~\cite{Michel03} and is based on a quadratic \hyperlink{a_SVM}{SVM} classifier which has been successfully applied for emotion classification in~\cite{Buciu2003}.

More specifically, the output of the \hyperlink{a_pHOG}{pHOG} descriptor is used as input to the BCS/BGCS classifier and the output of the \hyperlink{a_DTNnp}{DTNnp} descriptor is provided as input to the quadratic \hyperlink{a_SVM}{SVM} classifier. Regarding the novel origami features, since their length is too long, the t-Distributed Stochastic Neighbor Embedding (tSNE) technique is applied to reduce their dimensionality. 

Finally, the novel features obtained in the previous phase are combined to create new feature vectors. More specifically, a feature vector composed by $h$ \hyperlink{a_pHOG}{pHOG} features, $h_m$ magnified pHOG features, together with $n$ and $e$ reduced origami features is provided as input to the BCS/BGCS classifiers:
\begin{equation}
features_{pHOG}=[h_1,...,h_r,h_{m1},...,h_{mr},n_1,...,n_u,e_1,...,e_u].
\end{equation}
On the other hand, a feature vector formed by $d$ \hyperlink{a_DTNnp}{DTNnp}, $d_m$ magnified \hyperlink{a_DTNnp}{DTNnp} features, combined with $n$ and $e$ reduced origami features is used as input to the quadratic \hyperlink{a_SVM}{SVM} classifier:
\begin{equation}
features_{DTN_{np}}=[d_1,...,d_r,d_{m1},...,d_{mr},n_1,...,n_u,e_1,...,e_u].
\end{equation}
Finally, a combination of \hyperlink{a_DTNnp}{DTNnp} and \hyperlink{a_pHOG}{pHOG} descriptors were tested on the quadratic \hyperlink{a_SVM}{SVM} classifier in order to check if a combination of both methods' features will boost the results. The combined feature vector is composed by $d$ \hyperlink{a_DTNnp}{DTNnp} features, $h_m$ magnified \hyperlink{a_pHOG}{pHOG} features, together with $n$ and $e$ reduced origami features.
\begin{equation}
features{combined}=[d_1,...,d_r,h_{m1},...,h_{mr},n_1,...,n_u,e_1,...,e_u].
\end{equation}

A summary of the different combinations of features used as input to the classifiers is presented in Table~\ref{tab:method}.

\begin{table*}[!t]
  \centering
  \caption{Methods, features and classes. This tables shows the features and classes recognised by the BCS/BGCS and SVM classifiers}
  \begin{threeparttable}
    \begin{tabular}{l|l|l|l|l}
    DB & \multirow{1}{0.12\linewidth}{\centering Method} & \multicolumn{ 2}{c|}{Features} & \multirow{ 1}{0.15\linewidth}{\centering Classes} \\
    \noalign{\smallskip}
    \hline
    \noalign{\smallskip}
    \multirow{ 10}{*}{CK+} & \multirow{ 4}{0.12\linewidth}{\centering \hyperlink{a_BCS}{BCS} \hyperlink{a_BGCS}{BGCS}~\cite{Song2015}} & orig & \hyperlink{a_pHOG}{pHOG}  & \multirow{ 4}{0.15\linewidth}{\centering 24 Action Units} \\
        \cline{3-4}
        & & \multirow{ 3}{*}{prop} & \hyperlink{a_pHOG}{pHOG} + Mag \hyperlink{a_pHOG}{pHOG}\tnote{2} & \\
        & & & \hyperlink{a_pHOG}{pHOG} + red Ori \tnote{3}  &\\
        & & & \hyperlink{a_pHOG}{pHOG} + Mag \hyperlink{a_pHOG}{pHOG} \tnote{2} + red Ori \tnote{3} &\\
        & & & \hyperlink{a_DTNnp}{DTNnp} + red Mag \hyperlink{a_pHOG}{pHOG} \tnote{4} &\\
        & & & \hyperlink{a_DTNnp}{DTNnp} + red Mag \hyperlink{a_pHOG}{pHOG} + red Ori & \\
        \noalign{\smallskip}
        \cline{2-5}
        \noalign{\smallskip}
        & \multirow{ 6}{0.12\linewidth}{\centering Quad \hyperlink{a_SVM}{SVM}~\ref{distanceTOnose}} & orig & \hyperlink{a_DTNnp}{DTNnp} \tnote{1} & \multirow{ 6}{0.15\linewidth}{\centering 7 emotions} \\ \cline{3-4}
        & & \multirow{ 5}{*}{prop} & \hyperlink{a_DTNnp}{DTNnp} + Mag \hyperlink{a_DTNnp}{DTNnp} & \\
        & & & \hyperlink{a_DTNnp}{DTNnp} + red Ori & \\
        & & & \hyperlink{a_DTNnp}{DTNnp} + Mag \hyperlink{a_DTNnp}{DTNnp} + red Ori &\\
        & & & \hyperlink{a_DTNnp}{DTNnp} + red Mag \hyperlink{a_pHOG}{pHOG} &\\
        & & & \hyperlink{a_DTNnp}{DTNnp} + red Mag \hyperlink{a_pHOG}{pHOG} + red Ori & \\
        \noalign{\smallskip}
        \hline
        \noalign{\smallskip}
    \multirow{ 10}{*}{SEMdb}& \multirow{ 4}{0.12\linewidth}{\centering \hyperlink{a_BCS}{BCS}~\cite{Song2015}} & orig &  \hyperlink{a_pHOG}{pHOG} & \multirow{ 6}{0.15\linewidth}{\centering 4 autobiographical memory classes}\\ \cline{3-4}
        & & \multirow{ 3}{*}{prop} & \hyperlink{a_pHOG}{pHOG} + Mag \hyperlink{a_pHOG}{pHOG} & \\
        & & & \hyperlink{a_pHOG}{pHOG} + red Ori & \\
        & & & \hyperlink{a_pHOG}{pHOG} + Mag \hyperlink{a_pHOG}{pHOG} + red Ori &\\
        & & & \hyperlink{a_DTNnp}{DTNnp} + red Mag \hyperlink{a_pHOG}{pHOG} &\\
        & & & \hyperlink{a_DTNnp}{DTNnp} + red Mag \hyperlink{a_pHOG}{pHOG} + red Ori & \\
        \noalign{\smallskip}
        \cline{2-5}
        \noalign{\smallskip}
        & \multirow{ 6}{0.12\linewidth}{\centering Quad \hyperlink{a_SVM}{SVM}~\ref{distanceTOnose}} & orig & \hyperlink{a_DTNnp}{DTNnp} & \multirow{ 6}{0.15\linewidth}{\centering 4 autobiographical memory classes} \\ \cline{3-4}
        & & \multirow{ 5}{*}{prop} & \hyperlink{a_DTNnp}{DTNnp} + Mag \hyperlink{a_DTNnp}{DTNnp} & \\
        & & & \hyperlink{a_DTNnp}{DTNnp} + red Ori & \\
        & & & \hyperlink{a_DTNnp}{DTNnp} + Mag \hyperlink{a_DTNnp}{DTNnp} + red Ori & \\
        & & & \hyperlink{a_DTNnp}{DTNnp} + red Mag \hyperlink{a_pHOG}{pHOG} & \\
        & & & \hyperlink{a_DTNnp}{DTNnp} + red Mag \hyperlink{a_pHOG}{pHOG} + red Ori & \\
        \noalign{\smallskip}
        \hline
        \noalign{\smallskip}
    \end{tabular}%
    \begin{tablenotes}
    \item[1] \footnotesize \hyperlink{a_DTNnp}{DTNnp} is the Distance to nose (neutral vs peak)
    \item[2] \hyperlink{a_pHOG}{pHOG} features extracted from the magnified images
    \item[3] Origami features whose dimensionality was reduced using \hyperlink{a_tSNE}{t-SNE}
    \item[3] \hyperlink{a_tSNE}{t-SNE} reduced \hyperlink{a_pHOG}{pHOG} features extracted from the magnified images
    \end{tablenotes}
    \end{threeparttable}
  \label{tab:method}%
\end{table*}%

\section{Results}

Data from two different datasets (CK+ and \hyperlink{a_SEMdb}{SEMdb}) is used to validate the classification performance of the methods by calculating the classification accuracy and F1 score. The CK+ dataset contains 593 sequences of images from 123 subjects. Each sequence starts with a neutral face and ends with the peak stage of an emotion. The CK+ contains \hyperlink{a_AUs}{AU} labels for all the sequences but basic emotion labels for only 327. \hyperlink{a_SEMdb}{SEMdb} contains 810 recordings from nine subjects. The start of each recording is taken as a neutral face and the peak expression is obtained as explained in Chapter~\ref{Chapter4}. The peak frame is the frame whose landmarks vary most from the respective landmarks in the neutral face. \hyperlink{a_SEMdb}{SEMdb} contains labels for four classes related to autobiographical memories. These autobiographical memory classes are represented by spontaneous facial micro-expressions triggered by the observation of four different stimulations related to distant and recent autobiographical memories.

In order to validate the classification performance of the proposed novel features the accuracy and the F1 score are calculated. K-fold cross-validation is applied in order to prevent overfitting, therefore the final accuracies and F1 scores shown in this paper are the average of all the folds.

\begin{equation}\label{eq=accu}
  Accuracy = \frac{TP+TN}{TP+FP+FN+TN}
\end{equation}
\begin{equation}\label{eq=f1}
  F1 Score = 2* \frac{(Recall * Precision)}{ (Recall + Precision)}
\end{equation}

where  $Precision = \frac{TP}{TP+FP}$; $Recall = \frac{TP}{TP+FN}$ and $TP$ are the True Positives, $FP$ are the False Positives, $TN$ are the True Negatives and $FN$ are the False Negatives.

The purpose of our experiments is to prove the capacity of the new features to improve state-of-the-art results. Therefore, the results obtained using the features and classifiers from the state-of-the-art methods are compared with the ones obtained when the novel features are added.

Different set of classes (action units, emotions and reactions to autobiographical memories) will be classified according to the method and database used. When the \hyperlink{a_BCS}{BCS}/\hyperlink{a_BGCS}{BGCS} methods and CK+ database are used~\cite{Song2015,Lucey2010}, twenty-four action units are classified; but when the \hyperlink{a_BCS}{BCS}/\hyperlink{a_BGCS}{BGCS} methods are used with \hyperlink{a_SEMdb}{SEMdb} four classes related with autobiographical memories do. Nevertheless, when the \hyperlink{a_BCS}{BCS}/\hyperlink{a_BGCS}{BGCS} methods and \hyperlink{a_SEMdb}{SEMdb} are used, only \hyperlink{a_BCS}{BCS} is utilised since \hyperlink{a_BGCS}{BGCS} method is an optimisation of \hyperlink{a_BCS}{BCS} to detect action units by exploiting the sparsity and co-occurrence in action unit space. Therefore, since \hyperlink{a_SEMdb}{SEMdb} does not provide action units for classification, the use of \hyperlink{a_BGCS}{BGCS} is discarded. The quadratic \hyperlink{a_SVM}{SVM} classifier, whose input is the CK+ features is used to detect seven emotions (1=anger, 2=contempt, 3=disgust, 4=fear, 5=happy, 6=sadness, 7=surprise). Whereas, if its input is \hyperlink{a_SEMdb}{SEMdb} data, it is used to detect the four autobiographical memory classes. Table~\ref{tab:method} shows a summary of the features and methods used and classes to be recognised.

Before the analysis of the novel features for emotion classification on different databases a proof of concept example has been conducted to evaluate the novel origami features resistance to noise. Gaussian noise has been applied to CK+ database images on three different areas: the whole image, the top half and the bottom half of the face. Afterwards, twenty-four action units were classified using \hyperlink{a_BCS}{BCS} classifier and \hyperlink{a_pHOG}{pHOG} features and adding a \hyperlink{a_tSNE}{t-SNE} dimensionally reduced version of the origami features. Table~\ref{tab:noise} shows the improvement with noise of the results when adding the novel origami features.

\setlength{\tabcolsep}{4pt}
\begin{table}[htbp]
  \centering
  \caption{Origami features noise resistance. Gaussian noise was applied to the CK+ database images and pHOG and Origami features were extracted. Using only pHOG and pHOG plus origami features, BCS classifier was used to classify 24 action units. This table shows the classification F1 score and accuracy.}
  \begin{threeparttable}
    \begin{tabular}{r|r|r|r|r|r}
    \hline
    \noalign{\smallskip}
     \multirow{ 3}{*}{Noise Location} & \multirow{ 3}{*}{Classes} & \multirow{ 3}{*}{Method} & \multirow{ 3}{*}{Measure} & \multicolumn{ 2}{c}{Features}\\
     \cline{5-6}
     & &  &  & \multicolumn{1}{c|}{\hyperlink{a_pHOG}{pHOG}\tnote{1}} & \multicolumn{1}{c}{\hyperlink{a_pHOG}{pHOG}} \\
     & &  &  &  &  \textbf{red Ori\tnote{2}}\\
    \noalign{\smallskip}
    \hline
    \noalign{\smallskip}
    \multirow{ 2}{*}{Whole} & \multirow{ 2}{*}{24 \hyperlink{a_AUs}{AUs}} & \multirow{ 2}{*}{\hyperlink{a_BCS}{BCS}}  & F1   & 0.559  & \textbf{0.568}\\
     & &  & ACC  & 0.883 & \textbf{0.886}\\
     \noalign{\smallskip}
     \hline
     \noalign{\smallskip}
     \multirow{ 2}{*}{Top} & \multirow{ 2}{*}{24 \hyperlink{a_AUs}{AUs}} & \multirow{ 2}{*}{\hyperlink{a_BCS}{BCS}}  & F1   & 0.601 & \textbf{0.603}\\
     & &  & ACC  & 0.89 & \textbf{0.891}\\
     \noalign{\smallskip}
     \hline
     \noalign{\smallskip}
     \multirow{ 2}{*}{Bottom} & \multirow{ 2}{*}{24 \hyperlink{a_AUs}{AUs}} & \multirow{ 2}{*}{\hyperlink{a_BCS}{BCS}}  & F1   & 0.607 & \textbf{0.608}\\
     & &  & ACC  & 0.89 & \textbf{0.892} \\
     \noalign{\smallskip}
     \hline
   \end{tabular}%
   \begin{tablenotes}
    \item[1] Pyramid HOG
    \item[2] \hyperlink{a_tSNE}{t-SNE} reduced Origami features combined with \hyperlink{a_pHOG}{pHOG}
   \end{tablenotes}
  \end{threeparttable}
  \label{tab:noise}%
\end{table}%
\setlength{\tabcolsep}{1.4pt}

The next paragraphs explain the obtained results ordered by the objective classes (\hyperlink{a_AUs}{AUs}, seven basic emotions and four autobiographical emotions). Each table contains the accuracy and F1 scores which represent the performance of the emotion classifiers before and after the novel features are added.

The experiments are compared with results obtained using two state of the art methods: Bayesian Group-Sparse Compressed Sensing~\cite{Song2015} and landmarks to nose \ref{distanceTOnose}. Song et al. method compares two classification algorithms, i.e., the Bayesian Compressed Sensing (BCS) and their proposed improvement Bayesian Group-Sparse Compressed Sensing (BGCS). Both classifiers are used to detect 24 Action Units within the CK+ dataset using \hyperlink{a_pHOG}{pHOG} features. Landmarks to nose methods consist on using the landmarks distance to nose difference between the peak and neutral frame \ref{distanceTOnose} as input of an \hyperlink{a_SVM}{SVM} classifier to classify seven basic emotions in ~\cite{Michel03} and four autobiographical emotions in \ref{autobioEmo}.

\begin{figure}
\centering
  \includegraphics[scale = 0.5]{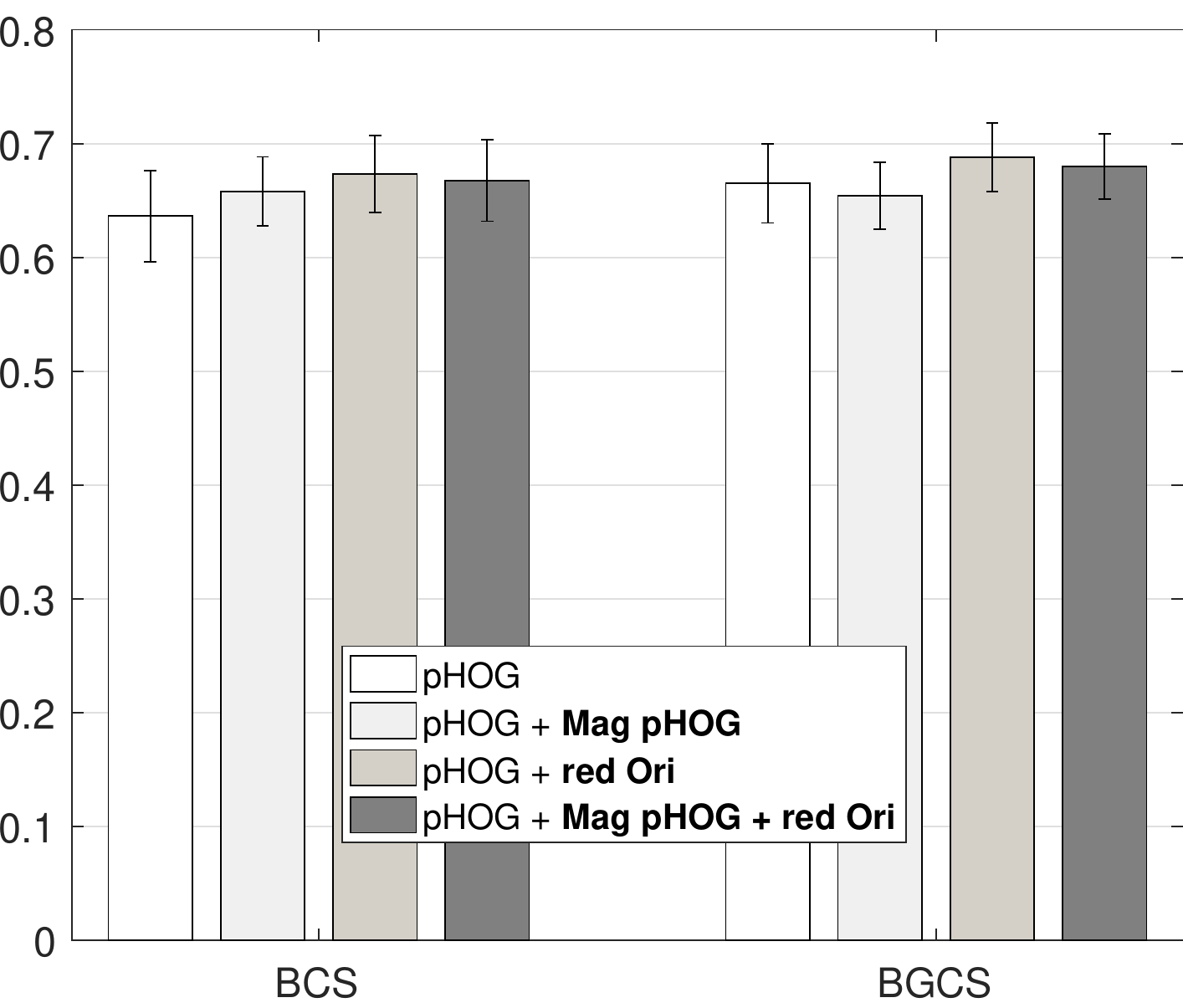}
  \caption{F1 score results of the 24 AUs classification using the state of the art methods BCS and BGCS and the new results when the novel features are added.}
  \label{fig:barsBCSBGCSresults}
\end{figure}

The 24 \hyperlink{a_AUs}{AUs} detection experiment involved the \hyperlink{a_BCS}{BCS} and the \hyperlink{a_BGCS}{BGCS} classifiers and CK+ database. The input features utilised included the state of the art ones and them combined with our proposed ones. Therefore, for the \hyperlink{a_AUs}{AUs} experiment the \hyperlink{a_pHOG}{pHOG} were tested independently and combined with the proposed magnified \hyperlink{a_pHOG}{pHOG} and origami features; and identically with the distance to nose features. The result of the different combinations are shown in Figure~\ref{fig:barsBCSBGCSresults}, Table~\ref{tab:BCSCK+} and Table~\ref{tab:BCScombCK}. They show that the contribution of the new descriptors improves the F1 score and the overall accuracy.

\begin{table}[!t]
  \centering
  \caption{Mean of k-fold F1 score and Accuracy values. 24 AU from CK+ dataset are classified using 4 combinations of features and BCS and BGCS classifier}
  \setlength\tabcolsep{2pt}
    \begin{tabular}{|r|r|r|r|r|r|}
    \noalign{\smallskip}
    \hline
    \noalign{\smallskip}
     \multirow{ 4}{*}{Method} & \multirow{ 4}{*}{Meas.} &  \multicolumn{ 4}{c|}{Features}\\  \cline{3-6}
     &  & \multicolumn{1}{c|}{pHOG} & \multicolumn{1}{c|}{pHOG}  & \multicolumn{1}{c|}{pHOG}   & \multicolumn{1}{c|}{pHOG} \\
     &  &  &  \textbf{red Ori} &  \textbf{Mag pHOG} &  \multicolumn{1}{c|}{\textbf{Mag pHOG}}\\
     &  &  &   &   &  \multicolumn{1}{c|}{\textbf{red Ori}}\\
    \noalign{\smallskip}
    \hline
    \noalign{\smallskip}
    \multirow{ 2}{*}{BCS}  & F1   & 0.636 & 0.658 & \textbf{ 0.673} & 0.667\\
      & ACC  & 0.901 & 0.909 & 0.909 & \textbf{0.912}\\
    \noalign{\smallskip}
    \hline
    \noalign{\smallskip}
    \multirow{ 2}{*}{BGCS} & F1   & 0.664 & 0.654 & \textbf{ 0.687} & 0.679\\
     & ACC  & 0.906 & 0.908 & \textbf{ 0.913} & 0.91 \\
    \noalign{\smallskip}
    \hline
    \noalign{\smallskip}
   \end{tabular}%
  \label{tab:BCSCK+}%
\end{table}%

\begin{table}[!t]
  \centering
  \caption{Mean of k-fold F1 score and Accuracy values. 24 Action Units from CK+ dataset are classified. A combination of the DTNnp~\ref{distanceTOnose}, the pHOG~\cite{Song2015} extracted from the magnified version of the data and the novel origami features are used as input to the BCS/BGCS classifiers.}
  \setlength\tabcolsep{2pt}
    \begin{tabular}{|r|r|r|r|r|}
    \hline
    \multirow{ 4}{*}{Method} & \multirow{ 4}{*}{Meas.} &  \multicolumn{ 3}{c|}{Features}\\  \cline{3-5}
      &  &  \multicolumn{1}{c|}{DTNnp} &  \multicolumn{1}{c|}{DTNnp} &  \multicolumn{1}{c|}{DTNnp} \\
    &  &  & \textbf{Mag pHOG} & \multicolumn{1}{c|}{\textbf{Mag pHOG}}\\
    &  &  &  &\multicolumn{1}{c|}{\textbf{red Ori}}\\
    \noalign{\smallskip}
    \hline
    \noalign{\smallskip}
    \multirow{ 2}{*}{BCS}  & F1   & 0.407 & \textbf{0.671} & 0.663\\
      & ACC  & 0.873 & 0.913 & \textbf{0.914}\\
    \noalign{\smallskip}
    \hline
    \noalign{\smallskip}
    \multirow{ 2}{*}{BGCS}  & F1   & 0.555 & \textbf{0.69} & 0.682\\
      & ACC  & 0.893 & \textbf{0.915} & 0.914\\
    \noalign{\smallskip}
    \hline
    \noalign{\smallskip}

    \end{tabular}%
  \label{tab:BCScombCK}%
\end{table}%

The second experiment's objective was the classification of seven basic emotions using a quadratic \hyperlink{a_SVM}{SVM} classifier, the distance to nose features combined with the proposed ones and the CK+ database. The results shown in Table~\ref{tab:SVMcombCK} the quadratic \hyperlink{a_SVM}{SVM} ones. The combination of the features did not provide any noticeable boost in the accuracy. The increment in the F1 score is not big enough to be taken into account.

\begin{table}[!t]
  \centering
  \caption{Mean of k-fold F1 score and Accuracy values. 7 emotion classes from CK+ dataset are classified. A combination of the DTNnp~\ref{distanceTOnose}, the pHOG~\cite{Song2015} extracted from the magnified version of the data and the novel origami features are used as input to the quadratic SVM classifier.}
  \setlength\tabcolsep{2pt}
    \begin{tabular}{|r|r|r|r|r|}
    \noalign{\smallskip}
    \hline
    \noalign{\smallskip}
    \multirow{ 4}{*}{Method} & \multirow{ 4}{*}{Meas.} &  \multicolumn{ 3}{c|}{Features}\\  \cline{3-5}
      &  &  \multicolumn{1}{c|}{DTNnp} &  \multicolumn{1}{c|}{DTNnp} &  \multicolumn{1}{c|}{DTNnp} \\
    &  &  &  \textbf{red Mag pHOG} &  \multicolumn{1}{c|}{\textbf{red Mag pHOG}}\\
    &  &  &  &  \multicolumn{1}{c|}{\textbf{red Ori}}\\
    \noalign{\smallskip}
    \hline
    \noalign{\smallskip}
    \multirow{ 2}{*}{qSVM}  & F1   & 0.861 & 0.857 & \textbf{0.865}\\
      & ACC  & \textbf{0.899} & 0.893 & \textbf{0.899}\\
    \noalign{\smallskip}
    \hline
    \noalign{\smallskip}

    \end{tabular}%
  \label{tab:SVMcombCK}%
\end{table}%

Table~\ref{tab:confMatrixSVMck} shows the confusion matrix of the emotions classified for the best F1 score experiments. This confusion matrix shows that fear is the emotion with poorer rate of success and surprise and happiness are the most accurately recognised.

\begin{table}[!t]
  \centering
  \caption{The obtained confusion matrix for the 7 emotions present in the CK+ dataset.}
  \setlength\tabcolsep{1pt}
    \begin{tabular}{l|ccccccc}
          & anger & contempt & disgust & fear & happy & sadness & surprise \\ \hline
    anger &38  &  2  &  3  &  0  &  0  &  2  &  0 \\
    contempt & 1  &  15   &  0   &  0  &   1   &  1  &  0 \\
    disgust & 5  &  0  &  54   &  0   &  0  &   0   &  0 \\
    fear & 0  &  0  &   0   & 19   &  4  &   0   &  2 \\
    happy & 0  &  2  &   0   &  1   & 66  &   0   &  0 \\
    sadness & 3  &  0  &   1   &  1   &  1  &  22   &  0 \\
    surprise & 0  &  2  &   0   &  0   &  1  &   0   & 80 \\
    \end{tabular}%
  \label{tab:confMatrixSVMck}%
\end{table}%

The third experiment (Table~\ref{tab:BCSSEMdb}) involved the \hyperlink{a_BCS}{BCS} and quadratic \hyperlink{a_SVM}{SVM} classifiers, using the \hyperlink{a_SEMdb}{SEMdb} dataset, \hyperlink{a_pHOG}{pHOG} features and a combination of \hyperlink{a_pHOG}{pHOG} and distance to nose features (both combined with the proposed ones) and detecting the corresponding four classes that are related to autobiographical memories. Table \ref{tab:BCSSEMdb} show that the novel descriptors applied to \hyperlink{a_pHOG}{pHOG} features help to improve both accuracy and F1 scores in the origami case, the magnification features increase the accuracy and their combination the overall F1 score. The results of the combination of both state of the art features (\hyperlink{a_DTNnp}{DTNnp} and \hyperlink{a_pHOG}{pHOG}) is presented in Table~\ref{tab:BCSmixSEMdb}. It shows results corresponding to the \hyperlink{a_BCS}{BCS} classifiers and to the quadratic \hyperlink{a_SVM}{SVM}. Both classifiers (\hyperlink{a_BCS}{BCS} and \hyperlink{a_SVM}{SVM}) provide improved classification estimates when the combination of all features is used to detect the four autobiographical memory classes.

\begin{table}[!t]
  \centering
  \caption{Mean of k-fold F1 score and Accuracy values. 4 classes from SEMdb dataset related with short/long term memories are classified using 4 combinations of features and BCS classifier.}
  \setlength\tabcolsep{2pt}
    \begin{tabular}{|r|r|r|r|r|r|}
    \hline
    \multirow{ 4}{*}{Method} & \multirow{ 4}{*}{Meas.} &  \multicolumn{ 4}{c|}{Features}\\  \cline{3-6}
     &  & \multicolumn{1}{c|}{pHOG} & \multicolumn{1}{c|}{pHOG}  & \multicolumn{1}{c|}{pHOG}  & \multicolumn{1}{c|}{pHOG} \\
     &  &  &  \textbf{red Ori} &  \textbf{Mag pHOG} &  \multicolumn{1}{c|}{\textbf{Mag pHOG}}\\
     &  &  &  &  &  \multicolumn{1}{c|}{\textbf{red Ori}}\\
    \noalign{\smallskip}
    \hline
    \noalign{\smallskip}
    \multirow{ 2}{*}{BCS}  & F1   & 0.389 & \textbf{ 0.401} & 0.388 & 0.393\\
      & ACC  & 0.741 & \textbf{0.742} & \textbf{0.746} & 0.739\\
    \noalign{\smallskip}
    \hline
    \noalign{\smallskip}
    \end{tabular}%
  \label{tab:BCSSEMdb}%
\end{table}%

\begin{table}[!t]
  \centering
  \caption{Mean of k-fold F1 score and Accuracy values. 4 autobiographical memory classes from SEMdb dataset are classified. A combination of the DTNnp~\ref{distanceTOnose}, the pHOG~\cite{Song2015} extracted from the magnified version of the data and the novel origami features are used as input to the BCS classifier.}
  \setlength\tabcolsep{2pt}
    \begin{tabular}{|r|r|r|r|r|}
    \hline
    \multirow{ 4}{*}{Method} & \multirow{ 4}{*}{Meas.} &  \multicolumn{ 3}{c|}{Features}\\  \cline{3-5}
     &  & \multicolumn{1}{c|}{DTNnp} & \multicolumn{1}{c|}{DTNnp} & \multicolumn{1}{c|}{DTNnp} \\
    &  &  &  \textbf{Mag pHOG} &  \multicolumn{1}{c|}{\textbf{Mag pHOG}}\\
    &  &  &  &  \multicolumn{1}{c|}{\textbf{red Ori}}\\
    \noalign{\smallskip}
    \hline
    \noalign{\smallskip}
    \multirow{ 2}{*}{BCS}  & F1   & 0.375 &  \textbf{0.384} & 0.383\\
      & ACC  & 0.727 & \textbf{0.73} & \textbf{0.73}\\
    \noalign{\smallskip}
    \hline
    \noalign{\smallskip}
    \multirow{ 2}{*}{qSVM}  & F1   & 0.725 & 0.729 & \textbf{0.739}\\
      & ACC  & 0.725 & 0.731 & \textbf{0.739}\\
    \noalign{\smallskip}
    \hline
    \noalign{\smallskip}

    \end{tabular}%
  \label{tab:BCSmixSEMdb}%
\end{table}%

To summarise, the results have proved that the proposed descriptors are improving the overall performance of the emotion classification methods. For example, the F1 score for the detection of \hyperlink{a_AUs}{AUs} has been increased from $66.4\%$ to $67.9\%$ (See Table \ref{tab:BCSCK+}), for the detection of seven basic emotions has been increased from $86.1\%$ to $86.5\%$ (See Table \ref{tab:SVMcombCK}) and for the detection of four autobiographical reactions has been increased from $72.5\%$ to $73.9\%$ (See Table \ref{tab:BCSmixSEMdb}).

\section{Discussion}

Human emotion analysis is widely studied in different areas. Current applications are requiring the addition of the best computational emotion recognition systems to provide novel functionalities. For example, \hyperlink{a_AI}{AI} assistants are currently being available to help people on their daily life and the addition of emotion recognition will move those \hyperlink{a_AI}{AI} systems a step ahead. Different sources of data are used to detect human emotions such as face images, audio sources or electroencephalogram. Since facial expressions are linked directly with human emotions, they are the most widely studied for emotion recognition. In these circumstances \hyperlink{a_EEG}{EEG} techniques for emotions recognition are more convenient since they can provide feedback from activities such as working memory, emotions, decision-making or mental workload. On the other hand, despite their emotion classification performance is swiftly improving over the years, currently the facial emotion approaches are still superior. In addition, the necessity of wearable devices from \hyperlink{a_EEG}{EEG} approaches makes facial expression techniques more adequate. Emotion recognition rates using Facial expression approaches vary depending on the used databases, being good in posed databases and spontaneous databases recorded on controlled scenarios and improvable in micro-expression and data in the wild.

In our work we present the improvement that novel preprocessing techniques and descriptors can provide in the classification of emotions through facial expressions and micro-expressions from facial images or videos, including those partially occluded or affected by noise. Our study proposes the use of Eulerian magnification in the preprocessing stage and a novel origami algorithm in the feature extraction stage. Results show that the addition of these techniques can help to increase the classification accuracy. The Eulerian magnification preprocessing generally helps to improve the performance more than the origami features.


\chapter{Conclusion} 

\label{Chapter6} 

\lhead{Chapter 6. \emph{Conclusion}} 


\section{Conclusion and Future Work}
Alzheimer's Disease early detection is essential for patients and family to receive practical information. Early treatment allow to slow down cognitive decline and to deal properly with symptoms such as depression. Early planning will also adjust the diagnosis for the individual and reduce societal costs. Currently, \hyperlink{a_AD}{AD} numbers in cost and of people affected are huge and they continue increasing every year. Therefore, the optimization of \hyperlink{a_AD}{AD} detection methods is indispensable to be able to diagnose \hyperlink{a_AD}{AD} as soon as possible. The most reliable tests, such as \hyperlink{a_MRI}{MRI} scans or biomarkers, are invasive, uncomfortable or expensive. Despite the validity of some non-invasive affordable methods has been proved for \hyperlink{a_AD}{AD} screening by many research works, most of them concur about the existence of a margin for improvement and about problems derived from the different IQ levels of the participants. Ceiling effect produces false negatives in patients with high IQ since the tests are usually too simple for them. Cognitive reserve is also related with this ceiling effect. It is the capacity of the brain to compensate damaged areas and it is usually higher in high IQ subjects. Finally, another frequent problem with common cognitive test is their tasks are usually similar and they usually evaluate particular symptoms from some of the affected cognitive domains. This thesis aims to address those issues taking advantage of new non-invasive affordable technologies; from Virtual Reality for task creation flexibility and ceiling effect attenuation to emotion analysis as novel approach to assess cognitive impairments, using \hyperlink{a_EEG}{EEG} and facial expression analysis. Next paragraphs outline these problems as well as the proposed solutions and future work.


\section{AD Screening Cognitive Tests}

The main tests used by doctors for \hyperlink{a_AD}{AD} screening are cognitive tests. These tests are usually short, around five to twenty minutes, and it is possible to complete them with only pen and paper and the supervision of a specialist. These tests usually examine cognitive domains such as learning and memory or executive function. The screening provided by these tests is used to know if any patient require further examinations but they do not provide a reliable diagnosis. The most commonly used and studied screening test is the \hyperlink{a_MMSE}{MMSE}, but other tests such as miniCog or \hyperlink{a_SLUMS}{SLUMS} have been proved more convenient for early detection. These tests are usually part of an exhaustive assessment that includes visual attention evaluation and physical examinations. If the assessment considers the patient as possible \hyperlink{a_AD}{AD} impaired, further analysis will include more reliable but invasive or uncomfortable and expensive techniques such as \hyperlink{a_MRI}{MRI}.

\subsection{Issues}
Commonly used cognitive \hyperlink{a_AD}{AD} screening techniques are not reliable to provide a diagnosis so expensive and sometimes painful techniques need to be applied. The detection performance of these tests need to be improved in order to avoid the application of further invasive methods. In order to do that, well-known problems should be solved or alleviated. For example, tasks should be adapted to subjects IQ and diverse in order to cover as many cognitive domains related to \hyperlink{a_AD}{AD} symptoms as possible.

Current research is including computerised tests due to their capability to automatically adapt to patients. Virtual reality technology has been boosted during last years allowing the creation of a diverse range of assessment tasks, including those ones which are complicated to recreate on a real scenario. Nevertheless, current proposals usually lack realism, immersion or the technology utilised is expensive and non-easily accessible.

\subsection{Proposed Solution}
Virtual Reality technology started to be used in many areas such as rehabilitation, phobia treatment or medical diagnosis. The appearance of many \hyperlink{a_VR}{VR} devices and the inclusion of \hyperlink{a_VR}{VR} on mobile phones have made this technology widely accessible and affordable. Diagnosis on virtual environments provides a safe environment where a wide range of tasks can be performed to evaluate the different cognitive domains affected by \hyperlink{a_AD}{AD}. In addition, computerised tasks include the capacity to be adaptable to the user, taking into account the user selections, being able to reduce the ceiling effect. The proposed tests have been developed using \hyperlink{a_VR}{VR} technology such as \hyperlink{a_VR}{VR} glasses and mobile phones, both affordable and easily accessible technologies. Four tests/tasks have been proposed based on different cognitive domains \cite{MontenegroIISA2015, Montenegro2015, Montenegro2016, Montenegro2017}. The first one evaluates memory and learning and perceptual-motor domains. The second task focuses on perceptual-motor and executive function. The third one analyses complex attention and memory and learning. The forth test is based on the inverse Turing problem, that is, instead of analysing how intelligent a computer is, the ability of the participant to discern real or incoherent information is evaluated. This challenges their executive function cognitive domain.

The tests have been adapted to high IQ, increasing  number of subtasks and their difficulty according to the educational level of the participant. The tasks are performed on a realistic virtual office and the participants are fully immersed using \hyperlink{a_VR}{VR} glasses. A Kinect sensor is used to track participants' movements so they can see their avatar moving accordingly, and to interact in a subtask of the second test. A mobile phone version has also been created to widely increase the accessibility and diffusion of the application.

Twenty participants, one of them diagnosed with mild \hyperlink{a_AD}{AD}, were evaluated using these novel tests and three state-of-the-art ones. The results were compared and the novel tests probed valid for \hyperlink{a_AD}{AD} screening. Nevertheless, since the amount of participants was small, they need to be further demonstrated. The main purpose of our proposal is the creation of a healthy model so any result out of the model will identify a cognition impairment, thus no additional \hyperlink{a_AD}{AD} patients would be required but the number of healthy participants need to be considerably increased.

\subsection{Future Work}
Further testing on \hyperlink{a_AD}{AD} patients is required to validate the diagnosis results obtained through the usage of the \hyperlink{a_VR}{VR} tests. The number of participants that tested the applications was reduced. The upload of the application on mobile phone applications stores will allow to obtain healthy participants scores but \hyperlink{a_AD}{AD} ones require ethics that will make more difficult the acquisition of results for comparison.

The interaction with the virtual environment can be more realistic. Currently, participants use a button on the headset or on the headset controller to select and proceed through the tests. The next step will include voice interaction, which would make the selection of objects from the screen more natural. A second update will include Brain Computer Interaction and eye tracking techniques. Eye tracking will help to navigate the \hyperlink{a_VEs}{VE}, using the gaze as the cursor during the tasks. The eye tracker could be added as a new attached device or it could be included to the \hyperlink{a_EEG}{EEG} sensors' functionality. The \hyperlink{a_EEG}{EEG} will be added as a reduced number of electrodes that will monitor mental workload, attention and emotions to adapt the game to the participants' mental state. Additionally, it can also be used to execute commands such as press or hold.


\section{BCI and EEG Emotion Recognition for AD Screening}
Brain Computer Interfaces as part of \hyperlink{a_HCI}{HCI} will integrate the techniques that allow to control external devices through mental processes. \hyperlink{a_EEG}{EEG} is the most used \hyperlink{a_BCI}{BCI} technique, it translates the bioelectric activity of the brain into actions or feedback understandable by computers. Navigation is one of the most common \hyperlink{a_BCI}{BCI}'s applications. It allows to perform simple actions such as pushing, grabbing or selecting, which will facilitate navigation in computers applications and virtual environments. This type of navigation will help subjects with reduced mobility to interact with new technologies, providing many advantages on health and care related areas. The feedback collected from brain activity will also be crucial in future applications and other areas such as marketing. Research has shown that amongst the brain signal recorded it is possible to study subject's mental state such as workload, attention or emotions. This feedback will allow to adapt applications to each user's mental state, making them more satisfying and boosting the immersion.

\subsection{Issues}
Most literature \hyperlink{a_EEG}{EEG} techniques for navigation purposes are usually implemented using sixteen to sixty-four electrodes/sensors. These devices extend through the subjects' scalp to receive signal from different areas of the brain. Depending on the application, certain areas of the brain do not provide relevant information so a reduced number of sensors could be utilised. A reduced amount of sensors would be more comfortable to wear and easier to combine with \hyperlink{a_VR}{VR} devices.

\hyperlink{a_EEG}{EEG} methods for mental assessment, navigation or emotion recognition are still not good enough to be incorporated into market applications. Emotion classification performance using \hyperlink{a_EEG}{EEG} signal is still below other modalities such as facial images. \hyperlink{a_EEG}{EEG} emotion descriptors that allow to perfectly differentiate amongst emotions have not been found yet.

When it comes to Alzheimer's Disease, there is not specific database that has been created for \hyperlink{a_AD}{AD} diagnosis containing multimodal information of participants while performing AD assessment tasks. Most of the research work analyse the capability of \hyperlink{a_AD}{AD} patients to identify emotions, however, there are not many works that study their emotional reactions.

\subsection{Proposed Solution}
Novel features using four frontal electrodes signal were proposed in order to detect the participants' gaze position on a monitor \cite{MontenegroGaze2016}. These features are based on quaternion \hyperlink{a_PCA}{PCA}. They help to reduce the loss of correlated information amongst the channels. They are used as input to different classifiers such as \hyperlink{a_kNN}{kNN} and \hyperlink{a_SVM}{SVM} in order to test their validity for navigation of virtual environments.

A database for emotions classification based on \hyperlink{a_AD}{AD} symptoms (\hyperlink{a_SEMdb}{SEMdb}) was created. Three different tasks related with cognitive domains impaired by \hyperlink{a_AD}{AD} were recorded. The first task was related with attention impairment. The resultant data could be also used for navigation training purposes. The second one analysed participants' searching ability. This task evaluated the complex attention and perceptual motor cognitive domains. The third one involves the main novelty of the database since it recorded spontaneous reactions to autobiographical stimuli. This task challenged the learning and memory and the perceptual-motor cognitive domains.

Using the third task and the \hyperlink{a_EEG}{EEG} quaternion features, a method was proposed to classify autobiographical emotions \cite{MontenegroFace2016, MontenegroEmo2016}. The classification performance was measured using the F1 score and it was compared with an \hyperlink{a_EEG}{EEG} state-of-the-art method. The proposed \hyperlink{a_EEG}{EEG} descriptors combined with \hyperlink{a_SVM}{SVM} and Adaboost outperformed the literature method. In addition, they were also compared with facial expression recognition approaches using both RGB and depth images. The overall results confirmed the superiority of \hyperlink{a_FER}{FER} techniques.

\subsection{Future Work}

\hyperlink{a_BCI}{BCI} could further improve the mental cognition assessment when combined with \hyperlink{a_AD}{AD} screening applications on Virtual Environment. \hyperlink{a_EEG}{EEG} techniques need to be improved in order to obtain more reliable feedback. Currently deep learning techniques are being effectively used in many areas. These techniques usually require big amount of data to be reliable. Therefore, increasing the amount of \hyperlink{a_SEMdb}{SEMdb} data, extracting the novel features and using them as input to deep learning networks, could help to increase the emotion classification rates. Due to the difficulty obtaining information from Alzheimer's patients mainly due to ethical reasons, the emotion classification models would be based on One Class Classification, since the amount of information will not be balanced. Using \hyperlink{a_OCC}{OCC} and increasing the amount of healthy participants it is possible to create a model that identifies the reactions of healthy participants to autobiographical and non-autobiographical memory related stimuli, thus being able to identify as cognitive impairment any reaction out of this model.

Afterwards, this method could be added for workload assessment, attention analysis, emotion recognition and navigational functions such as selecting options. Furthermore, the \hyperlink{a_AD}{AD} assessment tasks could be adapted to the mood of the participants, making them more effective.


\section{Facial Emotion Recognition}
Facial expression recognition approaches usually provide better emotion classification rates than \hyperlink{a_EEG}{EEG} ones. Reactions to specific stimuli can help to analyse different cognitive domain impairments. Current techniques provide high accuracy for emotion recognition on posed emotions and facial expressions data recorded in controlled environments.

\subsection{Issues}
Facial expression recognition techniques' performance is still not optimal in some scenarios. Classification of basic emotions or action units on data in the wild and micro-expression classification can be improved. The amount of research focused on micro-expressions is reduced, therefore there is room for new descriptors that accentuate facial micro-features. There is not any known approach with the aim of classifying emotions triggered by autobiographical memories. The application of existent \hyperlink{a_FER}{FER} methods, that are used to classify \hyperlink{a_AUs}{AUs} and basic emotions, results in improvable classification accuracies and F1 scores.

\subsection{Proposed Solution}
Enhancing preprocessing techniques and novel descriptors are introduced in order to emphasise micro-expressions. Eulerian magnification is used during preprocessing stages in order to increase facial expressions' motion on images so micro-expressions are accentuated. In addition, novel features based on origami crease patterns are used to classify emotions from CK+ \ref{db_CK} and \hyperlink{a_SEMdb}{SEMdb} databases. The results are compared with a state-of-the-art method that uses pyramid of \hyperlink{a_HOG}{HOG} features and Bayesian classifiers. The novel features combined with \hyperlink{a_SVM}{SVM} classifiers increase the accuracy and F1 score for classification of \hyperlink{a_AUs}{AUs}, basic and autobiographical related emotions. Additionally, a proof of concept experiment was performed proving the feature robustness to noise and partial occlusions.

\subsection{Future Work}
Novel features have proved to improve the performance of state-of-the-art methods for emotion classification, but they still leave room for improvement. Therefore, future work will include the use of deep learning techniques in combination with the proposed features. Deep learning for facial emotion recognition is reporting very high \hyperlink{a_FER}{FER} rates in controlled scenarios and posed databases and it has highly improved the classification performance on data in the wild and micro-expressions.

Once the emotion recognition method is reliable on data from \hyperlink{a_SEMdb}{SEMdb}, One Class Classification will be applied due to the imbalanced data (between healthy and \hyperlink{a_AD}{AD} patients). Even if the database is updated with new healthy and \hyperlink{a_AD}{AD} participants recordings, it is very likely the imbalance is still considerable. \hyperlink{a_OCC}{OCC} will provide a model that define micro-facial expressions of healthy subjects when reacting to autobiographical memories.


\section{Epilogue}
The goal of this thesis was the creation and improvement of methods for \hyperlink{a_AD}{AD} early diagnosis; taking into account the associated high cost, the limited accessibility, the lack of comfortability of the most reliable methods and the lack of reliability of the comfortable ones. Three different approaches have been proposed: design and creation of novel non-invasive cognitive tests on Virtual Environments, creation of a novel database based on \hyperlink{a_AD}{AD} symptoms, and analysis of emotions to distinguish reactions to specific autobiographical stimuli.

Virtual Reality has been one of the main research topics during last years. Its convenient usability for health related issues, such as phobias, rehabilitation, treatment and disease diagnosis was demonstrated. \hyperlink{a_VR}{VR} technologies allow a realistic immersion on \hyperlink{a_VEs}{VE} and provide a safe scenario where any kind of task can be performed. These technologies have also been optimised to be accessible from most of the current mobile phones, therefore, they are affordable and accessible. In addition, current advances on \hyperlink{a_BCI}{BCI} promise future realistic navigation on \hyperlink{a_VEs}{VEs} and the possibility to adapt the application to the mental state of each participant. The proposed \hyperlink{a_VR}{VR} tool for early \hyperlink{a_AD}{AD} screening has probed the wide range of tasks that could be created to evaluate different cognitive domains affected by \hyperlink{a_AD}{AD}. This part of the research provide researchers with ideas to create tests for the diagnosis of \hyperlink{a_AD}{AD} using affordable and accessible technology that could reduce the \hyperlink{a_AD}{AD} associated cost and could help patients to be diagnosed earlier.

The analysis of reactions to specific stimuli related with AD could be also used for \hyperlink{a_AD}{AD} screening. Lack of research regarding the analysis of \hyperlink{a_AD}{AD} patients' emotions is counteracted by the great amount of emotion recognition research. \hyperlink{a_EEG}{EEG} for emotion recognition is promising and it will be more adequate in future applications since they are not affected by environmental interferences like other approaches such as audio and images. Nevertheless, facial emotion recognition techniques are still superior to \hyperlink{a_EEG}{EEG} ones. Despite obtaining good emotion classification performance, current approaches leave room for improvement, mainly in data in the wild and micro-expressions. Nowadays, emotions research is focusing on deep learning techniques due to its success on other areas. Therefore, future work will concentrate on increasing the amount of data of specific databases for the purpose of improving deep learning techniques performance. The research related to the assessment of \hyperlink{a_AD}{AD} patients using emotions has provided a novel type of low cost tests that could be included to early \hyperlink{a_AD}{AD} screening battery tests. In order to accomplish this part of the research, a novel database based on \hyperlink{a_AD}{AD} symptoms was published which can be helpful to researchers that want to investigate emotions related with autobiographical memories. Moreover, novel features for emotion classification were proposed, one of them extracted from \hyperlink{a_EEG}{EEG} signal and the other from facial landmarks. These novel features can be useful for the general analysis of emotions or any other area that work with this type of data. For example, \hyperlink{a_EEG}{EEG} analysis for navigational or workload purposes, which can be useful for gaming.

As a summary, it is suggested the use of Virtual Environments for the diagnosis of dementias due to the wide range of tasks that can be created, the security and realism they provide, the reduced cost of the technology and its accessibility. It is also recommended the use of \hyperlink{a_EEG}{EEG} for computer interaction improvement and mental status assessment. Moreover, the study of emotions is also suggested as an extra task for \hyperlink{a_AD}{AD} diagnosis. Currently, the best emotion classification performance is obtained using facial images.




\addtocontents{toc}{\vspace{2em}} 

\appendix 



\chapter{Virtual Environment Tests Documents} 

\label{AppendixA} 

\lhead{Appendix A. \emph{Virtual Environment Tests Documents}} 

\section{VE Tool - Test Instructions} \label{AppAinstructions}

This test is formed by different subtests designed for dementia detection. Before the test starts, some general information about yourself is going to be collected (Age - Gender - Country - Type of Dementia - Dementia Level). Nevertheless, it is not going to be possible to identify you with that information.
If you do not have any inconvenience, we are going to start with the test. It is necessary to follow the instructions provided. In addition, the supervisor of the test will help you during the process.

1) General personal information:
It is necessary to fill the personal information section in order to generate a profile for your results.

2) Memory Test:
The first test is the Dr Oz Memory Quiz. In this test, you have to answer general questions, to solve simple problems and to do some memory tasks. There is a task where your time is going to be measured. When you reach that task, wait until the supervisor tells you to continue.

3) Visual test:
The second test is the Visual Association test. During this test, six images of 2 interacting objects are going to be shown. Memorise the objects.

4) Audio test:
The third test is the Dichotic Listening test. You are going to listen to two digits simultaneously, one on each ear. You are going to hear a total of 6 digits that you have to memorise, this is, 3 rounds of 2 digits. Afterwards, you have to write the numbers that you recall on the text areas.

This task is going to be done 8 times with different digits. Therefore, take your time between each round of numbers.

IMPORTANT: It is not necessary to remember the order of the digits.
You have to start writing in the left text area.

Now, we are going to start the test on a virtual environment. You are going to wear the Oculus Rift glasses during this test. Please, follow the instructions provided by your supervisor.

5) Objects memory test:
The first test is the memorization test. During this test, different objects are going to be shown. Your objective is to memorise them since you will need to recall them later. Follow the instructions on the screen. In order to select the different options you have to put the scope that will appear on the screen over the label you want to select.

6) Abnormal object test:
Follow the instructions provided by the test supervisor.

7) Sounds test:
Follow the instructions provided by the test supervisor.

8) Doctor Bot test:
During this test, you can find information about dementia. Different information is going to be provided by two different avatars. Follow the instructions on the screen. The selection mode is the same as in the Objects memory room, this is, place the scope over the chosen selection.

\section{VE Tool - Evaluation Results Sheet}
1) Explain the user that he/she is going to do some Alzheimer's tests on the computer. That the tests are guided and an explanation of each test is going to be given at the beginning of each test.\\
\\
2) Provide the User's guide and start the application.\\
\\
3) The main menu collects information about the patient.\\
\\
Age - Gender - Country - Type of Dementia - Dementia Level\\
	Age: \underline{\hspace{1cm}}\\		
	Gender: 	male	/    female \\
	Country:	\underline{\hspace{1cm}}\\
	Educational Background:  Under High School    /     High School    /    Over High School \\
	Type of Dementia:    Healthy    /    Alzheimer's \\
	Dementia Level:   None    /    Mild    /    Moderate     /     Severe \\

A text file, whose name is "YearMonthDay-HourMinuteSecond.txt", is created. This file will store the results from the next 3 tests.\\
\\
4) Memory test (Dr. Oz):
Tell the subject that there is a task where the time is going to be measured; and pen and paper will be provided (Animals recall task). Therefore, they have to wait in this task before they proceed to the next one.\\
	Number of correct:	\underline{\hspace{1cm}}\\
\\
5) Visual test:
Explain that he/she is going to see 6 images, each one with 2 interacting objects. They have to remember the objects.\\
	(chair - umbrella - balloon - saucepan - baby carriage - inkwell)\\
	Number of correct:	\underline{\hspace{1cm}}\\
\\
6) Audio test:
Explain the subject that they are going to hear two digits simultaneously, one on each ear. They are going to hear 6 digits that they have to memorise, this is, 3 rounds of 2 digits. Afterwards, they have to write the numbers that they recall on the text areas.\\
IMPORTANT: They do not have to recall the order of the digits.\\
They have to start writing in the left text area.\\
\\
	The second slide is an example.\\
		Number of Left correct:		\underline{\hspace{1cm}}\\
		Number of Right correct:	\underline{\hspace{1cm}}\\
		Number of First Right:		\underline{\hspace{1cm}}\\

---------------- HERE THE STATE-OF-THE-ART TESTS FINISH -------------------

Move the patient to a position where he/she is detected by Windows Kinect (always sitting in a chair).  Explain that during the next tests he/she will use the Oculus Rift glasses.

7)  Objects memory test:
	Explain the user that he/she has to follow the instructions on the screen and that she/he will need to select options on the screen using Oculus Rift. A scope will be in the centre of the screen and, in order to do the selection, the scope has to be positioned over the chosen label.\\
\\
	7.1) Objects recognition: 		\underline{\hspace{1cm}} \underline{\hspace{1cm}} \underline{\hspace{1cm}} \underline{\hspace{1cm}} \underline{\hspace{1cm}} \underline{\hspace{1cm}}\\
					(Apple, Mug, Glasses, Book, Keyboard, Monitor)\\
\\
	7.2) Press key 'Return' to go to the next task:\\
	        Location recall: 		\underline{\hspace{1cm}} \underline{\hspace{1cm}} \underline{\hspace{1cm}} \underline{\hspace{1cm}} \underline{\hspace{1cm}} \underline{\hspace{1cm}}\\
\\
	7.3) Press key 'Return' to go to the next task: Where the user has to find and name the interacting objects.\\
	        Press key 'n' to start the interacting object recall:\\
	        Interacting objects recall:		\underline{\hspace{1cm}} \underline{\hspace{1cm}} \underline{\hspace{1cm}}\\
					(Gorilla-Banana, Plant-Butterfly, Cat-Rat)\\
\\
	After the user has done all the tasks, ask him/her:\\
\\
	7.4) Ask for the time in the virtual room.\\
\\
	7.5) Ask them to quit the Oculus and ask them for the time again. These last two tasks are focused on making the user understand that the clock of the virtual room is working properly.\\
\\
	7.6) What were the first 6 objects over the table?\\
		First 6 objects recall:		\underline{\hspace{1cm}} \underline{\hspace{1cm}} \underline{\hspace{1cm}} \underline{\hspace{1cm}} \underline{\hspace{1cm}} \underline{\hspace{1cm}}\\
					(Apple, Mug, Glasses, Book, Keyboard, Monitor)\\
\\
8) Abnormal object test:\\
	Provide the instructions once the patient is inside the virtual room:\\
\\
	8.1) Can you find the abnormalities of the room? Name them and tell me what is wrong with them?\\
		6 abnormalities:\\
			Plant: 		\underline{\hspace{1cm}}	(upside down)\\
			Chair: 		\underline{\hspace{1cm}}	(moving alone)\\
			Mug:		\underline{\hspace{1cm}}	(spinning)\\
			Fake mirror:	\underline{\hspace{1cm}}	(the car is not reflecting in it)\\
			Car:		\underline{\hspace{1cm}}	\underline{\hspace{1cm}}	 (2 abnormalities*)\\
		* The Car has 2 abnormalities: it changes when it goes across the mirror and it can go through the wall. -> If they name the car, try to find if they detect both abnormalities.\\
\\
	8.2) If they do not detect that the time is wrong, ask them:\\
	What time is it? Ask them if the time light is according to that time in the morning (since the clock is analogical).\\
			Time:		\underline{\hspace{1cm}}\\
\\
	8.3) Once the abnormalities have been detected, press 'Return' to make appear the mirror avatar.\\
		8.3.1) Ask the patient to lift the right hand. Ask him if the avatar behaviour is like a mirror.\\\underline{\hspace{1cm}}	(It is not)\\
\\
8.3.2) Ask the patient to bend the body of the avatar in front of him towards the mug.\\		\underline{\hspace{1cm}}	(if the patient does it in the first attempt)\\
\\
8.3.3) Ask the patient to grab the mug.\\
					\underline{\hspace{1cm}}	(if the patient do it in the first attempt)\\
\\
9) Sounds test:\\
	IMPORTANT: The supervisor has to have his/her mobile phone ready in order to reproduce the mobile phone sound.\\
	Provide the instructions once the patient is inside the virtual room:	\\
\\
	9.1) Ask the user to name all the sounds he/she listens during the test.\\
		9.1.1) Sounds always present:   	Clock:	\underline{\hspace{1cm}}
			          			Fly:	\underline{\hspace{1cm}}\\
\\
		9.1.2) Mobile phone. When the supervisor thinks that it is convenient. Press key 'm' to make the virtual mobile phone move and AT THE SAME TIME make your mobile phone sound. Press key 'n' in order to make the mobile phone to stop moving.\\
						Mobile Phone: \underline{\hspace{1cm}}\\
\\		
		9.1.3) Door. Press key 'd' to reproduce the sound of the door closing. Check/ Add a note if the user turn around to see the door.\\
			Door:	\underline{\hspace{1cm}}		Note:\\
\\
		9.1.3) When the patient is not looking towards the mug in the room (make him/her look to another direction). Press the key 'g' in order the mug crashing sound is reproduced.\\
	Check/ Add a note if the user turn to see the mug.\\
			Mug: 	\underline{\hspace{1cm}}		Note:\\
\\
9.1.4) Rain: Press key 'r' in order to start the rain and the rain sound. Press key 's' in order to stop the rain and the rain sound.\\
			Rain:	\underline{\hspace{1cm}}\\
\\
10) Doctor Bot test:\\
	Explain the user that during this test two different avatars are going to provide different information about Alzheimer. They can choose a question selecting with the Oculus as they did during the Objects memory test.\\
	10.1) Which doctor gave the correct answers?\\
		a) The first one :	\underline{\hspace{1cm}}	(3 points)\\
		b) The second one:	\underline{\hspace{1cm}}	(0 points)\\
		c) Both of them:	\underline{\hspace{1cm}}	(1 point)\\
		d) None of them:	\underline{\hspace{1cm}}	(1 point)\\

\section{VE Tool - Interview} \label{Interview}

From 1 (Really bad) to 5 (Really good) describe and comment:

1. Mouse/Keyboard interaction.
1	2	3	4	5

2. Kinect/Oculus interaction.
1	2	3	4	5

3. Are the instructions easy to understand?
1	2	3	4	5

4. How comfortable is the Oculus Rift?
1	2	3	4	5

5. Comment about the application.


\chapter{SEMdb - Ethics Form} 

\label{AppendixB} 

\lhead{Appendix B. \emph{SEMdb - Ethics Form}} 

Faculty of Science, Engineering and Computing

Digital Imaging Research Centre

Informed Consent Form for RGB camera, Kinect, Eye tracker and EGG Recording - Strictly Confidential

Name:

Please answer the following questions truthfully and completely. The aim of this test is the recording of RGBDI video using the Kinect and RGB cameras, the movement of your eyes using an eye tracker and your brain signal using EEG sensors when different visual stimulus are presented.  Please answer questions 1-2 now and the remaining questions after the activity has taken place.

There are not side effects.  You do not need to take part in this study, and you can leave it at any time without affecting your education/relationship with the Faculty or University in any way.

Q.1) Do you know of any reason why you should not to participate in the proposed
testing protocol? If yes please give details.

Yes~~~~No

Details:
.............................................................................\\
.............................................................................\\

Q.2) Please confirm that you have fully understand the test procedure.

Yes~~~~No

Q.3) Do you permit the usage of the recorded video, brain signal, depth, infrared and eye tracker data for the purposes of research on Computer Vision algorithms?

Yes~~~~No

Q.4) Do you permit the publication of the recorded video, brain signal, depth, infrared and eye tracker data in scientific papers, conferences, workshops and websites for the purposes of research on Computer Vision algorithms?

Yes~~~~No

Q.5) Do you permit the use of captures of your video, infrared and depth data  to be used in scientific papers, conferences, workshops and websites for the purposes of research on Computer Vision algorithms?

Yes~~~~No

Q.6) Do you permit the recorded data to be used by other researchers? The data will be used for university-related research work and not for any commercial purposes.

Yes~~~~No

All information we gain from you will be maintained in a strictly confidential manner.  The only people who will have access to the information will be the researchers linked with this project.  After the project all raw data that can identify individuals will be keep in a locked hard drive whose access will be only accessed by authorised researchers.  In the reporting of the project, no information will be released which will enable the reader to identify who the respondent was unless you authorise it in this form. If you have any questions or problems, please contact me. My email is k1274433\@kingston.ac.uk

Yours sincerely

Juan Manuel Fernandez Montenegro

//
//
//

Statement by participant

I confirm that I have read and understood the information sheet/letter of invitation for this study. I have been informed of the purpose, risks, and benefits of taking part.

(Title of Study)-  Alzheimer’s Disease Monitoring And Diagnosis Based On Behaviour Analysis

I understand what my involvement will entail and any questions have been answered to my satisfaction.

I understand that my participation is entirely voluntary, and that I can withdraw at any time without prejudice.

I understand that all information obtained will be confidential.

I agree that research data gathered for the study may be published provided that I cannot be identified as a subject unless I have authorised it in this form.

Contact information has been provided should I (a) wish to seek further information from the investigator at any time for purposes of clarification (b) wish to make a complaint.

Participant’s Signature\\
---------------------------------------------------

//
Date: ~~/~~/~~

//
//
//

Statement by investigator

I have explained this project and the implications of participation in it to this participant without bias and I believe that the consent is informed and that he/she understands the implications of participation.

Name of investigator: -----  Juan Manuel Fernandez Montenegro  -----

Signature of investigator:// -------------------------------------------------------------------------------------------

Date: -----------------------------------------------------------------------------------
//
//
//

Faculty of Science, Engineering and Computing

Digital Imaging Research Centre

Health screening questionnaire

Name:	 		          Date of Birth: /  /

Contact Number:  	      Email:

Sex: 		 		      Left or right handed:

Wearing Glasses or Contact lenses:


\chapter{Motion Enhancing and Origami Algorithm} 

\label{AppendixC} 

\lhead{Appendix C. \emph{Motion Enhancing and Origami Algorithm}} 

\section{Eulerian Magnification}
Eulerian Magnification is used to reveal temporal variations in videos that are visually difficult to appreciate~\cite{Wu2012,Wadhwa2016}. In order to obtain the magnified video sequence, the original video goes through spatial decomposition and temporal filtering. Each sequence of images is decomposed into different spatial frequency bands. Afterwards a bandpass temporal filter is applied to each spatial band. The resultant filtered bands are then amplified and added to the original sequence.

The use of Eulerian magnification has been proved to increase the classification results for facial emotion analysis. The work presented in~\cite{Park2015} uses Eulerian magnification on a spatio-temporal approach that recognises five emotions using a \hyperlink{a_SVM}{SVM} classifier reaching a 70\% recognition rate on CASME II dataset.~\cite{LeNgo2016} also proves an improvement of 0.12 in the F1 score using a similar approach by Park et al.

\section{Lang's Universal Molecule Algorithm}
\begin{figure}
\centering
  \includegraphics[scale = 0.7]{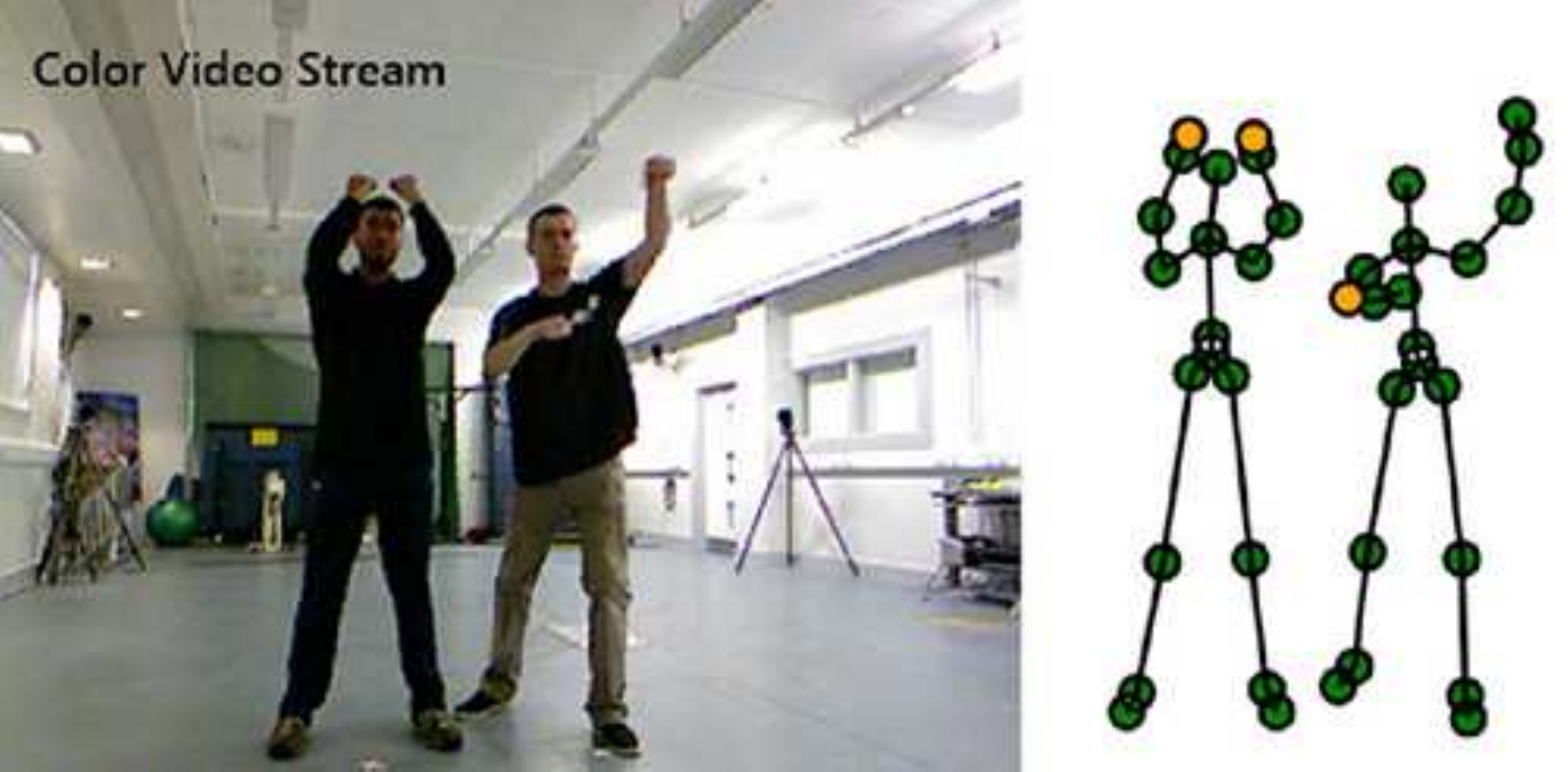}
  \caption{Example of a shadow tree projection from a human 3D model~\cite{Bloom2014,Bloom2015,Bloom2016}.}
  \label{fig:shadowTreeExp}
\end{figure}

A crease pattern is the underline blueprint for an origami figure. The universal molecule is a crease pattern constructed by the reduction of polygons until they are reduced to a point or a line. Lang~\cite{Lang1996} presented in 1996 a computational method to produce crease patterns with a simple uncut square of paper that describes all the folds necessary to create an origami figure. Lang's algorithm proved that it is possible to create a crease pattern from a shadow tree projection of a 3D model. This shadow tree projection is like a dot and stick molecular model where the joints and extremes of a 3D model are represented as dots and the connections as lines (see Figure~\ref{fig:shadowTreeExp}).

The universal molecule algorithm has been useful for other applications apart from the creation of origami. Most of these applications are related with the capability of folding big united areas into a smaller folded version, such as airbag flattening or a 100-meter aperture space telescope~\cite{Lang2004}.

\addtocontents{toc}{\vspace{2em}} 

\backmatter


\label{Bibliography}

\lhead{\emph{Bibliography}} 

\bibliographystyle{unsrtnat} 

\bibliography{Bibliography} 

\end{document}